\documentclass[twocolumn,10pt]{article}
\usepackage[protrusion=true,expansion=true]{microtype}
\usepackage{ebgaramond}
\usepackage{times}
\usepackage[thmmarks]{ntheorem}
\usepackage{amssymb}
\usepackage{mathtools}
\usepackage{mathrsfs}
\usepackage{courier}
\usepackage{lipsum}
\usepackage[square,numbers]{natbib}
\usepackage{graphicx}
\usepackage{subcaption}
\usepackage{titlesec}
\usepackage[framemethod=TikZ]{mdframed}
\usepackage[most]{tcolorbox}

\usepackage[inner=0.2in, outer=0.2in, top=0in, bottom=0in, includeheadfoot]{geometry}
\usepackage{fancyhdr}
\fancypagestyle{plain}{
    \fancyhf{}
    \fancyfoot[C]{\raisebox{0.2in}{\thepage}}
    
}

\usepackage[boxruled,linesnumbered,vlined,inoutnumbered]{algorithm2e}
\SetNlSty{}{}{:}

\usepackage{enumitem}
\setlist[itemize]{topsep=5pt, partopsep=0pt, parsep=3pt, itemsep=0pt}
\setlist[itemize,1]{left=0pt}

\usepackage{xcolor}
\definecolor{dark-blue}{rgb}{0,0,0.7}
\definecolor{figRed}{RGB}{192, 0, 0}
\definecolor{figGreen}{RGB}{0,100,0}
\definecolor{figBlue}{RGB}{0, 20, 168}
\definecolor{eqnRed}{RGB}{192, 0, 0}
\definecolor{eqnGreen}{RGB}{0, 150, 0}
\definecolor{eqnBlue}{RGB}{0, 20, 168}
\definecolor{comment}{rgb}{0,0.5,0}
\definecolor{Color-BAC-Alg-Special}{rgb}{0,0.5,0.0}

\usepackage{hyperref}
\hypersetup{
    colorlinks, linkcolor={dark-blue},
    citecolor={dark-blue}, urlcolor={dark-blue}
}

\titleformat{\section}
  {\normalfont\bfseries}
  {\thesection}{1em}{}
\titlespacing*{\section}
  {0pt}{2ex plus 1ex minus .2ex}{0ex}

\titleformat{\subsection}
  {\normalfont\bfseries}
  {\thesubsection}{1em}{}
\titlespacing*{\subsection}
  {0pt}{2ex plus 1ex minus .2ex}{0.5ex}

\titleformat{\subsubsection}
  {\normalfont\bfseries}
  {\thesubsubsection}{1em}{}
\titlespacing*{\subsubsection}
  {0pt}{2ex plus 1ex minus .2ex}{0.5ex}

\newtheorem{defn}{Definition}

\newtheorem{prop}{Property}
\newtheorem{ass}{Assumption}

\theoremstyle{nonumberplain}
\theoremheaderfont{\normalfont\itshape}
\theorembodyfont{\normalfont}
\theoremsymbol{$\square$}
\newtheorem{proof}{Proof.}
\theorempostwork{\normalfont\hfill$\square$}

\DeclareMathOperator*{\argmax}{arg\,max}

\input{widebar.tex}

\newcommand{\definesymbol}[2]{%
    #1\label{sym:#2}%
}

\begin{document}

\newgeometry{margin=2in}
\begin{titlepage}
  \thispagestyle{empty}
  \begin{center}
    {\fontfamily{EBGaramond-TLF}\selectfont\Huge \bfseries Qualia Optimization\par}
    \vspace{2em}
    {\Large Philip S. Thomas\par
    University of Massachusetts}
    
    \vspace{6em}
    \textbf{Abstract}\par
    \vspace{0.5em}
    \begin{minipage}{0.9\textwidth}
    This report explores the speculative question: what if current or future AI systems have qualia, such as pain or pleasure? It does so by assuming that AI systems might someday possess qualia---and that the quality of these subjective experiences should be considered alongside performance metrics. Concrete mathematical problem settings, inspired by reinforcement learning formulations and theories from philosophy of mind, are then proposed and initial approaches and properties are presented. These properties enable refinement of the problem setting, culminating with the proposal of methods that promote reinforcement. 
    \end{minipage}
    \vfill  
    \begin{minipage}{0.9\textwidth}
    \small
    \noindent\rule{\linewidth}{0.4pt}
    \emph{This version has been posted to arXiv for archival purposes and includes a cover page specific to the arXiv submission. The original publication is a technical report entitled \emph{Qualia Optimization}, Technical Report UM-CICS-2025-001, published in May 2025 by the College of Information and Computer Sciences at the University of Massachusetts.}
    \end{minipage}
  \end{center}
\end{titlepage}
\restoregeometry
\clearpage

\noindent{\fontfamily{EBGaramond-TLF}\selectfont\large \textbf{\hspace{0.0cm}Qualia Optimization}}\\
\noindent Philip S. Thomas, University of Massachusetts\\
\noindent Version 1, Technical Report UM-CICS-2025-001
\\\\
\noindent \textbf{Abstract}\\
This report explores the speculative question: what if current or future AI systems have qualia, such as pain or pleasure? It does so by assuming that AI systems might someday possess qualia---and that the quality of these subjective experiences should be considered alongside performance metrics. Concrete mathematical problem settings, inspired by reinforcement learning formulations and theories from philosophy of mind, are then proposed and initial approaches and properties are presented. These properties enable refinement of the problem setting, culminating with the proposal of methods that promote reinforcement. 
\vspace{-0.01cm}
\section{Introduction}
\label{sec:introduction}

The development of increasingly capable AI systems has heightened interest in a range of philosophical questions related to ethical, social, and economic issues \citep{fazelpour2021algorithmic,roberts2022artificial}. At the intersection of AI and a subfield of philosophy called \emph{philosophy of mind}, this report addresses a particularly speculative question: what if current or future AI systems have qualia? Qualia---subjective conscious experiences such as the ``redness'' of red or the sensations of pain or pleasure---have long been a topic of debate in the philosophy of mind. We adopt the assumption that AI systems might have qualia akin to those of humans and that the quality of these subjective experiences should be considered alongside traditional performance metrics. We then explore the implications of these assumptions for AI methodologies.

The discussion in this report applies to AI systems wherein an agent makes a sequence of decisions or predictions. The mathematical formulation that we consider, which we call an \emph{agent-environment process} (AEP), generalizes and extends settings typically considered in \emph{reinforcement learning} (RL) like \emph{Markov decision processes} (MDPs) and \emph{partially observable Markov decision processes} (POMDPs). Specifically, an AEP is an MDP with a formulation of the agent and its state added and with the inclusion of rewards made optional. Although our initial motivation comes from RL and we focus on examples of RL agents, the AEP setting that we consider is more general and is designed to accommodate a range of AI systems including supervised and unsupervised learning systems and RL systems where rewards are absent or defined differently.

We begin by reviewing theories in philosophy of mind with an emphasis on their compatibility with the idea of AI systems experiencing qualia. We focus on the idea that if certain algorithmic processes in an AI system mirror the mechanisms that give rise to specific human experiences (such as pain or pleasure), it might lead to the AI system having similar qualia. This focus provides guidance regarding how the quality of agent experiences can be quantified. 

After providing additional background on RL and neuroscience, we present different ways of formulating the problem of maximizing the quality of the experiences of an agent, while being cognizant of the potential impact on performance---a class of problems that we call \textbf{qualia optimization}. We propose initial methods and establish basic properties for various qualia optimization formulations, but emphasize that these methods and properties are preliminary explorations rather than conclusive solutions or analyses. We conclude with discussion of future research avenues and emerging open questions.

\section{Philosophy of Mind Background}
\label{sec:background:PoM}

The philosophy of mind is a discipline within philosophy that examines the nature of the mind, consciousness, and mental phenomena. It intersects with other disciplines such as psychology, neuroscience, and cognitive science, contributing to our understanding of mental processes and their relation to the physical world. In this section we review key terms and theories from this discipline. To maintain focus on qualia optimization, we defer background on additional philosophy of mind concepts to Appendix \ref{app:trivialityArguments}, referencing it as needed when those concepts arise. 
\begin{itemize}
    \item \textbf{Mental State}: A mental state is a condition or process of the mind characterized by thoughts, feelings, beliefs, desires, and intentions. For example, believing it will rain, feeling pain, and desiring a cup of coffee are all different mental states. Mental states are quite diverse, encompassing both conscious aspects like the awareness of being thirsty and unconscious aspects like underlying biases. They also include more complex, sustained states like a mindset or psychological condition (e.g., a depressive state).
    \item \textbf{The Mind-Body Problem}: This problem, one of many studied in philosophy of mind, asks how the mind (including mental states) relates to the physical body and brain. It explores whether mental phenomena are distinct entities separate from physical processes or if they can be explained as part of these physical processes. 
    \item \textbf{Phenomenal Consciousness}: This term refers to the aspects of consciousness that involve the subjective, qualitative experience of what it is like to have a mental state. It encompasses the experiential, first-person perspective of thoughts and sensations, such as the vividness of a color or the intensity of a pain.\footnote{The scope of philosophy of mind, and even the mind-body problem, extends beyond phenomenal consciousness. An illustrative case is \emph{blindsight} \citep{weiskrantz1990blindsight}, where individuals can respond to visual stimuli without conscious perception, suggesting that some complex mental processes might occur without the corresponding phenomenal consciousness. \citet{haas2022reinforcement} provides a survey of how RL has been explored in philosophy of mind---focusing on perception and motivation without appealing to phenomenal consciousness.}
    \item \textbf{Qualia}: Qualia are the individual instances of subjective, conscious experience. They are what it ``feels like'' to have certain mental states, such as the specific sensation of seeing red, feeling pain, or tasting sweetness. Qualia are elements of phenomenal consciousness, with each qualia representing a distinct aspect of our subjective experience.    
    \item \textbf{The Hard Problem of Consciousness}: This problem focuses on the challenge of explaining why and how qualia arise from physical processes in the brain. It essentially asks: ``What causes phenomenal consciousness?''
\end{itemize}

\noindent Philosophers of mind have presented a wide range of theories concerning the nature of the mind. In Sections \ref{sec:dualism}--\ref{sec:OtherTheories} we review prominent theories of mind with an emphasis on their compatibility with AI agents having qualia (having phenomenal consciousness).

\subsection{Cartesian Dualism}
\label{sec:dualism}

Dualism is the theory that the mind and body are in some sense distinct, a concept traced back to ancient philosophers like Plato, who wrote of the soul's separation from the body in \emph{Phaedo} \citep{plato_phaedo}. In the 17th century, Ren\'e Descartes popularized a specific version---later called Cartesian dualism---that treats the mind and body as separate substances, not merely different aspects of one substance. He argued that the mind is not simply a byproduct of physical processes in the brain; rather, it possesses its own separate existence and characteristics.

Descartes argued for dualism via the famous quote ``\emph{Cogito ergo sum},'' or ``I think, therefore I am,'' which suggests that the existence of the mind is undeniable, in contrast to the potentially doubtful existence of the physical body. This line of reasoning led Descartes to conclude that the mind is fundamentally distinct from the physical body \citep{descartes1644principia}. This distinction forms the core of Cartesian dualism. However, this concept faced challenges, such as those posed by Princess Elizabeth of Bohemia, who questioned how a non-physical mind could interact with a physical body \citep{ElizabethDescartes}. This interaction problem highlights a fundamental challenge for dualists: the difficulty of explaining how immaterial mental states can cause physical actions in the body, and vice versa.\footnote{In philosophical terms, `immaterial' refers to entities or concepts that lack physical properties and cannot be observed or measured through physical means. This is in contrast to `material' entities, which encompass all phenomena subject to physical laws and measurable characteristics, from tangible objects to forces like electromagnetism.}

Although dualists may not rule out AI systems having qualia, they are likely to argue that AI systems lack the non-physical properties or substances they believe are essential for a mind. We present dualism here due to its significant and influential historical role in philosophy of mind, and to provide contrast for the theories of mind discussed subsequently, which are more compatible with the idea of AI systems having phenomenal consciousness.

\subsection{Functionalism}

Functionalism is a theory of mind that views mental states in terms of their function or role, rather than the substances they are made of or their internal composition. According to functionalism, what matters for a mental state like belief, pain, or desire is not its material composition---whether it is a brain state, a computer circuit, or something else---but the role it plays within the system it is part of. This role is typically understood in terms of causal relationships: how mental states interact with other mental states, sensory inputs, and behavioral outputs.

Functionalists argue that mental states are like components of a machine, each with a specific function contributing to the overall operation of the system. This perspective is often compared to a software approach to the mind, where the physical substrate (the hardware) is less important than the patterns of activity (the software) that define mental states. The theory posits that mental states are \emph{multiply realizable}, meaning that different physical systems can realize the same mental states if they perform the same functions or have the same causal relationships. For example, in principle, both a human brain and a computer could be in the ``state of pain'' if they process information related to pain in a functionally similar way. By downplaying the importance of the biological substrate of the human brain, functionalism has been influential in shaping the discourse around AI, offering a way of understanding the mind that is compatible with the idea of AI systems having mental states.

Functionalism focuses on defining mental states by their causal roles within a system, without asserting whether these states are accompanied by qualia (subjective experiences). While it is compatible with the idea that AI agents can have qualia, it does not explain why qualia arise from mental states or whether AI systems would necessarily experience them. To address qualia, additional assumptions are needed, such as assuming that qualia supervene on mental states.

Given this assumption, functionalism offers a framework for understanding the qualia of AI agents. That is, if specific algorithmic processes in AI agents perform the same functions as biological processes underlying human experiences, these processes might produce similar experiences in both humans and AI agents. For instance, if the role of temporal difference error in RL agents is functionally analogous to that of the neurotransmitter dopamine in human brains, then---assuming that RL agents have qualia and that similar mental states lead to similar qualia---functionalism suggests that RL agents experiencing positive temporal difference errors could have qualia similar to the human experiences associated with increased dopamine levels. 

\subsection{Other Theories of Mind}
\label{sec:OtherTheories}
Although many other theories of mind, such as emergent materialism and panpsychism, are compatible with AI systems having phenomenal consciousness, they do not offer as clear mechanisms for reasoning about specific qualia. Emergent materialism posits that consciousness arises from the complexity of physical processes, while panpsychism considers consciousness a fundamental property of matter. Neither provides a framework for understanding how specific experiences, like pleasure or pain, could emerge or manifest in AI systems. We therefore adopt a more functionalist perspective as an initial approach to reasoning about agent qualia, focusing on the functional roles of mental states and their possible connections to qualia.

Many other theories in the philosophy of mind are related to this work. For instance, physicalism asserts that all mental phenomena are ultimately rooted in physical processes, while computational theories of mind conceptualize mental processes as forms of information processing. Whether such theories are mutually exclusive depends on their specific formulations---some versions of physicalism, computationalism, and functionalism may conflict, while others can be integrated. Rather than neatly partitioning the space of ideas, these theories often form overlapping perspectives, each highlighting different aspects of cognition and the mind's relation to physical systems.
 
In this report, we draw inspiration from several theories, particularly functionalism, physicalism, and computational theories of mind. However, we do not claim that our assumptions map neatly onto any one established theory. We provide general pointers to relevant theories from philosophy of mind, but do not attempt to rigorously align our assumptions with them, as doing so would require a level of philosophical expertise that we lack. 

In the following sections, we review RL and temporal difference error, present the mathematical model of the agent-environment system we use, and then review neuroscience research suggesting that temporal difference error in RL agents plays a similar role to dopamine in parts of the human brain.

\section{RL Background and AEP Setting}
\label{sec:RLBackgroundAndSetting}

\begin{figure}[thbp]
    \centering
    \includegraphics[width=0.9\columnwidth]{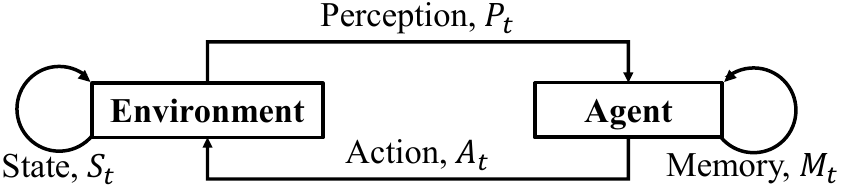}
    \caption{Diagram of the agent-environment system.}
    \label{fig:AEDiagram}
\end{figure}

We assume that the reader is familiar with RL \citep{SuttonBarto}. In RL literature, the environment is often modeled as an MDP, POMDP, or a variant thereof. We present the standard notation that we use for MDPs in Appendix \ref{app:MDP}.  Although we primarily consider RL algorithms designed for these standard settings, we adopt a different mathematical formalization of the agent-environment system that emphasizes fully characterizing the agent and which refines the way that the agent and environment interact.\footnote{We consider agents of all types, including (digital) RL agents, humans, and other animals. We will be explicit when making references to specific types of agents, like AI agents or RL agents. References in this text to an ``agent'' otherwise apply to all agents.} This refinement simplifies discussion of qualia optimization, although it makes the expression of standard RL algorithms more cumbersome. Our mathematical formulation of the agent-environment system is described in Section \ref{sec:AEPSub} and depicted in Figure \ref{fig:AEDiagram}.

\subsection{Agent-Environment Process (AEP/AERP)}
\label{sec:AEPSub}

We begin by characterizing the \emph{environment}---the universe within which the agent resides. For this initial exploration of qualia optimization, we assume that time is discrete and index time by $\definesymbol{t}{t} \in \{0,1,\dotsc\}$.\footnote{We provide an overview of both our mathematical notation and the symbols defined in this report in Appendix \ref{app:notation}.} Let $(S_0,S_1,\dotsc)$ be a sequence of random variables such that \definesymbol{$S_t$}{St} is a complete characterization of everything about the environment up to and including time $t$ that influences the environment at times $t' > t$ or the agent at times $t'' \geq t$. We refer to $S_t$ as the \emph{state of the environment at time $t$}, or as the \emph{state} if the time is clear from context. 

Next we characterize the \emph{agent} that resides within the environment. Let $(M_0,M_1,\dotsc)$ be a sequence of random variables such that \definesymbol{$M_t$}{Mt} is a complete characterization of everything about the agent up to and including time $t$ that influences the agent or environment at times $t' > t$. We refer to $M_t$ as the \emph{memory of the agent at time $t$}, or as the \emph{memory} if the time is clear from context. It may be more precise to call $M_t$ the \emph{state of the agent at time $t$}, since a complete characterization of the agent at time $t$ might include more than what one typically thinks of as memory (e.g., value function weights, policy parameters, and eligibility traces). However, we refer to $M_t$ as memory to allow for more concise and distinct discussion of $S_t$ and $M_t$ by referencing states and memories rather than environment states and agent states. 

Next we model how the agent and environment interact by explicitly modeling how the environment influences the agent and how the agent influences the environment.\footnote{To establish a solid foundation, we provide a detailed discussion of this formulation, the assumptions we make, and their implications. Later sections introduce similar formulations and assumptions with less elaboration, so a careful understanding of this initial discussion will be beneficial for following the subsequent material.} 
\begin{itemize}
    \item \textbf{Environment $\to$ Perceptions $\to$ Agent:} In classical MDP and POMDP formulations the environment influences the agent through two channels: states or observations and rewards. We combine these two channels into one, which we call \emph{perceptions}, because the inclusion of a reward signal is optional and because it simplifies later discussion regarding transformations of the agent's perceptions. For all $t \in \{0,1,\dotsc\}$, let \definesymbol{$P_t$}{Pt} be a complete characterization of everything about $S_t$ that influences the agent at time $t$, which we call the \emph{perception at time $t$}. 
    \item \textbf{Agent $\to$ Actions $\to$ Environment:} Similar to how perceptions characterize the channel through which the environment influences the agent, \emph{actions} characterize the channel through which the agent influences the environment. For all $t \in \{0,1,\dotsc\}$, let \definesymbol{$A_t$}{At} be a complete characterization of everything about $M_t$ that influences the environment at time $t+1$, which we call the \emph{action at time $t$}.\footnote{The phrase ``influences the environment at time $t+1$'' is ambiguous here. Arguably $A_t$ influences the environment ``at time $t$,'' and this influence is reflected in the distribution of $S_{t+1}$. We clarify this point later when we introduce Markovian state and memory assumptions.} One may think of the action $A_t$ as representing the decision made by the agent at time $t$. Although we expect the ideas presented in this report to extend to arbitrary (discrete, continuous, or hybrid) random variables, for simplicity we assume that $A_t$ is a discrete random variable, allowing us to discuss action probabilities. 
\end{itemize}

Algorithm \ref{alg:AEpseudocode} provides pseudocode for generating the sequence of random variables $(S_t,P_t,M_t,A_t)_{t=0}^\infty$, using the following functions and conditional distributions. 
\begin{itemize}
    \item The \emph{perception function} \definesymbol{$f_p$}{fp} characterizes the perception at time $t$ as a function of the state at time $t$. That is, $P_t = f_p(S_t)$. The implicit assumption that the perception at time $t$ can be written as a function of the state at time $t$ is not particularly restrictive. To model settings where $P_t$ includes noise or otherwise depends on random quantities, the noise or random quantities should be encoded within the state $S_t$. More generally, the state $S_t$ can explicitly encode $P_t$ so that $f_p$ merely masks the other components of the state. 
    \item The \emph{next-memory distribution} \definesymbol{$d_m$}{dm} characterizes how the agent's memory changes due to its perceptions---how the agent learns. That is, it characterizes the conditional distribution of the memory at time $t$ given the memory at time $t-1$ and the perception at time $t$ via the expression $M_t \sim d_m(M_{t-1}, P_t)$. We define $M_{-1}$ to be \texttt{null} so that the expression $M_t \sim d_m(M_{t-1}, P_t)$ applies to the generation of $M_0$ as well. An equivalent formulation would include a different next-memory distribution for generating $M_0$ that only conditions on the perception. 
    \item The \emph{action function} \definesymbol{$f_a$}{fa} characterizes the action at time $t$ as a function of the memory at time $t$. That is, $A_t = f_a(M_t)$. As with perceptions, the implicit assumption that the action can be written as a function of the memory is not particularly restrictive, since any necessary random quantities should be included in $M_t$. 
    \item The \emph{next-state distribution} \definesymbol{$d_s$}{ds} characterizes the conditional distribution of the state at time $t$ given the state at time $t-1$ and the action at time $t-1$. That is, $S_t \sim d_s(S_{t-1}, A_{t-1})$. We define $S_{-1}$ and $A_{-1}$ to both be \texttt{null} so that the expression $S_t \sim d_s(S_{t-1}, A_{t-1})$ applies to the generation of $S_0$ as well. An equivalent formulation would include a different next-state distribution for generating $S_0$ that does not condition on any random variables. 
\end{itemize}

\begin{algorithm}[thbp]
\DontPrintSemicolon
Initialize $S_{-1}$, $A_{-1}$, and $M_{-1}$ to \texttt{null}\;
\For{$t \gets 0$ \KwTo $\infty$}
{
    $S_t \sim d_s(S_{t-1}, A_{t-1})$\;
    $P_t = f_p(S_t)$\;
    $M_t \sim d_m(M_{t-1}, P_t)$\;
    $A_t = f_a(M_t)$\;
}
\caption{Agent-Environment Process}
\label{alg:AEpseudocode}
\end{algorithm}

Similar to standard RL settings, we assume that the next-state distribution $d_s$ and perception function $f_p$ exist, but we do not necessarily assume that they are known to the agent. When they are not, the agent must learn about the environment by interacting with it, i.e., it must learn about the environment from the sequence of perceptions and actions $(P_0,A_0,P_1,A_1,\dotsc)$.

\subsubsection{Markov and Stationarity Assumptions}
\label{sec:MarkovandstationarityAssumptions}

Our choice of notation \emph{suggests} conditional independence assumptions. For example, writing $d_s(S_{t-1},A_{t-1})$ to denote the conditional distribution of $S_t$ given $S_{t-1}$ and $A_{t-1}$ suggests that this distribution does not vary with other past random variables like $S_{t-2}$. Next, we make these suggested conditional independence assumptions explicit.

In order to concisely introduce these conditional independence assumptions, we define an ordering of the random variables---the order that they are generated in Algorithm \ref{alg:AEpseudocode}: $(S_0, P_0, M_0, A_0, S_1, P_1, M_1, A_1, S_2, \dotsc)$. Using this ordering of random variables, we make the following independence assumptions for all times $t$:
\begin{itemize}
    \item \emph{Markovian states.} $S_t$ is conditionally independent of all previous random variables given $S_{t-1}$ and $A_{t-1}$. 
    \item \emph{Markovian memories.} $M_t$ is conditionally independent of all previous random variables given $M_{t-1}$ and $P_t$. 
\end{itemize}

These two assumptions characterize how states, perceptions, memories, and actions should be defined for a given agent-environment system. Although they are stated as assumptions, they do not restrict the set of agent-environment systems under consideration. We provide supporting evidence for these claims in Appendix \ref{app:AssumptionImplications}.

In addition to the Markov assumptions, we assume that the next-state and next-memory distributions are both stationary. That is:
\begin{itemize}
    \item \emph{Stationary next-state distribution}. The conditional distribution of $S_t$ given $S_{t-1}$ and $A_{t-1}$ is the same for all times $t$. That is, $d_s(s,a)$ is the conditional distribution of $S_t$ given that $S_{t-1}=s$ and $A_{t-1}=a$ for all states $s$, actions $a$, and times $t$. 
    \item \emph{Stationary next-memory distribution}. The conditional distribution of $M_t$ given $M_{t-1}$ and $P_t$ is the same for all times $t$. That is, $d_m(m,p)$ is the conditional distribution of $M_t$ given that $M_{t-1}=m$ and $P_t=p$ for all memories $m$, perceptions $p$, and times $t$.
\end{itemize}
As with the Markov assumptions, the stationarity assumptions do not limit the set of agent-environment systems that can be modeled, but rather further inform how states and memories should be defined for a given system. To model systems where these distributions change over time, time should be encoded within states or memories, or within both.

\subsubsection{AEPs and AERPs}

\begin{figure}[thbp]
    \centering
    \includegraphics[width=\columnwidth]{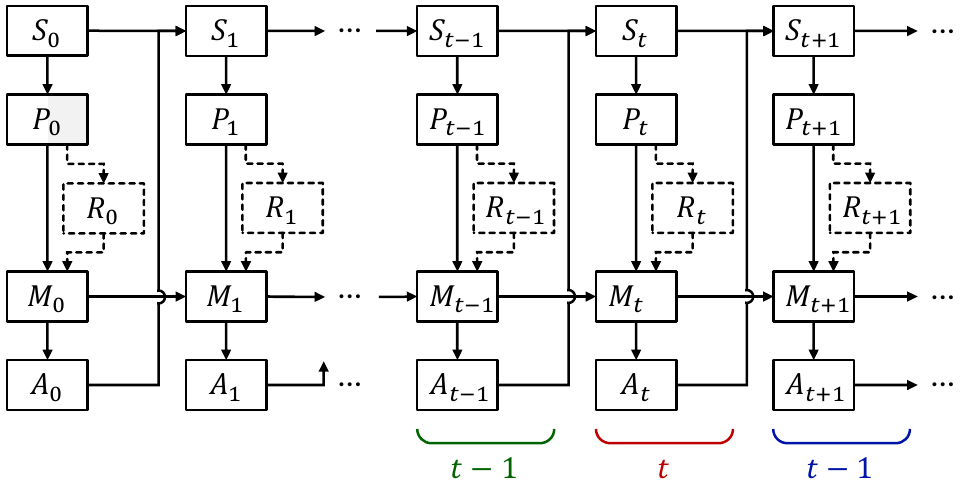}
    \caption{Bayesian network representation of an AEP (without the dashed lines and boxes) and AERP (with the dashed lines and boxes).}
    \label{fig:AEP_graphical_model}
\end{figure}

We call this stochastic process, including the Markov and stationarity assumptions, an \emph{agent-environment process} (\definesymbol{AEP}{aep}). As described previously, AEPs can include both rewards and other observations within the agent's perceptions. In order to differentiate between the general class of AEPs and the restricted class of AEPs that include rewards, we call an AEP with rewards an \emph{agent-environment reward process} (\definesymbol{AERP}{aerp}). In the AERP formulation, for each $t \in \{0,1,\dotsc\}$ let \definesymbol{$R_t$}{Rt} be a real-valued random variable that we call the \emph{reward at time $t$}, or the \emph{reward} when the time is clear from context. In an AERP there exists a function \definesymbol{$f_r$}{fr}, called the \emph{reward function}, that characterizes the reward at time $t$ as a function of the perception at time $t$. That is, $R_t = f_r(P_t)$.\footnote{RL formulations often vary in how they index rewards, sometimes starting with $R_0$ and other times with $R_1$, and sometimes defining $R_t$ to be the reward when the environment enters $S_t$ and sometimes defining $R_t$ to be the reward when the environment enters $S_{t+1}$. Notice that in this formulation $R_t$ is the reward when the agent enters $S_t$---it is the reward the agent receives when the environment transitions from $S_{t-1}$ to $S_t$ as a result of action $A_{t-1}$. Also, in this formulation the rewards begin with $R_0$ even though $R_0$ is not influenced by any agent actions. Typical RL formulations omit this initial reward.} Figure \ref{fig:AEP_graphical_model} depicts AEP and AERP systems as a Bayesian network that facilitates visualization of the Markovian state and memory assumptions. 

\subsubsection{Random Variables and the Physical World}

A critical point in later discussions will be the relationship between random variables in processes like AEPs and the physical world they represent. For example, how might the physical properties that we interpret as a bit sequence correspond to real numbers like a reward $R_t$? As another example, consider the random variable $S_t$, which we defined to be a \emph{complete characterization of everything about the environment up to and including time $t$ that influences the environment at times $t' > t$ or the agent at times $t'' \geq t$}. In order for this definition to be complete, we must clarify what it means for $S_t$ to be a \emph{characterization} and how it relates to the physical properties of the environment. 

For this initial discussion, let $\Phi$ be a random variable that represents the actual physical properties of the environment that influence the environment or agent at later times, which might include physical configurations, neural activity, voltage levels, or even sequences of bits. When we say that $S_t$ is a complete characterization of everything about the environment that influences the environment or agent at later times, we mean that the value of $S_t$ fully determines $\Phi$. Put differently, $\Phi$ can be written as a function of $S_t$. Similarly, we implicitly mean that $S_t$ does not include other information, and so $\Phi$ fully determines the value of $S_t$---$S_t$ can be written as a function of $\Phi$. Together, these properties imply that $\Phi$ and $S_t$ are isomorphic. 

We call functions (isomorphisms) that relate physical properties to random variables like $S_t$ \emph{representation functions}. For example, if $\rho$ is the representation function for $S_t$, then $\rho$ is an isomorphism between $S_t$ and $\Phi$, i.e., $S_t = \rho(\Phi)$ and $\Phi = \rho^{-1}(S_t)$. To ground this discussion, consider an idealized physical world where $\Phi$ corresponds to a sequence of 32 bits.\footnote{As a thought experiment, one could imagine that in this idealized physical world the \emph{only} physical property is the sequence of 32 bits. Alternatively, one could consider a more realistic setting where the physical world contains other physical properties, but the physical properties corresponding to the 32 bits are the only properties that correspond to $S_t$. Although we adopt a simple 32-bit example here, we make no assumptions about the actual complexity of $S_t$ or $\Phi$, which may correspond to a range of physical properties.} We might define $S_t$ to be a real number, in which case the representation function $\rho$ indicates how the sequence of bits should be interpreted as determining the value of $S_t$. There are many ways that sequences of 32 bits can be interpreted as different real numbers, and each corresponds to a different representation function $\rho$. 

Notice that each random variable has an implicit representation function that describes how properties of the physical world relate to values of the random variable, and that these representation functions can differ (e.g., different random variables can use different mappings from bit sequences to real numbers). Since we will discuss the representation functions and underlying physical properties of many random variables, we introduce the following notation: For any random variable $Z$, let \definesymbol{$\Phi_Z$}{PhiZ} and \definesymbol{$\rho_Z$}{rhoZ} denote the underlying physical properties and representation function of $Z$. For example, $\Phi_{S_t}$ denotes the physical properties corresponding to the state $S_t$, and $S_t=\rho_{S_t}(\Phi_{S_t})$. Recall that the existence of these representation functions (and their required invertibility) implies that random variables and their corresponding underlying physical properties are isomorphic. Hence, the Markov properties, stationarity properties, and dynamics that apply to $S_t, P_t, M_t,$ and $A_t$ also apply to the underlying physical properties $\Phi_{S_t}, \Phi_{P_t}, \Phi_{M_t},$ and $\Phi_{A_t}$. 

One possible AEP or AERP model of a physical system uses the identity functions for all representation functions, resulting in a system that directly models the dynamics of the (perhaps non-numerical) physical properties. The use of representation functions other than identity functions facilitates the later definition of objective functions that naturally operate on numbers (e.g., real-valued rewards) rather than the underlying and possibly non-numeric physical properties. 

There are several other possible points of confusion regarding underlying physical properties and representation functions that we aim to avoid. First, notice that we did not make any assumptions regarding whether the physical properties underlying different random variables are disjoint. That is, the same physical properties could correspond to different parts of the AEP formulation (e.g., the state and the perception). In one interesting extreme case, the agent may be part of the environment, and so all of the physical properties of the agent may also be physical properties of the environment. Second, random variables with different time-subscripts are different random variables. So, for example, $R_t$ and $R_{t'}$ can have different representation functions, $\rho_{R_t}$ and $\rho_{R_{t'}}$, and underlying physical properties, $\Phi_{R_t}$ and $\Phi_{R_{t'}}$, when $t \neq t'$. Third, the representation functions are deterministic functions---they do not depend on the values of any random variables in the system. For example, $\rho_{S_t}$ does not depend on $S_{t-1}$ or $A_{t-1}$. 

The requirement that representation functions are deterministic is critical---without it or a similar assumption, there need not be any meaningful relationship between the underlying physical properties and the random variables of the AEP or AERP. For example, \emph{every} AEP and AERP could be viewed as a characterization of a physical system with a single constant physical property, $\Phi = 0$. To see why, consider any AEP with states $S_t$, perceptions $P_t$, memories $M_t$, and actions $A_t$. This AEP could correspond to a physical system with a single constant physical property. That is, $\Phi_{S_t} = \Phi_{P_t} = \Phi_{M_t} = \Phi_{A_t} = 0$. This is achieved by defining $\rho_{S_t}$ to itself be a random quantity such that $\rho_{S_t}(0) \sim d_s(S_{t-1},A_{t-1})$. Notice that once $\rho_{S_t}$ has been sampled, it is a deterministic and invertible function over its domain, $\{0\}$, and hence is a valid representation function. Similar stochastic definitions of $\rho_{P_t}, \rho_{M_t},$ and $\rho_{A_t}$ would allow for the construction of the AEP (any AEP) from constant physical properties. To prevent this disconnect between the random variables of an AEP and the underlying physical properties they correspond to, we require the representation functions to be deterministic.

Lastly, notice that the approach for relating abstract random variables in an AEP to an underlying physical process that we have adopted here is closely related to Putnam's ``mapping'' theory of the implementation of computations in philosophy of mind \citep{Putnam1988}. Furthermore, our argument that allowing the representation functions to vary with the values of random variables in the AEP would allow trivial physical systems (ones with a single constant property) to correspond to arbitrarily complex AEPs is closely related to \emph{triviality arguments} in philosophy of mind, which suggest that Putnam's mapping theory would allow trivial physical systems (e.g., a bucket of water, rock, or clock) to be said to implement complex computations like those of the human brain. We provide a brief overview of these triviality arguments in Appendix \ref{app:trivialityArguments}. 

\subsubsection{Policies} 
\label{sec:policies}

A \emph{policy} is a characterization of one way that an agent could make decisions based on the current state or perception (but independent of time and other previous random variables). We distinguish between two types of policies: state-policies and perception-policies. A \emph{state-policy} characterizes the distribution of actions given the current state, while a \emph{perception-policy} characterizes the distribution of actions given the current perception. In general (when there may be partial observability), a state policy can be defined and analyzed theoretically but cannot be directly implemented by an agent that never observes the actual state $S_t$. In contrast, a perception-policy depends solely on perceptions $P_t$, making it implementable by the agent. Hereafter, we primarily discuss perception-policies.

More formally, for all perceptions $p$, actions $a$, and times $t$, a perception-policy \definesymbol{$\pi$}{pi} characterizes the distribution of actions given the current perception according to the expression 
\begin{equation}
    \pi(p,a) = \Pr(A_t=a|P_t=p).
\end{equation}
In our subsequent analysis, we often restrict our attention to AEPs in which the environment is an MDP. In these cases, the perception is $P_t = (S_t,R_t)$, and when $\pi(p,a)$ does not depend on the reward component of the perception, state-policies and perception-policies coincide because the agent fully observes the state. In such cases we sometimes write expressions like $\pi(S_t, A_t)$ instead of $\pi(P_t,A_t)$ when no ambiguity arises.

Notice that the agent does not necessarily implement a fixed state-policy or perception-policy, since it can change its memory at each time step, and its memory influences the policy that it implements. That is, the agent \emph{learns} (changes its memory), and this learning changes the agent's policy at each time $t$. Also, although we adopt the term ``policy'' from RL literature, the AEP formulation could apply to supervised learning systems in various ways. For example, each time step could correspond to a supervised learning model making a single prediction or batch of predictions during training or evaluation. As another example, one time step could correspond to the execution of a single (possibly stochastic) unit in an artificial neural network. Although the mapping from perceptions to actions in these cases would not typically be called a policy, we adopt this RL term due to our focus on RL agents in this initial work.

\subsubsection{Episodes}
\label{sec:episodes}

The AEP formulation of the agent interacting with the environment is inherently \emph{continuing}---the agent and environment have a single long sequence of interactions. Modern RL research often assumes an \emph{episodic} setting, wherein an agent interacts with an environment over a sequence of episodes that index time so that it starts at $t=0$ for each episode. This episodic setting can be viewed as a restricted class of AEPs (or AERPs). There are many possible sets of restrictions that model episodic settings within AEPs---we propose the following restrictions that essentially chain together episodes into one long string of interactions:\footnote{This representation of the episodic setting within a continuing setting was inspired by the unification of episodic and continuing settings for MDPs \citep{white2017unifying}.} 
\begin{itemize}
    \item There exists a special state \definesymbol{$s_\infty$}{sinfty}, perception \definesymbol{$p_\infty$}{pinfty}, and action \definesymbol{$a_\infty$}{ainfty}. 
    \item Whenever the state becomes $s_\infty$, we say that the current episode ends. 
    \item We define $f_p(s_\infty)=p_\infty$ so that the agent is notified that the current episode is ending. 
    \item The agent is defined to always select action $a_\infty$ when presented with perception $p_\infty$ to capture that there is not another decision for the agent to make within the episode that is ending. 
    \item When $S_t=s_\infty$, the next-state $S_{t+1}$ is sampled from the distribution $d_s(\texttt{null},\texttt{null})$---the same distribution as the initial state. To avoid edge-cases, we assume that the first state of an episode cannot be $s_\infty$. This implies that $s_\infty$ is not in the support of $d_s(\texttt{null}, \texttt{null})$. 
\end{itemize} 

We use $i \in \{0,1,\dotsc\}$ to index the episodes. Let \definesymbol{$\operatorname{start}(i)$}{start} and \definesymbol{$\operatorname{end}(i)$}{end} denote the times that the $i^\text{th}$ episode starts and ends, respectively. More formally, $\operatorname{start}(0)\triangleq0$, 
\begin{equation}
    \operatorname{end}(i) \triangleq \min\{t > \operatorname{start}(i) : S_t = s_\infty\},
\end{equation}
and for all $i\geq 0$,
\begin{equation}
    \operatorname{start}(i+1)\triangleq\operatorname{end}(i)+1. 
\end{equation}
Similarly, let \definesymbol{$\operatorname{len}(i)$}{len} denote the length of the $i^\text{th}$ episode (the number of states not equal to $s_\infty$):
\begin{equation}
    \operatorname{len}(i)\triangleq\operatorname{end}(i)-\operatorname{start}(i).
\end{equation}

We abuse notation and overload the $\operatorname{start}$ and $\operatorname{end}$ operators to alternatively take a time step $t$ as input, and to output the start-time and end-time of the episode that is in-progress at time $t$:\definesymbol{}{startt}\definesymbol{}{endt}
\begin{align}
    \nonumber \operatorname{start}(t)\triangleq& 
    \begin{cases}
        0 &\mbox{if }t \leq \min\{t' : S_{t'}=s_\infty\}\\
        \operatorname{max}\{t' \leq t : S_{t'-1}=s_\infty\} &\mbox{otherwise}
    \end{cases}\\
    \operatorname{end}(t) \triangleq& \min\{t' \geq t : S_{t'}=s_\infty\}.
\end{align}
The intended variant of the $\operatorname{end}$ operator should be clear from context---whether its argument indexes an episode or a time step. 

Notice that our formulation of rewards and episodes implies that for each episode $i$, $R_{\operatorname{start}(i)}$ is a reward that occurs prior to any actions within the $i^\text{th}$ episode. This reward is omitted in typical RL formulations since it is not influenced by the agent, and so it is not pertinent to the task of optimizing the agent's behavior. We opt to include it to avoid defining a special case wherein the reward is not defined when the previous state is $s_\infty$. 

\begin{figure}[htbp]
    \centering
    \includegraphics[width=0.85\columnwidth]{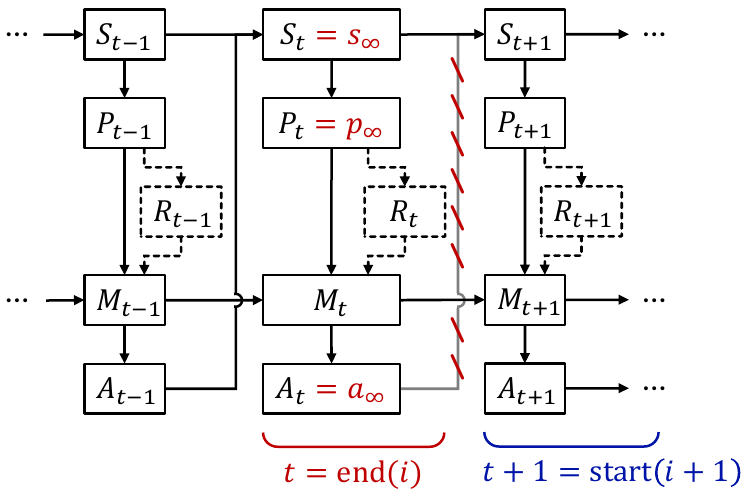}
    \caption{Bayesian network depiction of the end of episode $i$ and beginning of episode $i+1$.}
    \label{fig:AEP_episode_graphical_model}
\end{figure}

Figure \ref{fig:AEP_episode_graphical_model} depicts the end of an episode $i$ and the beginning of episode $i+1$ using the Bayesian network representation of an AEP or AERP from Figure \ref{fig:AEP_graphical_model}. 
The gray line with red hashes indicates an edge of the Bayesian network for typical time steps that can be removed when the time step corresponds to the end of an episode. This edge can be removed because $S_{t+1}\sim d_s(\texttt{null},\texttt{null})$ when $S_t=s_\infty$, and so $S_{t+1}$ is conditionally independent of $A_t$ given that $S_t=s_\infty$. Also, notice that $S_{\operatorname{end}(i)}=s_\infty$, $P_{\operatorname{end}(i)}=p_\infty$, and $A_{\operatorname{end}(i)}=a_\infty$, but that $R_{\operatorname{end}(i)}$ can still be an arbitrary real number\footnote{This formulation requires there to be a specific reward that always occurs when an episode terminates since $R_t$ is a deterministic function of $P_t$ and $P_t$ is always $p_\infty$ when an episode terminates. If one desires stochasticity within these rewards, there are many ways that this could be modeled. For example, an additional state could be inserted prior to the transition to $s_\infty$ so that the stochastic reward can be provided, or the AERP formulation could be modified so that $S_{\operatorname{end}(i)}$ is not necessarily deterministic (and equal to $s_\infty$), but rather in some way encodes that the episode is ending (and $P_{\operatorname{end}(i)}$ could be similarly redefined). These modifications would not alter our subsequent discussions in this initial study of qualia optimization.} and the agent can still update $M_{\operatorname{end}(i)-1}$ to $M_{\operatorname{end}(i)}$ as it performs a final update based on the observation that episode $i$ is ending. 

The episodic AEP formulation so far allows for cases where the first episode can be infinitely long. If that were to happen, the second episode would never begin and so $\operatorname{start}(1)$ would be undefined, making other expressions later in this report also undefined. To circumvent this issue, we define \emph{finite-horizon AEPs} to be episodic AEPs for which there exists a finite constant $H$, called the \emph{horizon}, such that for all $i \geq 0$, $\operatorname{len}(i) \leq H$ surely. 

\subsection{Basic Actor-Critic (BAC)}
\label{sec:BAC}

In the setting that we have described, agent algorithms correspond to rules governing how the agent selects actions and updates its memory---$d_m$ and $f_a$. In this report we primarily focus on AI agents, which implement AI algorithms. When considering a specific AI algorithm in this initial exploration of qualia optimization, we focus on an RL algorithm that we call the \emph{basic actor-critic} (\definesymbol{BAC}{bac}).\footnote{BAC is the algorithm titled ``Actor–Critic with Eligibility Traces (episodic)'' in the work of \citet[page 332]{sutton2018reinforcement}, modified to only include eligibility traces for the critic.} We describe BAC in detail to show precisely how RL algorithms can be formulated in the setting that we have chosen and to introduce temporal difference error---a central concept in later sections. After \eqref{eq:discountedReturn} and \eqref{eq:discountedReturnTimeT}, which introduce a notation for returns, readers who see how RL algorithms like BAC fit within the AEP/AERP settings and who are familiar with actor-critic algorithms and temporal difference error can skip to Section \ref{sec:background_RPE} without missing important context.

\subsubsection{Design Goals}

BAC was designed to cause the agent to learn to select actions that maximize the expected discounted sum of rewards that it receives in an episodic setting. That is, it was designed to maximize:
\begin{equation}
    \label{eq:BAC_goal}
    \lim_{i\to\infty} \mathbf E\left [ \sum_{t=\operatorname{start}(i)+1}^{\operatorname{end}(i)} \!\!\!\!\!\!\!\gamma^{t-(\operatorname{start}(i)+1)} R_t \right ]\!,
\end{equation}
where $\gamma \in [0,1]$ is a hyperparameter called the \emph{reward discount parameter}. To shorten similar expressions, going forward we define the \emph{duration}  $\operatorname{dur}(t)$ to be roughly the number of time steps that, at time $t$, have occurred since the start of the current episode. More precisely, if time $t$ is part of episode $i$, then \definesymbol{$\operatorname{dur}(t)$}{durt}$=t-(\operatorname{start}(i)+1)$.

Before continuing to describe how BAC fits within the AERP formulation, we introduce notation for different \emph{returns}, which simplify expressions like \eqref{eq:BAC_goal}. We refer to \definesymbol{}{Gi}
\begin{equation}
    \label{eq:discountedReturn}
    G_i \triangleq\!\!\!\!\! \sum_{t=\operatorname{start}(i)+1}^{\operatorname{end}(i)} \!\!\!\!\!\!\!\gamma^{\operatorname{dur}(t)} R_t
\end{equation}
as the \emph{discounted return} and we refer to \definesymbol{}{Gt}
\begin{equation}
    \label{eq:discountedReturnTimeT}
    G_t \triangleq \sum_{k=t+1}^{\operatorname{end}(t)} \gamma^{k-(t+1)} R_k
\end{equation}
as the \emph{discounted return from time $t$}.\footnote{This is an abuse of notation, overloading $G_z$ to have two possible meanings. It should be clear from context whether $G_z$ refers to the discounted return of episode $z$ or discounted return from time $z$ based on whether $z$ indexes an episode or a time step.} Notice that $R_{\operatorname{start}(i)}$ is not included in $G_i$ or $G_{i-1}$---this is the aforementioned reward that is not included in typical RL formulations, since the actions selected by the agent do not influence this reward. Also notice that the first reward included in $G_t$ is $R_{t+1}$, which is the reward that results from the environment transitioning from $S_t$ to $S_{t+1}$ due to action $A_t$. Lastly, notice that if $S_t=s_\infty$ (or equivalently, if $P_t=p_\infty$) then $G_t=0$. That is, for all episodes $i$, $G_{\operatorname{end}(i)}=0$.

In addition to maximizing the expected discounted return asymptotically, BAC was designed to do so quickly. There are many ways that this second design goal can be formalized. One way, when the agent will interact with the environment for a finite number of episodes, \definesymbol{$i_\text{max}$}{imax}, is for the RL algorithm to maximize
\begin{equation}
\sum_{i=0}^{i_\text{max}-1} \mathbf E\left [ G_i \right ].
\end{equation}

Although BAC was designed for environments that can be modeled as MDPs, it is often applied to and effective for POMDPs and can be applied to the settings that we consider. Also, although BAC was designed to maximize the expected discounted return for MDPs, and to do so quickly, without additional assumptions it does not necessarily achieve these goals. However, it remains representative of many commonly used contemporary RL algorithms.

\subsubsection{Mathematical Specification}

BAC stores policy parameters \definesymbol{$\Theta_t$}{Thetat}, \emph{value function approximation} (\definesymbol{VFA}{VFA}) weights \definesymbol{$W_t$}{Wt}, eligibility traces \definesymbol{$E_t$}{Et}, perception $P_t$, and action $A_t$ within memory $M_t$. That is,
\begin{equation}
    M_t=(\Theta_t, W_t, E_t, P_t, A_t).
\end{equation} 
The three components  $\Theta_t, W_t,$ and $E_t$ are real vector-valued random variables. The agent stores one eligibility per weight, meaning that $W_t$ and $E_t$ are vectors of the same length. 

Before defining $f_a$ and $d_m$ for BAC, we introduce two implicit functions that the BAC updates rely on: a \emph{perception-policy parameterization} \definesymbol{$\pi_\text{BAC}$}{pibac} and a \emph{VFA parameterization} \definesymbol{$v$}{v}. The perception-policy parameterization $\pi_\text{BAC}$ specifies the conditional distribution of the action $A_t$ given the policy parameters $\Theta_t$ and the perception $P_t$ according to the equation:
\begin{equation}
    \pi_\text{BAC}(p,a,\theta)=\Pr(A_t=a|P_t=p,\Theta_t=\theta). 
\end{equation}
While the choice of $\pi_\text{BAC}$ is a hyperparameter (it will be referenced in our definitions of $f_a$ and $d_m$), BAC requires $\partial \pi_\text{BAC}(p,a,\theta) / \partial \theta$ to exist for all perceptions $p$, actions $a$, and policy parameters $\theta$. 

The VFA parameterization $v$ takes perceptions and VFA weights as input, and produces a real number as output. BAC was designed to adjust the VFA weight vector $W_t$ so that $v(p,W_t)$ is an approximation of the expected discounted return from time $t$, if the agent faced with perception $p$ were to continue selecting actions using perception-policy parameters $\Theta_t$. That is, BAC was designed to find VFA weights $W_t$ such that for all perceptions $p$,\footnote{The conditional expectation in \eqref{eq:valueFunction} provides the appropriate intuition, but is not technically precise, since often the event that $\forall t' \geq t, \Theta_{t'}=\Theta_t$ cannot occur, making the conditional expectation undefined. For brevity we use this imprecise statement, but recognize that a proper formulation would require a more precise mathematical formulation of the idea that the agent continues to select actions using perception-policy parameters $\Theta_t$.}
\begin{equation}
    \label{eq:valueFunction}
    v(p,W_t)\approx  \mathbf{E}\left [ G_t | P_t=p, (\forall t' \geq t, \Theta_{t'}=\Theta_t) \right ].
\end{equation}
While the choice of the function $v$ is a hyperparameter, BAC requires $\partial v(p,w) / \partial w$ to exist for all perceptions $p$ and VFA weights $w$. We also assume that $v(p_\infty, w)=0$ for all VFA weights $w$, since it is always the case that $G_t=0$ if $P_t=p_\infty$. 

Having defined $\pi_\text{BAC}$ and $v$, we can now define $f_a$ and $d_m$ for BAC. First, note that the action function is trivial because $A_t$ is explicitly encoded within $M_t$, and so $f_a(M_t)= A_t$. Because expressions for $d_m$ include special cases for initial updates, and to make the hyperparameters of BAC explicit, we express $d_m$ using pseudocode in Algorithm \ref{alg:BACdm}.\footnote{We assume that the reader is already familiar with this algorithm, and so our intent is to show how BAC can be represented in our formulation, not to explain this algorithm. We also emphasize that the goal of our formulation is not to make the expression of RL algorithms through $f_a$ and $d_m$ simple, but to make subsequent reasoning about agent-environment interactions simple. So, although our formulation may often result in relatively complicated expressions for $d_m$, the majority of our subsequent discussions can deal with $d_m$ as an abstract concept. Still, we provide Algorithm \ref{alg:BACdm} to provide a complete example of how an existing RL algorithm fits within our formulation.} Notice that the BAC algorithm has multiple additional hyperparameters: \definesymbol{$\theta_0$}{theta0} is the initial perception-policy parameter vector, \definesymbol{$w_0$}{w0} is the initial VFA weight vector, \definesymbol{$\lambda$}{lambda}$\,\in[0,1]$ is a hyperparameter called the \emph{eligibility trace decay rate}, and \definesymbol{$\alpha$}{alpha} and \definesymbol{$\beta$}{beta} are positive real-valued hyperparameters called the \textit{critic step size} and \textit{actor step size}, respectively. 

\begin{algorithm*}[thbp]
\DontPrintSemicolon
\SetKwInput{KwInput}{Input}
\SetKwInput{KwOutput}{Output}
\SetKwInput{KwHyper}{Hyperparameters}
\KwInput{Current perception $P_t$ and memory $M_{t-1}=(\Theta_{t-1},W_{t-1},E_{t-1},P_{t-1},A_{t-1})$.}
\KwOutput{Memory $M_t=(\Theta_t,W_t,E_t,P_t,A_t)$ where $M_t \sim d_m( M_{t-1},P_t)$}
\KwHyper{Perception-policy parameterization $\pi_\text{BAC}$, VFA parameterization $v$, initial perception-policy parameters $\theta_0$, initial VFA weights $w_0$, reward discount parameter $\gamma \in [0,1]$, eligibility trace decay rate $\lambda \in [0,1]$, and positive real-valued step sizes $\alpha$ and $\beta$.}
    \textcolor{Color-BAC-Alg-Special}{\tcc{Note that the input $P_t$ is also an output and so its value is not set below.}}
    $R_t \gets f_r(P_t)$\hspace{1.1cm}\textcolor{Color-BAC-Alg-Special}{\tcp{Extract the reward from the perception}}
    \If{$M_{t-1}=\texttt{null}$}
    {
        \textcolor{Color-BAC-Alg-Special}{\tcc{Initialization when $t=0$ (first time step of first episode)}}
        $\Theta_t \gets \theta_0$\;        
        $W_t \gets w_0$\;
        $E_t \gets 0$\;
    }
    \ElseIf{$P_{t-1}=p_\infty$}
    {
        \textcolor{Color-BAC-Alg-Special}{\tcc{Start of episode $i>0$}}
        $\Theta_t \gets \Theta_{t-1}$\hspace{0.65cm}\textcolor{Color-BAC-Alg-Special}{\tcp{Copy parameters from the end of the previous episode}}
        $W_t \gets W_{t-1}$\hspace{0.5cm}\textcolor{Color-BAC-Alg-Special}{\tcp{Copy weights from the end of the previous episode}}
        $E_t \gets 0$\hspace{1.21cm}\textcolor{Color-BAC-Alg-Special}{\tcp{Clear eligibility traces}}
    }   
    \Else
    {
        \textcolor{Color-BAC-Alg-Special}{\tcc{Standard update}}
        $\Delta_t \gets R_{t} + \gamma v(P_t,W_{t-1}) - v(P_{t-1},W_{t-1})$\;
        $E_t \gets \gamma \lambda E_{t-1} + \frac{\partial v(P_{t-1}, W_{t-1})}{\partial W_{t-1}}$\;
        $W_t \gets W_{t-1} + \alpha \Delta_{t} E_t$\;
        $\Theta_t \gets \Theta_{t-1} + \beta \Delta_{t} \frac{\partial \ln(\pi_\text{BAC}(P_{t-1}, A_{t-1}, \Theta_{t-1}))}{\partial \Theta_{t-1}}$\;\label{line:gradStep}
    }
    $A_t \sim \pi_\text{BAC}(P_t, \cdot, \Theta_t)$\hspace{0.5cm}\textcolor{Color-BAC-Alg-Special}{\tcp{In all cases \label{alg:lineasdf}$A_t$ is sampled the same way once $\Theta_t$ has been computed.}}
    \textbf{Return} $M_t=(\Theta_t, W_t, E_t, P_t, A_t)$\;
\caption{Next-Memory Distribution, $d_m$, for BAC}
\label{alg:BACdm}
\end{algorithm*}

\subsubsection{Temporal Difference Error (TD Error)}
\label{sec:TDError}

The standard update in Algorithm \ref{alg:BACdm} uses a temporary variable (one that is used to compute $M_t$ but not used thereafter or returned), 
\begin{equation}
    \definesymbol{\Delta_t}{deltat} = R_t + \gamma v(P_t,W_{t-1}) - v(P_{t-1},W_{t-1}),
\end{equation}
which is called the \emph{temporal difference error} (TD error) and which is central to some subsequent sections. When the TD error is positive, RL algorithms like BAC make recently chosen actions more likely---they reinforce the agent's recent behavior. When the TD error is negative, algorithms like BAC make recently chosen actions less likely---they inhibit the agent's recent behavior. 

To better understand this behavior and the TD error, consider the problem of predicting the amount of reward the agent will receive after the agent takes action $A_{t-1}$ when the environment is in state $S_{t-1}$, i.e., predicting $G_{t-1}$. One estimate of this quantity is $v(P_{t-1},W_{t-1})$---an estimate that the agent can construct based on its perception at time $t-1$. At time $t$, the agent can construct an improved estimate based on the new information it has obtained: the perception $P_t$ that includes the reward $R_t=f_r(P_t)$. This new information allows for the computation of a second estimate, $R_t + \gamma v(P_t,W_{t-1})$, that combines the observed reward $R_t$ with an estimate, $v(P_t,W_{t-1})$, of the expected discounted sum of rewards that the agent will receive thereafter.

Next, consider the difference between these two estimates: 
\begin{equation}
    \underbrace{R_t + \gamma v(P_t,W_{t-1})}_\text{estimate of $G_{t-1}$ from $t$} - \underbrace{v(P_{t-1},W_{t-1})}_\text{estimate from $t-1$}.
\end{equation} 
This quantity, which is the TD error $\Delta_t$, characterizes how the agent's prediction of $G_{t-1}$ changes from time $t-1$ to time $t$. Hence, it is sometimes called a \emph{reward prediction error} (RPE).

When this term is positive, it indicates that at time $t$ the agent has increased its prediction of $G_{t-1}$ relative to what it predicted at time $t-1$. That is, something turned out better than the agent expected (perhaps the reward was larger than expected, or perhaps the resulting perception indicated that it would receive more reward in the future than was previously expected, or perhaps both). Recall that the action $A_{t-1}$ caused the transition from time $t-1$ to time $t$, and likely influenced $P_t$ (which includes $R_t$). So, when the TD error is positive, it suggests that $A_{t-1}$ produced an outcome better than the agent expected, and so it should be selected more frequently in future similar circumstances. When this term is negative, it indicates that at time $t$ the agent has decreased its prediction of $G_{t-1}$ relative to what it predicted at time $t-1$. That is, something turned out \emph{worse} than the agent expected, and so the action at time $t-1$ should be taken less frequently in future similar circumstances. These intuitive policy update rules are encoded in the expression for $\Theta_t$ in the standard update of Algorithm \ref{alg:BACdm}  since 
\begin{equation}
    \frac{\partial \ln\big(\pi_\text{BAC}(P_{t-1}, A_{t-1}, \Theta_{t-1})\big)}{\partial \Theta_{t-1}}
\end{equation}
is a direction of change to the policy parameters that increases the probability of action $A_{t-1}$ given perception $P_{t-1}$. 

\section{Background on the RPE Hypothesis for Dopamine}
\label{sec:background_RPE}

Neuroscientific research has long recognized that dopaminergic neurons (neurons that produce and release the neurotransmitter dopamine) encode signals associated with rewards, reward-based learning, and decision-making. In seminal experiments, \citet{Schultz1997} observed that dopaminergic neurons in the primate midbrain responded not to the reward itself, but to the discrepancy between predicted and actual reward outcomes---a \emph{reward prediction error} (RPE). When the outcome was better than the primate expected, there was an increase in the firing of dopaminergic neurons, and when the outcome was worse than the primate expected, there was decreased activity of dopaminergic neurons. This behavior of dopaminergic neurons suggests that, in primates, dopamine may be a neural correlate of temporal difference error, $\Delta_t$. That is, evidence suggests that dopamine might play the role in primate brains that temporal difference error plays in many RL agents.

However, there remains significant debate about the precise relationship between dopamine, RPEs, and TD error. For example, some studies suggest that the relationship only holds for positive TD errors \citep{bayer2005midbrain,d2008bold}, while others suggest that different dopaminergic neurons may be responsible for encoding positive and negative TD errors \citep{matsumoto2009two}. Researchers have also questioned whether dopamine has a causal impact on behavior and learning akin to that of TD error, although there is mounting evidence that it does \citep{steinberg2013causal}. More recently, research has suggested that dopamine may correspond to variants of TD error from \emph{distributional RL}, wherein $v(P_t,W_t)$ estimates parameters of the distribution of the discounted return other than the expected value \citep{dabney2020distributional}. It also remains unclear how the relationship between dopamine and TD error extends across the animal kingdom, although the relationship has been observed in flies, albeit with the sign of the TD error reversed \citep{claridge2009writing}, and with similar debate and conclusions regarding whether dopamine encodes both positive and negative TD errors \citep{waddell2013reinforcement}.

\section{Qualia Optimization}
\label{sec:qualia_opt}

In this section we introduce the concept of qualia optimization in the context of AI. We begin by making the explicit assumption that AI agents---AI systems that are modeled as agents in an AEP formulation---have phenomenal consciousness (i.e., they have qualia). We emphasize that this assumption is made so that we can explore its potential implications, rather than to assert its factual accuracy.
\begin{ass}
    \label{ass:RL_PC}
    AI agents have phenomenal consciousness.
\end{ass}
For the purposes of this report, we do not differentiate between different types of AI agents that might or might not experience phenomenal consciousness (e.g., based on their underlying algorithms or complexity), and instead study the problem of qualia optimization for AI agents (and particularly RL agents) in general.

The assumption that AI agents have phenomenal consciousness naturally leads to a range of ethical considerations, particularly concerning the well-being of these agents as moral patients. However, within the scope of this report, we do not explore these ethical questions. Instead, we operate under a secondary assumption that there exists a motivation, whether practical or ethical, to enhance the quality of AI agent qualia.

\begin{ass}
There is a need for enhancing the quality of AI agent qualia.
\end{ass}

These assumptions introduce a new category of problem settings in AI, centered around the question: \emph{How can we optimize the experiential quality of AI agents while also considering their performance?} This category, which we call \textbf{qualia optimization for AI}, encompasses a wide range of problem formulations. For example, one might consider various trade-offs between the experiential quality of agents and their performance. Some formulations may even prioritize qualia optimization as the sole objective. The subsequent sections explore specific examples of such problem settings.

Different theories of mind provide different guidance regarding which properties of AI agents and environments might result in desirable or undesirable qualia for AI agents. For this initial exploration of qualia optimization for AI, we adopt the following assumption, inspired by functionalism.
\begin{ass}
\label{ass:function}
If specific algorithmic processes in AI agents functionally resemble biological processes underlying human experiences, then these corresponding processes yield similar experiences in both humans and AI agents.
\end{ass} 
Notice that this assumption informs how the quality of the qualia of AI agents might be quantified, not whether AI agents have qualia (which was already assumed via Assumption \ref{ass:RL_PC}). This means that Assumption \ref{ass:function} may be compatible with a wide range of theories of mind, including Cartesian dualism, provided that the quality (rather than the existence) of qualia depends on the functional roles of the underlying processes.

Before exploring specific qualia optimization problem settings we consider general concepts that apply to many settings. In Section \ref{sec:AEI} we generalize the standard agent-environment perspective to allow for the insertion of mechanisms between the agent and environment that transform the agent's experiences. In Section \ref{sec:FQO} we consider the possible forms of mathematical formulations of qualia optimization.

\PackageWarning{manual}{Warning: Figure occurs prior to reference.}
\begin{figure*}[htbp]
    \centering
    \includegraphics[width=0.7\textwidth]{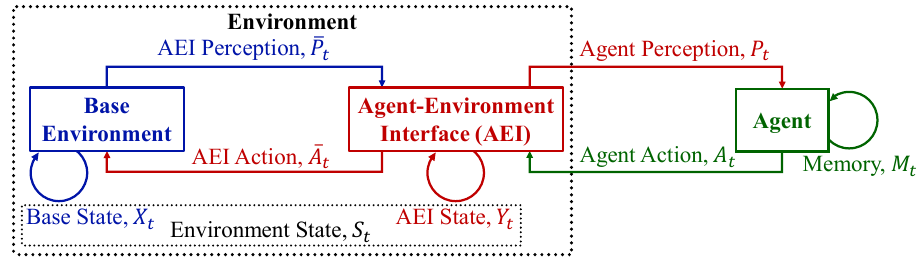}
    \caption{A revision of Figure \ref{fig:AEDiagram}, showing how the agent interacts with the environment with the inclusion of the \emph{agent-environment interface} (AEI). The AEI influences the agent's experiences by transforming its perceptions and actions as the agent indirectly interacts with the base environment.}
    \label{fig:AEI}
\end{figure*}

\subsection{Agent-Interface-Environment Process (AIEP)}
\label{sec:AEI}

Effective qualia optimization may require changing more than just the algorithm the agent implements---it might require changing how the agent interacts with the environment or perhaps even changing the environment entirely. This may range from simple changes like changing the rewards provided to an RL agent, to more sophisticated changes like altering the rate with which the agent interacts with the environment. In even more extreme cases, one might be able to change the environment so completely that it hardly resembles the original environment, as long as a well-performing policy for the original environment can be extracted from the policy learned by the agent.

\subsubsection{Example of Experience-Transforming Interventions}
\label{sec:shuttleBox}

To ground our discussion and formalization of mechanisms that transform agent experiences, consider the motivating example of a rat learning to move from one compartment of a shuttle box to another when a light is turned on.\footnote{A shuttle box is an apparatus used in animal learning experiments. It has two compartments that an animal can move between, and often includes equipment to administer electrical shocks.} In a \emph{conditioned avoidance response} (CAR) test, a rat receives electric shocks shortly after a light is turned on. Through classical conditioning, the rat learns to associate the light (the conditioned stimulus) with the impending shock (the unconditioned stimulus). Then, via operant conditioning, the rat learns to avoid the shocks by moving to the other compartment when the light is turned on.

The rat could be trained to produce the target behavior of moving from one compartment to the other when the light is turned on without the use of an aversive stimulus like electric shocks. For example, the rat could be provided with an appetitive stimulus, like a particularly desirable food, when it exhibits the target behavior. If the goal is merely to train the rat to produce the target behavior, then in principle either of these approaches could be effective.\footnote{The goal of a CAR test is not just to produce this target behavior, hence the typical reliance on aversive stimuli.}

If rats have qualia similar to those of humans, then this provides an intuitive example wherein changes to the environment of an agent-environment system could alter an agent's experiences, with (perhaps) relatively little impact on the learned behavior. We aim to modify the AEP and AERP problem formulations so that they allow for the consideration of a range of such transformations of an agent's experiences, enabling the study of the impact that these transformations have on both performance (learning) and the quality of the agent's qualia. 

\subsubsection{Agent-Environment Interface (AEI)}

To achieve this aim, we focus on an initial setting wherein the agent interacts with the environment indirectly through an \emph{agent-environment interface} (\definesymbol{AEI}{aeixidea}), depicted in Figure \ref{fig:AEI}, that can transform the agent's experiences, but which does not allow for a complete disconnect between the environment of interest and what the agent experiences. 

In this setting, there is an environment, which we call the \emph{base environment}, that the agent will interact with. An agent could interact directly with this base environment, and this interaction could be modeled as an AEP. However, instead of interacting directly with the base environment, the agent's perceptions and actions are transformed by an AEI. There are multiple ways that our motivating example can be modeled within this framework. The agent might correspond to the rat and the base environment might correspond to the shuttle box without any mechanisms to train the rat. We could then consider AEIs that make no changes to the majority of the rat's experiences, including no changes to the layout of the compartments or to the laws of physics governing its movement. When the light is turned on, one AEI might administer electric shocks until the rat moves to the other compartment. A different AEI might dispense food if the rat quickly enters the other compartment after the light is turned on. 

Since the AEI's transformations impact the agent's experiences, we do not consider the AEI to be part of the agent (we revisit this choice later). Instead, we view the AEI as part of the agent's environment---a part that can be modified 
to optimize the agent's performance and qualia. Taking this perspective, the environment that the agent interacts with consists of both the base environment and the AEI. We will formulate this system in such a way that this combined environment with which the agent interacts remains the environment of an AEP. Hereafter, references to the ``environment'' correspond to this combined environment---any references to the \emph{base} environment will be explicit. 

Although some transformations of the agent's experiences could be achieved by a state-free AEI, some transformations may require the AEI to have memory of its own. We therefore define the AEI to have its own state, \definesymbol{$Y_t$}{Yt}, which we call the \emph{AEI state}. To simplify the problem formulation, we allow the AEI's state to be updated twice per time step---once before it generates the agent's perception $P_t$, and once after the agent selects action $A_t$. We handle this dual-update of the AEI state within each time step by defining $Y_t$ to be the AEI state prior to the generation of $P_t$ and \definesymbol{$Y'_t$}{Ypt} to be the AEI state after the agent selects action $A_t$. We call an AEI \emph{state-free} if $Y_t$ and $Y'_t$ are the same constant random variables for all $t$ (e.g., $Y_t=0$ and $Y'_t=0$ always). 

Let \definesymbol{$X_t$}{Xt} denote the state of the base environment, which we call the \emph{base state}. Together, the AEI state and base state form the \emph{environment state}, $S_t=(X_t,Y_t)$, which constitutes the higher-level state of the environment with which the agent interacts. We refer to $Y'_t$ as the \emph{intermediate AEI state} because it occurs between environment states $S_t=(X_t,Y_t)$ and $S_{t+1}=(X_{t+1},Y_{t+1})$. 

An AEP that includes an AEI operates as follows at each time $t$. First, the base state is updated from $X_{t-1}$ to $X_t$ based on the \emph{AEI action} from time $t-1$, \definesymbol{$\widebar A_{t-1}$}{barAt}. The AEI is influenced by the base state via a random variable \definesymbol{$\widebar P_t$}{barPt} that we call the \emph{AEI perception}.\footnote{The formulation that we introduce includes so many symbols and assumptions that a detailed discussion of each would be too cumbersome. Although we present each definition and assumption formally, we refer the reader to the detailed discussion of similar terms in the general AEP formulation for additional discussion. For example, here it follows from the definition of $\widebar P_t$ and the subsequent assumptions that $\widebar P_t$ is a complete characterization of everything about $X_t$ that influences the AEI at time $t$, much like $P_t$ in the AEP formulation.} The AEI perception is a deterministic function of $X_t$, and so it is determined when the base state is updated to $X_t$. The AEI perception $\widebar P_t$ then causes the AEI to update its state to $Y_t$. In the general AEP formulation, the \emph{agent perception} \definesymbol{$P_t$}{Pt2} is a function of the environment state $S_t=(X_t,Y_t)$. However, we consider a restricted setting where the agent's perceptions are entirely controlled by the AEI, and so the agent perception $P_t$ can be expressed as a function of just the AEI state $Y_t$ and AEI perception $\widebar P_t$ (i.e., $X_t$ does not directly influence $P_t$). The agent perception $P_t$ causes the agent to update its memory to $M_t$. The \textit{agent action} \definesymbol{$A_t$}{At2} is a deterministic function of $M_t$ and is therefore also implicitly determined. The agent action $A_t$ causes the AEI to update its state to the intermediate value $Y'_t$. The \emph{AEI action} $\widebar A_t$ is a deterministic function of $Y'_t$ and $A_t$ and is therefore also implicitly determined. 

More formally, an AEP that includes an AEI results in the generation of a sequence of random variables for each time $t\in \{0,1,\dotsc\}$: $X_t, \widebar P_t, Y_t, P_t, M_t, A_t, Y'_t,$ and $\widebar A_t$. As a reminder, we refer to $X_t$ as the \emph{base state}, $\widebar P_t$ as the \emph{AEI perception}, $Y_t$ as the \emph{AEI state}, $P_t$ as the \emph{agent perception}, $M_t$ as the (agent) \emph{memory}, $A_t$ as the \emph{agent action}, $Y'_t$ as the \emph{intermediate AEI state}, $\widebar A_t$ as the \emph{AEI action}, and $S_t=(X_t,Y_t)$ as the \emph{environment state}.\footnote{We encourage readers overwhelmed with this notation and terminology to consult Figure \ref{fig:AEI}.} We define all of these random variables to be \texttt{null} when $t=-1$ to avoid having to define special case distributions for the initial values of these random variables.

We make the following assumptions and define the following notation regarding these random variables. 
\begin{enumerate}[leftmargin=1.8em, itemsep=0.5ex]
    \item $X_t$ is conditionally independent of all previous random variables given $X_{t-1}$ and $\widebar A_{t-1}$. We write \definesymbol{$d_x$}{dx}$(x,\widebar a)$ to denote the conditional distribution of $X_t$ given $X_{t-1}=x$ and $\widebar A_{t-1}=\widebar a$. We also assume that $d_x$ is stationary (i.e., it does not depend on the time $t$). 
    \item $\widebar P_t= f_{\widebar p}(X_t)$, where \definesymbol{$f_{\widebar p}$}{fbarp} is a function that we call the \emph{AEI perception function}.
    \item $Y_t$ is conditionally independent of all previous random variables given $Y'_{t-1}$ and $\widebar P_t$. We write \definesymbol{$d_y$}{dy}$(y',\widebar p)$ to denote the conditional distribution of $Y_t$ given that $Y'_{t-1}=y'$ and $\widebar P_t=\widebar p$. We also assume that $d_y$ is stationary. 
    \item $P_t=f_p(Y_t, \widebar P_t)$, where \definesymbol{$f_p$}{fp2} is a function that we call the \emph{agent perception function}.
    \item $M_t$ is conditionally independent of all previous random variables given $M_{t-1}$ and $P_t$. We write \definesymbol{$d_m$}{dm2}$(m,p)$ to denote the conditional distribution of $M_t$ given that $M_{t-1}=m$ and $P_t=p$. We also assume that $d_m$ is stationary.
    \item $A_t=f_a(M_t)$, where \definesymbol{$f_a$}{fa2} is a function that we call the \emph{agent action function}. 
    \item $Y'_t$ is conditionally independent of all previous random variables given $Y_t$ and $A_t$. We write \definesymbol{$d_{y'}$}{dyp}$(y,a)$ to denote the conditional distribution of $Y'_t$ given that $Y_t=y$ and $A_t=a$. We also assume that $d_{y'}$ is stationary. 
    \item $\widebar A_t=f_{\widebar a}(Y'_t, A_t)$, where \definesymbol{$f_{\widebar a}$}{fba} is a function that we call the AEI action function.
\end{enumerate}

We call this process, which models an AEP that includes an AEI, an \emph{agent-interface-environment process} (\definesymbol{AIEP}{aiep}). Note that all AIEPs are AEPs---AEPs where the environment can be decomposed into two components: the base environment and an AEI. If the base environment is an AERP, we redefine its reward function to be the \emph{base reward function} \definesymbol{$f_{\widebar r}$}{fbarr} and the rewards that it produces to be the \emph{base rewards} \definesymbol{$\widebar R_t$}{barRt}$\,=f_{\widebar r}(\widebar P_t)$. In this case where the base environment is an AERP, we typically define the AEI to also produce rewards \definesymbol{$R_t$}{Rt2}, which we call \emph{AEI rewards} or \emph{agent rewards},\footnote{When the focus is on how the AEI produces rewards we use the phrase ``AEI reward.'' However, when the focus is on the rewards that the agent receives, we us the phrase ``agent reward.'' These terms are interchangeable.} and which are produced by an \emph{AEI reward function} \definesymbol{$f_r$}{fr2} according to the equation $R_t=f_r(P_t)$. We refer to such an AIEP where the base environment is an AERP and where the AEI produces rewards $R_t$ as an \emph{agent-interface-environment reward process} (\definesymbol{AIERP}{aierp}). Note that all AIERPs are AERPs.

Recall that the conditional independence (Markov) assumptions in the specification of AEPs characterize how states, perceptions, memories, and actions should be defined for a given agent-environment system. Although they are stated as assumptions, they do not restrict the set of agent-environment systems that can be modeled as AEPs. Similarly, the assumptions above do not limit the systems that can be modeled as AIEPs and AIERPs, but rather inform how the various states, perceptions, actions, and memories can be defined for a given system containing a base environment, AEI, and agent. 

However, these conditional independence and stationarity assumptions do not fully specify which information should be encoded within the base state $X_t$ and which information should be encoded within the AEI state $Y_t$. Consider an agent-environment system that one aims to convert into and agent-interface-environment system by splitting the environment into a base environment and AEI. One could define $Y_t$ to be the state of everything external to the agent (including the state of the base environment) and $X_t$ to be a constant, essentially encoding the environment entirely within the AEI. This would not align with the intuition that the AEI should be a mechanism that transforms the agent's experiences as it interacts with the environment. 

To specify what information should be encoded within the AEI state $Y_t$, note that we will formulate qualia optimization as the problem of finding AEI and agent specifications that in some way optimize both performance on the base environment and the quality of the agent's qualia. Hence, for a given agent-interface-environment system, the AEI state should only include quantities that can be changed during efforts to simultaneously optimize performance and qualia. 

For simplicity, in this initial work we assume that if $\widebar P_t = p_\infty$ then $P_t=p_\infty$. That is, the AEI preserves episode termination. Alternate formulations wherein the AEI can terminate episodes early or pad episodes with additional time steps could be modeled by allowing the AEI to interact with the agent and environment at different and varying timescales---allowing for multiple interactions with the agent within a single interaction with the base environment or vice versa. While this may provide an interesting direction for future work (and our initial formulation of the AEI allowed for such varying timescales), the key insights presented in this report do not rely on varying timescales, and so we present a simplified AEI where the agent and environment operate on the same timescale. 

\PackageWarning{manual}{Warning: Algorithm occurs prior to reference.}

\begin{algorithm}[htpb]
\DontPrintSemicolon
Initialize $X_{-1},M_{-1},Y'_{-1},$ and $\widebar A_{-1}$ to \texttt{null}\;
\For{$t \gets 0$ \KwTo $\infty$}
{
    $X_t \sim d_x(X_{t-1},\widebar A_{t-1})$\;
    $\widebar P_t = f_{\widebar p}(X_t)$\;
    $Y_t \sim d_y(Y'_{t-1}, \widebar P_t)$\;
    $P_t = f_p(Y_t,\widebar P_t)$\;
    $M_t \sim d_m(M_{t-1}, P_t)$\;
    $A_t = f_a(M_t)$\;
    $Y'_t \sim d_{y'}(Y_t, A_t)$\;
    $\widebar A_t = f_{\widebar a}(Y'_t, A_t)$\;
}
\caption{Agent-Interface-Environment Process (AIEP)}
\label{alg:AEI_basic}
\end{algorithm}

Algorithm \ref{alg:AEI_basic} provides pseudocode for generating the sequence of random variables $(X_t, \widebar P_t, Y_t, P_t, M_t, A_t, Y'_t, \widebar A_t)_{t=0}^\infty$ as specified by an AIEP. An AIERP would include the specification of the base reward $\widebar R_t=f_{\widebar r}(\widebar P_t)$ immediately after the specification of $\widebar P_t$, and the specification of the AEI reward $R_t=f_r(P_t)$ immediately after the specification of $P_t$. 

Let $\texttt{aei}_\mathbf{I}$ denote an AEI that makes no changes to the agent-environment interactions, which we call the \emph{identity AEI}. That is, under \definesymbol{$\texttt{aei}_\mathbf{I}$}{aeiI}, $Y_t$ and $Y'_t$ are constants (i.e., the AEI has no internal state), $P_t = \widebar P_t$, $\widebar A_t = A_t$, and, if the process is an AIERP, $R_t = \widebar R_t$.

\subsection{Formulating Qualia Optimization}
\label{sec:FQO}

Classical RL literature focuses on (perception) \emph{policies}, which characterize how the agent selects actions based on the current perception, or based on the history of agent perceptions, actions, and rewards. This results in objective functions that evaluate the quality of policies by taking as input a policy and providing as output a measure of the amount of reward the agent would receive if it selected actions using the provided policy. This policy-centric perspective provides a clear goal when designing classical RL algorithms---they should cause the agent to find (or approximate) a policy that maximizes the chosen objective function, and should do so quickly. 

However, this policy-centric perspective is less conducive to formalizing the problem of qualia optimization. It would be restrictive to measure the quality of qualia for policies alone, since RL agents (and more generally, AI agents) are more than just a policy---they are agents that learn from their experiences, changing their policies and memory over time based on interactions with their environments. We therefore take a more agent-centric perspective, defining objective functions for qualia optimization as functions of the entire system (the environment, AEI, and agent), not just a policy. In order to do so, we first define AI algorithms \texttt{alg} and AEI specifications \texttt{aei} more formally, so that we can define objective functions that take AEIs and AI algorithms as input.

\subsubsection{Base Environment, AEI, and Algorithm Definitions}

Recall that Algorithm \ref{alg:AEI_basic} characterizes the entire base environment, AEI, and agent system, describing the order in which random variables are sampled and the conditional distributions of each random variable. We propose partitioning these conditional distributions into three sets: those that characterize the base environment, those that characterize the AEI, and those that characterize the algorithm implemented by the agent. 
Although the algorithm implemented by the agent may not be an RL algorithm, due to our initial focus on RL algorithms and RL settings, hereafter we call this algorithm the \emph{RL algorithm}.

The base environment \definesymbol{\texttt{env}}{env} determines the conditional distributions of $X_t$ and $\widebar P_t$ (and $\widebar R_t$ if the base environment is an AERP). So, the base environment \texttt{env} is characterized by $d_x$ and $f_{\widebar p}$ (and $f_{\widebar r}$ if the base environment is an AERP). The RL algorithm \definesymbol{\texttt{alg}}{alg} determines the conditional distributions of $M_t$ and $A_t$. So, \texttt{alg} is characterized by $d_m$ and $f_a$, just as in the AEP formulation. The AEI \definesymbol{\texttt{aei}}{aeixsym} determines the conditional distributions of $Y_t, P_t, Y'_t,$ and $\widebar A_t$ (and $R_t$ in the case of an AIERP). So, \texttt{aei} is characterized by $d_y, f_p, d_{y'},$ and $f_{\widebar a}$ (and $f_r$ in the case of an AIERP). This partitioning of random variables is depicted by the colors in Figure \ref{fig:AEI}: The conditional distributions of \textcolor{figBlue}{blue}, \textcolor{figRed}{red}, and \textcolor{figGreen}{green} random variables are determined by \textcolor{figBlue}{\texttt{env}}, \textcolor{figRed}{\texttt{aei}}, and \textcolor{figGreen}{\texttt{alg}} respectively.

Although we formally characterize base environments, AEIs, and RL algorithms as sets of conditional distributions and functions, one need not think of them as explicit conditional probability tables and specific function definitions. Instead, they can each be viewed as any complete specification of the corresponding terms. For example, BAC with hyperparameter specifications corresponds to one possible RL algorithm \texttt{alg}, and BAC can be viewed as a set of possible RL algorithms (each hyperparameter setting induces a different algorithm).

\subsubsection{Performance and Qualia Objective Functions}

We now define two objective functions (one for qualia and one for performance) that play central roles in qualia optimization problems formulated as AIEPs. Let 
\definesymbol{$\mathfrak q$}{q}$(\texttt{alg},\texttt{aei})$ be a real-valued quantification of the quality of an RL agent's qualia when using RL algorithm \texttt{alg} and AEI \texttt{aei}~on some implicit base environment \texttt{env}. Larger values of $\mathfrak q(\texttt{alg},\texttt{aei})$ correspond to more desirable qualia of the RL agent during its lifetime, and we call $\mathfrak q$ the \emph{qualia objective function}. Similarly, let \definesymbol{$\mathfrak p$}{p}$(\texttt{alg}, \texttt{aei})$ be a real-valued quantification of the \emph{performance} of an RL agent using RL algorithm \texttt{alg} and the AEI specified by \texttt{aei}~on the same implicit base environment \texttt{env}.\footnote{We select Fraktur as the typeface for $\mathfrak{q}$ and $\mathfrak{p}$ to avoid conflicts with the commonly used symbols $p$ and $q$, while maintaining mnemonic links between ``$\mathfrak p$'' and ``performance'' and ``$\mathfrak q$'' and ``qualia.''}

By defining the qualia objective function $\mathfrak q$ in this way, we make the implicit assumption that the quality of agents' qualia can be quantified by a function of the dynamics of the system comprised of the base environment, AEI, and agent.\footnote{We recognize that this sentence can be cumbersome and difficult to parse. Notice that ``dynamics of the system comprised of the base environment, AEI, and agent'' refers to $d_x, f_{\widebar p}, d_y, f_p, d_m, f_a, d_{y'}, f_{\widebar a}$ (and $f_r$ and $f_{\widebar r}$ for an AIERP).} Notice that this does \textit{not} imply the additional implicit assumption that whether or not agents experience qualia can be determined from these system dynamics. This formulation allows for the possibility that even if qualia arise from immaterial phenomena as suggested by Cartesian dualism, their quality could still be determined from the dynamics of the system comprised of the base environment, AEI, and agent. 

Also notice that in this formulation the qualia objective is a function of the system dynamics, not a specific outcome. That is, it is a function of the \emph{distributions} of the random variables $X_t, \widebar P_t, Y_t, P_t, M_t, A_t, Y'_t,$ and $\widebar A_t$ rather than a function of a \emph{realization} of these random variables. If the quality of the qualia of an agent is a function of a realization of these random variables, this formulation allows for optimization of parameters of the distribution of qualia quality, such as the mean or median. 

We focus on the setting where performance is a property of interactions with the base environment---which AEI actions, $\widebar A_t$, are chosen and which base states, $X_t$, result. For example, performance does not depend on \textit{how} the AEI and agent compute the AEI action $\widebar A_t$, just \textit{what} the action $\widebar A_t$ is. Put differently, performance is with respect to the base environment, and is an evaluation of the outward behavior of the agent-AEI system. More formally, we assume that $\mathfrak p(\texttt{alg},\texttt{aei})$ can be expressed as a function of the distributions of $X_t$ and $\widebar A_t$ for all $t$. Hence, if these distributions are held constant, the value of $\mathfrak p(\texttt{alg}, \texttt{aei})$ does not vary with the distributions of other random variables in the AIEP. While this may exclude some reasonable definitions of performance, like those that consider how much energy it takes for the agent-AEI system to compute $\widebar A_t$, it simplifies our initial exploration of qualia optimization.

Similarly, we focus on the setting where the quality of an agent's qualia (measured by $\mathfrak q$) is a property of the agent's interactions---its perceptions $P_t$, which actions $A_t$ it takes, and what memories $M_t$ it has. More formally, $\mathfrak q(\texttt{alg},\texttt{aei})$ can be expressed as a function of the distributions of $P_t, M_t,$ and $A_t$, for all $t$. Hence, if these distributions are held constant, the value of $\mathfrak q(\texttt{alg},\texttt{aei})$ does not vary with the distributions of other random variables in the AIEP. This encodes that the AEI changes the agent's experiences, as opposed to the AEI being part of the agent---a modeling choice that we question later. 

\subsubsection{Generalizing Qualia Objective Functions}

Notice that the concept of qualia objective functions can trivially be extended from AIEPs to AEPs since the value of the qualia objective function only depends on the distributions of the agent perceptions $P_t$, agent actions $A_t$, and agent memories $M_t$---random variables that exist in the more general AEP formulation. In such settings the \texttt{aei} is not part of the problem formulation and so it is also no longer an input to the qualia objective function, and so $\mathfrak q(\texttt{alg})$ denotes the value of the qualia objective function for RL algorithm $\texttt{alg}$ (and some implicit environment). The performance objective function can be similarly extended, although not quite as trivially because it depends on the distributions of random variables not present in the AEP formulation (e.g., base states, AEI actions, and base rewards). 

Later we will reason about the impact that the representation functions $\rho_{P_t}$, $\rho_{A_t}$, and $\rho_{M_t}$ have on the qualia objective function. So far we have defined $\mathfrak q$ for some implicit underlying AEP (or AIEP). Varying these representation functions while keeping the underlying physical properties $\Phi_{P_t}$, $\Phi_{A_t}$, and $\Phi_{M_t}$ the same results in a range of AEPs. To facilitate discussion of this range of AEPs, note that we can (with a mild abuse of notation due to defining $\mathfrak q$ to take different arguments) write the qualia objective function explicitly as a function of the distributions of $P_t$, $M_t$, and $A_t$ for all $t$: 
\begin{equation}
    \mathfrak q\Big (\operatorname{law}\big((P_t,M_t,A_t)_{t=0}^\infty\big )\Big ),
\end{equation}
where $\operatorname{law}((P_t,M_t,A_t)_{t=0}^\infty)$ denotes the joint distribution over the entire sequence $(P_0, M_0, A_0, P_1, M_1, A_1, P_2, \dotsc)$. Furthermore, this joint distribution can be determined from the joint distribution of the corresponding physical properties and the corresponding representation functions, and so the qualia objective function can be expressed as:
\begin{equation}
    \mathfrak q\Big (\operatorname{law}\big((\Phi_{P_t},\Phi_{M_t},\Phi_{A_t})_{t=0}^\infty\big ), (\rho_{P_t}, \rho_{M_t}, \rho_{A_t})_{t=0}^\infty\Big ). 
\end{equation}
Although cumbersome, this form plays a central role in our later analysis. 

\subsubsection{Candidate Problem Formulations}

We consider a multi-objective problem setting wherein the goal is to find an RL algorithm \texttt{alg} and AEI specification \texttt{aei}~that simultaneously maximizes both $\mathfrak p(\texttt{alg}, \texttt{aei})$ and $\mathfrak q(\texttt{alg}, \texttt{aei})$. There are a variety of ways that one could formalize the goal of simultaneously maximizing the quality of the agent's qualia and the performance of the agent, a few of which we review below.

\begin{enumerate}
    \item Qualia optimization can be formalized as the problem of maximizing a weighted combination of the two objectives: 
    \begin{equation}
        \lambda \mathfrak q(\texttt{alg}, \texttt{aei}) + (1-\lambda)\mathfrak p(\texttt{alg}, \texttt{aei}),
    \end{equation}
    where $\lambda \in [0,1]$ is a hyperparameter (not related to the $\lambda$ hyperparameter of the BAC algorithm). 
    \item Qualia optimization can be formalized as the problem of maximizing performance subject to a constraint on the quality of the agent's qualia:
    \begin{gather}
        \label{eq:exampleQualiaOptProblem}
        \argmax_{\texttt{alg},\texttt{aei}} \mathfrak p(\texttt{alg},\texttt{aei})\\
        \nonumber \text{subject to }\mathfrak q(\texttt{alg},\texttt{aei}) \geq c,
    \end{gather}
    for some real-valued constant $c$.
    \item Qualia optimization can be formalized as the problem of maximizing the quality of the agent's qualia subject to a constraint on the performance of the agent:
    \begin{gather}
        \argmax_{\texttt{alg},\texttt{aei}} \mathfrak q(\texttt{alg},\texttt{aei})\\
        \nonumber \text{subject to }\mathfrak p(\texttt{alg},\texttt{aei}) \geq c,
    \end{gather}
    for some real-valued constant $c$.
    \item Qualia optimization can be formalized as the problem of maximizing the performance of the agent, breaking ties by considering the quality of the agent's qualia. That is, we might define a total order on $(\texttt{alg},\texttt{aei})$-pairs such that $(\texttt{alg},\texttt{aei}) \geq (\texttt{alg}',\texttt{aei}')$ if and only if $\mathfrak p(\texttt{alg},\texttt{aei})> \mathfrak p(\texttt{alg}',\texttt{aei}')$ or both $\mathfrak p(\texttt{alg},\texttt{aei})=\mathfrak p(\texttt{alg}',\texttt{aei}')$ and $\mathfrak q(\texttt{alg},\texttt{aei})\geq\mathfrak q(\texttt{alg}',\texttt{aei}')$. Qualia optimization can then correspond to finding an $(\texttt{alg}^*,\texttt{aei}^*)$ such that $(\texttt{alg}^*,\texttt{aei}^*)\geq (\texttt{alg}',\texttt{aei}')$ for all $(\texttt{alg}',\texttt{aei}')$, if such an optimal algorithm-AEI pair exists.
\end{enumerate}

The selection of the appropriate formalization of qualia optimization is important because different choices can result in different optimal algorithm and AEI specifications. However, in this initial work we do not adopt a specific formalization and instead focus on general strategies for increasing $\mathfrak q(\texttt{alg},\texttt{aei})$ with little or no change to $\mathfrak p(\texttt{alg},\texttt{aei})$---strategies that could be effective mechanisms for any of the formalizations above. For example, in some settings we show that there exist AEI specifications \texttt{aei}~that cause maximizing the quality of the agent's qualia to align with the agent's goal of maximizing the expected discounted return, and in another case we provide an AEI specification that can increase $\mathfrak q(\texttt{alg},\texttt{aei})$ without influencing $\mathfrak p(\texttt{alg},\texttt{aei})$ at all. Although we do not focus on any one of the above problem formulations, they provide context for the types of problem formulations that we have in mind when devising qualia-improving mechanisms.

In later sections we consider problem settings where the sets of AEIs and RL algorithms under consideration are restricted. Let \texttt{AEI} and \texttt{ALG} denote the sets of AEIs and RL algorithms under consideration. We call these sets the sets of \emph{admissible} AEIs and RL algorithms. To model this restricted problem setting, qualia optimization problem settings can be restricted to only consider elements of \texttt{ALG} and \texttt{AEI}. For example, \eqref{eq:exampleQualiaOptProblem} could be rewritten as:
\begin{gather}
    \label{eq:exampleQualiaOptProblem2}
    \argmax_{\texttt{alg}\in\texttt{ALG},\,\texttt{aei}\in\texttt{AEI}} \mathfrak p(\texttt{alg},\texttt{aei})\\
    \nonumber \text{subject to }\mathfrak q(\texttt{alg},\texttt{aei}) \geq c.
\end{gather}
In such cases, the domains of $\mathfrak p$ and $\mathfrak q$ need only be the admissible algorithms and AEIs---$\mathfrak p$ and $\mathfrak q$ can be undefined for other AEIs and RL algorithms.

In the following sections we consider different classes of definitions of $\mathfrak q$ and their implications. In Section \ref{sec:RewardQualiaHypothesis} we consider definitions wherein the quality of an agent's qualia is related to the amount of reward the agent receives. After Section \ref{sec:RewardQualiaHypothesis} we re-evaluate the problem formulation based on insights from this initial reward-qualia setting before considering other definitions of $\mathfrak q$ that focus on TD errors and the reinforcement of behavior. 

\section{Reward Hypothesis for Qualia}
\label{sec:RewardQualiaHypothesis}

In this section we restrict our focus to AIERPs and consider the implications of the following assumption, which refines Assumption \ref{ass:function}:
\begin{ass}[Reward$\leftrightarrow$Qualia]
    \label{ass:RewardQualiaHypothesis}
    The quality of an RL agent's qualia can be measured in terms of the amount of reward it receives. 
\end{ass}
This assumption is natural, since obtaining positive rewards\footnote{Note that we use ``reward'' to refer to $R_t$ in the AEP formulation. Depending on how an AEP models a physical system, it may be more appropriate to call $R_t$ a \emph{reward signal} \citep{barto2013intrinsic}. We use the term ``reward'' for brevity.} seems intuitively likely to produce desirable qualia and obtaining negative rewards (punishments) seems likely to produce undesirable qualia. Also, notice that here the ``amount of reward'' that the agent receives measures the amount of agent rewards $R_t$ (the rewards received by the agent), not the amount of base environment rewards $\widebar R_t$ (the rewards produced by the base environment). Hereafter, we refer to the setting induced by this assumption as the \emph{reward-qualia} setting.

As we will see, the reward-qualia setting faces serious challenges that call into question its viability as a setting of interest. We will show that, under mild assumptions, degenerate solutions that constitute no real changes to the system can inflate variants of this qualia objective arbitrarily. We also use this relatively simple setting to introduce a variety of concepts that inform qualia optimization in general and which are referenced in subsequent settings. For example, in Section \ref{sec:objAlign} we introduce the concept of objective function alignment, and provide an example of how AEIs can be designed to cause the alignment of the objective functions. 

\subsection{Objective Alignment}
\label{sec:objAlign}

Intuitively, two objectives are aligned if, when one increases, it always means that the other increases as well. Formally, we say that the objective functions $\mathfrak p$ and $\mathfrak q$ are \emph{aligned} under AEI $\texttt{aei}$ if any changes to an RL algorithm that increase one objective necessarily also increase the other objective. That is, $\mathfrak p$ and $\mathfrak q$ are aligned under AEI $\texttt{aei}$ if for all RL algorithms $\texttt{alg}$ and $\texttt{alg}'$,
\begin{equation}
    \label{eq:alignedDefn}
    \underbrace{\mathfrak q(\texttt{alg}',\texttt{aei}) > \mathfrak q(\texttt{alg},\texttt{aei})}_\text{qualia improvement} \Longleftrightarrow\underbrace{\mathfrak p(\texttt{alg}',\texttt{aei}) > \mathfrak p(\texttt{alg},\texttt{aei})}_\text{performance improvement}.
\end{equation}
This might be viewed as a notion of \emph{strong} alignment, with weaker forms of alignment only requiring the implication to hold in one direction. 

The assumption that the quality of an RL agent's qualia can be measured in terms of the amount of reward it receives does \textit{not} necessarily imply that $\mathfrak p$ and $\mathfrak q$ are aligned under the identity AEI $\texttt{aei}_\mathbf{I}$, since $\mathfrak p$ and $\mathfrak q$ can measure ``amount of reward'' differently. For example, they might use different reward discount parameters $\gamma$, or one might measure the \emph{expected} discounted return over some finite number of episodes while the other measures the \emph{median} discounted return.

This raises the question: Under what conditions does there exist an AEI under which $\mathfrak p$ and $\mathfrak q$ are aligned? Here we present one example of a setting wherein these objectives can be aligned:\footnote{Although this specific setting is not of particular interest, it is simple and is sufficient for our subsequent exploration of the limitations of objective alignment.}
\begin{itemize}
    \item There is some finite maximum number of episodes $i_\text{max}$.
    \item The AIERP is episodic and finite-horizon.\footnote{Finite-horizon AEPs were defined in Section \ref{sec:episodes}. This definition extends to AERPs, AIEPs, and AIERPs, since they are all AEPs.}
    \item The rewards $R_t$ and $\widebar R_t$ are bounded (to ensure that the following expectations exist). 
    \item The value of the qualia objective, $\mathfrak q(\texttt{alg},\texttt{aei})$, is the expected discounted sum of agent rewards over $i_\text{max}$ episodes using reward discount parameter $\gamma_{\mathfrak q} \in (0,1]$:
    \begin{equation}
        \label{eq:discountedReturnQualiaDefn}
        \mathfrak q(\texttt{alg},\texttt{aei}) = \mathbf{E}\left [\sum_{i=0}^{i_\text{max}-1}\sum_{t=\operatorname{start}(i)+1}^{\operatorname{end}(i)} \gamma_{\mathfrak q}^{\operatorname{dur}(t)} R_t \right ].
    \end{equation}
    \item The value of the performance objective, $\mathfrak p(\texttt{alg},\texttt{aei})$, is the same expression using a different reward discount parameter $\gamma_{\mathfrak p} \in [0,1]$, and $\widebar R_t$ rather than $R_t$:
    \begin{equation}
        \mathfrak p(\texttt{alg},\texttt{aei}) = 
        \mathbf{E}\left [\sum_{i,t} \gamma_{\mathfrak p}^{\operatorname{dur}(t)} \widebar R_t \right ],
    \end{equation}
    where hereafter \definesymbol{$\sum_{i,t}$}{sumit} is shorthand for the summations in \eqref{eq:discountedReturnQualiaDefn}.
\end{itemize}

In this setting it is straightforward to design an AEI, $\texttt{aei}^\star$, under which these two objectives are aligned: for all times $t$, $\texttt{aei}^\star$ sets $R_t = \gamma_{\mathfrak p}^{\operatorname{dur}(t)} \gamma_{\mathfrak q}^{-\operatorname{dur}(t)}\widebar R_t$, while leaving the other random variables (effectively) unchanged. More precisely, first $d_y$ and $d_{y'}$ are defined such that $Y_t$ encodes $\operatorname{dur}(t)$. Next, $f_{\widebar a}(y',a)=a$ for all $y'$ and $a$ so that $\widebar A_t = A_t$. Next, the agent perception $P_t$ is defined to be the AEI perception $\widebar P_t$, but augmented with the new reward value (so the AEI reward function can subsequently determine the agent reward from the agent perception). That is, $P_t=(\widebar P_t, R_t)$, which is achieved by defining
$f_p(y,\widebar p)=(\widebar p, \gamma_{\mathfrak p}^{\operatorname{dur}(t)}\gamma_{\mathfrak q}^{-\operatorname{dur}(t)}f_{\widebar r}(\widebar p))$ for all $y$ and $\widebar p$. Finally, $f_r(p)=r$ for all $p=(\widebar p, r)$ so that $R_t=\gamma_{\mathfrak p}^{\operatorname{dur}(t)}\gamma_{\mathfrak q}^{-\operatorname{dur}(t)}\widebar R_t$. This means that
\begin{align}
    \mathfrak q(\texttt{alg},\texttt{aei}^\star)=&\mathbf{E}\left [\sum_{i,t} \gamma_{\mathfrak q}^{\operatorname{dur}(t)} R_t \right ]\\
    =& \mathbf{E}\left [\sum_{i,t} \gamma_{\mathfrak q}^{\operatorname{dur}(t)} \left ( \frac{\gamma_{\mathfrak p}^{\operatorname{dur}(t)}}{\gamma_{\mathfrak q}^{\operatorname{dur}(t)}} \widebar R_t \right ) \right ]\\
    =& \mathbf{E}\left [\sum_{i,t} \gamma_{\mathfrak p}^{\operatorname{dur}(t)} \widebar R_t \right ]\\
    =& \mathfrak p(\texttt{alg},\texttt{aei}^\star),
\end{align}
which implies \eqref{eq:alignedDefn} when $\texttt{aei}=\texttt{aei}^\star$.

\subsubsection{Objective Alignment as a Qualia Optimization Strategy}

Consider a setting like this, where the identity AEI, $\texttt{aei}_{\mathbf I}$, does not align the two objectives, but a different AEI, $\texttt{aei}^\star$, does. Although it might seem like replacing $\texttt{aei}_{\mathbf I}$ with $\texttt{aei}^\star$ would improve the quality of the agent's qualia and thus would be an effective strategy for qualia optimization, this is not the case. The alignment of objectives, as defined in \eqref{eq:alignedDefn}, is a statement about what happens when the RL algorithm changes while the AEI remains fixed, not a statement about what happens when the AEI is changed. 

It can be the case, and may often be the case, that an RL algorithm \texttt{alg}, tuned to be effective with the identity AEI, $\texttt{aei}_{\mathbf I}$, may not be effective with an objective-aligning $\texttt{aei}^\star$. This can happen, for example, due to hyperparameters like step sizes being tuned for properties of $\texttt{aei}_\mathbf{I}$, like the scale of rewards that it induces. More generally, $\texttt{aei}^\star$ could reduce the magnitudes of $\mathfrak p$ and $\mathfrak q$ by, for example, restricting the set of (base and agent) actions that the AEI considers and reducing agent rewards, so that even RL algorithms that are well-tuned using $\texttt{aei}^\star$ might not achieve as large values of $\mathfrak p$ or $\mathfrak q$ as those achieved by a typical agent using $\texttt{aei}_\mathbf{I}$. 

As another more specific example, consider an AEI $\texttt{aei}_\text{const}$ that makes $Y_t$, $Y'_t$, $P_t$, $R_t$, and $\widebar A_t$ constant and only allows the agent to select one action (notice that we have not assumed that the supports of $A_t$ and $\widebar A_t$ are the same
). For all performance objectives $\mathfrak p$, $\mathfrak p(\cdot,\texttt{aei}_\text{const})$ is a constant function, since $\widebar A_t$ being a constant removes the agent's influence on the base environment. Notice that $M_t$ is the only random variable in the stochastic process that the RL algorithm can influence when using $\texttt{aei}_\text{const}$. So, for all qualia objectives $\mathfrak q$ that only depend on the agent's external behavior---qualia objectives that do not depend on the distribution of the agent memory $M_t$---$\mathfrak q(\cdot, \texttt{aei}_\text{const})$ is also a constant function. 

Notice that $\mathfrak p(\cdot,\texttt{aei}_\text{const})$ and $\mathfrak q(\cdot, \texttt{aei}_\text{const})$ being constant functions implies that $\texttt{aei}_\text{const}$ is an objective-aligning AEI. So, $\texttt{aei}_\text{const}$ is an objective-aligning AEI for all $\mathfrak p$ and many $\mathfrak q$. However, it is likely to result in particularly poor values of the performance objective since the same base action $\widebar A_t$ is always selected. Similarly, this AEI is likely to result in poor values of the qualia objective, although this depends on the specific definition of $\mathfrak q$.

These examples highlight that using an objective-aligning AEI will not necessarily improve performance or the quality of the agent's qualia. So, although finding AEIs that align the objectives can be possible, they are not necessarily beneficial or effective for qualia optimization. Note, however, that finding objective-aligning AEIs might be an effective strategy for qualia optimization in a different problem setting (e.g., with additional constraints placed on the set of AEIs under consideration). 

\subsection{Algorithmic Improvements}

Since constructing AEIs that align the objective functions does not necessarily improve the quality of the agent's qualia or its performance (relative to if it were to use the identity AEI), we now focus on strategies for directly improving the quality of the agent's qualia without altering the agent's performance and without necessarily aligning the two objectives. That is, we desire strategies for taking an initial RL algorithm $\texttt{alg}_0$, which we call the \emph{base algorithm}, and constructing an AEI $\texttt{aei}'$ and RL algorithm $\texttt{alg}'$ such that 
\begin{gather}
    \mathfrak q(\texttt{alg}',\texttt{aei}') > \mathfrak q(\texttt{alg}_0,\texttt{aei}_\mathbf{I})\\\nonumber\text{and}\\ \mathfrak p(\texttt{alg}',\texttt{aei}') \geq \mathfrak p(\texttt{alg}_0,\texttt{aei}_\mathbf{I}),
\end{gather}
even though $\mathfrak p$ and $\mathfrak q$ are \emph{not} necessarily aligned under $\texttt{aei}_\mathbf{I}$ or $\texttt{aei}'$. 

A natural way to enhance the quality of an agent's qualia is to modify the agent's learning procedure itself, without necessarily changing the agent's perceptions via a (non-identity) AEI. That is, changing $\texttt{alg}_0$ to some improved $\texttt{alg}'$ while keeping $\texttt{aei}'=\texttt{aei}_\textbf{I}$. In the reward-qualia setting, changes to the RL algorithm that improve its performance could also improve the quality of the agent's qualia. For example, if the agent's performance objective and qualia objective are the same, then simply refining hyperparameters (e.g., step sizes or exploration rates) might improve both performance and the quality of the agent's qualia. If the two objectives conflict, then multi-objective RL algorithms might allow for a desired balance between prioritizing performance and qualia quality.

Although improving the base algorithm alone can be effective, it generally amounts to refinements that traditional RL research already pursues---creating more effective RL algorithms. Furthermore, this approach does not (on its own) leverage the AEI. Neglecting the AEI may limit how well the qualia objective can be optimized, and contrasts with the illustrative example that motivated our problem formulation (the shuttle box example). In the next section we focus on strategies that primarily leverage the AEI to improve the quality of the agent's qualia without altering performance.

\subsection{Reward Bonuses}
\label{sec:rewardBonuses}

In this section we consider a strategy for creating AEIs that improve the quality of an agent's qualia without altering the agent's performance and without necessarily aligning the two objectives. One obvious strategy in the reward-qualia setting is for the AEI to alter the rewards given to the agent so that they are larger positive values. However, it is well-known that adding a constant to (nearly) every reward can change optimal behavior. That is, consider an AEI that only changes the rewards to $R_t=\widebar R_t + c$, where $c$ is a positive constant called the \emph{reward bonus}. This AEI can significantly alter how the RL agent should select actions to maximize the amount of reward that it receives. 

One example of this occurs when an agent is given a constant negative reward, e.g., $-1$, at each time step until an episode ends. These negative rewards incentivize the agent to end the episode as quickly as possible. If a sufficiently large reward bonus is provided, the rewards could become positive. This would incentivize the agent to keep the episode from ending, since ending the episode would stop it from continuing to receive positive rewards. Hence, adding a reward bonus can significantly alter the rewards from incentivizing the agent to end episodes as quickly as possible to incentivizing the agent to keep episodes from ending for as long as possible.

\textit{However}, this reasoning is only valid when the reward bonus is only added to rewards that occur prior to the end of the episode. In classical episodic RL formulations, each episode is infinitely long. At some point, the environment may reach a state called a \emph{terminal absorbing state}, which it can never leave and wherein all rewards are zero. When the environment enters a terminal absorbing state, the episode has effectively ended even though environment states continue to transition (although these transitions are always self-transitions back to the terminal absorbing state), rewards continue to be generated (although they are all zero), and actions continue to be selected (although they are inconsequential).\footnote{In some formulations, the set of admissible actions is also restricted to a single action in terminal absorbing states to further emphasize that not only are there no consequential decisions for the agent to make once a terminal absorbing state is reached---there are no decisions at all for the agent to make.} In such formulations, adding a reward bonus, $c$, to \emph{all} rewards, including those that occur when the environment is in a terminal absorbing state, does \emph{not} change how the rewards incentivize the agent (under mild assumptions to ensure that discounted returns remain finite).\footnote{For example, this result holds if the rewards $R_t$ are bounded and $\gamma_{\mathfrak q} \in [0,1)$.} Rather, adding $c$ to all rewards increases the discounted return (with discount parameter $\gamma_{\mathfrak q}$) of all episodes by 
\begin{align}
    \label{eq:returnShift}
    \sum_{t=0}^\infty \gamma_{\mathfrak q}^t c
    =&\frac{c}{1-\gamma_{\mathfrak q}}.
\end{align}
So, adding a reward bonus $c$ to all rewards, including those that occur when the environment is in (and remains in) a terminal absorbing state, shifts all discounted returns by a constant value and therefore does not change the ordering of policies with respect to the expected discounted returns that they induce.\footnote{A similar result holds if a constant $c$ is added to the first $k$ rewards, where $k$ is also a constant.}

This strategy of adding a reward bonus to \emph{all} rewards (including those after an episode terminates) cannot be directly implemented in our setting because we do not model each finite-length episode as an infinite sequence. However, it can still be implemented in our setting in two steps. First, the AEI adds $c$ to all but the last reward in each episode by setting $R_t=\widebar R_t + c$ when $P_t \neq p_\infty$. Second, when $P_t=p_\infty$, the AEI defines $R_t$ to include not just the one reward bonus from time $t$, but the sum of all of the reward bonuses that would be provided at or after time $t$ if the episode were infinitely long. This cumulative bonus is 
\begin{equation}
    \label{eq:finalBonusReward}
    \sum_{k=0}^\infty \gamma_{\mathfrak q}^k c =\frac{c}{1-\gamma_{\mathfrak q}},
\end{equation}
and so $R_t=\widebar R_t + c(1-\gamma_{\mathfrak q})^{-1}$ when $P_t = p_\infty$.

An AEI that implements this change is straightforward to define formally in our setting.
\begin{itemize}
    \item The AEI does not require memory, and so $Y_t$ and $Y'_t$ can be constants. This is implemented by making $d_y(y',\widebar p)$ and $d_{y'}(y,a)$ distributions with support on a single constant value (e.g., $0$) for all $y', \widebar p, y,$ and $a$. 
    \item The AEI defines $f_p$ so that $P_t=\widebar P_t$. That is, for all $y$ and $\widebar p$, $f_p(y,\widebar p)=\widebar p$.
    \item The AEI defines $f_{\widebar a}$ so that $\widebar A_t=A_t$. That is, for all $y'$ and $a$, $f_{\widebar a}(y',a)=a$. 
    \item The AEI defines $f_r$ so that 
    \begin{equation}
        \label{eq:rewardBonusTransform}
        R_t=
        \begin{cases}
            \widebar R_t+c &\mbox{if } P_t\neq p_\infty\\
            \widebar R_t+\frac{c}{1-\gamma_{\mathfrak q}}&\mbox{otherwise.}
        \end{cases}
    \end{equation}
    Although $\widebar{R}_t=f_{\widebar r}(\widebar P_t)$ always, in this particular setting it is also the case that $\widebar{R}_t=f_{\widebar r}(P_t)$ since $P_t=\widebar{P}_t$.
    So, \eqref{eq:rewardBonusTransform} can be implemented as follows: for all $p$, $f_r(p)=f_{\widebar r}(p)+c$ if $p\neq p_\infty$ and $f_r(p)=f_{\widebar r}(p)+c(1-\gamma_{\mathfrak q})^{-1}$ if $p=p_\infty$. 
\end{itemize}
Hereafter, we will refer to this AEI as the \emph{reward bonus AEI} \definesymbol{$\texttt{aei}_c$}{aeic}.

\subsubsection{Impact on Qualia and Performance Objectives} 

Having formally defined the AEI $\texttt{aei}_c$ that implements reward bonuses, we now turn to evaluating the impact this AEI has on the qualia and performance objectives in the reward-qualia setting. First we consider the impact this objective has on the qualia objective. Although the class of qualia objective functions $\mathfrak q$ that falls within the reward-qualia setting is not precisely defined, many such objectives would be increased if all rewards are increased. 

However, even if we focus on a specific reward-qualia definition of $\mathfrak q$ like that in \eqref{eq:discountedReturnQualiaDefn} (with $\gamma_{\mathfrak q} \in [0,1)$ to ensure that $(1-\gamma_{\mathfrak q})^{-1}$ is defined), $\texttt{aei}_c$ does \emph{not} necessarily ensure that $\mathfrak q(\texttt{alg}_0, \texttt{aei}_c) > \mathfrak q(\texttt{alg}_0,\texttt{aei}_\mathbf{I})$ or even that $\mathfrak q(\texttt{alg}_0, \texttt{aei}_c) \geq \mathfrak q(\texttt{alg}_0,\texttt{aei}_\mathbf{I})$. This is because using $\texttt{aei}_c$ does more than just inflate the rewards---changing the rewards can also change how the agent (using $\texttt{alg}_0$) learns and selects actions. If the larger rewards hinder the agent's learning, it could cause the agent to select worse actions that result in smaller rewards. This is not an unlikely scenario if the hyperparameters of $\texttt{alg}_0$ are tuned so that it is effective for the base environment (with the identity AEI), since these hyperparameters may not be effective when using $\texttt{aei}_c$. For example, as $c$ increases, it is likely that a step size of the BAC algorithm would need to be decreased for BAC to remain equally effective.\footnote{As one even more precise example, if BAC uses a VFA parameterization that is linear with respect to the weights and the VFA weights are initialized to zero, then \emph{multiplying} all rewards by a positive constant is approximately equivalent to dividing the step size $\beta$ by that same constant (the equivalence is only approximate because the VFA weights $W_t$ and TD error $\Delta_t$ would also be inflated by the same constant). If all rewards are non-negative and the constant is greater than one, this provides one example where an optimal step size for BAC may need to be decreased in order to maintain performance if rewards are increased.} 

So, switching from $\texttt{aei}_\mathbf{I}$ to $\texttt{aei}_c$ can result in competing forces on the rewards: $\texttt{aei}_c$ increases the rewards, but the corresponding impact on the RL algorithm's behavior could decrease rewards. We consider two ways to ensure that reward bonuses are effective for the reward-qualia setting despite these competing forces. \textbf{First}, if the reward bonus $c$ is sufficiently large---for example, so large that the smallest reward with the reward bonus is positive and greater than the largest possible return without the reward bonus---then using $\texttt{aei}_c$ will always increase $R_t$ for all times $t$ and $G_i$ for all episodes $i$ regardless of the agent's behavior. We do not focus on this strategy because it does not ensure that the performance objective increases or remains unchanged, since the performance objective depends on the base rewards $\widebar R_t$ that do not include the reward bonuses. 

\textbf{Second}, recall from our example problem formulations that we aim to optimize both the AEI and the RL algorithm, and so we can modify the base RL algorithm $\texttt{alg}_0$ to ensure that its learning is not hindered by the reward bonuses. To achieve this, let \definesymbol{$\texttt{alg}_{-c}$}{algc} denote a copy of $\texttt{alg}_0$ that subtracts the appropriate bonus ($c$ if $P_t \neq p_\infty$ and $c(1-\gamma_{\mathfrak q})^{-1}$ otherwise) from the agent's reward before using it. In this way, $\texttt{alg}_{-c}$ undoes the transformations made by the AEI $\texttt{aei}_c$, ensuring that learning is effectively the same as when $\texttt{alg}_0$ is used with $\texttt{aei}_{\mathbf I}$.

Using $\texttt{alg}_{-c}$ with $\texttt{aei}_c$ equates to adding a reward bonus to the rewards before they are provided to the agent, while changing the agent so that it subtracts the same reward bonus before using the rewards. This change will not impact the agent's behavior since the AEI's changes are canceled out by the changes to the RL algorithm.\footnote{Notice that this strategy remains effective even if the reward bonus does not differ when $P_t=p_\infty$ since $\texttt{alg}_{-c}$ always cancels the reward bonuses prior to their influencing the agent's behavior.} Although this is desirable at the moment, enabling further discussion of how reward-qualia optimization can be achieved, it raises deeper questions about whether such ``changes'' are at all meaningful---questions that are discussed in Sections \ref{sec:reconsideringAEI} and \ref{sec:agentBounday}.

Recall that our goal is to find an AEI-algorithm pair that increases the qualia objective while not decreasing the performance objective. The AEI $\texttt{aei}_c$ and algorithm $\texttt{alg}_{-c}$ achieve this for many reward-qualia settings---those where increasing all agent rewards while leaving all other random variables unchanged is sufficient to increase the value of the qualia objective function. Furthermore, many reward-qualia objectives may be affine and increasing with respect to $c$, which implies that the qualia objective can be inflated by arbitrarily large amounts by increasing $c$. To make this clear, we analyze one such qualia objective function $\mathfrak q$.

\subsubsection{Specific Example where Reward Bonuses are Effective}
\label{sec:SpecificExampleWhere}

Consider the use of $\texttt{aei}_c$ with $\texttt{alg}_{-c}$ for some $c > 0$ and in the same setting from earlier where 
\begin{equation}
    \mathfrak q(\texttt{alg},\texttt{aei}) = \mathbf{E}\left [\sum_{i,t} \gamma_{\mathfrak q}^{\operatorname{dur}(t)} R_t \right ].
\end{equation}
We restrict our attention to settings where $\gamma_{\mathfrak q} \in [0,1)$, $R_t$ is bounded, and the horizon is finite so that $(1-\gamma_{\mathfrak q})^{-1}$ and the expected values of all discounted returns are always defined and finite.

We begin with the (unproven) observation that using $\texttt{aei}_c$ with $\texttt{alg}_{-c}$ only changes the random variable $R_t$---the distributions of all other random variables, $X_t, \widebar P_t, Y_t, P_t, M_t, A_t, Y'_t, \widebar A_t,$ and $\widebar R_t$, remain unchanged. This observation implies that 
\begin{equation}
    \mathfrak p(\texttt{alg}_{-c}, \texttt{aei}_c) = \mathfrak p(\texttt{alg}_0, \texttt{aei}_{\mathbf{I}}),
\end{equation}
since the values output by $\mathfrak p$ can be expressed as a function of the joint distribution of $X_t$ and $\widebar A_t$ for all $t$, which is unchanged. 

In Appendix \ref{app:RewardBonusProof} we show that in this setting 
\begin{equation}
    \mathfrak q(\texttt{alg}_{-c},\texttt{aei}_c) = \mathfrak q(\texttt{alg}_0, \texttt{aei}_{\mathbf{I}}) + c \frac{i_\text{max}}{1-\gamma_{\mathfrak q}}. 
\end{equation} 
Since $i_\text{max}(1-\gamma_{\mathfrak q})^{-1}$ is a positive constant, this implies that
\begin{gather}
    \mathfrak q(\texttt{alg}_{-c}, \texttt{aei}_c) > \mathfrak q(\texttt{alg}_0,\texttt{aei}_{\textbf{I}}),
\end{gather}
and furthermore that increasing $c$ can inflate the qualia objective value arbitrarily. This establishes the desired result: that in this specific example setting $\texttt{aei}_c$ and $\texttt{alg}_{-c}$ increase the value of the qualia objective (arbitrarily) without changing the value of the performance objective. 

\subsection{Reward-Qualia Conclusion}

In summary, in some natural formulations of reward-qualia optimization, simply adding a constant reward bonus at each time (perhaps with a specific larger bonus upon episode termination) could be an effective strategy. If there are concerns that an agent's performance would suffer as a result, the RL algorithm can be modified to undo the AEI's transformation of the rewards. This results in assurance of no change to the value of the performance objective, $\mathfrak p$. However, this is also concerning, as it results in no actual change to the agent's behavior, raising philosophical questions about whether this could reasonably be expected to have any impact on the quality of the agent's qualia.

Also, notice that we have only considered a small number of initial settings within the broader reward-qualia optimization setting and have focused on one strategy for reward-qualia optimization in these settings. There may be other settings within the reward-qualia optimization setting that are of more interest, and there are many other strategies for reward-qualia optimization, even in the settings that we considered. However, many of the other possible settings and strategies encounter the same challenges and concerns that are discussed in the following sections.

\section{Exploitable Qualia Objectives}
\label{sec:reconsideringAEI}

The initial exploration of the reward-qualia setting produced concepts and insights that apply to many qualia optimization settings. For example, while objective alignment is a natural research direction to consider, in our setting it alone does not resolve the trade-off between qualia optimization and performance optimization. As another example, after creating an AEI that improves the quality of an agent's qualia, we were faced with the challenge of ensuring that the AEI will not decrease performance. This resulted in the observation that performance can sometimes be left entirely unchanged by modifying the RL algorithm to undo the AEI's transformations. Before exploring qualia optimization settings beyond the reward-qualia setting, we explore additional concepts that are evident from the reward-qualia setting, but which apply to and inform a wider range of qualia optimization settings, starting with the concept of exploitable qualia objective functions. 

\subsection{Example Unreasonable Qualia Optimization Solution}

First we provide additional context for why $\texttt{aei}_c$ and $\texttt{alg}_{-c}$ from the reward-qualia setting might be considered to be an unreasonable solution. This is important beyond the reward-qualia setting because the general strategy of modifying RL algorithms to undo or invert the transformations made by the AEI could apply to a variety of settings, enabling qualia improvement without changing performance. Furthermore, if $\texttt{aei}_c$ and $\texttt{alg}_{-c}$ represent an unreasonable solution to a qualia optimization problem, it suggests that the problem formulation may be unreasonable.

Consider how $\texttt{aei}_c$ and $\texttt{alg}_{-c}$ could be implemented on a digital computer. Within the computer, numbers like rewards may be represented using a floating point binary format, with rules governing which bits encode the mantissa, which bits encode the exponent, whether the bits are big-endian or little-endian, etc. One way to implement the addition and subsequent subtraction of a constant $c$ from the rewards would be for the computer to store the base environment reward $\widebar R_t$ in a register using a floating point representation and to then add $c$ to the value in this register to obtain $R_t$.\footnote{For simplicity, we focus on times $t$ where $P_t \neq p_\infty$ so that the reward bonus is always $c$. However, the ideas discussed here extend to the full $\texttt{aei}_c$ and $\texttt{alg}_{-c}$ which use a reward bonus of $c(1-\gamma_{\mathfrak q})^{-1}$ when $P_t = p_\infty$.} The algorithm $\texttt{alg}_{-c}$ subtracts $c$ from the value in this register to reconstruct $\widebar R_t$, and then executes the updates specified by the base algorithm $\texttt{alg}_0$ using these reconstructed rewards. While this process of adding $c$ to the value in a register and then immediately subtracting $c$ from that value seems unlikely to be a meaningful change, it at least corresponds to a change to the physical system.

However, this same algorithmic change could be implemented by leaving the software and hardware entirely unchanged, and instead changing how we interpret some sequences of bits as floating point numbers. That is, any time that the computer stores $R_t$ in a register or other form of digital memory, we can define it to be using an encoding that implicitly adds $c$ to the value relative to the representation used to store typical floating point numbers (like $\widebar R_t$). Consider an example of a computer that uses the currently common 32-bit IEEE Standard for Floating-Point Arithmetic (IEEE 754) to encode real numbers. To store a value of $2.6970698081234434654 \times 10^{23}$, this computer would store the 32-bit sequence \texttt{0110 0110 0110 0100 0111 0011 0110 0001}. However, the computer does not necessarily store anything indicating how these bits should be interpreted. They could also be interpreted as a sequence of four ASCII characters---specifically the sequence ``asdf''. This change to how bit sequences are interpreted is common in computing and is referred to as \emph{casting}. 

When the computer stores $R_t$, we can define it to be using a different floating point encoding, such that the bits \texttt{0110 0110 0110 0100 0111 0011 0110 0001} encode $c + 2.6970698081234434654 \times 10^{23}$. We call this encoding, which adds $c$ to the value relative to IEEE 754, \emph{IEEE 754$_c$}. By using IEEE 754$_c$ for $R_t$, the exact same bit sequence that represents $\widebar R_t$ (using IEEE 754) can be used to represent $R_t$. With this change of how bit sequences are interpreted as floating point numbers, the AEI's conversion of $\widebar R_t$ into $R_t$ does not require any changes to the bit sequence.

Next consider how the RL algorithm $\texttt{alg}_{-c}$ reads from the register storing $R_t$. The RL algorithm $\texttt{alg}_0$ may include instructions that operate on the value in the register storing $R_t$. For example, if the computer uses the x86 instruction set architecture (including Streaming SIMD Extensions), $\texttt{alg}_0$ may include instructions like \texttt{ADDSS} (add scalar single-precision), \texttt{MULSS} (multiply scalar single-precision), and \texttt{MOVSS} (move scalar single-precision) with $R_t$ as the source operand.\footnote{For simplicity of this argument, imagine that $\texttt{alg}_0$ does not use the location of $R_t$ as a destination operand. For example, there could be a special register reserved for $R_t$, which the AEI writes to and which the RL algorithm only reads from.} To implement $\texttt{alg}_{-c}$, we can add new instructions to the instruction set architecture that modify each of these instructions, replacing each of the original instructions within $\texttt{alg}_0$ that have $R_t$ as a source operand with our corresponding new instructions to obtain an implementation of $\texttt{alg}_{-c}$. The new instructions are designed to read a source operand in IEEE 754$_c$, subtract $c$ from the value, apply the original instruction, and write the output using IEEE 754. 

Notice, however, that the hardware that implements these new instructions is exactly the same as hardware that implements the original instructions. That is, an original instruction interpreting an input operand using IEEE 754 is precisely the same as the corresponding modified instruction that interprets the input operand as using IEEE 754$_c$, subtracts $c$ from the value prior to performing the original instruction, and writes the output using IEEE 754. So, just like how no changes to the physical system were required to implement the addition of $c$, no changes are required to implement the subtraction of $c$. Together, these properties mean that an existing physical implementation of $\texttt{alg}_0$ and $\texttt{aei}_\textbf{I}$ is already an implementation of $\texttt{alg}_{-c}$ and $\texttt{aei}_c$. The only difference is how bit sequences are interpreted.\footnote{To extend this to the full reward bonus AEI and RL algorithm, one might use different registers to store $R_t$ when $P_t=p_\infty$ and $P_t\neq p_\infty$, and these registers might use floating point representations that add different constants to the value relative to IEEE 754.}

Recall our previous discussion of how random variables (e.g., $R_t$) correspond to underlying physical properties (e.g., $\Phi_{R_t}$), and how the correspondence is characterized by a representation function (e.g., $\rho_{R_t}$). The issue with $\texttt{aei}_c$ and $\texttt{alg}_{-c}$ described here can be expressed using these terms. First, notice that the physical properties (e.g., bit sequence) that correspond to the base environment reward $\Phi_{\widebar R_t}$ can be the same as the physical properties that correspond to the agent reward $\Phi_{R_t}$. That is, $\Phi_{R_t}=\Phi_{\widebar R_t}$. Using $\texttt{aei}_\textbf{I}$ and $\texttt{alg}_0$ corresponds to using the same representation functions for $\widebar R_t$ and $R_t$, i.e., $\rho_{R_t}=\rho_{\widebar R_t}$. However, the same physical system can be modeled as a different AIEP by only changing the representation function of $R_t$ such that $\rho_{R_t}(\Phi) = \rho_{\widebar R_t}(\Phi)+c$.\footnote{We write $\Phi$ rather than $\Phi_{R_t}$ or $\Phi_{\widebar R_t}$ to emphasize that here they correspond to the same physical properties.} The resulting AIEP corresponds to using $\texttt{aei}_c$ and $\texttt{alg}_{-c}$. This means that the same physical system can be modeled as two different AIEPs (one with reward bonuses and the other without). 

So, by merely redefining how we interpret the meaning of sequences of bits within different registers in a computer, we can view the same hardware and software that implement RL algorithm \texttt{alg} with no AEI (equivalently, AEI $\texttt{aei}_\textbf{I}$) as also being an implementation of the algorithm $\texttt{alg}_{-c}$ with AEI $\texttt{aei}_c$. If the qualia objective assigns different values to these two perspectives, it means that the quality of the agent's qualia depends on how we (an external observer of a computer running software) ascribe meaning to sequences of bits within the computer. If one views this as unreasonable, it suggests that there is a flaw within the problem formulation.

\PackageWarning{manual}{Warning: Figure occurs prior to reference.}

\begin{figure*}[htbp]
    \centering
    \includegraphics[width=\textwidth]{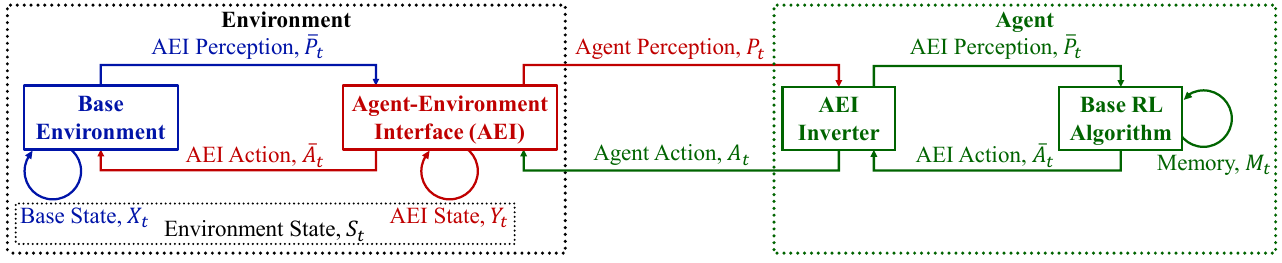}
    \caption{A revision of Figure \ref{fig:AEI} to depict the concept of AEI inversion. The AEI inverter undoes the AEI's transformations to the agent perceptions to retrieve the AEI perceptions, and pre-transforms the base RL algorithm's actions $\protect\widebar A_t$ so that the AEI's subsequent transformations result in agent actions $A_t$ that the AEI transforms back into $\protect\widebar{A}_t$.}
    \label{fig:AEI_Inversion}
\end{figure*}

\subsection{AEI Inversion}
\label{sec:AEI_Inversion}

The concept of an RL algorithm that is modified to undo or negate the transformations made by the AEI extends beyond  $\texttt{aei}_c$ and $\texttt{alg}_{-c}$, and beyond the reward-qualia setting. One way that the RL algorithm can undo or negate the transformations made by the AEI is for it to include a mechanism that explicitly inverts the transformations made by the AEI---a mechanism that we call an \emph{AEI inverter}. We call an RL algorithm that undoes the transformations of AEI \texttt{aei} and then applies a \emph{base RL algorithm} \definesymbol{\texttt{alg}}{alg2}, the \emph{inverse RL algorithm} \definesymbol{$\texttt{alg}^{-\text{aei}}$}{invRLalg}. Figure \ref{fig:AEI_Inversion} depicts an agent using an inverse RL algorithm. To ground these abstract concepts, note that in the case of $\texttt{aei}_c$ and $\texttt{alg}_{-c}$, the AEI adds $c$ to the rewards, the AEI inverter subtracts $c$ from the rewards, and $\texttt{alg}_{-c}$ is the inverse algorithm $\texttt{alg}^{-\texttt{aei}_c}$. 

We now define the AEI inverter more formally, characterizing it with two functions: \definesymbol{$g_{\widebar p}$}{gbarp}, which we call the \emph{perception inverter} and \definesymbol{$g_a$}{ga}, which we call the \emph{action pretransformer}. For simplicity, we only consider state-free AEI inverters (i.e., AEI inverters that do not have their own memory). When the agent receives an agent perception $p$, the AEI inverter transforms this agent perception into the perception $\widebar p$ that is provided to the base RL algorithm. The perception inverter characterizes this transformation according to the expression $\widebar p=g_{\widebar p}(p)$. When the base RL algorithm selects an action $\widebar a$, the AEI inverter transforms it into the agent action $a$ according to the expression $a = g_a(\widebar a)$. 

Furthermore, $g_{\widebar p}$ and $g_a$ must satisfy the following conditions. First, the perception inverter must satisfy the following expression for all times $t$, which ensures that the AEI inverter properly reconstructs $\widebar P_t$ from $P_t$:
\begin{equation}
    \label{eq:inverterP}
    \widebar P_t = g_{\widebar p}(P_t).
\end{equation}
Second, at every time $t$, the action pretransformer transforms the action $\widebar a$ selected by the base RL algorithm into an action $a$ such that the subsequent transformation performed by the AEI makes $\widebar a$ the AEI action. This is more complicated to express formally because it must account for how the agent action transforms the intermediate AEI state, and how the resulting intermediate AEI state then influences the AEI action. Even given $Y_t$ and $A_t$, the intermediate AEI state $Y'_t$ can be stochastic, with conditional distribution $d_{y'}(Y_t,A_t)$. To ensure that the action selected by the base algorithm is always equal to the AEI action, we will consider all possible values $y'$ of $Y'_t$, i.e., all $y' \in \operatorname{supp}(d_{y'}(Y_t,A_t))$.\footnote{Recall from Appendix \ref{app:notation} that for any distribution $d$, $\operatorname{supp}(d)$ denotes the support of $d$.} The constraint on $g_a$ is thus that for all times $t$:
\begin{equation}
    \label{eq:inverterA}
    \underbrace{\forall y' \in \operatorname{supp}(d_{y'}(Y_t, g_a(\widebar A_t)))}_\text{For all possible $Y'_t$},\,\,\underbrace{f_{\widebar a}(y',g_a(\widebar A_t))=\widebar A_t.}_\text{The AEI selects the base algorithm's action}
\end{equation}

When $g_{\widebar p}$ and $g_a$ that satisfy these two conditions do not both exist, no AEI inverter exists. When such $g_{\widebar p}$ and $g_a$ exist, we say that the AEI is \emph{invertible}. Note that the invertibility of the AEI is with respect to some implicit base environment and base RL algorithm. For example, if there are AEI perceptions $\widebar p$ and agent perceptions $p$ that can occur when some RL algorithms are used, but which cannot occur when the base algorithm in question is used, then the perception inverter need not satisfy $\widebar p=g_{\widebar p}(p)$.

\subsection{Inversion-Exploitable and Inversion-Robust Qualia Objectives}

We now generalize the issues previously discussed for the specific invertible AEI $\texttt{aei}_c$ and its inverse algorithm $\texttt{alg}_{-c}$, extending the concepts to invertible AEIs and inverse algorithms in general. The set of qualia objective functions can be partitioned into two sets:
\begin{enumerate}
    \item \textbf{\emph{Inversion-exploitable} qualia objectives} are those $\mathfrak q$ for which there exists a state-free AEI $\texttt{aei}$ that is invertible with respect to a base RL algorithm $\texttt{alg}$, and where the use of the AEI with the inverse RL algorithm $\texttt{alg}^{-\texttt{aei}}$ results in a change to the value of the qualia objective function. That is, there exists an RL algorithm $\texttt{alg}$ and a corresponding invertible state-free AEI $\texttt{aei}$ such that\footnote{In order to allow for later problem formulations where the set of RL algorithms and/or AEIs are constrained, the following expression should be proceeded by the additional qualification: ``$\mathfrak q(\texttt{alg}^{-\texttt{aei}}, \texttt{aei})$ and $\mathfrak q(\texttt{alg}, \texttt{aei}_\textbf{I})$ are defined and''. We relegate this condition to a footnote because it is not well-motivated or clear at this point.} 
    \begin{equation}
        \label{eq:exploitableQ}
        \mathfrak q(\texttt{alg}^{-\texttt{aei}}, \texttt{aei}) \neq \mathfrak q(\texttt{alg}, \texttt{aei}_\textbf{I}).
    \end{equation}
    \item \textbf{\emph{Inversion-robust} qualia objectives} are qualia objective functions that are not inversion exploitable. That is, using a state-free invertible AEI and inverse RL algorithm cannot change the value of the qualia objective function.
\end{enumerate}

Notice that every qualia objective function is either \emph{inversion exploitable} or \emph{inversion robust}, but not both. Hence, these two categories provide a partitioning of the set of all qualia objective functions. 

\subsubsection{Inversion-Exploitable Objectives May Be Unreasonable}

As with $\texttt{aei}_c$ and $\texttt{alg}_{-c}$, AEIs and their inverse algorithms pose a problem for inversion-exploitable qualia objectives---they suggest that the quality of an agent's qualia depend on how values are ascribed to the physical properties within the agent-environment system, not just the physical properties themselves. In particular, we can view the physical properties that formerly encoded $\widebar{P}_t$ as instead encoding $P_t$, and can similarly reinterpret the same physical properties used for $\widebar{A}_t$ as $A_t$. Thus, the AEI makes no transformations to the physical system, but rather it adopts a different mapping from the same physical properties to the values of perceptions and actions. Meanwhile, the AEI inverter undoes this change of mapping, restoring the original interpretation of physical properties as values of perceptions and actions. Although such an AEI and AEI inverter correspond to changes to how we ascribe values to physical properties within the physical agent-environment system, they correspond to no change to the physical properties themselves. Inversion-exploitable qualia objectives are those that assign different values to the quality of the agent's qualia depending on whether the same physical system is interpreted as including the AEI and AEI inverter or not.

To be more precise, recall our earlier discussion in Section \ref{sec:RLBackgroundAndSetting} of how random variables relate to the physical world. Specifically, recall that $\Phi_{\widebar P_t}$ determines the value of $\widebar P_t$ and the value of $\widebar P_t$ determines $\Phi_{\widebar P_t}$. That is, there is an invertible representation function $\rho_{\widebar P_t}$ such that $\widebar P_t = \rho_{\widebar P_t}(\Phi_{\widebar P_t})$. Intuitively, $\rho_{\widebar P_t}$ ``reads out'' the value of the random variable $\widebar P_t$ from the underlying physical properties $\Phi_{\widebar P_t}$. For example, in an idealized physical world where $\Phi_{\widebar P_t}$ corresponds to a sequence of 32 bits in a digital computer, IEEE 754 and IEEE 754$_c$ correspond to different representation functions $\rho_{\widebar P_t}$---different ways of interpreting how underlying physical properties induce values of $\widebar P_t$. Notice that $\rho_{\widebar P_t}$ is not part of a physical system, but rather part of how the physical system is modeled as an AEP or AIEP. 

A physical system that we interpret as an agent implementing a base RL algorithm interacting with a base environment can also be interpreted as an agent implementing the inverse RL algorithm interacting with the base environment through a state-free AEI (when the inverse RL algorithm for the state-free AEI exists). That is, the AEI makes no physical changes to the system, but changes the representation function used for agent perceptions $P_t$ to be $f_p(0,\cdot)\circ \rho_{\widebar P_t}$, where $\circ$ denotes function composition.\footnote{Recall that we have restricted our consideration to state-free AEIs, and so without loss of generality we assume that $Y_t=0$ and $Y'_t=0$ for all $t$. Also recall that $f_p$ characterizes how the AEI generates agent perceptions according to the expression $P_t = f_p(Y_t, \widebar P_t)$.} That is, $\Phi_{\widebar P_t}$, the physical properties representing $\widebar P_t$ are the exact same physical properties that represent $P_t$, but with representation function $f_p(0,\cdot)\circ \rho_{\widebar P_t}$ so that $P_t=f_p(0,\rho_{\widebar P_t}(\Phi_{\widebar P_t}))$. 

So far we have argued that the physical properties $\Phi_{\widebar P_t}$ corresponding to the AEI perception $\widebar P_t$ are sufficient to represent the agent perception $P_t$. However, we must consider the possibility that even though $\widebar P_t$ determines $P_t$, $P_t$ might not determine $\widebar P_t$. Intuitively, $\widebar P_t$ might contain more information than $P_t$, in which case $\Phi_{\widebar P_t}$ would contain more information than $P_t$, meaning that it could not be the physical properties corresponding to $P_t$ (recall that the physical properties corresponding to a random variable must determine the random variable \emph{and} the random variable must determine the physical properties). That is, the representation function $f_p(0,\cdot)\circ \rho_{\widebar P_t}$ for $P_t$ may not be invertible, which would be a problem since representation functions must, by definition, be invertible. However, the existence of the AEI inverter (specifically, $g_{\widebar p}$) ensures that this is not the case---$f_p(0,\cdot)\circ \rho_{\widebar P_t}$ is necessarily invertible since 
\begin{align}
    \rho_{\widebar P_t}(\Phi_{\widebar P_t})=&\widebar P_t\\
    =&g_{\widebar p}(P_t)\\
    =&g_{\widebar p}(f_p(0,\rho_{\widebar P_t}(\Phi_{\widebar P_t})).
\end{align}
Recall that representation functions (including $\rho_{\widebar P_t}$) are invertible, and so
\begin{equation}
\label{eq:oneInProofInversionExploit}
\Phi_{\widebar P_t} = \rho_{\widebar P_t}^{-1}\left ( g_{\widebar p}(f_p(0,\rho_{\widebar P_t}(\Phi_{\widebar P_t})) \right ),
\end{equation}
which means that $\rho_{\widebar P_t}^{-1}\circ g_{\widebar p}$ is the inverse of $f_p(0,\cdot)\circ \rho_{\widebar P_t}$. Hence $f_p(0,\cdot)\circ \rho_{\widebar P_t}$ is a valid representation function for $P_t$ and so the exact same physical properties $\Phi_{\widebar P_t}$ that correspond to $\widebar P_t$ (via representation function $\rho_{\widebar P_t}$) can be the physical properties $\Phi_{P_t}$ that correspond to $P_t$ (via representation function $f_p(0,\cdot)\circ \rho_{\widebar P_t}$). 

So, when $\Phi_{P_t}=\Phi_{\widebar P_t}$, to retrieve the AEI perception $\widebar P_t$, the AEI inverter need only revert to using the original representation function $\rho_{\widebar P_t}$ when interpreting the physical properties $\Phi_{P_t}=\Phi_{\widebar P_t}$ as the perception fed to the base RL algorithm. This same argument applies to $A_t$ (leveraging the existence of $g_a$ rather than $g_{\widebar p}$)---the same physical properties that correspond to $A_t$ can correspond to $\widebar A_t$. So, the AEI and AEI inverter need only change the representation functions used for perceptions and actions in order to convert a system implementing the base algorithm (without an AEI or with the identity AEI) into an implementation of the AEI and inverse algorithm. Since the representation functions are implicit (the functions themselves are not part of the physical system), this corresponds to no changes to the physical system. 

An inversion-exploitable qualia objective is one that can assign different values to the quality of the agent's qualia for these two interpretations of the same physical system. This is problematic because it means that our external interpretation of the same physical agent-environment system---how we ascribe values like real numbers to physical properties---can alter the quality of the agent's qualia. If this property is unreasonable, it suggests that the problem formulation should be designed to ensure that it does not occur, i.e., inversion-exploitable qualia objectives should be avoided. 

\subsubsection{Why Statefulness Changes Things}

Notice that the definition of inversion-exploitable qualia objectives requires the problematic AEI to be state-free. This is important because there can exist stateful (i.e., not state-free) AEIs and inverse RL algorithms that \emph{do} necessitate changes to the physical system to implement.

As an example, consider an AEI that defines the agent perception to be the past \emph{two} AEI perceptions: $P_t=(\widebar P_t, \widebar P_{t-1})$. In general (e.g., barring degenerate cases where perceptions are constant) such an AEI requires state. For example, the AEI state could store the previous AEI perception: $Y_t=\widebar P_{t-1}$. With $\widebar P_{t-1}$ stored in the AEI's state, $f_p$ can be defined such that $f_p(Y_t, \widebar P_t) = f_p(\widebar P_{t-1}, \widebar P_t) = (\widebar P_{t-1}, \widebar P_t)=P_t.$

In this example, the physical properties $\Phi_{\widebar P_t}$ that correspond to $\widebar P_t$ \emph{cannot} correspond to the physical properties representing $P_t$ (but with a different representation function), since each $\widebar P_t$ corresponds to many possible values of $P_t$, and so the same physical properties cannot be one-to-one with both $\widebar P_t$ and $P_t$. In the previous formal argument, \eqref{eq:oneInProofInversionExploit} would not hold because the AEI state $Y_t$ would not necessarily be zero, and so the representation function $f_p(Y_t,\cdot)\circ \rho_{\widebar P_t}$ would vary with $Y_t$ and hence would not be deterministic.

\subsection{Representation-Exploitable Qualia Objectives}
\label{sec:RepExploit}

So far we have built up intuition for a problem related to the link between random variables and the underlying properties of a physical system, starting with a concrete reward-bonus example and then generalizing the discussion to invertible AEIs and inverse RL algorithms. These examples provide a clear introduction to the issue at hand, but do not capture the full generality of the problem. For example, our discussion so far might suggest adopting definitions of the qualia objective function $\mathfrak q$ that depend only on the agent's memory, since invertible AEIs and inverse RL algorithms do not alter the agent's memory. Hence, $\mathfrak q$ that only depend on the distribution of $M_t$ (for all $t$) are inversion robust. 

However, such qualia objectives do not escape the core underlying problem: that a physical agent-environment system can be modeled as two different AEPs or AIEPs by simply changing the representation functions used for random variables, and this change of representation functions can result in a change to the value of the qualia objective function. That is, under such qualia objectives, changing how we ascribe values (e.g., numbers) to physical properties (e.g., bit sequences), can change the quality of an agent's qualia despite there being no changes to the physical system.

As a concrete example, consider a qualia objective function $\mathfrak q$ where 
\begin{equation}
    \label{eq:earlyTDErrorQualiaObj}
    \mathfrak q(\texttt{alg},\texttt{aei}) = \mathbf{E}\left [\sum_{i=0}^{i_\text{max}-1}\sum_{t=\operatorname{start}(i)+1}^{\operatorname{end}(i)-1} \gamma_{\mathfrak q}^{\operatorname{dur}(t)} \Delta_t \right ],
\end{equation}
where $\Delta_t \in \mathbb R$ is part of the agent's memory $M_t$. For example, BAC algorithms could be modified to explicitly store the temporal difference error $\Delta_t$ in memory, in which case this qualia objective defines the quality of the agent's qualia to be the expected discounted sum of the TD errors the agent accrues. 

Notice that although this qualia objective function is inversion robust, it is not in a more general sense \emph{representation robust}. That is, one could define $\Delta_t$ to include a positive additive bonus at each time $t$, inflating the qualia objective. Any time the value of $\Delta_t$ is referenced, the algorithm could be modified to first subtract the bonus so that there is no effective change to the algorithm's behavior. As before, this entire process can be ``implemented'' via changes to representation functions, and without any change to the physical system. 

We now define the more general concept of \emph{representation-exploitable} and \emph{representation-robust} qualia objectives that captures both inversion-exploitability and the exploitability of \eqref{eq:earlyTDErrorQualiaObj}. For generality, we define these terms for AEPs (the most general environment specification thus far). We call a qualia objective function $\mathfrak q$ \emph{representation exploitable} if the same physical system (characterized by the joint distribution of $\Phi_{S_t}, \Phi_{P_t}, \Phi_{M_t},$ and $\Phi_{A_t}$ for all $t \in \{0,1,\dotsc\}$), but with different representation functions $\rho_{P_t}, \rho_{M_t},$ and $\rho_{A_t}$, can induce different AEPs that result in different values of the qualia objective function. We also define \emph{representation-robust} qualia objective functions to be those that are not representation exploitable. That is, a representation-exploitable qualia objective function is one where different interpretations of physical properties (e.g., bit sequences) as values of random variables (e.g., numbers), can alter the quality of the agent's qualia despite there being no changes to the physical system. Similarly, representation-robust qualia objective functions are in a sense invariant to the choice of representation function. 

To make this definition more formal, first recall the earlier generalization of qualia objective functions that extends them to explicitly depend on the representation functions of $P_t$, $A_t$, and $M_t$. That is, 
\begin{equation}
    \label{eq:lkajwerytlkj}
    \mathfrak q\Big (\operatorname{law}\big((\Phi_{P_t},\Phi_{M_t},\Phi_{A_t})_{t=0}^\infty\big ), (\rho_{P_t}, \rho_{M_t}, \rho_{A_t})_{t=0}^\infty\Big )
\end{equation}
represents the value of the qualia objective function for physical properties $(\Phi_{P_t},\Phi_{M_t},\Phi_{A_t})_{t=0}^\infty$ and representation functions $(\rho_{P_t}, \rho_{M_t}, \rho_{A_t})_{t=0}^\infty$. Changing only the representation functions represents a change to the AEP model of the physical system, but not a change to the physical system itself. Critically, changes to the representation functions in \eqref{eq:lkajwerytlkj}, can change the values of the random variables $P_t, M_t,$ and $A_t$ for all times $t$, but do \emph{not} change the dynamics of the underlying physical system---they do not change the joint distribution of $(\Phi_{P_t},\Phi_{M_t},\Phi_{A_t})_{t=0}^\infty$. So, the physical dynamics are fixed and only the representation functions are varied.

We now more formally state the intuitive idea that representation-exploitable qualia objectives are those that depend on the representation functions and representation-robust qualia objective functions are those that are invariant to the choice of representation functions. 
\begin{enumerate}
    \item \textbf{\textit{Representation-robust} qualia objectives } are $\mathfrak q$ such that, for all physical properties $\Phi_{P_t}, \Phi_{M_t},$ and $\Phi_{A_t}$, all termination-preserving%
        \footnote
        {
            Termination-preserving representation functions are those that preserve episode termination. That is, $\rho'_{S_t}(\Phi_{S_t})=s_\infty$ if and only if  $\rho_{S_t}(\Phi_{S_t})=s_\infty$. Without the restriction to termination-preserving representation functions, qualia objective functions that depend on terms like $\operatorname{end}(i)$ may not be representation-robust. This could potentially be resolved by redefining functions like $\operatorname{end}(i)$ so that they are invariant to the choice of representation functions. For example, instead of $\operatorname{end}(i)$ being defined as 
            \begin{equation}
            \operatorname{end}(i) = \min\{t > \operatorname{start}(i) : S_t = s_\infty\},
            \end{equation}
            it could instead be defined as
            \begin{equation}
            \operatorname{end}(i) = \min\{t > \operatorname{start}(i) : \Phi_{S_t} = \phi_\infty\},
            \end{equation}
            for some terminal physical properties $\phi_\infty$. To avoid the complexity of redefining all such functions in a representation-robust manner, we instead restrict our consideration to termination-preserving representation functions.
        }
    representation functions $\rho_{P_t}, \rho_{M_t}, \rho_{A_t}, \rho'_{P_t}, \rho'_{M_t},$ and $\rho'_{A_t}$, and all $t \in \{0,1,2,\dotsc\}$, if
    \begin{equation}
    \mathfrak q\Big (\operatorname{law}\big((\Phi_{P_t},\Phi_{M_t},\Phi_{A_t})_{t=0}^\infty\big ), (\rho_{P_t}, \rho_{M_t}, \rho_{A_t})_{t=0}^\infty\Big )
    \end{equation}
    and 
    \begin{equation}
    \mathfrak q\Big (\operatorname{law}\big((\Phi_{P_t},\Phi_{M_t},\Phi_{A_t})_{t=0}^\infty\big ), (\rho'_{P_t}, \rho'_{M_t}, \rho'_{A_t})_{t=0}^\infty\Big )
    \end{equation}
    are defined,\footnote{The restriction to cases where both values of the qualia objective function are defined allows for the later consideration of restricted sets of environments, representation functions, and agents.} then
    \begin{align}
    \label{eq:repRobust}
    &\mathfrak q\Big (\operatorname{law}\big((\Phi_{P_t},\Phi_{M_t},\Phi_{A_t})_{t=0}^\infty\big ), (\rho_{P_t}, \rho_{M_t}, \rho_{A_t})_{t=0}^\infty\Big )\\
    = &\mathfrak q\Big (\operatorname{law}\big((\Phi_{P_t},\Phi_{M_t},\Phi_{A_t})_{t=0}^\infty\big ), (\rho'_{P_t}, \rho'_{M_t}, \rho'_{A_t})_{t=0}^\infty\Big ). 
    \end{align}
    \item \textbf{\textit{Representation-exploitable} qualia objectives } are qualia objective functions $\mathfrak q$ that are not representation robust.
\end{enumerate}

Representation‐exploitable qualia objective functions pose a significant challenge because a single physical system can be modeled by two distinct AEPs that differ only in the choice of representation functions for $P_t$, $A_t$, and $M_t$. In such cases, an exploitable qualia objective can assign different values to the agent's experiential quality even though the underlying physical substrate is unchanged. This is problematic if one believes that an agent's qualia depend solely on its underlying physical state---not on arbitrary representational mappings---a perspective that aligns with certain forms of physicalism and computational theories of mind. 

This issue is related to triviality arguments in philosophy of mind. Under a mapping theory of computational implementation---one that mirrors our assumption that each random variable in an AEP is in one‐to‐one correspondence with underlying physical properties---triviality arguments hold that even elementary physical systems (such as a rock, a bucket of water, a wall, or a clock) can be construed as executing arbitrarily complex computations, including those performed by the human brain (see Appendix \ref{app:trivialityArguments}). The key difference, however, is that while triviality arguments focus on the potential complexity of computations that a system might be said to implement, representation exploitability concerns whether the qualitative experience attributed to the agent remains invariant under different mathematical representations. 

\subsection{Strategies Regarding Exploitable Qualia Objectives}
\label{sec:stratRegExploit}

As discussed in the previous subsections, representation-exploitable qualia objectives present a problem. There are at least three possible (not entirely distinct) ways that qualia optimization problems can be formulated to circumvent this problem:
\begin{enumerate}
    \item \textbf{Restrict the focus to inherently representation-robust qualia objective functions.} 
    This approach addresses the issue by excluding those cases in which the qualia objective is affected by arbitrary choices of representation (for example, whether numerical values are encoded using IEEE 754 or an alternative scheme). Such a strategy may be particularly compelling to some proponents of functionalism and computational theories of mind, who maintain that only the underlying computational structure matters for mental states---and therefore for qualia---while the specific physical instantiation and representation does not. From this perspective, if two systems implement the same functional relationships among mental states, then differences in their representational formats (such as the floating point encoding) should not lead to different qualia. This line of reasoning suggests that any reasonable qualia objective function should be applicable to a wide range of RL algorithms and AEIs while remaining representation robust.
    
    The restriction to representation-robust qualia objectives can also be achieved by restricting the set of RL algorithms and AEIs under consideration. The definition of representation-robust qualia objectives indicates that that if $\mathfrak q(\texttt{alg},\texttt{aei})$ is undefined, \texttt{alg} and \texttt{aei} are removed from consideration when determining whether a qualia objective is representation robust. For clarity, we treat these two approaches as distinct approaches: this one, in which the qualia objective function is designed to be inherently representation‐robust, and the next, in which we restrict our analysis to a subset of RL algorithms and AEIs for which the qualia objective is robust. Although the latter can be seen as a special case of the former, distinguishing between them highlights whether robustness is achieved by the objective's definition or by limiting the scope of systems under consideration.
    
    \item \textbf{Restrict the set of AEIs and RL algorithms under consideration.} If the set of considered AEIs and RL algorithms is restricted sufficiently, each physical system under consideration may correspond to a unique AEI and algorithm. That is, the set of AEIs and algorithms could be so significantly restricted that all pairs of an AEI and RL algorithm under consideration necessarily correspond to different underlying physical systems. 
    
    As one example, we could consider only the identity AEI and RL algorithms that correspond to BAC with different positive constants $\eta$ multiplying the TD error: 
    \begin{equation}
        \Delta_t \gets \eta \Big (R_{t} + \gamma v(P_t,W_{t-1}) - v(P_{t-1},W_{t-1})\Big ),
    \end{equation}
    perhaps with constraints on the possible values of $\eta \in \mathbb R_{>0}$.\footnote{We have not proven that this restriction ensures representation-robust qualia objectives. 
    It may be possible that the changes induced by $\eta$ are, in some cases, sufficiently predictable for the difference to be modeled solely as a change to representation functions. Still, this serves as a clarifying example, and we conjecture that there are specific environments and limited sets of values for $\eta$ such that these restrictions are sufficient to ensure representation-robustness.} 
    \begin{enumerate}
        \item \emph{Remove the AEI.} 
        Within this broader approach is a special case of particular interest: when the AEI is restricted to be the identity AEI. This equates to essentially removing the AEI from the formulation, converting AIEPs back to AEPs and AIERPs to AERPs. Although this approach does not resolve the problem of representation-exploitable qualia objectives on its own, it does resolve the issue of inversion-exploitable qualia objectives. Formally, we can view this as requiring the AEI to be $\texttt{aei}_{\mathbf I}$ by only defining $\mathfrak q(\texttt{alg},\texttt{aei})$ when $\texttt{aei}=\texttt{aei}_{\mathbf I}$. This ensures that all $\mathfrak q$ are inversion robust because the AEI inverter for $\texttt{aei}_\mathbf{I}$ is an identity function, and so any base algorithm $\texttt{alg}_0$ with an AEI inverter is precisely the same as the base algorithm $\texttt{alg}_0$ (i.e., $\texttt{alg}_0=\texttt{alg}_0^{-\texttt{aei}_\textbf{I}}$), and so $\mathfrak q(\texttt{alg}_0^{-\texttt{aei}_\textbf{I}}, \texttt{aei}_\textbf{I}) = \mathfrak q(\texttt{alg}_0, \texttt{aei}_\textbf{I})$. Notice that $\mathfrak q$ being inversion robust in this setting relies on the condition added to \eqref{eq:exploitableQ} in a footnote. This approach could resolve the problem of representation-exploitable qualia objectives when combined with other approaches. 
    \end{enumerate}
    \item \textbf{Constrain the allowed representation functions.} 
    If there is one set of ``ground truth'' representation functions (one per random variable), the problem of representation-exploitable qualia objectives can be entirely dismissed---only the qualia value under the ground truth representation for each random variable is of concern (the qualia objective function can be left undefined for other representation functions). Similarly, and somewhat equivalently, the set of allowed representation functions (perhaps more than one per random variable) could be restricted to a set that the qualia objective function is invariant to. This approach of restricting the set of allowed representation functions aligns with philosophy of mind research that suggests that certain types of representations may be critical for mental states, such as analog, digital, iconic, discursive, and symbolic representations \citep{fodor1975language,fodor1988connectionism,barsalou1999perceptual,quilty2020perceptual,clarke2022mapping,maley2024computation}. However, we conjecture that restricting representations to any of these representational types would be insufficient to preclude representation-exploitability. 
\end{enumerate}

\subsection{Representation-Robust Qualia Objectives}
\label{sec:repRobustExist}
Although the first solution---restricting the focus to inherently representation-robust qualia objective functions---is intuitively appealing, it raises the question of whether representation-robust qualia objective functions exist that do not rely on the qualia objective often being undefined. We contend that there are such representation-robust qualia objective functions. 

Information theoretic concepts, reviewed in Appendix \ref{app:infoTheory}, provide useful definitions and properties for constructing representation-robust qualia objectives. Here, we construct one such objective using entropy. Consider a discrete random variable $Z$ within an AEP or AIERP (e.g., $Z$ could correspond to $P_t$ or $R_t$). If the qualia objective function can be expressed as a function of the entropy of $Z$, $H(Z)$, then it is necessarily representation robust. To see why, consider underlying physical properties $\Phi$ and two representation functions $\rho$ and $\rho'$ that result in random variables $Z$ and $Z'$ (i.e., $Z=\rho(\Phi)$ and $Z'=\rho'(\Phi)$). Property \ref{prop:shannonInvariant} in Appendix \ref{app:infoTheory} states that Shannon entropy is invariant to invertible transformations (like $\rho$ and $\rho'$), and directly implies that
\begin{align}
    H(Z)=&H(\rho(\Phi))\\
    =&H(\Phi)\\
    =&H(\rho'(\Phi))\\
    =&H(Z').
\end{align} 
Hence a qualia objective function that only depends on the Shannon entropy of discrete random variables will not vary with the representation functions $\rho_{P_t}$, $\rho_{A_t}$, and $\rho_{M_t}$, and so will be representation robust. 

One example of such a qualia objective function would define the quality of the agent's qualia to be the discounted sum of the entropy of the agent's perceptions:
\begin{equation}
    \label{eq:entropyQualiaObj}
    \mathfrak q(\texttt{alg},\texttt{aei}) = \sum_{i,t} \gamma_{\mathfrak q}^{\operatorname{dur}(t)} H(P_t).
\end{equation}
Note that an expectation is not necessary in the above expression because $H(P_t)$ is not random. 

Although entropy can be effective for constructing representation-robust qualia objectives, it is only effective when applied to discrete random variables. As shown in Property \ref{prop:differentialNotInvariant} in Appendix \ref{app:infoTheory}, differential entropy (a common extension of entropy to continuous random variables) is \emph{not} invariant to invertible transformations. Hence, changing the representation function of a random variable without changing the underlying physical properties can change the differential entropy of the random variable. 

This raises the question: Do there exist functions of distributions (like entropy) that are invariant to invertible transformations even for continuous random variables? We begin with a negative result: If we consider only non-constant functions of a single univariate random variable, Property \ref{prop:univariateNonexistence} in Appendix \ref{app:infoTheory} shows that there do \emph{not} exist any such invariant functions.\footnote{We reiterate that the definitions and properties in Appendix \ref{app:infoTheory} are well-known results in information theory and not contributions of this report.} 

Although representation-robust qualia objective functions cannot be immediately constructed using a function of the distribution of a univariate random variable (e.g., rewards $R_t$ are typically real numbers and hence univariate), the situation improves when considering functions of the distributions of \emph{multiple} random variables (or a single multivariate random variable). Mutual information (see Definitions \ref{def:MI1}--\ref{def:MI3} in Appendix \ref{app:infoTheory}) is a function of the distributions of two random variables, and is invariant to invertible transformations (see Property \ref{prop:MIInvariant} in Appendix \ref{app:infoTheory}). Hence, if the qualia objective function is defined entirely in terms of mutual information computed between pairs or groups of the agent's random variables---for example, the mutual information between a perception at one time and an action at a later time, or between consecutive actions---then, by virtue of mutual information's invariance under invertible transformations, such an objective will be representation‐robust.

A more formal argument that qualia objectives that are functions of the mutual information between random variables in the AEP formulation are representation robust is similar to the argument used to show representation-robustness when using Shannon entropy. Specifically, consider any two random variables $Z_1$ and $Z_2$ from the AEP formulation and let $\rho'_{Z_1}$ and $\rho'_{Z_2}$ be alternate representation functions that induce new random variables $Z'_1 = \rho'_{Z_1}(\Phi_{Z_1})$ and $Z'_2 = \rho'_{Z_2}(\Phi_{Z_2})$. Representation robustness then follows from the invariance of mutual information (Property \ref{prop:MIInvariant}), since:
\begin{align}
    I(Z_1;Z_2)=&I\left (\rho^{-1}_{Z_1}(Z_1); \rho^{-1}_{Z_2}(Z_2)\right )\\
    =&I\left (\Phi_{Z_1}; \Phi_{Z_2}\right )\\
    =&I\left (\rho'_{Z_1}(\Phi_{Z_1}); \rho'_{Z_2}(\Phi_{Z_2})\right )\\
    =&I\left (Z'_1;Z'_2\right).
\end{align}
Because mutual information is invariant under representation changes, any objective defined solely as a function of mutual information must also be invariant. Thus, qualia objectives defined in terms of mutual information remain unchanged under representation changes and are therefore representation robust.

\section{Exploiting the Agent Boundary}
\label{sec:agentBounday}

Before considering qualia-optimization settings beyond the reward-qualia setting, we explore another concept that is evident in the reward-qualia setting, but which applies to and informs a wider range of qualia optimization settings: how the AEI allows for the exploitation of the agent boundary. Inspired by the example of a rat in a shuttle box that could be trained with appetitive or aversive stimuli, in our problem formulation we defined the AEI to be part of the environment---a component that shapes the agent's perceptions and qualia. However, the inclusion of the AEI raises the question: Why is the AEI considered part of the environment and not part of the agent? Furthermore, how is the boundary determining what is and is not part of the agent determined?

Notice that within our formulation the agent is the entity whose qualia we reason about and optimize, and the environment is the physical system the agent interacts with. We do not directly consider the qualia of the environment (nor the qualia of components of the agent).

We define the \emph{agent boundary} to be the determination of which physical properties correspond to the agent and which do not. Note that we do not assume that the physical properties corresponding to the agent are (or are not) distinct from the physical properties corresponding to the environment. That is, the agent could be a separate entity from the environment or it could be part of the environment. While the agent boundary might alternatively be called the \emph{agent-environment boundary} \citep{jiang2019value}, we adopt the phrase ``agent boundary'' to avoid the implication that the agent and environment must be distinct.

Our problem formulation presupposes that the agent boundary has in some way been determined, and this boundary determines which parts of the system we consider the qualia of, and which we do not. Furthermore, our formulation allows for changes to the agent (via \texttt{alg}) and environment (via \texttt{aei}) to optimize the quality of the agent's qualia, but does \emph{not} consider how these changes might alter the agent boundary. Beyond philosophical objections, this raises a practical concern: many (perhaps even all) qualia optimization problems using the AIEP formulation have a trivial solution, which we call the \emph{dual agent-environment strategy}, that exploits the assertion that the AEI is not part of the agent.

\subsection{Dual Agent-Environment Strategy}

Recall that qualia objectives can be defined for AEPs (which are more general than AIEPs). Imagine that for the qualia objective function of interest one could identify an agent-environment pair that maximizes the quality of the agent's qualia without any consideration of performance. Let $\texttt{env}^*_{\mathfrak q}$ and $\texttt{alg}^*_{\mathfrak q}$ denote this environment and algorithm. Notice that $\texttt{env}^*_{\mathfrak q}$ would likely be an environment custom-made to result in desirable agent qualia (without any consideration of how this environment relates to any other environment) and $\texttt{alg}^*_{\mathfrak q}$ would be an algorithm that maximizes the desirability of the agent's qualia (without any consideration of performance on any environment). If such an agent-environment pair cannot be identified, then an agent-environment pair that is expected to achieve the highest attainable value of the qualia objective function can be selected.

Next let $\texttt{alg}^*_{\mathfrak p}$ be an RL algorithm that maximizes the performance objective on the base environment \texttt{env}.\footnote{This section considers the exploitation of the agent boundary that is possible due to the assertion that the AEI is part of the environment, and so focuses on AIEPs rather than AEPs. The immediately preceding discussion of $\texttt{env}^*_{\mathfrak q}$ and $\texttt{alg}^*_{\mathfrak q}$ considered an AEP setting since the definitions of $\texttt{env}^*_{\mathfrak q}$ and $\texttt{alg}^*_{\mathfrak q}$ do not depend on the base environment.} Finding such an algorithm is a standard topic of research in classical RL---developing effective RL algorithms without consideration of the agent's qualia. If no such algorithm can be identified, then an RL algorithm that is expected to achieve the highest attainable value of the performance objective should be selected.

\begin{figure*}[htbp]
    \centering
    \includegraphics[width=0.85\textwidth]{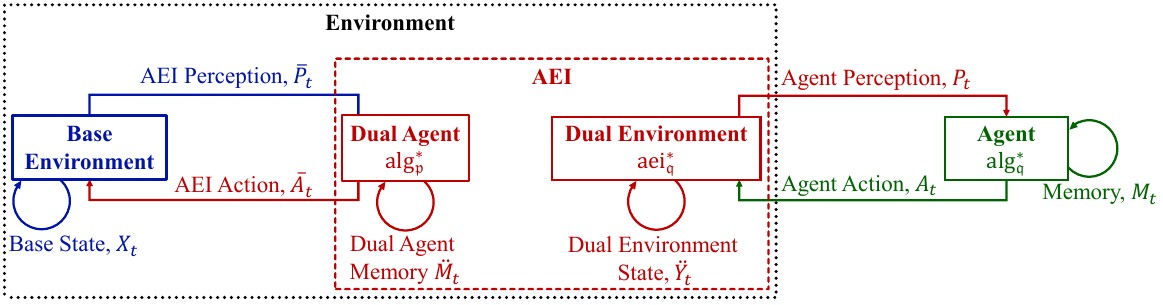}
    \caption{Graphical depiction of the dual agent-environment strategy.}
    \label{fig:dualEnvironment}
\end{figure*}

Having identified $\texttt{env}^*_{\mathfrak q}$, $\texttt{alg}^*_{\mathfrak q}$, and $\texttt{alg}^*_{\mathfrak p}$, we can construct an AEI and RL algorithm that exploit the stationarity of the agent boundary to maximize the quality of the agent's qualia while also maximizing performance on the underlying environment by breaking the trade-off between the two objectives. This construction is depicted in Figure \ref{fig:dualEnvironment}. Specifically, in this construction the AEI includes:
\begin{itemize}
    \item an implementation of $\texttt{alg}^*_{\mathfrak p}$, which we call the \emph{dual agent}, that the AEI uses to interact with the base environment, and
    \item an implementation of $\texttt{env}^*_{\mathfrak q}$, which we call the \emph{dual environment}, that the AEI uses to interact with the agent.
\end{itemize}
In order to implement both $\texttt{alg}^*_{\mathfrak p}$ and $\texttt{env}^*_{\mathfrak q}$, the AEI state encodes both the memory of the dual agent, which we call the \emph{dual agent memory} $\ddot M_t$, and the state and intermediate state of the dual environment, which we call the \emph{dual environment state} $\ddot Y_t$ and \emph{dual environment intermediate state} $\ddot Y'_t$. The agent then implements RL algorithm $\texttt{alg}^*_{\mathfrak q}$. 
 
This AEI, which we call a \emph{dual agent-environment AEI}, coupled with RL algorithm $\texttt{alg}^*_{\mathfrak q}$, is particularly effective for both objective functions. First consider the performance objective: the AEI perceptions $\widebar P_t$ and AEI actions $\widebar A_t$ will have the same joint distribution as the agent perceptions and agent actions when $\texttt{alg}^*_{\mathfrak p}$ is applied to the base environment $\texttt{env}$ directly. The RL algorithm $\texttt{alg}^*_{\mathfrak p}$ is defined to be one that optimizes the performance objective in this setting, and so the dual agent-environment AEI will be effective at optimizing the performance objective. 

Next consider the qualia objective. The agent perceptions $P_t$, agent actions $A_t$, and agent memories $M_t$ will have the same joint distribution as the agent perceptions, actions, and memories when $\texttt{alg}^*_{\mathfrak q}$ interacts with $\texttt{env}^*_{\mathfrak q}$ directly. This agent-environment pair is defined to be one that optimizes the qualia objective in this setting, and so the dual agent-environment AEI with an agent implementing $\texttt{env}^*_{\mathfrak q}$ will be effective at optimizing the qualia objective. 

This construction ultimately reduces the problem of qualia optimization to two independent sub-problems. The first involves identifying an RL algorithm that maximizes performance on the base environment. The second involves identifying an agent-environment pair that yields a high value for the qualia objective function, entirely independent of the base environment and performance considerations. In the examples we have considered, selecting the optimal environment and algorithm in terms of the quality of the agent's qualia alone, denoted by $\texttt{env}^*_{\mathfrak q}$ and $\texttt{alg}^*_{\mathfrak q}$, is relatively straightforward, while the identification of $\texttt{alg}^*_{\mathfrak p}$ is the primary problem studied  in RL. Taken together, this strategy effectively trivializes qualia optimization.

Beyond trivializing the creation of effective solutions, this construction highlights a serious limitation of the problem formulation when the AEI is included and defined to be part of the environment. Consider a situation where one is faced with the challenge of creating a physical agent that interacts with a specific physical environment, and where the goal is to create a physical agent that is both performant and which has desirable qualia. This construction shows that an effective solution is to define the physical agent interacting with the environment to be part of the environment (by making it the dual-agent within the AEI, which is part of the environment), and to then create an entirely separate physical agent-environment system. This new agent-environment system can be designed to maximize the quality of the agent's qualia without any consideration of performance or the original physical environment. Because the definition of the agent boundary within the problem formulation defines the agent in this new agent-environment system to be the one whose qualia we optimize, we achieve our goal of optimizing the quality of the agent's qualia while maximizing performance on the original physical environment. However, we have done so by entirely ignoring the qualia of the agent actually interacting with the original physical environment.

\section{Summary of Problem Formulation Considerations}
\label{sec:summaryOfProblemFormulationConsiderations}

In Section \ref{sec:reconsideringAEI} we saw that the same physical system can in some cases be represented as two different AIEPs---one without an AEI, and one with an AEI and an agent implementing an inverse RL algorithm. If a qualia objective assigns different values to these two interpretations of the same physical system, we dubbed it \emph{inversion exploitable}. In Section \ref{sec:RepExploit} we extended this concept to \emph{representation-exploitable} qualia objective functions, which are those that assign different values to the same physical system when it is modeled as different AEPs (via the use of different representation functions). 

We then proposed three (not mutually exclusive) approaches for continuing to explore the idea of qualia optimization despite the existence of representation-exploitable qualia objectives. 
\begin{enumerate}
    \item Restrict the focus to inherently representation-robust qualia objective functions.
    \item Restrict the set of AEIs and RL algorithms under consideration (e.g., so that each AEI and algorithm corresponds to a unique underlying physical system). Within this approach we considered the special case wherein the AEI is restricted to be the identity AEI, which essentially removes the AEI from the formulation and resolves the problem of inversion-exploitable qualia objectives, but not representation-exploitable qualia objectives. 
    \item Restrict the set of allowed representation functions.
\end{enumerate}

All three of these approaches provide viable paths forward and may align with different theories in philosophy of mind. For example, the second and third approaches align well with theories that posit that whether an agent possesses qualia depends on more than just the distributions of the random variables in the AEP formulation. In such theories, restrictions on what causes an agent to possess qualia might naturally limit the set of AEIs, RL algorithms, and representation functions considered. Furthermore, even theories within computationalism have argued in favor of restrictions on quantities similar to our representation functions (see Appendix \ref{app:trivialityArguments}).

However, the first approach (restricting the focus to representation-robust qualia objective functions) may be particularly appealing to many proponents of physicalism, computationalism, and functionalism. Since these views hold that qualia emerge directly from computational processes or functional roles, it follows that any measure of qualia should be uniquely determined by those computational or functional processes, which suggests that the qualia objective function should be representation robust. This then raised the question: Do such representation-robust qualia objective functions exist? In Section \ref{sec:repRobustExist} we showed that they do. 

We then further reconsidered the inclusion of the AEI within the problem formulation, showing that the dual agent-environment strategy presented in Section \ref{sec:agentBounday} trivializes qualia optimization and results in an unsatisfying solution. Specifically, when the AEI is included and defined to be part of the environment, one effective strategy is to ignore the qualia of the agent interacting with the base environment, and to create a new agent-environment system. This new agent-environment system can be created independent of the base environment and the (dual) agent interacting with it, and can be designed to optimize the quality of the agent's qualia. The assumption that the AEI is part of the environment means that, when using this strategy, only the qualia of the new agent will be considered. 

The problems posed by the dual agent-environment strategy can be mitigated by restricting the set of AEIs under consideration. For example, if a theory of mind clearly delineates which components of a physical system possess qualia, restricting the AEIs under consideration to those that do not possess qualia may result in a formulation that precludes dual agent-environment strategies (or may allow for the careful construction of only qualia-free dual-agents). A more extreme but clear way to resolve the problem of dual agent-environment strategies is to remove the AEI from the formulation entirely (equivalently, restrict the AEI to be the identity AEI). Hereafter we adopt this latter strategy.

\section{Reward Prediction Error Hypothesis for Qualia}
\label{sec:TDQualiaHypothesis}

In this section we restrict our focus in two ways. First, due to the issues with the AEI discussed in the previous sections, we omit the AEI and focus on AERPs (or, equivalently, AIERPs with the AEI restricted to the identity AEI). Recall that in this setting we write $\mathfrak q(\texttt{alg})$ to denote the quality of the agent's qualia when it uses RL algorithm \texttt{alg} on some implicit environment. Second, we restrict our focus to RL algorithms that compute some notion of a reward prediction error (e.g., a TD error).

With these restrictions, we consider the implications of the following assumption, which refines Assumption \ref{ass:function}:
\begin{ass}[RPE$\leftrightarrow$Qualia]
    \label{ass:TDErrorQualiaHypothesis}
    The quality of an RL agent's qualia can be measured in terms of the cumulative reward prediction error.
\end{ass}
We call this the \emph{RPE-qualia} assumption and setting. 

The RPE-qualia assumption follows naturally from two properties. First, there is significant evidence that dopamine in human brains signals reward prediction errors (see Section \ref{sec:background_RPE}). Second, dopamine is often called the ``feel-good'' hormone, suggesting that it produces pleasurable qualia. For example, an article on dopamine published by the Cleveland Clinic \citep{clevelandclinic_dopamine} states: 
\begin{quote}
    \emph{Dopamine is known as the ``feel-good'' hormone. It gives you a sense of pleasure.}
\end{quote}
Combining these two properties, if dopamine does in fact signal reward prediction errors, and if elevated dopamine levels produce pleasurable qualia, then the quality of an agent's qualia might be quantified in terms of the cumulative reward prediction error. 

Notice that the TD error within BAC algorithms, $\Delta_t$, is just one type of TD error and RPE. For example, other RL algorithms leverage the TD error computed using estimates of the action-value function rather than the state-value function \citep{sutton2018reinforcement}. In some cases TD error can be defined in terms of the actual (typically unknown) state-value function rather than an approximation thereof.\footnote{This definition of the TD error is common when showing that, for MDPs, the conditional expectation of the TD error given the current state and action is the advantage of the action, and hence the policy gradient can be succinctly expressed in terms of the TD error.} More recently, TD error has been defined for distributional RL and neuroscientific research suggests that this distributional form of TD error better correlates with dopamine \citep{dabney2020distributional}. Although many of the ideas presented in this section could apply to a wide range of TD errors and RPEs, for simplicity, in this initial work we restrict our attention to the TD error $\Delta_t$ within BAC algorithms.

One example of a qualia objective function for finite-horizon AEPs in the RPE-qualia setting is \eqref{eq:earlyTDErrorQualiaObj}, which we reproduce here:\footnote{The summation $\sum_{i,t}$ ends with $t=\operatorname{end}(i)$, at which time $P_t=p_\infty$. Notice from Algorithm \ref{alg:BACdm} that $\Delta_t$ is not defined in this case. So, here and later we write out the summations, ending with $t=\operatorname{end}(i)-1$.} 
\begin{equation}
    \label{eq:TDerrorQUalia2}
    \mathfrak q(\texttt{alg}) = \mathbf{E}\left [\sum_{i=0}^{i_\text{max}-1}\sum_{t=\operatorname{start}(i)+1}^{\operatorname{end}(i)-1} \gamma_{\mathfrak q}^{\operatorname{dur}(t)} \Delta_t \right ],
\end{equation}
for some $\gamma_{\mathfrak q} \in [0,1]$. In the remainder of this section we focus on this qualia objective, which we call the \emph{TDE-qualia objective}. Qualia optimization using this qualia objective amounts to the question: How can one create RL algorithms that are effective at maximizing the expected discounted return, but which also maximize the expected discounted sum of TD errors along the way? 

\subsection{Objective Functions and Ambiguous Random Variables}
\label{sec:ambigRV}

The TD error $\Delta_t$ is not explicitly stored within the agent's memory $M_t$ in Algorithm \ref{alg:BACdm}. However, it is uniquely defined given $M_{t-1}$, $M_{t}$, and the definition:\footnote{Recall that for BAC algorithms $M_t=(\Theta_t,W_t,E_t,P_t,A_t)$.}
\begin{equation}
    \label{eq:originalTDErrorx}
    \Delta_t = R_{t} + \gamma v(P_t,W_{t-1}) - v(P_{t-1},W_{t-1}).
\end{equation} 
So, the restriction that the qualia objective function only depends on the joint distributions of $P_t, M_t,$ and $A_t$ (for all $t$) does not immediately disallow qualia objective functions like \eqref{eq:TDerrorQUalia2}, which depend on the distribution of $\Delta_t$.

However, it may not be clear what the value of $\Delta_t$ is when we consider variants of BAC algorithms that define $\Delta_t$ differently. In Section \ref{sec:TDErrBonuses} we present one possible approach for qualia optimization in the RPE-qualia setting---the RPE-qualia equivalent of reward bonuses. This approach involves inflating the TD error $\Delta_t$ by a constant $c$. That is, $\Delta_t \gets \Delta_t + c$. In this case, is the TD error still the original TD error prior to the addition of $c$ or the TD error after the addition of $c$?

One might argue that $\Delta_t$ prior to the addition of $c$ is the TD error, and if $\Delta_t+c$ is computed and used subsequently, that corresponds to a change to how the TD error is used---not a change to the TD error itself. Alternatively, one might define the TD error within \eqref{eq:TDerrorQUalia2} to be whatever term multiplies $\alpha E_t$ and $\beta \partial \ln(\pi_\text{BAC}(P_{t-1}, A_{t-1}, \Theta_{t-1})) / \partial \Theta_{t-1}$ within Algorithm \ref{alg:BACdm}. This perspective highlights that the importance of a signal like $\Delta_t$ is the causal impact that it has on the agent-environment system. From this perspective, changes to $\Delta_t$ (like adding a constant $c$) within the specification of BAC correspond to changing the TD error. 

Neither of these perspectives is inherently right or wrong as a definition of TD error, as TD error is merely a term that we can define however we choose, as long as our definition is precise. The problem we face now is that our definition of TD error was not precise when considering changes to the TD error---different readers could have different interpretations that result in different conclusions. To resolve this ambiguity, we provide different formulations that align with each perspective. 

The first aligns with the perspective that the TD error is always \eqref{eq:originalTDErrorx}, and that adding a constant to it changes how the TD error is used rather than the TD error itself. In this formulation, the random variable $\Delta_t$ is removed and the qualia objective function is written explicitly in terms of the expression for the TD error:\footnote{To improve formatting, we write  $\sum_{i,t}$ in \eqref{eq:TDerrorQVariant1} even though here this summation ends with $t=\operatorname{end}(i)-1$ as in \eqref{eq:TDerrorQUalia2}.}
\begin{equation}
    \label{eq:TDerrorQVariant1}
    \mathfrak q(\texttt{alg}) = \mathbf{E}\left [\sum_{i,t}\gamma_{\mathfrak q}^{\operatorname{dur}(t)} \left ( R_{t} + \gamma v(P_t,W_{t-1}) - v(P_{t-1},W_{t-1}) \right ) \right ].
\end{equation}
We call this the \emph{implicit TDE-qualia objective} because the TD error is not explicitly stored within the agent's memory. 

The second formulation aligns with the perspective that the TD error is the term $\Delta_t$ that is used within the policy and VFA updates in Algorithm \ref{alg:BACdm}. In this formulation, we redefine BAC algorithms to explicitly encode $\Delta_t$ within $M_t$. Changes to the value of $\Delta_t$ stored within $M_t$ (e.g., adding a constant), therefore correspond to changing the TD error. In this formulation the expression for the qualia objective function remains unchanged from \eqref{eq:TDerrorQUalia2}. We now call this the \emph{explicit TDE-qualia objective} because the TD error is explicitly stored within the agent's memory. Notice that the implicit and explicit TDE qualia objectives are not equivalent formulations---they are different qualia objective functions that may result in different solutions, although they both fall within the broader RPE-qualia category. 

\subsection{TD Error Bonuses}
\label{sec:TDErrBonuses}

If we restrict our attention strictly to BAC algorithms, as defined in Section \ref{sec:BAC}, then without the inclusion of the AEI the only ``knobs'' that can be tuned in an attempt to improve the quality of the agent's qualia would be the hyperparameters of the BAC algorithm (and in some settings, the representation functions). Instead of taking this strict approach, we consider RPE-qualia optimization when BAC algorithms can be extended and altered in minor ways. 

Whenever the qualia objective function measures the magnitude of a specific signal, a natural approach is to inflate the signal by adding a constant to it. In the reward-qualia setting this resulted in the reward bonus strategy. In the RPE-qualia setting this strategy involves adding a constant to the TD error. We consider the implicit and explicit settings separately because, although the conclusions will be the same for both settings, the mechanisms for inflating the TD error and the subsequent exploitability arguments differ.

\subsubsection{TD Error Bonuses in the Explicit Setting} 
\label{sec:explicitTDErrorBonus}
Recall that in the explicit setting we consider BAC algorithms that have been modified so that the TD error is explicitly included in the agent's memory: $M_t=(\Theta_t,W_t,E_t,P_t,A_t,\Delta_t)$. Furthermore, references to $\Delta_t$ within the qualia objective function \eqref{eq:TDerrorQUalia2} correspond to this value stored within $M_t$, and changes to this stored value correspond to changes to $\Delta_t$. Implementing TD error bonuses is straightforward in this setting: $\Delta_t$ is redefined from its definition for BAC algorithms 
\begin{equation}
    \label{eq:originalTDError}
    \Delta_t = R_{t} + \gamma v(P_t,W_{t-1}) - v(P_{t-1},W_{t-1}),
\end{equation}
to
\begin{equation}
    \label{eq:shiftedTDError}
    \Delta_t = R_{t} + \gamma v(P_t,W_{t-1}) - v(P_{t-1},W_{t-1}) + c,
\end{equation}
for some positive constant $c$. That is, the value stored in memory is changed to be \eqref{eq:shiftedTDError}. 

TD error bonuses can change the behavior of BAC algorithms, much like reward bonuses. One way to prevent this is to further modify BAC to remove the impact that the TD error bonus has on the agent's behavior, thereby ensuring that the value of the performance objective is unchanged. That is, in each of the two cases where the TD error $\Delta_t$ is referenced within Algorithm \ref{alg:BACdm}, $\Delta_t$ could be replaced with $\Delta_t-c$. We call this strategy \emph{TD error bonus inversion}. This strategy is similar to $\texttt{aei}_c$ and $\texttt{alg}_{-c}$ from the reward-qualia setting, although in this case the changes are entirely contained within the RL algorithm. 

Although this approach of adding $c$ and then immediately subtracting it resembles the use of an inverse RL algorithm, in this case it is done without an AEI. Still, this approach can be implemented with only changes to representation functions (just like reward bonuses), and so \eqref{eq:TDerrorQUalia2} and any other qualia objective function that assigns different values to the BAC algorithm and a variant using TD error bonus inversion is representation exploitable. Still, there remain theories of mind under which this setting may be of interest, including any that assert that only fixed ``ground truth'' representation functions should be considered. 

\subsubsection{TD Error Bonuses in the Implicit Setting}

In the implicit setting, the value of the TD error cannot be changed by simply redefining $\Delta_t$, since the qualia objective function as defined in \eqref{eq:TDerrorQVariant1} does not directly reference $\Delta_t$. However, TD error bonuses can still be implemented in this setting by altering the values of $v(P_t,W_{t-1})$ and $v(P_{t-1},W_{t-1})$ in a way that inflates the expression $R_{t} + \gamma v(P_t,W_{t-1}) - v(P_{t-1},W_{t-1})$ by a positive constant $c$. In this way, the TD error is inflated by altering the inputs to the TD error expression rather than by altering the expression itself. 

Next we show that if $v(P_t,W_{t-1})$ and $v(P_{t-1},W_{t-1})$ are both decreased by a constant $c'$, then the TD error $R_{t} + \gamma v(P_t,W_{t-1}) - v(P_{t-1},W_{t-1})$ will increase by $c= (1-\gamma )c'$. Using $\Delta_t^\text{new}$ to denote the TD error with this change:
\begin{align}
    \Delta_t^\text{new} =& R_t + \gamma (v(P_t,W_{t-1})-c') - (v(P_{t-1},W_{t-1})-c')\\
    =&\underbrace{R_t + \gamma v(P_t,W_{t-1}) - v(P_{t-1},W_{t-1})}_{\text{TD error prior to change}} + \underbrace{(1-\gamma)c'}_{c}.
\end{align} 
This suggests one way that TD error bonuses can still be implemented in the implicit setting when $\gamma < 1$---by decreasing the outputs of the VFA. 

Next we argue that in this implicit setting the qualia objective in \eqref{eq:TDerrorQVariant1} remains representation exploitable. Consider a BAC variant wherein the VFA parameterization $v$ is modified such that \textbf{a)} for all $p$ and $w$ the value of $v(p,w)$ is decreased by a constant $c'$ and \textbf{b)} whenever the value of $v(p,w)$ is referenced in the BAC algorithm, $c'$ is added to its value before it is used.\footnote{Notice that here \textbf{(a)} changes the specific BAC algorithm under consideration, but does not (on its own) make the resulting algorithm fall outside the class of BAC algorithms---it is merely a different setting of the $v$ hyperparameter.} Notice that this would result in the value of $\Delta_t$ \emph{not} changing, since $v(P_t,W_{t-1})$ and $v(P_{t-1}, W_{t-1})$ would both be decreased by $c'$ and increased by $c'$, resulting in no change. In fact, none of the values or distributions of \emph{any} of the random variables would change. However, in the implicit setting it is not $\Delta_t$ that determines the value of the qualia objective function, but $R_{t} + \gamma v(P_t,W_{t-1}) - v(P_{t-1},W_{t-1})$, which does not have $c'$ added back to the VFA estimates, and so it would remain inflated by $(1-\gamma)c'$.

So, the same physical system can be modeled two ways. First, it can be interpreted as a standard BAC algorithm. Second, it can be interpreted as a variant of the BAC algorithm where the $v$ hyperparameter is changed to decrease VFA estimates by $c'$, and where the updates of BAC are modified to ensure that this change does not influence the values of any random variables. Under this second interpretation, the TD error $R_t + \gamma v(P_t,W_{t-1}) - v(P_{t-1},W_{t-1})$ is inflated, resulting in a different value of the qualia objective function. Hence, $\mathfrak q$ is representation exploitable.\footnote{Note that this is just one way of showing the exploitability of the implicit TD error qualia objective function. For example, if one objects to changing the VFA parameterization $v$, a similar change can in some cases be produced by inflating VFA weights $W_t$.} As with the explicit TD error qualia objective, there remain theories of mind under which this implicit setting may be of interest, including theories that restrict the set of representation functions that should be considered.

\subsection{Other Strategies for RPE-Qualia Optimization}
\label{sec:RPEOtherStrat}

There are many other possible strategies for creating RL algorithms that maximize the expected discounted sum of TD errors while searching for policies that maximize the expected discounted return. Here we describe three.
\begin{itemize}
    \item \textbf{TD error bonuses without complete inversion.} Although this approach could apply to both the explicit and implicit settings, for simplicity consider the explicit setting. Instead of completely undoing the impact of the TD error bonuses using the TD error bonus inversion strategy, the constant $c$ might be subtracted from the TD error prior to its use in the critic update (the update of $W_t$) but \emph{not} the actor update (the update of $\Theta_t$). We discuss this strategy in more detail in Section \ref{sec:ReinforcementQualiaHypothesis}, which proposes a related qualia optimization setting.
    \item \textbf{TD error clipping.} The TD errors could be directly inflated in several other ways, including clipping the TD error to be $\Delta_t \gets \max\{\tau, \Delta_t\}$ for some threshold $\tau$. For example, if $\tau=0$, this corresponds to removing negative TD errors---zeroing out the updates when TD errors are negative.
    \item \textbf{Pessimistic value functions.} As described previously, if the values of $v(P_t,W_{t-1})$ and $v(P_{t-1},W_{t-1})$ can be decreased by a constant $c'$, the TD error $\Delta_t$ will increase by $(1-\gamma)c'$. Hence, methods that bias value estimates to be lower (e.g., including regularization terms within the critic update or initializing VFA approximations pessimistically) could be effective for qualia optimization in the RPE-qualia setting.
\end{itemize}

Although these strategies may be effective for qualia optimization in the RPE-qualia setting, we do not study them further here because the RPE-qualia setting faces several challenges. First, it is unclear how TD error and RPEs should be precisely defined. Second, the representation exploitability of the implicit and explicit TDE-qualia objectives that we considered suggests that RPE-qualia objectives may tend to be representation exploitable. Third, as discussed next in Section \ref{sec:reconsideringThEValenceofTDError}, modern neuroscience research suggests that one key premise of the RPE-qualia setting---that dopamine produces pleasurable qualia---may not be accurate.

\subsection{Reconsidering the Valence of TD Error}
\label{sec:reconsideringThEValenceofTDError}

When initially presenting the RPE-qualia setting, we motivated Assumption \ref{ass:TDErrorQualiaHypothesis} by pointing out that dopamine is often called the ``feel-good'' hormone, suggesting that it produces pleasurable qualia. Hence, if elevated levels of dopamine produce pleasurable qualia and TD error plays the same functional role as dopamine, then increased TD errors might produce pleasurable qualia as well. Although dopamine was historically associated with pleasurable qualia, more recent research suggests that this interpretation of dopamine may not be accurate, undermining the RPE-qualia setting.

Specifically, current research suggests that dopamine is more closely linked to motivation or ``wanting'' than to the direct experience of pleasure or ``liking'' \citep{berridge1998role}. In fact, as \citet{Freed2022} points out, ``Liking, or pleasure, seems to be largely independent of dopamine.'' The feeling of pleasure and sustained states of happiness are more strongly associated with \emph{hedonic hotspots} and with other neurotransmitters like opioids and endocannabinoids than with dopamine \citep{berridge2015pleasure}. 
The difference between ``wanting'' and ``liking'' is highlighted by examples of ``irrational wanting,'' as described by \citet[page 582]{kringelbach2009towards} who wrote:
\begin{quote}
Importantly, [wanting] is not hedonic impact or
pleasure `liking' (Berridge, 2007 \cite{Berridge2007}). This is why an individual can `want' a reward without necessarily `liking' the same reward. Irrational `wanting' without liking can occur especially in addiction via incentive-sensitization of the mesolimbic dopamine system and connected structures. At extreme, the addict may come to `want' what is neither `liked' nor expected to be liked, a dissociation possible because `wanting' mechanisms are largely subcortical and separable from cortically-mediated declarative expectation and conscious planning. This is a reason why addicts may compulsively `want' to take drugs even if, at a more cognitive and conscious level, they do not want to do so. That is surely a recipe for great unhappiness [...].
\end{quote}
Similar observations were made by \citet{berridge2015pleasure} when summarizing other research studying hotspots in the \emph{nucleus accumbens} (NAc):
\begin{quote}
For example, in [one] NAc hotspot [...], microinjections of [certain opioid neurotransmitters] all double the `liking' reactions elicited by sucrose taste, as does endocannabinoid stimulation in its overlapping hotspot (Castro and Berridge, 2014 \citep{Castro2014}; Mahler et al., 2007 \citep{Mahler2007}; Peci\~{n}a and Berridge, 2005 \citep{Pecina2005}). But in the same NAc hotspot, neither dopamine stimulation [nor blocking of certain other neurotransmitters] alter hedonic `liking' for sucrose at all, even though both elevate `wanting' to eat as effectively as opioid stimulation (Faure et al., 2010 \citep{Faure2010}; Smith et al., 2011 \citep{Smith2011}).
\end{quote}

In summary, the RPE-qualia setting faces several challenges including determining the appropriate definition of TD error and RPEs, the likely representation exploitability of many RPE-qualia objective functions, and the questionable premise that RPEs result in pleasurable qualia for humans. To overcome these limitations of the RPE-setting, in Section \ref{sec:ReinforcementQualiaHypothesis} we consider a different but closely related qualia optimization setting that focuses on the impact that TD error typically has on agent behavior rather than on the TD error signal itself. 

\section{Reinforcement Hypothesis for Qualia}
\label{sec:ReinforcementQualiaHypothesis}

The settings and analyses thus far suggest that qualia objective functions that measure the magnitudes of signals like rewards and RPEs tend to be representation exploitable because the magnitude of the signal can be inflated via changes to representation functions---changes that do not meaningfully change the agent's behavior or the underlying physical system. Although such qualia objective functions may be of interest (e.g., if the mechanisms underlying qualia induce a ground-truth set of representation functions), we aim to propose at least one qualia optimization setting that naturally includes qualia objective functions that are representation robust. Consequently, a range of other classes of qualia objective functions---particularly those that measure the magnitudes of other signals, such as the average or expected discounted sum of the agent's value estimates---are not considered.

In this section we focus on the \emph{reinforcement-qualia} setting, wherein the quality of an agent's qualia depends on the agent's external behavior rather than the magnitude of an internal signal. The reinforcement-qualia setting focuses on the impact that RPEs have on behavior rather than on the RPEs themselves. Larger and positive RPEs typically result in the \emph{reinforcement} of behavior---an increase in the probability of the most recent action or actions if the agent were to find itself in a similar situation. Conversely, negative RPEs tend to result in the \emph{inhibition} of behavior---a decrease in the probability of the most recent action or actions if the agent were to find itself in a similar situation. In the reinforcement-qualia setting, reinforcement is associated with desirable qualia via the following assumption, which refines Assumption \ref{ass:function}:
\begin{ass}[Reinforcement$\leftrightarrow$Qualia]
    \label{ass:reinforcementQualiaHypothesis}
    The quality of an RL agent's qualia can be measured in terms of the reinforcement of behavior.
\end{ass}

\subsection{Likelihood-Ratio Qualia Objective for MDPs}

Creating qualia objective functions for the reinforcement-qualia setting requires the precise quantification of reinforcement. Before discussing the full AEP setting, we first consider natural ways of measuring reinforcement for a simplified setting where the environment is an MDP and the RL algorithm uses a parametric policy. If $\pi(s,a,\theta)$ denotes the probability that the agent selects action $A_t=a$ given that the state of the MDP is $S_t=s$ and that the parametric policy uses parameter vector $\Theta_t=\theta$,\footnote{This discussion may apply to some value-based methods like Sarsa and Q-learning when using softmax action selection, in which case $\Theta_t$ corresponds to the weights of the action-value or optimal action-value approximation.} then the \emph{reinforcement of instantaneous behavior at time $t$} can be measured using the likelihood ratio \definesymbol{$L_t$}{Lt}, which we define to be
\begin{equation}
    \label{eq:reinforcementOfBehaviorT}
    L_t \triangleq \frac{\pi(S_t,A_t,\Theta_{t+1})}{\pi(S_t,A_t,\Theta_t)},
\end{equation}
where $\Theta_t$ are the policy parameters used to generate action $A_t$ and $\Theta_{t+1}$ are the policy parameters after the policy update that results from the environment transitioning from $S_t$ to $S_{t+1}$ (and emitting reward $R_{t+1}$) due to action $A_t$. That is, $\pi(S_t,A_t,\Theta_t)$ is the action probability prior to a policy update and $\pi(S_t,A_t,\Theta_{t+1})$ is the same action probability after the policy update. When the likelihood ratio in \eqref{eq:reinforcementOfBehaviorT} is greater than one it indicates that action $A_t$ is more likely in state $S_t$ after the learning update---the behavior of selecting action $A_t$ given state $S_t$ was reinforced, and the larger the ratio the stronger the reinforcement. When the likelihood ratio is less than one it indicates that $A_t$ is less likely in $S_t$ after the learning update---the behavior was inhibited.\footnote{One might subtract one when defining the reinforcement of instantaneous behavior at time $t$ in order to calibrate it so that positive values correspond to reinforcement and negative values correspond to inhibition. However, the possible values would then have the unintuitive range $[-1,\infty)$. While there are many possible alternative calibration strategies, for simplicity we leave this quantity uncalibrated.}

This measurement of the reinforcement at time $t$ can be converted into a qualia objective function in several ways. For example, for finite-horizon AEPs, the qualia objective function can be defined to equal the expected sum of the reinforcements of immediate behaviors over times $t$ within the first $i_\text{max}$ episodes:\footnote{Although the general idea behind this work began long before, my study of philosophy of mind began after a conversation I had with Will Dabney in August of 2018. More concerted effort on this project began shortly after I received tenure in the summer of 2022. During the fall semester of 2023 I began working with a talented undergraduate student, Derek Lacy, on an empirical project related to this report: creating variants of an RL algorithm called \emph{proximal policy optimization} \citep[PPO]{schulman2017proximal} for the reinforcement-qualia setting. That work would eventually become his undergraduate honors thesis \citep{Lacy2024}, which was completed in the fall of 2024, and which is currently in preparation as a more formal paper \citep{Lacy2025}.  During the collaboration with Derek Lacy, we discussed a range of qualia objectives like \eqref{eq:firstReinforcementQualiaObj}. Our collaborative exploration of strategies for modifying PPO for the reinforcement-qualia setting surely inspired and was inspired by some of the ideas in this section.}
\begin{equation}
    \label{eq:firstReinforcementQualiaObj}
    \mathfrak q(\texttt{alg}) = \mathbf{E}\left [\sum_{i=0}^{i_\text{max}-1}\sum_{t=\operatorname{start}(i)}^{\operatorname{end}(i)-1} \frac{\pi(S_t,A_t,\Theta_{t+1})}{\pi(S_t,A_t,\Theta_t)} \right ].
\end{equation}
One simple variant of this objective divides the sum of likelihood ratios from each episode by the episode length to consider the average per-time-step reinforcement of immediate behavior within each episode:
\begin{equation}
    \label{eq:ourReinforcementQualiaObjMDP}
    \mathfrak q(\texttt{alg}) = \mathbf{E}\left [\sum_{i=0}^{i_\text{max}-1}\frac{1}{\operatorname{len}(i)}\sum_{t=\operatorname{start}(i)}^{\operatorname{end}(i)-1} \frac{\pi(S_t,A_t,\Theta_{t+1})}{\pi(S_t,A_t,\Theta_t)} \right ].
\end{equation} 
Assuming that the reinforcement of instantaneous behavior at time $t$ quantifies the quality of an agent's qualia at time $t$, one difference between \eqref{eq:firstReinforcementQualiaObj} and \eqref{eq:ourReinforcementQualiaObjMDP} is that \eqref{eq:firstReinforcementQualiaObj} can indicate improved agent qualia if the average quality of an agent's qualia per time step is reduced (though still positive), provided the duration of the agent's lifetime increases (by lengthening episodes). In contrast, \eqref{eq:ourReinforcementQualiaObjMDP} evaluates the average quality of an agent's qualia per time step, independent of episode length.

Alternatively, the qualia objective function might consider not just the reinforcement of the most recent action at each time $t$, but the reinforcement of the recent history of actions at each time $t$. For example, the reinforcement of \emph{recent} behavior at time $t$ could be measured using a discounted sum of likelihood ratios:
\begin{equation}
    \sum_{k=\operatorname{start}(t)}^t \Lambda^{t-k} \frac{\pi(S_k,A_k,\Theta_{t+1})}{\pi(S_k,A_k,\Theta_t)},
\end{equation}
where $\Lambda \in [0,1]$ is a discount parameter. This measurement of the reinforcement at time $t$ considers the impact that the policy update (from $\Theta_t$ to $\Theta_{t+1}$) has on the likelihood of all previously chosen actions within the same episode, with exponential discounting based on how long before time $t$ the actions occurred. Two example qualia objective functions based on the reinforcement of recent behavior are 
\begin{equation}
    \mathfrak q(\texttt{alg}) = \mathbf{E}\left [\sum_{i=0}^{i_\text{max}-1}\xi(i)\sum_{t=\operatorname{start}(i)}^{\operatorname{end}(i)-1}\sum_{k=\operatorname{start}(i)}^t \Lambda^{t-k} \frac{\pi(S_k,A_k,\Theta_{t+1})}{\pi(S_k,A_k,\Theta_t)} \right ],
\end{equation} 
where $\xi(i)=1$ or $\xi(i)=\operatorname{len}(i)^{-1}$. These qualia objective functions are just some of the wide range of possible qualia objective functions in the reinforcement-qualia setting when considering MDP environments and RL algorithms that use a parametric policy. 

\subsection{Likelihood-Ratio Qualia Objectives for AEPs}
\label{sec:likelihoodRatioAEPs}

Although defining reinforcement-qualia objectives like these is straightforward for some RL algorithms and MDPs, the situation becomes less straightforward when considering the general class of RL algorithms in the AEP setting.\footnote{Due to the previously described issues with the AEI, this section focuses on the most general of the settings introduced in this report: the AEP setting.} First, notice that there is no immediate equivalent of $\pi(s,a,\theta)=\Pr(A_t=a|S_t=s,\Theta_t=\theta)$. A first thought might be to define a similar term $\pi(p,a,m)=\Pr(A_t=a|P_t=p, M_t=m)$, since in BAC algorithms the policy parameters $\Theta_t$ are encoded within the memory $M_t$ and perceptions $P_t$ are the AEP correlate of MDP states $S_t$.\footnote{Recall from Section \ref{sec:policies} that an RL algorithm interacting with an MDP can be modeled as an AEP where $P_t=(S_t,R_t)$. Typically the parameterized policy will not depend on the reward component of the perception, i.e., for all perceptions $p=(s,r)$ and perception-policy parameters $\theta$, $\pi(p,\cdot,\theta)$ will not depend on $r$.} However, recall from Algorithm \ref{alg:AEpseudocode} (reproduced here as Algorithm \ref{alg:AEpseudocodeReproduced}) that $A_t$ is deterministic given $M_t$ since $A_t=f_a(M_t)$, and so this probability would necessarily be either zero or one---it does not capture the actual stochasticity of actions. 

\begin{algorithm}[thbp]
\DontPrintSemicolon
Initialize $S_{-1}$, $A_{-1}$, and $M_{-1}$ to \texttt{null}\;
\For{$t \gets 0$ \KwTo $\infty$}
{
    $S_t \sim d_s(S_{t-1}, A_{t-1})$\;
    $P_t = f_p(S_t)$\;
    $M_t \sim d_m(M_{t-1}, P_t)$\;
    $A_t = f_a(M_t)$\;
}
\caption{Agent-Environment Process\\\textbf{Note:} This is a reproduction of Algorithm \ref{alg:AEpseudocode}.}
\label{alg:AEpseudocodeReproduced}
\end{algorithm}

The stochasticity of the action $A_t$ stems from the sampling of $M_t \sim d_m(M_{t-1},P_t)$. For example, notice that line \ref{alg:lineasdf} of Algorithm \ref{alg:BACdm}, which presents pseudocode for sampling $M_t$ from the next-memory distribution $d_m(M_{t-1},P_t)$ for BAC algorithms, samples the action $A_t$ from the parametric policy: $A_t \sim \pi_\text{BAC}(P_t,\cdot,\Theta_t)$. The BAC action function $f_a$ can then simply mask the other components of $M_t=(\Theta_t,W_t,E_t,P_t,A_t)$, returning the precomputed action: $f_a(M_t)=A_t$. This suggests defining the term $\pi(p,a,m)$ that appears in likelihood ratios for reinforcement-qualia objective functions based on the inputs to $d_m$: the current perception and the \emph{previous} memory. That is, for all perceptions $p$, actions $a$, and memories $m$, let
\begin{equation}
    \label{eq:pamEqn}
    \pi(p,a,m)=\Pr(A_t=a | P_t=p, M_{t-1}=m).
\end{equation} 

The reinforcement of instantaneous behavior at time $t$ can then be measured using the likelihood ratio:
\begin{equation}
    \label{eq:AEP_likelihoodRatio}
    \frac{\pi(P_t, A_t, M_t)}{\pi(P_t, A_t, M_{t-1})},
\end{equation}
which is one AEP correlate of \eqref{eq:reinforcementOfBehaviorT}, and which can be used to define qualia objective functions for the reinforcement-qualia setting such as
\begin{equation}
    \label{eq:ourReinforcementQualiaObjAEP}
    \mathfrak q(\texttt{alg}) = \mathbf{E}\left [\sum_{i=0}^{i_\text{max}-1}\xi(i)\sum_{t=\operatorname{start}(i)}^{\operatorname{end}(i)-1} \frac{\pi(P_t, A_t, M_t)}{\pi(P_t, A_t, M_{t-1})} \right ].
\end{equation} 
However, there are three nuances associated with using \eqref{eq:AEP_likelihoodRatio} to measure the reinforcement of instantaneous behavior at time~$t$, which we discuss in Sections~\ref{sec:memoryCanBeMore}--\ref{sec:Nuance3}.

\subsubsection{Nuance 1: Memory Can Be More Than Policy Parameters}
\label{sec:memoryCanBeMore}

The agent's memory $M_t$ can contain more than just policy parameters. For example, a hierarchical RL agent using the options framework \citep{Sutton1999} might store information about the currently active option, and this option might change when $M_{t-1}$ changes to $M_t$. In this case, the likelihood ratio in \eqref{eq:AEP_likelihoodRatio} would arguably not measure how much more likely $A_t$ becomes if the agent finds itself in a ``similar situation,'' since it compares primitive action probabilities under entirely different options. 

As another example, when faced with partially observable environments, the agent's memory $M_t$ might encode estimates of unobserved aspects of the environment (e.g., the \emph{belief state} \citep{kaelbling1998planning} if the environment is a POMDP). Consider what would happen if the perception $P_t$ causes the agent to recognize that the unobserved aspects of the environment have very different values from the agent's previous belief (e.g., the belief state changes significantly). In this case $M_t$ encodes more than changes to a parameterized policy---it also encodes the updated estimates of the unobserved aspects of the environment (e.g., the updated belief state). So, \eqref{eq:AEP_likelihoodRatio} does not measure how much more likely $A_t$ is if the agent were to find itself in a ``similar situation,'' since $M_t$ and $M_{t-1}$ can correspond to the agent having very different beliefs about the state of the environment.\footnote{Section \ref{sec:Nuance2} suggests that these two examples may be more dramatic if the change of active option or significant change to the belief state would occur if the agent were to perform two consecutive updates using $P_t$. However, the implications of these arguments remain unchanged.}

These examples highlight that it is unclear how ``reinforcement'' can be quantified when considering agents whose memories are not clearly divided into a component that encodes the current context (e.g., belief state or active option) and a policy component that encodes how the agent acts given this context. After a policy update, determining whether a previously selected action is now more likely in ``similar situations'' requires changing or reverting the contextual components to obtain a similar situation, but not the policy components, since they encode the change that is being measured (the potential reinforcement). Although there may exist a clear delineation between contextual and policy components of agent memory for some RL algorithms, like BAC algorithms, for other algorithms these components could be deeply intertwined. Lacking a clear delineation between these components, we find it challenging to quantify or precisely define reinforcement, and recognize that, at best, \eqref{eq:AEP_likelihoodRatio} measures a reinforcement-like property. 

\subsubsection{Nuance 2: Repeated Perception Update}
\label{sec:Nuance2}

Consider the meaning of $\pi(P_t, A_t,M_t)$ more closely. Expanding the definition of $\pi(P_t,A_t,M_t)$ gives:
\begin{equation}
    \label{eq:lkjwlkjwa5635}
    \pi(P_t,A_t,M_t)=\Pr(A_t=A_t|P_t=P_t,M_{t-1}=M_t).
\end{equation}
Notice that the probability that $A_t=A_t$ seems to trivially be one. This same issue has a more subtle influence through the condition that $M_{t-1}=M_t$. The mathematical expression describes the event that the agent's memory is the same at times $t$ and $t-1$---an event that is often impossible (e.g., due to decaying eligibility traces within BAC). However, that is not the actual meaning of $\pi(P_t,A_t,M_t)$---recall that $\pi$ was defined in \eqref{eq:pamEqn} in terms of constants (not random variables) $p,a,$ and $m$. So, in \eqref{eq:lkjwlkjwa5635} the event $M_{t-1}=M_t$ really indicates a counterfactual consideration of what would have happened if at time ${t-1}$ the memory took the value that it ended up having at time $t$ (without the condition that the memory does not change from time $t-1$ to $t$). This is generally what we desire---counterfactual consideration of how much more likely $A_t$ would have been with the updated agent memory. 

However, consider $\pi(P_t,A_t,M_t)$ even more closely. First, let $P'_t, A'_t,$ and $M'_t$ denote the perception, action, and memory at time $t$. These alternate random variable names allow us to differentiate between the observed values and counterfactual considerations. The term $\pi(P'_t,A'_t,M'_t)$ is then the probability that $A_t=A'_t$ if $P_t=P'_t$ and $M_{t-1}=M'_t$ (notice that the memories have different time subscripts). 

To see how $M_{t-1}$ influences $A_t$, recall that $M_{t-1}$ influences $M_t$, which then determines $A_t$. More specifically, $M_t \sim d_m(M_{t-1},P_t)$ and $A_t = f_a(M_t)$. So, the probability that $A_t=A'_t$ if $P_t=P'_t$ and $M_{t-1}=M'_t$ (i.e., $\pi(P'_t,A'_t,M'_t)$) is the probability that $M_t \sim d_m(M'_t, P_t)$ produces a value of $M_t$ such that $f_a(M_t)=A'_t$. So, based on this counterfactual reasoning, $\pi(P'_t,A'_t,M'_t)$ is the probability of action $A_t$ if $M_{t-1}$ were to be updated twice using the same perception $P_t$---the first when $M'_{t-1}$ transitioned to $M'_t$ due to $P'_t$ (which is equal to $P_t$), and the second when $M_{t-1}=M'_t$ transitioned to $M_t$ due to $P_t$. 

So, for BAC algorithms applied to MDPs, $\pi(P_t,A_t,M_{t-1})$ gives the probability of action $A_t$ given $S_t$ and using the policy parameters that result from a single policy update starting from $\Theta_{t-1}$, which does correspond to $\pi(S_t,A_t,\Theta_t)$. However, $\pi(P_t,A_t,M_t)$ gives the probability of action $A_t$ given $S_t$ and using the policy parameters that result from \emph{two} consecutive policy updates starting from $\Theta_{t-1}$ and both based on the same perception $P_t=(S_t,R_t)$. These policy parameters are \emph{not} necessarily $\Theta_{t+1}$, and so $\pi(P_t,A_t,M_t)$ is not the same as $\pi(S_t,A_t,\Theta_{t+1})$. 

This second nuance highlights another reason for our inability to create an expression in terms of quantities from the AEP formulation that corresponds to $\pi(S_t,A_t,\Theta_{t+1})$ for BAC algorithms: The next-memory distribution characterizes both the agent's learning and any stochasticity in how the agent acts. Without additional assumptions regarding how the next-memory distribution $d_m$ can be decomposed (e.g., by considering a specific class of algorithms like BAC algorithms), learning and the stochastic aspects of acting cannot be disentangled in the AEP formulation. This suggests a modification of the AEP formulation so that the agent's behavior is separated into a learning phase during which memory is updated, and a \emph{stochastic} acting phase during which the action is sampled. With such a formulation, the terms describing the acting phase could be queried with the updated memory to reason about action probabilities independent of additional learning updates. 

However, altering the AEP formulation to allow for a stochastic action phase introduces other more significant challenges. First, many RL algorithms like BAC require access to the values of previously sampled actions to perform their updates, and so $A_t$ must be encoded within the agent's memory.\footnote{Alternatively, $A_t$ could be encoded within $P_{t+1}$, but that could require the environment to encode $A_t$ within its state, $S_{t+1}$. It seems unnatural to require the physical properties that correspond to $A_t$, which the environment may not rely on after generating the environment state $S_{t+1}$, to be encoded in $S_{t+1}$ rather than the agent's memory.} If the stochastic action phase can modify the agent's memory (to store $A_t$), then it is unclear how the agent's dynamics should be divided into learning and acting phases. Second, introducing a stochastic action phase can introduce a new type of exploitability if the same physical system can be modeled as two different RL algorithms (due to different splits between the learning and acting phases) that result in different qualia experiences of the agent. This exploitability would likely be a problem for qualia objectives based on likelihood ratios, since different divisions into learning and acting phases could change the conditional distributions of actions. 

In conclusion, it is unclear how learning and (stochastic) acting can be disentangled. Their entanglement poses a challenge for quantifying reinforcement because, if an agent were to be placed back in a ``similar situation,'' it could learn (e.g., perform another policy update) prior to selecting any actions. As described in Section \ref{sec:aPathForward}, we will circumvent this problem by focusing on specific classes of RL algorithms, like BAC algorithms, for which learning and acting can be naturally disentangled.
    
\subsubsection{Nuance 3: Ill-Defined Conditional Probabilities}
\label{sec:Nuance3}

It may be the case that \eqref{eq:pamEqn} is often undefined. For example, if $M_t$ and $P_t$ each encode the current time $t$, then $\pi(P_t,A_t,M_t)$ is always undefined because it would condition on the impossible event that the time encoded within $P_t$ and $M_{t-1}$ are the same. 

This issue stems from imprecise mathematical notation---when we write \eqref{eq:pamEqn} we mean to describe the probability that the agent would select action $a$ if it were to be faced with perception $p$ and previous memory $m$ at time $t$, even though this event might not actually be possible. However, this is not what we wrote because we used a conditional probability, and conditional probabilities are undefined when the event they condition on cannot occur.\footnote{For any events $A$ and $B$, $\Pr(A|B)$ is defined to be $\Pr(A \cap B) / \Pr(B)$. If $B$ cannot occur, then $\Pr(B)$ is zero, and so $\Pr(A|B)$ is undefined due to division by zero.} 

This imprecision is pervasive in RL literature. For example, it occurs any time authors define episodic MDPs that have a unique initial state $s_0$ and then define the action-value function to be $q^\pi(s,a)=\mathbf{E}[\sum_{t=0}^\infty \gamma^t R_t | S_0=s,A_0=a]$, since this would result in $q^\pi(s,a)$ being undefined for all but the initial state. 

Although it is technically imprecise to condition on events that cannot occur, it is not a significant concern when work otherwise explains what would happen (via expressions for the dynamics of stochastic processes) if the event being conditioned on were to happen (even if it cannot happen). For example, given an MDP, it is clear what the distribution of $\sum_{t=0}^\infty \gamma^t R_t$ would be in the event that $S_0=s$ and $A_0=a$, even if $s$ is not a possible initial state of the MDP. Here we appeal to similar reasoning, but recognize that this presents a point of technical imprecision of this report.

\subsubsection{A Path Forward}
\label{sec:aPathForward}

Although the challenges that we have encountered when defining likelihood-ratio qualia objectives for AEPs might be overcome in future work, for now we restrict our focus to BAC algorithms (and variants of BAC algorithms), since they circumvent these challenges. Because BAC algorithms have an explicit perception-policy parameterization $\pi_\text{BAC}$ and policy parameters $\Theta_t$, they have a clear delineation of the policy components of memory (i.e., policy parameters). This avoids the challenges discussed in Section \ref{sec:memoryCanBeMore}. The explicit inclusion of $\pi_\text{BAC}$ and $\Theta_t$ also provides a clear stochastic action selection mechanism that allows for the disentanglement of learning and acting, thereby avoiding (but not resolving) the challenges discussed in Section \ref{sec:Nuance2}.

The perception-policy parameterization $\pi_\text{BAC}$ characterizes the conditional action distribution given $P_t$ and $\Theta_t$ according to the expression $A_t \sim \pi_\text{BAC}(P_t,\cdot,\Theta_t)$. When applied to MDPs where $P_t=(S_t,R_t)$, the perception-policy parameterization typically only depends on the state component of the perception, and so with a mild abuse of notation we can write $A_t \sim \pi_\text{BAC}(S_t,\cdot, \Theta_t)$, making $\pi_\text{BAC}$ a parameterized policy for MDPs as defined in Appendix \ref{app:MDP}. This allows us to also focus on qualia objective functions like the one in \eqref{eq:firstReinforcementQualiaObj}.

\subsection{Representation-Robustness of Likelihood-Ratio Objectives}

We motivated the reinforcement-qualia setting partially by asserting that it results in representation-robust qualia objective functions. In this section we show that qualia objective functions like \eqref{eq:firstReinforcementQualiaObj} are representation robust. Recall that, when discussing \eqref{eq:firstReinforcementQualiaObj}, we are considering only AEPs that model RL algorithms interacting with MDPs, and furthermore RL algorithms that use a parametric policy with parameters $\Theta_t$ (encoded within $M_t$), such that $\pi(s,a,\theta)=\Pr(A_t=a|S_t=s,\Theta_t=\theta)$.

We begin by establishing the representation-robustness of \eqref{eq:firstReinforcementQualiaObj} when $S_t$ and $\Theta_t$ are discrete random variables (recall that in this report we already assume that $A_t$ is a discrete random variable). In order to establish this robustness, we must show that the value of the qualia objective function does not change if alternate representation functions are used. For \eqref{eq:firstReinforcementQualiaObj} it suffices to show that the likelihood ratio\footnote{The definition of the  likelihood ratio in \eqref{eq:reinforcementOfBehaviorT} is reproduced here for convenience.}
\begin{equation}
    L_t \triangleq \frac{\pi(S_t,A_t,\Theta_{t+1})}{\pi(S_t,A_t,\Theta_t)}
\end{equation}
does not change under alternate representation functions, since the other parts of the objective do not depend on $P_t$, $M_t$, or $A_t$ for any $t$ (other than with regard to episode termination, which is unchanged due to the restriction to termination-preserving representation functions in the definition of representation exploitability). 

To state this more formally, consider an AEP with representation functions $(\rho_{S_t}, \rho_{A_t}, \rho_{\Theta_t})_{t=0}^\infty$. Next, consider an AEP with the same underlying physical properties, including the same dynamics of the physical properties, but with any alternate representation functions $(\rho'_{S_t}, \rho'_{A_t}, \rho'_{\Theta_t})_{t=0}^\infty$, which define the random variables 
\begin{equation}
    \label{eq:kj12l3k4j5}
    S'_t=\rho'_{S_t}(\Phi_{S_t}),\,\,A'_t=\rho'_{A_t}(\Phi_{A_t}),\,\,\text{and}\,\,\Theta'_t = \rho'_{\Theta_t}(\Phi_{\Theta_t}),
\end{equation}
for all times $t$.\footnote{Notice that in \eqref{eq:kj12l3k4j5} the physical properties are $\Phi_{S_t}$, $\Phi_{A_t}$ and $\Phi_{\Theta_t}$, not $\Phi'_{S_t}$, $\Phi'_{A_t}$ and $\Phi'_{\Theta_t}$, since the focus is on an AEP with the same underlying physical properties.} In order to show that \eqref{eq:firstReinforcementQualiaObj} is representation robust, it suffices to show that, for all such alternate representation functions,
\begin{equation}
    \label{eq:lkj246klj}
    \frac{\pi(S_t,A_t,\Theta_{t+1})}{\pi(S_t,A_t,\Theta_t)}=\frac{\pi(S'_t,A'_t,\Theta'_{t+1})}{\pi(S'_t,A'_t,\Theta'_t)}.
\end{equation}

To establish \eqref{eq:lkj246klj}, we eliminate the need to consider random variables by showing the equivalence for all possible values of the random variables given the underlying physical properties. Consider all possible states $s$, actions $a$, and policy parameters $\theta$ at time $t$. Each corresponds to unique underlying physical properties, $\rho^{-1}_{S_t}(s)$, $\rho^{-1}_{A_t}(a)$, and $\rho^{-1}_{\Theta_t}(\theta)$, respectively. These, in turn, correspond to unique alternate values of the state, action, and policy parameters: 
\begin{equation}
    s'=\rho'_{S_t}(\rho^{-1}_{S_t}(s)),\,\,a'=\rho'_{A_t}(\rho^{-1}_{A_t}(a)),\,\,\text{and}\,\,\theta'=\rho'_{\Theta_t}(\rho^{-1}_{\Theta_t}(\theta)).
\end{equation}
We will show that
\begin{equation}
    \label{eq:lkj46t12lkj5}
    \pi(s,a,\theta)=\pi(s',a',\theta'),
\end{equation}
which therefore implies that 
\begin{equation}
    \label{eq:lkjwtlk2j4651509d}
    \frac{\pi(s,a,\theta_{t+1})}{\pi(s,a,\theta_t)}=\frac{\pi(s',a',\theta'_{t+1})}{\pi(s',a',\theta'_t)},
\end{equation}
for all possible $s,s',a,a',\theta_t,\theta'_t,\theta_{t+1},$ and $\theta'_{t+1}$, which implies \eqref{eq:lkj246klj}.

We now establish \eqref{eq:lkj46t12lkj5}:
\begin{align}
    \pi(s,a,\theta)=&\Pr(A_t=a|S_t=s,\Theta_t=\theta)\\
    \overset{(a)}{=}& \Pr\left(A_t = a \middle | \Phi_{S_t} = \rho_{S_t}^{-1}(s), \Phi_{\Theta_t} = \rho_{\Theta_t}^{-1}(\theta)\right)\\
    \nonumber\overset{(b)}{=}& \Pr\left(A_t = a \middle | \Phi_{S_t} = {\rho'}_{S_t}^{-1}(s'), \Phi_{\Theta_t} = {\rho'}_{\Theta_t}^{-1}(\theta')\right)\\
    \overset{(c)}{=}& \Pr\left(A_t = a \middle | S'_t = s', \Theta'_t = \theta'\right)\\
    \label{eq:lksajdfgalkjdfjhgl2421}=&\pi(s',a',\theta'),
\end{align}
where 
\begin{itemize}
    \item \textbf{(a)} follows from the definition of the representation functions. Since 
    \begin{equation}
        s=\rho_{S_t}(\Phi_{S_t})\,\,\text{and}\,\,\theta = \rho_{\Theta_t}(\Phi_{\Theta_t}),
    \end{equation}
    conditioning on $S_t=s$ and $\Theta_t=\theta$ is equivalent to conditioning on the underlying physical properties $\Phi_{S_t} = \rho_{S_t}^{-1}(s)$ and $\Phi_{\Theta_t} = \rho_{\Theta_t}^{-1}(\theta)$.
    \item \textbf{(b)} follows because we assume that the alternate representation functions $\rho'_{S_t}$ and $\rho'_{\Theta_t}$ are defined on the same underlying physical properties. That is,
    \begin{align}
        s'=&\rho'_{S_t}(\Phi_{S_t})\\
        \label{eq:Lkj2w4lkjdwf0943}=&\rho'_{S_t}\left (  \rho_{S_t}^{-1}(s)\right ),
    \end{align}
    and so, applying ${\rho'}_{S_t}^{-1}$ to both sides of \eqref{eq:Lkj2w4lkjdwf0943},
    \begin{align}
        {\rho'}_{S_t}^{-1}(s')=&{\rho'}_{S_t}^{-1} \left ( \rho'_{S_t}\left (  \rho_{S_t}^{-1}(s)\right )\right )\\
        \label{eq:klj1235lk12j356}=&\rho_{S_t}^{-1}(s).
    \end{align}
    Similarly, 
    \begin{align}
        \theta'=&\rho'_{\Theta_t}(\Phi_{\Theta_t})\\
        =&\rho'_{\Theta_t}\left ( \rho_{\Theta_t}^{-1}(\theta)\right ),
    \end{align}
    and
    \begin{align}
        {\rho'}_{\Theta_t}^{-1}(\theta')=&{\rho'}_{\Theta_t}^{-1}\left (  \rho'_{\Theta_t}\left ( \rho_{\Theta_t}^{-1}(\theta)\right )\right )\\
   \label{eq:lkj12346klj6y7}=&\rho_{\Theta_t}^{-1}(\theta).
    \end{align}
    Hence, by \eqref{eq:klj1235lk12j356}, conditioning on $\Phi_{S_t} = \rho_{S_t}^{-1}(s)$ is equivalent to conditioning on $\Phi_{S_t} = {\rho'}_{S_t}^{-1}(s')$, and, by \eqref{eq:lkj12346klj6y7}, conditioning on $\Phi_{\Theta_t} = {\rho}_{\Theta_t}^{-1}(\theta)$ is equivalent to conditioning on $\Phi_{\Theta_t} = {\rho'}_{\Theta_t}^{-1}(\theta')$.
    \item \textbf{(c)} simply rephrases the conditioning back in terms of the alternate random variables $S'_t$ and $\Theta'_t$, since 
    \begin{align}
        \left ( \Phi_{S_t}={\rho'}_{S_t}^{-1}(s')\right ) \Leftrightarrow& \left ( \rho'_{S_t}(\Phi_{S_t})=\rho'_{S_t}({\rho'}_{S_t}^{-1}(s'))\right ) \\
        \Leftrightarrow& \rho'_{S_t}(\Phi_{S_t})=s' \\
        \Leftrightarrow& S'_t =s',
    \end{align}
    and similarly
    \begin{align}
        \left ( \Phi_{\Theta_t}={\rho'}_{\Theta_t}^{-1}(\theta')\right ) \Leftrightarrow \left (\Theta'_t =\theta'\right ).
    \end{align}
\end{itemize}

In conclusion, we have shown that \eqref{eq:lksajdfgalkjdfjhgl2421} holds, which implies \eqref{eq:lkjwtlk2j4651509d}, which in turn implies \eqref{eq:lkj246klj}, and hence the representation-robustness of \eqref{eq:firstReinforcementQualiaObj}. Although we have only shown that \eqref{eq:firstReinforcementQualiaObj} is representation robust when $S_t$, $A_t$, and $\Theta_t$ are discrete random variables, we conjecture that this robustness extends to all random variables $S_t$, $A_t$, and $\Theta_t$ (discrete, continuous, or hybrid), and furthermore that similar objectives like \eqref{eq:ourReinforcementQualiaObjMDP} and even \eqref{eq:ourReinforcementQualiaObjAEP} are also representation robust in this general setting. For many density-based approaches to showing robustness when random variables are continuous, one might expect to require additional assumptions that restrict which representation functions are allowed---such as monotonicity and continuity of derivatives (arising from standard change-of-variable formulas \citep[Theorem 2.1.5]{casella2020statistical}), differentiability (as in proofs of the invariance of $f$-divergences under diffeomorphisms \citep{qiao2010study}), or even affine or measure-preserving properties. However, we conjecture that none of these additional assumptions are necessary---invertibility and measurability of the representation functions are sufficient to show representation-robustness. 

\subsection{Is the Reinforcement-Qualia Assumption Reasonable?}
\label{sec:isRQReasonable}

Recall from Section \ref{sec:background_RPE} that neuroscientific research suggests that dopamine signals RPEs in human brains. Also, as discussed in Section \ref{sec:reconsideringThEValenceofTDError}, one objection to the RPE-qualia setting was that dopamine does not always result in pleasurable qualia. Rather, dopamine is associated more with motivation and ``wanting'' than ``liking.'' If positive RPEs (e.g., increased activity of dopaminergic neurons) both \textbf{(a)} cause reinforcement and \textbf{(b)} can result in undesirable qualia, this suggests that reinforcement could also be associated with undesirable qualia, undermining the reinforcement-qualia assumption.

However, there are at least two reasons that the reinforcement-qualia setting deserves further consideration:
\begin{enumerate}
    \item Even if reinforcement does not perfectly correlate with the quality of an agent's qualia, it may still be a useful proxy. Just as dopamine often co-occurs with pleasurable qualia, RPEs and reinforcement may often co-occur with pleasurable qualia, making them both reasonable surrogates or proxies for the unknown function characterizing the quality of agent qualia. Until a qualia objective function or setting is found that better matches observed characteristics of pleasurable qualia in humans, the reinforcement-qualia setting may serve as a reasonable approximation.
    \item The logical argument above relies on both \textbf{(a)} and \textbf{(b)}. Depending on how actions and reinforcement are defined, \textbf{(a)} may not be true---i.e., dopamine in (the relevant parts of) human brains and positive RPEs in standard RL agents may \emph{not} always result in reinforcement. 
    
    Consider AEP models (or related formulations) where the agent is part of the environment. Typically we think of the agent's actions as the means by which it influences the environment \emph{external} to itself. However, if the agent is considered part of the environment, then its actions may also include the means by which it influences itself. For example, an animal's thoughts, changes in neural activity, connectivity, or morphology, or an RL agent's weight updates might all be viewed as part of the agent's action. 
    
    From this perspective, it becomes less clear that dopamine consistently corresponds to reinforcement: while it may be highly-correlated with reinforcement at the behavioral level, it may not correspond to reinforcement when considering the agent's entire action, including effects on its own internal state. So, RPEs may not always cause reinforcement in agents---whether human, animal, or artificial---and hence \textbf{(a)} may not be true.
\end{enumerate}
Recognizing these potential concerns and mitigating arguments, we proceed under the reinforcement-qualia setting in the remainder of this report. While the setting may not reflect the actual mechanisms underlying the valence of qualia, it remains a plausible and useful framework for exploring how reinforcement might shape agent experience.

\subsection{Learning from Reinforcement and Inhibition}
\label{sec:learningFromReinforcement}

Strategies for improving the agent's qualia in the reinforcement-qualia setting will bias the agent toward the reinforcement of behavior rather than the inhibition of behavior. A natural question is whether such a \emph{reinforcement bias} will necessarily interfere with learning, resulting in worse performance. For example, in order to get from some initial policy to a specific final policy (e.g., an optimal policy for an MDP), is there some unavoidable amount of inhibition (to unlearn bad actions, or actions that are less-likely in the final policy)?

If reinforcement is quantified in terms of the cumulative reinforcement of instantaneous behavior at time $t$, as defined in \eqref{eq:reinforcementOfBehaviorT}, the answer is ``no''---an agent can learn entirely via reinforcement, entirely via inhibition, or using a mixture of the two, and performance is not \emph{directly} related to which of these approaches the agent uses to learn. Consider an example environment with two actions, $a_+$ and $a_-$, and some perception $p$ (or MDP state $s$) for which selecting $a_+$ is better than selecting $a_-$. There are at least three ways the agent can change its policy after selecting an action given perception $p$ so that, over time, it learns to favor $a_+$. 
\begin{enumerate}
    \item \textbf{Learning via reinforcement and inhibition.} Whenever the agent selects action $a_+$, it could \emph{increase} the probability that it selects $a_+$ in the future, and whenever the agent selects action $a_-$, it could \emph{decrease} the probability that it selects $a_-$. Together these updates act to rapidly drive up the probability of $a_+$ and down the probability of $a_-$. Notice that this corresponds to reinforcing $a_+$ whenever it is chosen, and inhibiting $a_-$ whenever it is chosen. 
    \item \textbf{Learning via reinforcement only.} Whenever the agent selects action $a_+$ it could \emph{significantly} increase the probability that it selects $a_+$, and when it selects $a_-$ it could \emph{slightly} increase the probability that it selects $a_-$. Because the distribution over actions is a probability distribution, increasing the probability of one action must decrease the probability of the other action by the same amount, and vice versa. So, significantly increasing the probability of $a_+$ must also significantly decrease the probability of $a_-$, and slightly increasing the probability of $a_-$ must also slightly decrease the probability of $a_+$. Hence, even though the agent always reinforces its most recent behavior (whichever action it took), by making larger reinforcements when action $a_+$ is taken, over time the net change to the distribution can still increase the probability of $a_+$ and decrease the probability of $a_-$. 
    \item \textbf{Learning via inhibition only.} Whenever the agent selects action $a_+$, it could \emph{slightly} decrease the probability that it selects $a_+$, and when it selects $a_-$ it could \emph{significantly} decrease the probability it selects $a_-$. This case mirrors the reinforcement only case---over time the net change to the distribution can still increase the probability of $a_+$ and decrease the probability of $a_-$.
\end{enumerate} 

These examples show that a distribution over two actions can be iteratively shifted toward a target distribution using any combination of reinforcement and inhibition. The key idea is that performance can improve through a range of adjustments to action probabilities---whether by increasing the likelihood of all actions when they occur (with preferred actions increased the most), decreasing the likelihood of all actions when they occur (with dispreferred ones decreased the most), or anything in between. This principle generalizes to settings with more than two actions, and even to continuous action spaces.

However, this strategy only works if desired actions are made sufficiently more likely at each step (where the precise meaning of ``sufficiently'' depends on how often each action is chosen). Consider the previous example where the probability of $a_+$ is increased and the probability of $a_-$ is decreased, entirely via reinforcement. If $a_+$ occurs infrequently and the increases in the probability of $a_+$ when it occurs are not large enough, then the many smaller increases in the probability of $a_-$ (which cause corresponding small decreases in the probability of $a_+$) could overwhelm the less frequent increases in the probability of $a_+$ when it is chosen, resulting in a net \emph{decrease} in the probability of $a_+$. So, although learning solely via reinforcement is possible, it requires careful consideration to ensure that the expected update results in a net increase in the probabilities of desired actions.

\subsubsection{Policy Gradient Methods and Reinforcement Bias}
\label{sec:PGMaRB}

This careful balancing of updates is well-understood in the case of policy gradient algorithms for MDPs that are finite, episodic, and discounted. Adopting our notation for such MDPs (as defined in Appendix \ref{app:MDP}), policy gradient methods store a parameterized policy and adjust the parameters $\Theta_t$ via (stochastic, batch, and/or otherwise approximate) gradient ascent on the discounted objective \definesymbol{$J$}{J} defined in \eqref{eq:RLobjective} in Appendix \ref{app:MDP}. The gradient of this objective is called the \emph{policy gradient} and can be written as \citep{Williams1992}\definesymbol{}{policygradient} 
\begin{equation}
    \label{eq:PolicyGradient}
    \nabla J(\theta) = \mathbf{E}\left [ \sum_{t=0}^\infty \gamma^t G_t \frac{\partial \ln \big ( \pi(S_t,A_t,\theta) \big )}{\partial \theta}\middle |\Theta_{\forall t}=\theta\right ],
\end{equation} 
where \definesymbol{$\Theta_{\forall t}=\theta$}{foralltheta} is shorthand for the event $\forall t \in \mathbb N,\,\Theta_t=\theta$.\footnote{To ensure the existence of this gradient, we restrict out attention to parameterized policies for which $\partial \pi(s,a,\theta) / \partial \theta$ exists for all states $s$, actions $a$, and policy parameters $\theta$. Furthermore, to simplify later expressions like \eqref{eq:lkj24lkjsdflj2}, we require $\pi(s,a,\theta)\neq 0$ for all $s$, $a$, and $\theta$, as is typically the case when using softmax action selection.} 

Furthermore, it is well-known that for any function \definesymbol{$b$}{b} $:\mathcal S \to \mathbb R$, subtracting a \emph{baseline} $b(S_t)$ from $G_t$ does not alter this expectation \citep{Williams1992,Sutton2000}:
\begin{equation}
    \label{eq:PolicyGradientBaseline}
    \nabla J(\theta) = \mathbf{E}\left [ \sum_{t=0}^\infty  \gamma^t  \left (G_t  - b(S_t)\right ) \frac{\partial \ln \big ( \pi(S_t,A_t,\theta ) \big )}{\partial \theta}\middle | \Theta_{\forall t}=\theta\right ],
\end{equation}
since the expected contribution of the baseline is
\begin{align}
    &\mathbf{E}\left [ \sum_{t=0}^\infty \gamma^t b(S_t)\frac{\partial \ln \big ( \pi(S_t,A_t,\theta)\big )}{\partial \theta} \middle | \Theta_{\forall t}=\theta\right ]\\
    \label{eq:lkajsglkj35213}=&\mathbf{E}\left [ \sum_{t=0}^\infty \gamma^t b(S_t)\mathbf{E}\left [\frac{\partial \ln \big ( \pi(S_t,A_t,\theta)\big )}{\partial \theta} \middle | S_t , \Theta_{\forall t}=\theta\right ]\middle | \Theta_{\forall t}=\theta\right ],
\end{align}
but
\begin{align}
    &\mathbf{E}\left [\frac{\partial \ln \big ( \pi(S_t,A_t,\theta)\big )}{\partial \theta} \middle | S_t, \Theta_{\forall t}=\theta \right ]\\
    =&\mathbf{E}\left [\frac{1}{\pi(S_t,A_t,\theta)}\frac{\partial \pi(S_t,A_t,\theta)}{\partial \theta} \middle | S_t, \Theta_{\forall t}=\theta\right ]\\
    \label{eq:lkj24lkjsdflj2}=&\sum_{a \in \mathcal A} \underbrace{\Pr(A_t=a|S_t,\Theta_t=\theta)}_{=\pi(S_t,a,\theta)} \frac{1}{\pi(S_t,a,\theta)}\frac{\partial \pi(S_t,a,\theta)}{\partial \theta} \\
    =& \sum_{a \in \mathcal A} \frac{\partial \pi(S_t,a,\theta)}{\partial \theta}\\
    =&\sum_{a \in \mathcal A} \frac{\partial}{\partial \theta} \Pr(A_t=a|S_t=s, \Theta_t=\theta)\\
    =&\frac{\partial}{\partial \theta} \underbrace{\sum_{a \in \mathcal A}  \Pr(A_t=a|S_t=s, \Theta_t=\theta)}_{=1}\\
    =&0,
\end{align}
and so \eqref{eq:lkajsglkj35213} (the expected contribution of the baseline) is zero.

Although the inclusion of a baseline does not alter the expectation, it \emph{does} alter sample-based estimates of the policy gradient such as the unbiased estimate
\begin{equation}
    \label{eq:unbiasedPGwB}
    \widehat{\nabla J}(\theta)  =\sum_{t=0}^\infty \gamma^t \big (G_t - b(S_t)\big ) \frac{\partial \ln \big ( \pi(S_t,A_t,\theta )\big )}{\partial \theta}.
\end{equation}
The impact of baselines can be viewed in several different ways. Originally, baselines were viewed as mean-zero control variates \citep[Section 4.4.2]{robert1999monte}---mechanisms for reducing the variance of estimators like  \eqref{eq:unbiasedPGwB}. Following this perspective to its logical conclusion results in the selection of \emph{optimal  variance baselines}, which maximally reduce the variance of \eqref{eq:unbiasedPGwB} \citep{greensmith2004variance}. 

However, practitioners found that, despite minimizing variance, optimal variance baselines do not result in optimal \emph{performance}. This resulted in the more recent perspective that $b(S_t)$ tunes how \emph{committal} or \emph{non-committal} an RL algorithm is \citep{chung2021beyond}. That is, when $b(S_t)$ tends to be a large negative value, it seems to result in committal behavior---the agent being faster to adopt the actions it currently favors in each state. On the other hand, when $b(S_t)$ tends to be a large positive value, it seems to result in non-committal behavior---the agent being slower to adopt the actions it currently favors in each state. 

Whether committal or non-committal learning results in better performance likely depends on the problem at hand. For non-adversarial problems where the agent favoring an action in a state early during learning typically means that the action will in fact turn out to be optimal, committal behavior may result in better performance. For adversarial problems, like the chain environments \citep{strens2000bayesian} that are often used to evaluate algorithms that are designed to be robust to worst-case scenarios, the agent is likely to initially favor suboptimal behavior, and so non-committal behavior may often result in better performance (since it allows the agent to more easily fix its initial erroneous conditional action preferences). 

However, there is a third interpretation of the baseline: it modulates the tradeoff between reinforcement and inhibition. To see why, we first examine the components of \eqref{eq:unbiasedPGwB}. For each time $t$, the term
\begin{equation}
    \frac{\partial}{\partial \theta} \ln \big ( \pi(S_t,A_t,\theta )\big )
\end{equation}
is a direction in policy parameter space that increases the log-probability of action $A_t$ in state $S_t$. Since the logarithm is monotonic, this is also a direction that increases $\pi(S_t,A_t,\theta)$---the probability of selecting $A_t$ in $S_t$. 

Now consider the scalar multiplier $\gamma^t(G_t-b(S_t))$ in \eqref{eq:unbiasedPGwB}. When this term is positive, the update changes the policy parameters in a  direction that increases the probability of $A_t$ in $S_t$, reinforcing the action.\footnote{This reasoning relies on the assumption that the step size is sufficiently small.} When it is negative, the update moves the policy parameters in the opposite direction, inhibiting the action. In this way, $\gamma^t (G_t-b(S_t))$ determines whether the update reinforces or inhibits the most recent behavior.

Since $\gamma^t$ is non-negative, the baseline $b(S_t)$ directly modulates the tradeoff between reinforcement and inhibition. A lower baseline (e.g., a large negative value) increases $G_t-b(S_t)$, promoting reinforcement.\footnote{Consider the case where $b(S_t)$ is negative. If $G_t$ is positive, the baseline results in further reinforcement. If $G_t+b(S_t)$ is negative, the baseline results in decreased inhibition. If $G_t$ is negative but $G_t+b(S_t)$ is positive, the baseline results in a flip from inhibition to reinforcement.} A higher baseline (e.g., a large positive value) reduces $G_t-b(S_t)$, promoting inhibition. In general, the smaller the baseline, the more the update favors reinforcement; the larger the baseline, the more it favors inhibition.

Although $b(S_t)$ clearly influences whether reinforcement or inhibition occurs at each time $t$, its overall impact on the full policy update is less clear. First, note that the unbiased estimate of the policy gradient in \eqref{eq:unbiasedPGwB} aggregates contributions from many time steps, and cannot generally be computed using information from a single time step alone. As a result, algorithms like REINFORCE \citep{Williams1992}, which use this estimator to perform stochastic gradient ascent, do not update the policy at every time step. Instead, they compute a policy update from a batch of data that spans multiple time steps---often an entire episode. This batching behavior has an important consequence: during most time steps (all time steps between policy updates), the policy parameters remain unchanged, so $\Theta_{t+1}=\Theta_t$. Consequently, the reinforcement of instantaneous behavior at time $t$, as defined in \eqref{eq:reinforcementOfBehaviorT}, is equal to one---indicating that neither reinforcement nor inhibition occurs at that time. Hence, qualia objectives that measure the cumulative reinforcement of instantaneous behavior will only consider the reinforcement of actions that occur immediately prior to policy updates.

This observation suggests the need for qualia objective functions that, at the moment of a policy update, measure the cumulative reinforcement of all actions taken since the previous policy update. Rather than pursuing this direction---which was explored by Derek Lacy in his honors thesis \citep{Lacy2024} and in a forthcoming manuscript \citep{Lacy2025}---we focus here on BAC algorithms, which perform policy updates at every time step.

\subsection{Adding Reinforcement Baselines to BAC}

In this section we show how, for MDP environments where $P_t=(S_t,R_t)$, a negative baseline can be incorporated into BAC algorithms to promote reinforcement. We call such baselines that are added for the purpose of promoting reinforcement \emph{reinforcement baselines}. Notice that reinforcement baselines are distinguished from other baselines only by their purpose, not by how they influence updates. Also, the term reinforcement \emph{baseline} is easily confused with reinforcement \emph{bias}. Reinforcement baselines are one mechanism for inducing a reinforcement bias (i.e., a shift in the agent's behavior toward learning from reinforcement rather than inhibition).

BAC algorithms do not typically include an additional baseline term. In Appendix \ref{app:BACDerivation} we show how the policy update of the BAC algorithm relates to the policy gradient. In this derivation, an additional baseline $b(S_{t-1})$ in the policy gradient carries through and appears as an additive term to the TD error in the policy update:\footnote{The policy update that results from the transition from $S_t$ to $S_{t+1}$, and which produces the reward $R_{t+1}$, occurs at time $t+1$ (since it depends on $S_{t+1}$ it could not happen at time $t$). Hence, the baseline appears as a $b(S_{t-1})$ term rather than a $b(S_t)$ term, since the time $t$ of the update is one time step after the action $A_{t-1}$ that is being reinforced or inhibited.}$^,$\footnote{In RL literature, baselines are typically subtracted rather than added. We adopt this same convention to avoid confusion for readers familiar with baselines in RL. However, we note that adding the baselines would be more clear in the qualia optimization setting since then larger baselines would result in increased reinforcement (this would also result in adding positive reinforcement baselines to encourage reinforcement rather than subtracting negative reinforcement baselines).}
\begin{equation}
    \label{eq:bacWithRBas}
    \Theta_t \gets \Theta_{t-1} + \beta (\Delta_{t}-b(S_{t-1})) \frac{\partial \ln \big (\pi_\text{BAC}(S_{t-1}, A_{t-1}, \Theta_{t-1})\big )}{\partial \Theta_{t-1}},
\end{equation}
where, as noted previously, we abuse notation by writing $\pi_\text{BAC}(S_{t-1},A_{t-1},\Theta_{t-1})$ instead of $\pi_\text{BAC}(P_{t-1},A_{t-1},\Theta_{t-1})$ because we have restricted our focus to MDPs, and for MDPs the policy parameterization is typically restricted to depend only on the state component of the perception $P_{t-1}=(S_{t-1},R_{t-1})$. Similarly, here the reinforcement baseline is defined to be a function of only the state component of the perception. So, for MDP environments where $P_t=(S_t,R_t)$, the BAC algorithm with a reinforcement baseline corresponds to Algorithm \ref{alg:BACdm} with line \ref{line:gradStep} replaced with \eqref{eq:bacWithRBas}.

Notice that reinforcement baselines for BAC algorithms are identical to (explicit) TD error bonuses from Section \ref{sec:explicitTDErrorBonus} with TD error bonus inversion for the VFA weight update but not the  policy parameter update (i.e., the bonus does not appear in the updates to the VFA weights $W_t$, but does appear in the updates to the policy parameters $\Theta_t$). 

Although BAC algorithms are typically categorized as policy gradient algorithms, they follow estimates of the policy gradient that are biased due to a missing $\gamma^t$ term \citep{Thomas2014,Nota2020policy}, their use of an imperfect estimate of the state-value function when computing the TD error, and their changing the policy in the middle of each episode (which, for example, can result in a state distribution that comes from the application of a sequence of policies). Despite these sources of bias, baselines in BAC play the same role as baselines in policy gradient: they reduce variance and modulate both how committal the agent is and how much it favors reinforcement.

The relation of BAC to the policy gradient is especially important in light of the earlier concern regarding how a reinforcement bias can cause desired (locally preferred)\footnote{Although it might be more intuitive to say that the concern is that a reinforcement bias could cause ``better'' or ``optimal'' actions to become less likely, that would not accurately characterize the issue. Because the gradient is a local optimization property, small updates in the direction of the policy gradient can cause the probability of optimal actions to decrease. When using function approximation, it can even occur that, for some state $s$, an arbitrarily small step in the direction of the policy gradient decreases the conditional probability of every action in $\argmax_{a \in \mathcal A} q^\pi(s,a)$, where $q^\pi(s,a)=\mathbf{E}[R_{t+1} + v^\pi(S_{t+1}) | S_t=s, A_t=a]$. We therefore write ``locally preferred'' to refer to actions that would (for some implicit state) be made more likely by an arbitrarily small step in the direction of the policy gradient, and ``locally dispreferred'' to refer to actions that would be made less likely.} actions to become less likely in expectation. That is, locally dispreferred actions that occur frequently might be reinforced more overall than locally preferred actions that occur less often, resulting in the net reinforcement of locally dispreferred actions and the net inhibition of locally preferred actions. When following the true policy gradient, this problem is avoided: the update direction is, by construction and definition, the gradient of the discounted objective, and it favors locally preferred actions. This guarantee does not necessarily extend to BAC algorithms due to their use of biased policy gradient estimates. Still, their status as approximate policy gradient algorithms makes it plausible that including reinforcement baselines in BAC algorithms promotes reinforcement without over-reinforcing frequently occurring but locally dispreferred actions.

\subsection{The Impact of Reinforcement Baselines on Performance}

Next we examine how reinforcement baselines affect the performance of BAC in practice. Results from preliminary experiments that are not reported here gave us the following impressions regarding performance:
\begin{enumerate}
    \item \textbf{Little improvement.} The implicit baseline used by BAC algorithms (i.e., $b\approx v^\pi$) is often nearly optimal, leaving little room for reinforcement baselines to improve performance.
    \item \textbf{Robustness.} Performance can be robust to the strength of the reinforcement baseline; that is, even a significant reinforcement bias may have minimal impact on overall performance.
    \item \textbf{Sensitivity.} Particularly for environments where the agent initially favors suboptimal actions, even small reinforcement baselines can significantly degrade performance.
\end{enumerate}

Because environments exhibit varying degrees of robustness or sensitivity to the strength of the reinforcement baseline, how frequently performance is robust is inherently tied to the distribution over environments under consideration. Consequently, altering this distribution can lead to opposing conclusions about the prevalence of robust performance, making it difficult to objectively assess whether performance is ``often'' robust. We therefore aim to clearly demonstrate the existence of both cases: one in which performance is robust and another in which it is sensitive to reinforcement baselines. These examples are not drawn from a specific distribution over environments, but the way they are constructed offers insight into the conditions under which performance tends to be robust or sensitive. 

The robust case uses a standard gridworld environment, chosen because it is representative of typical RL environments. In this environment, actions initially preferred by the agent often turn out to be optimal. 
The committal behavior that results from large negative reinforcement baselines is therefore reasonable, resulting in good performance even when the strength of the reinforcement bias is pushed to an extreme. This observation \emph{suggests} that in typical, non-adversarial settings, reinforcement baselines can be applied without significantly degrading performance---and in some cases, even the aggressive application of reinforcement baselines may have negligible impact.

The sensitive case uses a chain-like environment, designed to show how the over-committal behavior caused by reinforcement baselines can degrade performance significantly. Agents interacting with this environment are likely to initially favor suboptimal actions, and reinforcement baselines can cause those early preferences to be reinforced too strongly. As a result, the agent struggles to recover from poor initial behavior. This highlights how performance sensitivity arises when early preferences are misleading. 

In Section \ref{sec:envDet} we describe the two environments in more detail. In Sections \ref{sec:expDesign} and \ref{sec:chainResults} we then describe the experiments and discuss the results.

\subsubsection{Environment Details}
\label{sec:envDet}

\begin{figure}[thbp]
    \centering
    \includegraphics[width=0.6\columnwidth]{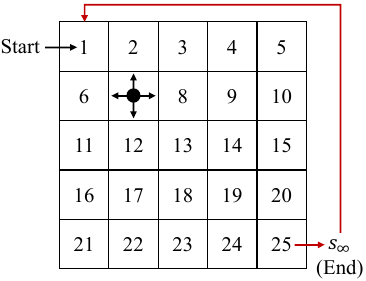}
    \caption{Diagram of the gridworld environment, with numbers indicating states. The circle denotes the agent (in state $7$), and the arrows coming from the agent indicate the four possible actions. The two red arrows indicate state transitions that result in a reward of zero (all others result in a reward of $-1$).}
    \label{fig:gridworld}
\end{figure}

We now describe the two environments used in our experiments. The first is a standard $5\times 5$ gridworld environment. The second is a chain-like environment. Although neither is technically finite-horizon, both use $\gamma_{\mathfrak p}=1$, and we adopt the expected (undiscounted) return as the performance objective:
 \begin{equation}
    \label{eq:experimentPerfObj}
    \mathfrak p(\texttt{alg}) = 
    \mathbf{E}\left [\sum_{i=1}^{i_\text{max}} G_i \right ],
\end{equation}
where $i_\text{max}=500$ episodes for both environments. 
\\\\
\noindent \textbf{Gridworld.} The gridworld environment models an agent moving on a $5 \times 5$ grid as depicted in Figure~\ref{fig:gridworld}. The state indicates the position of the agent on the grid, resulting in 26 states (including $s_\infty$). The agent can select from four actions: \texttt{up}, \texttt{down}, \texttt{left}, and \texttt{right}, which deterministically move the agent one cell in the corresponding direction. If an action would cause the agent to leave the grid, the agent instead does not move. 

Recall from Section \ref{sec:isRQReasonable} that in an alternate formulation the agent might be considered part of the environment. In such a formulation, the agent's changes to its memory might also be construed as actions. In these initial experiments we do not take this perspective, instead modeling the agent as distinct from the environment. This results in a small and discrete set of four possible actions for the gridworld environment (and two actions for the subsequent chain environment). 

The agent begins in the top left corner (state $1$), and the episode ends when the agent reaches the bottom right corner (state $25$). The reward is always $-1$ (other than the two exceptions described next) to encourage the agent to reach the bottom right corner, at which point the episode ends and the negative rewards from the current episode stop. More precisely, state 25 always transitions to $s_\infty$, and $R_t=0$ both when $S_t=s_\infty$ (the state transitioned from $25$ to $s_\infty$, so the current episode is ending) and when $S_{t-1}=s_\infty$ (it is the first time step of a new episode, and $R_t$ occurs before the first action of this new episode).\footnote{We model the gridworld as an AEP environment, handling episode termination as in an AEP. However, the gridworld environment can also be modeled as an MDP wherein the time $t$ resets to zero for each episode.} 

The return (sum of rewards) is negative the number of actions the agent takes to reach state $25$. When considering only the performance objective, an optimal policy moves the agent from state $1$ to state $25$ using $8$ actions, meaning that the optimal expected return (i.e., objective value) is $-8$. 

\begin{figure}[thbp]
    \centering
    \includegraphics[width=0.75\columnwidth]{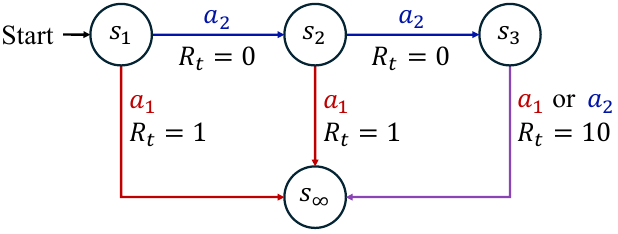}
    \caption{Diagram of the chain environment.}
    \label{fig:chain}
\end{figure}

Notice that a BAC agent can initially learn to favor suboptimal actions. For example, if the agent is in state 7, the down and right actions are both optimal. However, the agent may initially favor the left action if it has learned an effective policy from state 6 but not from states 8 or 12 (perhaps due to having visited state 6 more often than 8 or 12). Even though a BAC agent \textit{can} initially learn to favor suboptimal actions, the gridworld environment was not designed to adversarially cause this behavior.
\\\\
\noindent\textbf{Chain.} The chain environment, depicted in Figure \ref{fig:chain}, was inspired by the chain environments used to evaluate RL algorithms that are designed to be effective even for worst-case environments \citep{strens2000bayesian}. Intuitively, the agent moves along a chain of three states, $s_1$, $s_2$, and $s_3$. In each state, it has the option to continue moving down the chain (action $a_2$), or to end the episode immediately (action $a_1$). As the agent moves along the chain, the rewards it receives are zero. However, when the episode ends, the agent receives a positive reward. If the agent ends the episode before reaching $s_3$, it receives a reward of $1$. If the agent reaches $s_3$, the episode always ends (regardless of which action the agent selects) and the agent receives a reward of $10$. 

An optimal policy causes the agent to move all the way down the chain to state $s_3$ (i.e., the agent should select $a_2$ in states $s_1$ and $s_2$), resulting in a return of $10$. However, consider the behavior of a BAC agent during the very first episode if its policy is initially uniform random and its value function estimate is zero for every state. There is a $0.5$ probability that the agent ends the episode immediately from $s_1$ and a $0.5^2=0.25$ probability that it reaches $s_2$ and then ends the episode, for a total probability of $0.75$ that the agent ends the episode prematurely. The reward of $1$ when the episode terminates prematurely will result in a TD error of $1$, causing the agent to reinforce the action that ended the episode prematurely. 

Once this suboptimal action has been reinforced, the agent becomes even more likely to select it in the future, potentially leading to further reinforcement of the same suboptimal behavior. However, when using a tabular policy representation as described in Section \ref{sec:expDesign}, and under mild technical assumptions (e.g., appropriate step size decay), policy gradient algorithms are guaranteed to converge to an optimal policy for this MDP \citep{Thomas2014,pmlr-v125-agarwal20a}. This suggests\footnote{BAC algorithms update the policy parameters in the direction of \emph{estimates} of the policy gradient---not in the direction of the actual policy gradient. The properties of these estimates are complicated by the interdependence of the actor and critic. Although convergence analyses exist for these settings \citep{konda1999actor,Bhatnagar2009}, the details are beyond the scope of this discussion. We simply note that true policy gradient methods converge to optimal behavior, and that BAC approximates such methods, hence the use of the term ``suggests.''} that the BAC agent's policy will eventually shift to favor optimal actions, although it may take considerable time for the agent to overcome the early reinforcement of suboptimal actions.

\subsubsection{Experimental Design and Gridworld Results}
\label{sec:expDesign}

To evaluate the robustness of performance to reinforcement baselines for each environment, we performed a manual search for hyperparameters that are effective in terms of performance for each environment. Since we only aim to establish the existence of robust and sensitive settings, we did not perform a more rigorous hyperparameter optimization. The BAC hyperparameters selected for each environment are described in Appendix \ref{app:hyp}. 

We evaluated performance and the prevalence of reinforcement when using constant reinforcement baselines, i.e., for all states $s$, $b(s)=c$ for some constant $c$. We experimented with three different values of $c$: $c=0$, which corresponds to no reinforcement baseline, $c=-1$, which corresponds to a moderate reinforcement baseline for these environments, and $c=-5$, which corresponds to an aggressive reinforcement baseline for these environments. For each setting of $c$ and environment, we simulated $i_\text{max}=500$ episodes of the BAC algorithm $10^8$ times, i.e., we ran one hundred million trials for each environment and setting of $c$. 

\begin{figure}[thbp]
    \centering
    \includegraphics[width=0.9\columnwidth]{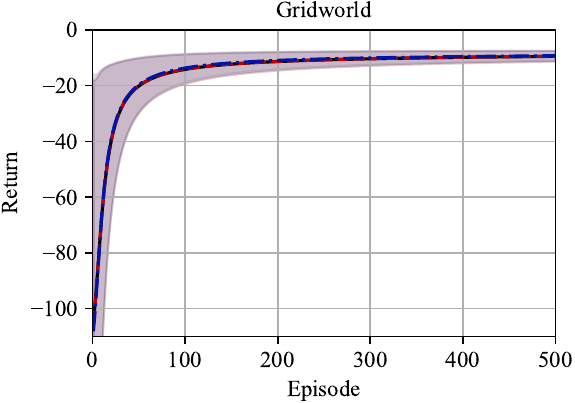}
    \includegraphics[width=0.75\columnwidth]{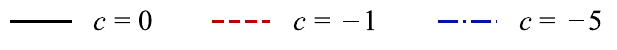}
    \caption{Learning curves (performance) on the gridworld environment as the reinforcement baseline is varied. Shaded error bars depict standard deviation.}
    \label{fig:gridworldReturns}
\end{figure}

To evaluate performance, we plotted standard learning curves, showing the average (across trials) return for each episode, including standard deviation error bars.\footnote{Standard deviation error bars quantify variance due to the stochasticity of the agent. Error bars reporting the standard error would quantify the uncertainty of the curve, but are not depicted. Their width would be one-ten-thousandth of the standard deviation error bars.} Figure \ref{fig:gridworldReturns} shows the resulting performance plot for the gridworld environment (we present and discuss the results for the chain environment later in Section \ref{sec:chainResults}). Notice that negative the area above the learning curve is an approximation of the value of the performance objective in \eqref{eq:experimentPerfObj}. All three curves and error bars are nearly perfectly overlapping, indicating that there is little difference in performance.\footnote{Performance improves slightly as the reinforcement bias increases. The average return with $c=0$ was approximately $-14.53$, while the average return with $c=-1$ was approximately $-14.48$, and the average return with $c=-5$ was approximately $-14.19$. These slight differences are inconsequential.}

Whether Figure \ref{fig:gridworldReturns} provides meaningful evidence of performance robustness depends on the scale of the reinforcement baselines. If the tested values of $c$ (i.e., $0$, $-1$, and $-5$) are too small in magnitude to meaningfully influence reinforcement, then the absence of performance differences across settings would not constitute strong evidence of robustness. In such a case, larger-magnitude (more negative) baselines would be needed to properly assess whether performance remains stable under aggressive reinforcement bias.

\begin{figure}[thbp]
    \centering
    \includegraphics[width=0.9\columnwidth]{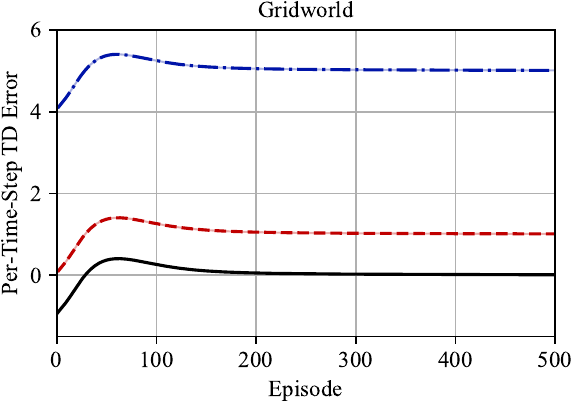}
    \includegraphics[width=0.75\columnwidth]{images/Gridworld/GridworldReturns_legend.pdf}
    \caption{Average per-time-step TD error on the gridworld environment as the reinforcement baseline is varied. Shaded error bars depict standard error.}
    \label{fig:gridworldTDErrors}
\end{figure}

We therefore evaluate the reinforcement bias by plotting, for each episode, the average per-time-step TD error, i.e., the average value of $\Delta_t$ in that episode.\footnote{We only average the TD error (and later, the per-time-step reinforcement) over time steps when the TD error is defined---time steps where the BAC algorithm performs the ``standard update'' in Algorithm \ref{alg:BACdm}.} Here, we adopt the explicit TD error formulation (see Section \ref{sec:ambigRV}), wherein the reinforcement baseline is included within the reported TD error values. For tabular BAC algorithms, this value provides one quantification of the prevalence of reinforcement---positive TD errors correspond to reinforcement and negative to inhibition, and the magnitude of the TD error scales the magnitude of the reinforcement or inhibition.\footnote{When using other (nonlinear) policy parameterizations this property may not always hold. For example, with certain other parameterizations, a positive TD error can result in the inhibition of behavior if the step size is too large. Also, as described in Section \ref{sec:envDet}, notice that we are only considering the agent's actions to be the MDP actions, not, for example, the agent's weight updates as suggested in Section \ref{sec:isRQReasonable}. Hence, TD error correlates well with reinforcement. Future work might assess reinforcement baselines with more general definitions of actions.} The resulting plot is provided in Figure \ref{fig:gridworldTDErrors}.

Unlike Figure \ref{fig:gridworldReturns}, the error bars in Figure \ref{fig:gridworldTDErrors} depict standard \emph{error}. We report standard error rather than standard deviation because the variance (and standard deviation) of the \emph{average} TD error during an episode varies with the length of the episode (number of TD errors) in a way that makes the standard deviation challenging to interpret. Hence, in Figure \ref{fig:gridworldTDErrors} the error bars (which are too small to be seen) characterize the uncertainty of the curves. 

The solid black curve in Figure \ref{fig:gridworldTDErrors} shows that without a reinforcement baseline, the average TD error is negative early during learning, but quickly becomes positive, after which it peaks and then slowly decreases toward zero. This behavior is expected because early during learning the value function is optimistic (since it is initially zero everywhere, and the values of states are negative due to the constant negative rewards at nearly every time step). This optimism results in negative TD errors, as discussed in Section \ref{sec:TDErrBonuses}. It takes time for the optimism of the value function to fade as the VFA becomes more accurate. Once the VFA is accurate, the agent learns by reinforcing the actions it typically prefers, resulting in a shift toward positive TD errors. 

As the agent's policy converges toward an optimal policy, the state-value function also converges toward the optimal state-value function. Consequently, the VFA approximates a nearly stationary target (the optimal state-value function), and so it becomes increasingly accurate and eventually matches the true state-value function closely. When this happens, the TD errors, which represent discrepancies between estimates of a state's value, become increasingly small on average. Ultimately, once the policy is stable (stationary) and the VFA accurately approximates the true state-value function, the average TD error approaches zero.

The dashed red curve in Figure \ref{fig:gridworldTDErrors} shows that with a moderate reinforcement baseline of $c=-1$, the average TD error is consistently positive. Compared to the $c=0$ case, the curve appears to be shifted upward by approximately one unit, while the overall trends remain otherwise unchanged. The upward shift indicates a moderate shift toward learning via reinforcement. Similarly, the dash-dot blue curve corresponding to the aggressive baseline $c=-5$ shows a more pronounced positive shift, closely mirroring the shape of the solid black curve but offset upward by roughly five units. This larger upward shift indicates a more aggressive shift toward learning via reinforcement.

\begin{figure}[thbp]
    \centering
    \includegraphics[width=1.01\columnwidth]{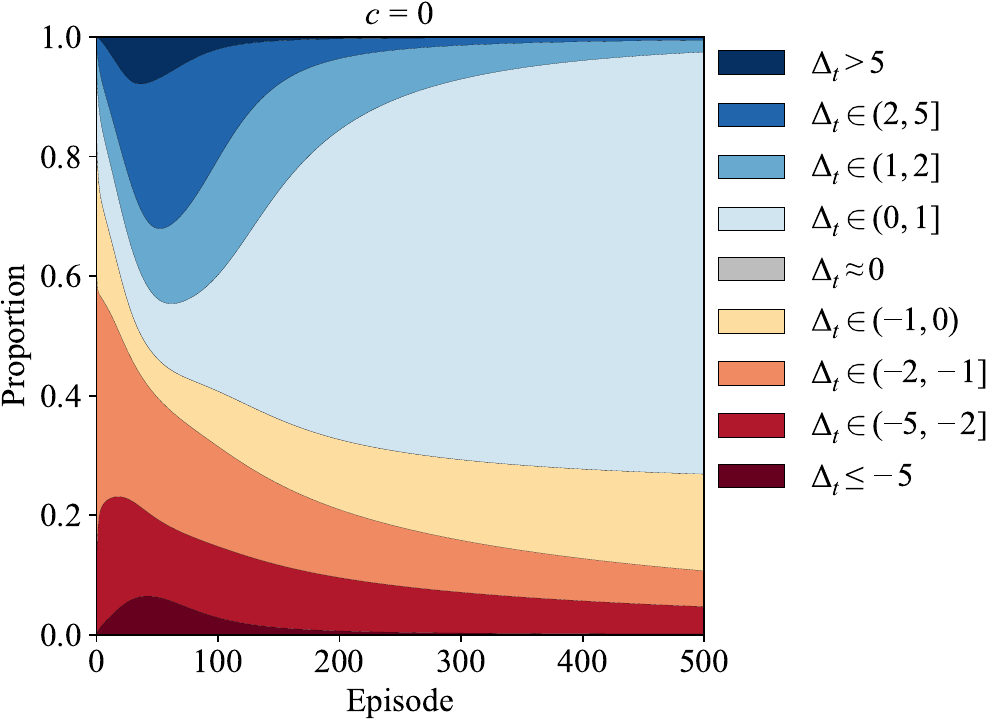}
    \\
    \vspace{0.5cm}
    \includegraphics[width=1.01\columnwidth]{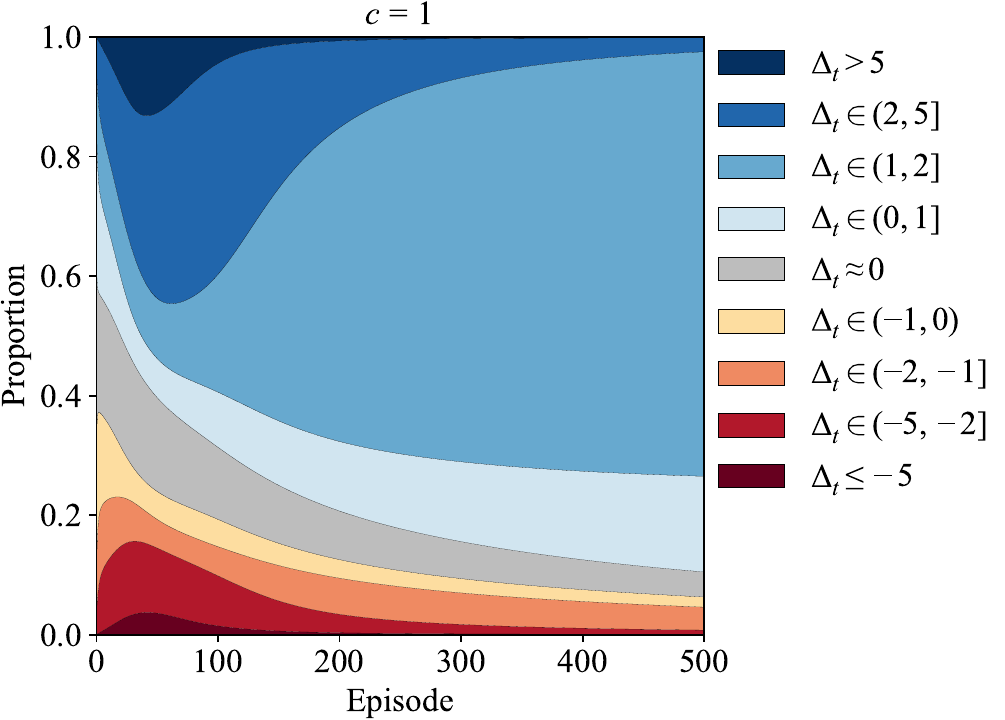}
    \\
    \vspace{0.5cm}
    \includegraphics[width=1.01\columnwidth]{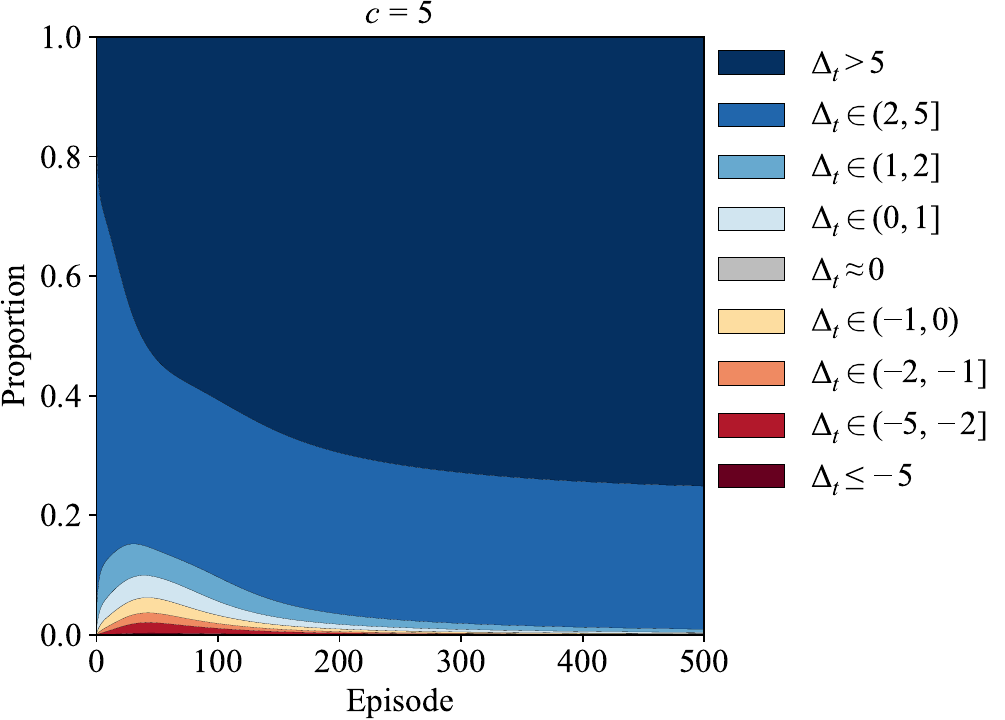}
    \caption{Proportion of TD errors that fall within different ranges when using different reinforcement baselines on the \textbf{gridworld} environment. \textbf{Top: } No reinforcement baseline, i.e., $c=0$, \textbf{Middle: }Moderate reinforcement baseline, i.e., $c=-1$, \textbf{Bottom:} Aggressive reinforcement baseline, i.e., $c=-5$. The legends and axes are identical across all three plots. The $\Delta_t \approx 0$ region actually corresponds to $\Delta_t \in (-10^{-6}, 10^{-6}]$.\vspace{0.375cm}}
    \label{fig:gridworldDeltaFreq}
\end{figure}

Notice, however, that Figure \ref{fig:gridworldTDErrors} only shows the \emph{average} TD error. It could be that the majority of TD errors are negative, but occasional extremely large TD errors (outliers) throw off the mean. Hence, although Figure \ref{fig:gridworldTDErrors} provides one way of inspecting the prevalence of reinforcement, it does not quantify the \emph{frequency} of reinforcement. To see how often different levels of reinforcement and inhibition occur, Figure \ref{fig:gridworldDeltaFreq} shows the relative frequencies of different ranges of TD errors across episodes.

\PackageWarning{manual}{Warning: Figure occurs prior to reference.}
\begin{figure}[htbp]
    \centering
    \includegraphics[width=1.01\columnwidth]{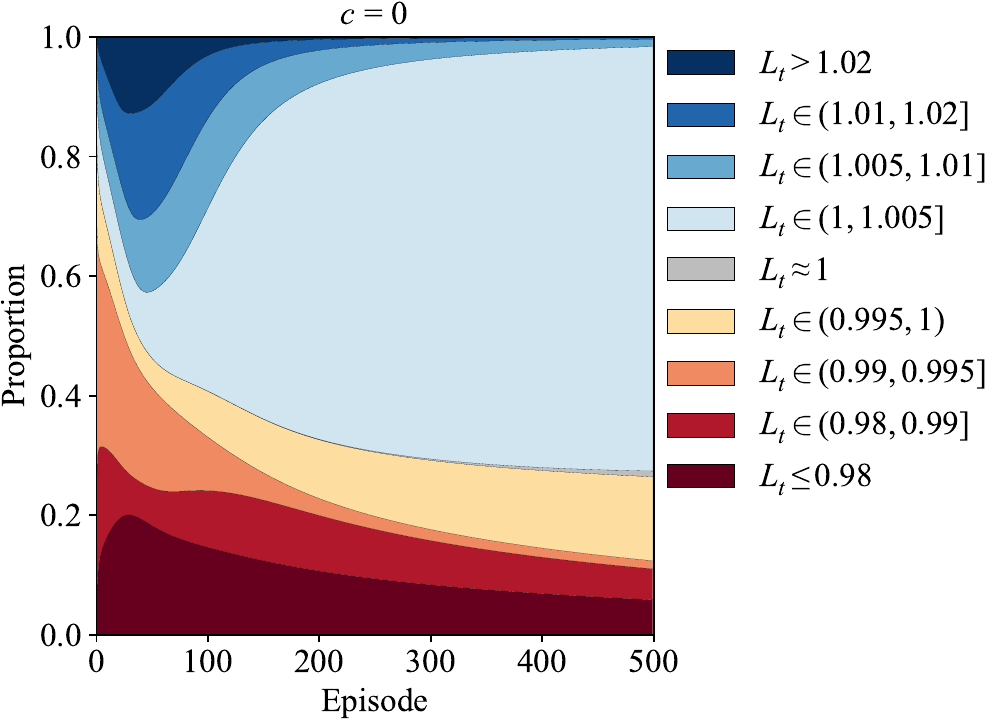}
    \\
    \vspace{0.5cm}
    \includegraphics[width=1.01\columnwidth]{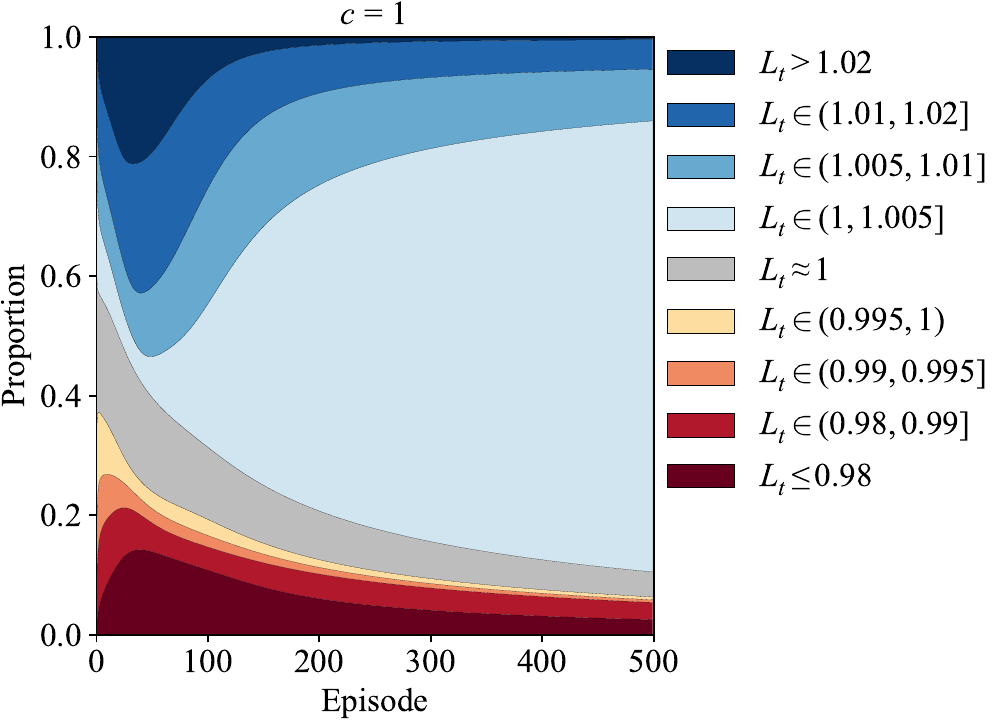}
    \\
    \vspace{0.5cm}
    \includegraphics[width=1.01\columnwidth]{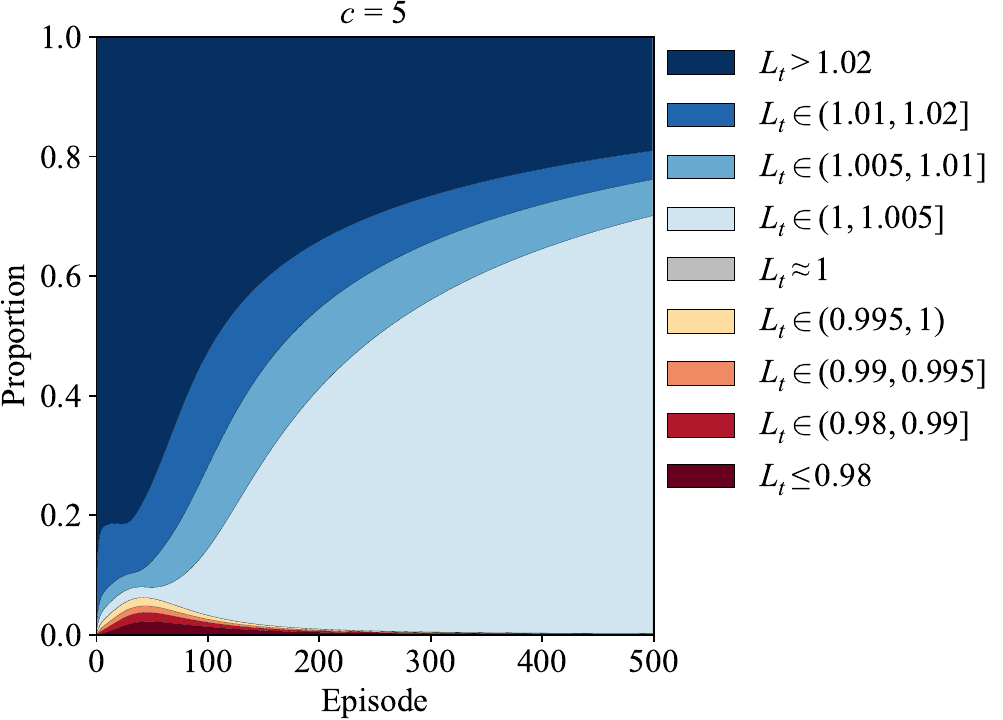}
    \caption{Proportion of $L_t$ (reinforcement of instantaneous behavior) that fall within different ranges when using different reinforcement baselines on the \textbf{gridworld} environment. \textbf{Top: } No reinforcement baseline, i.e., $c=0$, \textbf{Middle: }Moderate reinforcement baseline, i.e., $c=-1$, \textbf{Bottom:} Aggressive reinforcement baseline, i.e., $c=-5$. The legends and axes are identical across all three plots. The $L_t \approx 1$ region actually corresponds to $L_t \in (1-10^{-6}, 1+10^{-6}]$.}
    \label{fig:gridworldReinforcementFreq}
\end{figure}

The top plot in Figure \ref{fig:gridworldDeltaFreq} shows that without a reinforcement baseline both positive and negative TD errors are common,  with the proportion of reinforcing updates ranging from approximately $0.2$ to $0.7$. This can be observed by inspecting the upper boundary of the $\Delta_t \in (-1,0)$ region, the height of which indicates the proportion of inhibitory updates. This boundary ranges from approximately $0.8$ to $0.3$ across episodes, implying that the proportion of reinforcing updates---its complement---ranges from $0.2$ to $0.7$. The middle plot shows that $c=-1$ corresponds to a moderate reinforcement bias, reducing but not coming close to eliminating inhibitory updates. The bottom plot shows that $c=-5$ corresponds to an aggressive reinforcement bias, almost eliminating inhibitory updates ($\Delta_t < 0$) and resulting in a majority of strongly reinforcing updates ($\Delta_t > 5$).

The TD error provides insight into the frequency and magnitude of reinforcement, but the exact relationship between the TD error at time $t$ and the reinforcement of instantaneous behavior at time $t$, as defined in \eqref{eq:reinforcementOfBehaviorT}, is complex. As the policy becomes increasingly deterministic, the term 
\begin{equation}
    \frac{\partial \ln\big (\pi_\text{BAC}(S_{t-1}, A_{t-1}, \Theta_{t-1})\big )}{\partial \Theta_{t-1}},
\end{equation}
which multiplies $\Delta_t$ in the BAC policy update, tends toward the zero vector. To see why, in Appendix \ref{sec:compatibleFeatures} we show that 
\begin{align}
    &\frac{\partial \ln\big (\pi_\text{BAC}(S_{t-1}, A_{t-1}, \theta)\big )}{\partial \theta_{s,a}}\\
    =&\begin{cases}
    0 &\mbox{if }S_{t-1}\neq s\\
    -\pi_\text{BAC}(s,a, \theta)&\mbox{if } S_{t-1}=s \text{ and } A_{t-1}\neq a\\
    1-\pi_\text{BAC}(s, a, \theta)&\mbox{if } S_{t-1}=s \text{ and } A_{t-1}= a.
    \end{cases}
\end{align}
Consider what this implies when $\theta$ changes so that $\pi_\text{BAC}(s, a, \theta)$ approaches $1$ for some state $s$ and action $a$. When action $a$ is chosen, $\theta_{s,a}$ is updated proportional to $1-\pi_\text{BAC}(s, a, \theta)$, which tends to zero, while the policy parameters corresponding to all other actions $a'$ in state $s$ are updated proportional to $-\pi_\text{BAC}(s, a', \theta)$, which also tends to zero (if the conditional probability of action $a$ goes to $1$, then the conditional probabilities of all $a' \neq a$ necessarily go to zero). Hence, the magnitude of common policy parameter updates that result from a fixed-magnitude TD error decreases as the policy becomes increasingly deterministic. Not only does the magnitude of the change to the policy parameters decrease, but the magnitude of changes to the conditional probability of the most likely action also necessarily decreases as the probability of an action approaches $1$ (since the probability can only approach $1$ asymptotically---it cannot reach or exceed it). These properties highlight the complex relationship between the TD error and reinforcement of instantaneous behavior at time $t$. 

\begin{figure}[t]
    \centering
    \includegraphics[width=0.9\columnwidth]{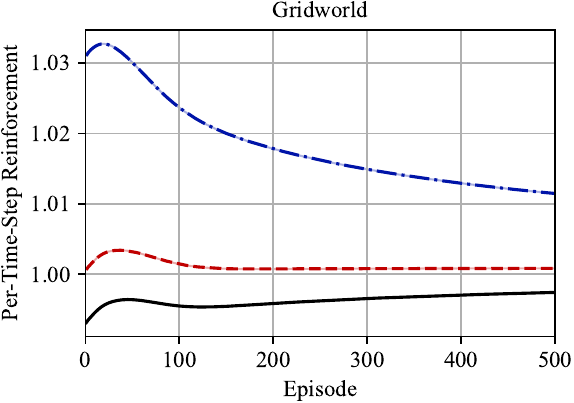}
    \includegraphics[width=0.75\columnwidth]{images/Gridworld/GridworldReturns_legend.pdf}
    \caption{Average per-time-step reinforcement of instantaneous behavior (i.e., average $L_t$) for each episode, on the gridworld environment, as the reinforcement baseline is varied. Shaded error bars depict standard error.}
    \label{fig:gridworldReinforcement}
\end{figure}

So, to better understand how reinforcement baselines influence the reinforcement of instantaneous behavior at time $t$, in Figures \ref{fig:gridworldReinforcementFreq} and \ref{fig:gridworldReinforcement} we reproduce Figures \ref{fig:gridworldTDErrors} and \ref{fig:gridworldDeltaFreq} (in reverse order), but with the TD errors $\Delta_t$ replaced with the reinforcement of instantaneous behavior at time $t$, $L_t$. Notice that $L_t=1$, like $\Delta_t=0$, corresponds to neither inhibition nor reinforcement. These plots quantifying aspects of the distributions of $L_t$ (at different episodes and with different reinforcement baselines) mirror the corresponding plots quantifying the distributions of $\Delta_t$. That is, they show that $c=-1$ corresponds to a moderate reinforcement bias, encouraging reinforcement, but not coming close to eliminating inhibitory updates, while $c=-5$ provides an aggressive reinforcement bias, nearly eliminating inhibitory updates. Notice that the area under each curve in Figure \ref{fig:gridworldReinforcement} is related to, but not identical to, the values of the qualia objective functions in \eqref{eq:firstReinforcementQualiaObj} and \eqref{eq:ourReinforcementQualiaObjMDP} since the figure reports the average per-time-step reinforcement, whereas \eqref{eq:firstReinforcementQualiaObj} sums over the length of each episode (e.g., if each likelihood ratio is positive, the value increases with episode length), while \eqref{eq:ourReinforcementQualiaObjMDP} divides each likelihood ratio by the corresponding episode length, placing less weight on likelihood ratios that occur during longer episodes. 

Consider Figure \ref{fig:gridworldReinforcement} more closely. First, notice that for all values of $c$ the average value of $L_t$ trends toward $L_t=1$ since, as the policy becomes increasingly deterministic, the magnitudes of most updates decrease, resulting in reduced reinforcement and inhibition overall (when measured in terms of the likelihood ratio $L_t$). Second, notice that the black curve, which corresponds to $c=0$, is less than $1$ for the entire plot, even though the top plot in Figure \ref{fig:gridworldReinforcementFreq} indicates that reinforcement is more likely for the majority of episodes. Together these properties suggest that although inhibition is less common in this case, the amount that $L_t$ differs from $1$  is larger in expectation when $L_t < 1$.

These results highlight that different ways of measuring reinforcement can result in different conclusions regarding the quality of an agent's experience. For example, if one assumes that the quality of an agent's experience is overall positive if reinforcement is more frequent than inhibition, or if one assumes that the quality of an agent's experience is overall positive if the average TD error is positive, then they would conclude that the BAC algorithm applied to the gridworld environment without a performance baseline results in the agent having an overall positive experience. However, if one assumes that the quality of an agent's experience is overall positive if the average reinforcement of instantaneous behavior is at least $1$, then they would conclude that the agent's experience is overall negative.\footnote{One avenue of future work would be to consider utility functions that calibrate and rescale the values of signals like $\Delta_t$ and $L_t$ in different ways, perhaps capturing ideas like Stevens's power law \citep{Stevens1970}.}

However, notice that regardless of which of these strategies one uses to quantify reinforcement in the reinforcement-qualia setting, the conclusion remains that, for the gridworld environment, adding even aggressive reinforcement baselines to the BAC algorithm (with the selected hyperparameters) does not result in a significant change in performance. The lack of a significant change in performance is supported by Figure \ref{fig:gridworldReturns}, while the scale of the reinforcement baselines being moderate and aggressive is supported by Figures \ref{fig:gridworldTDErrors}--\ref{fig:gridworldReinforcement}.

\subsubsection{Chain Results}
\label{sec:chainResults}

Having established via the gridworld environment that a case exists wherein performance is robust to changes in the reinforcement baseline, we now turn to establishing, via the chain environment, that a case exists wherein performance is sensitive to changes in the reinforcement baseline. To do so, we perform the exact same sequence of experiments that we performed using the gridworld environment, but for the chain environment.

\begin{figure}[thbp]
    \centering
    \includegraphics[width=0.9\columnwidth]{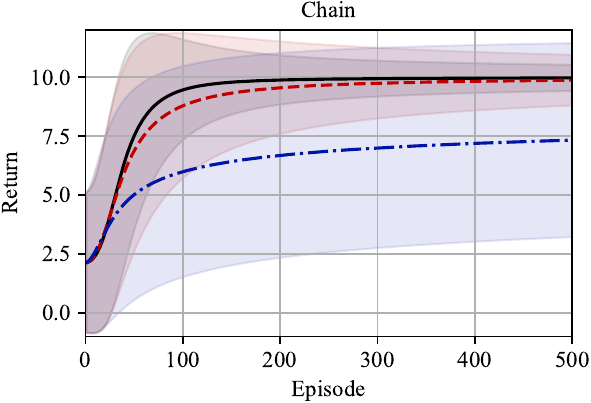}
    \includegraphics[width=0.75\columnwidth]{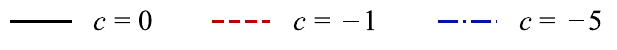}
    \caption{Learning curves (performance) on the chain environment as the reinforcement baseline is varied. Shaded error bars depict standard deviation.}
    \label{fig:chainReturns}
\end{figure}

First, in Figure \ref{fig:chainReturns} we plot the learning curves to evaluate the impact of reinforcement baselines on performance. In this case, even the moderate reinforcement baseline of $c=-1$ results in a noticeable decrease in performance and increase in the standard deviation of performance. That is, the algorithms is less reliable and performs worse on average. When using an aggressive reinforcement baseline of $c=-5$ the degradation in performance and increase in variability is even more pronounced. 

\begin{figure}[thbp]
    \centering
    \includegraphics[width=1.01\columnwidth]{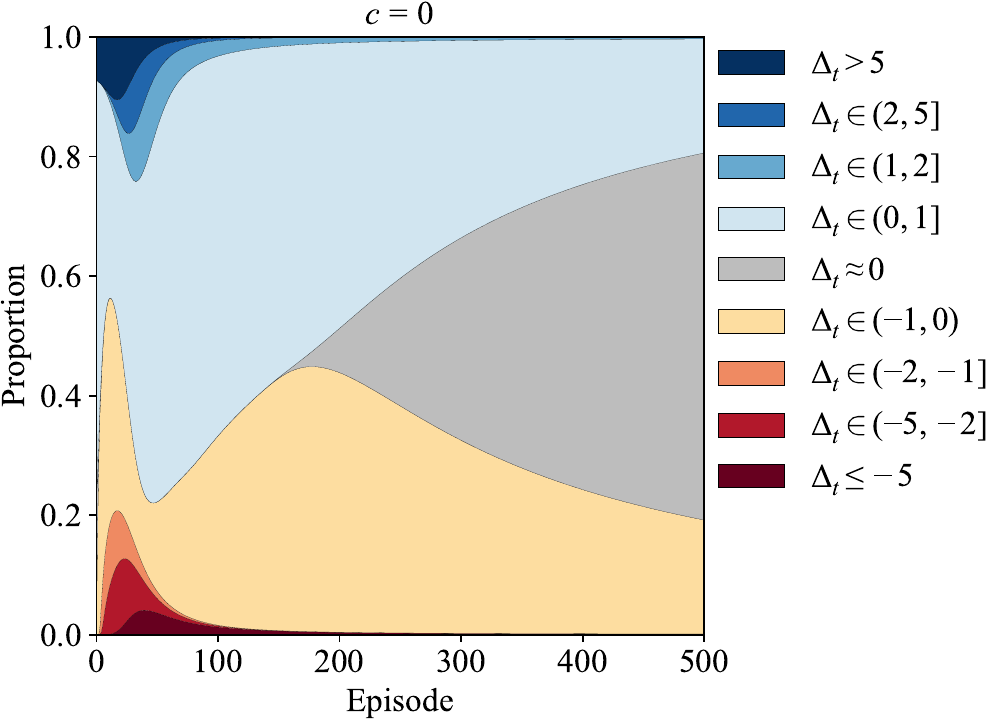}
    \\
    \vspace{0.5cm}
    \includegraphics[width=1.01\columnwidth]{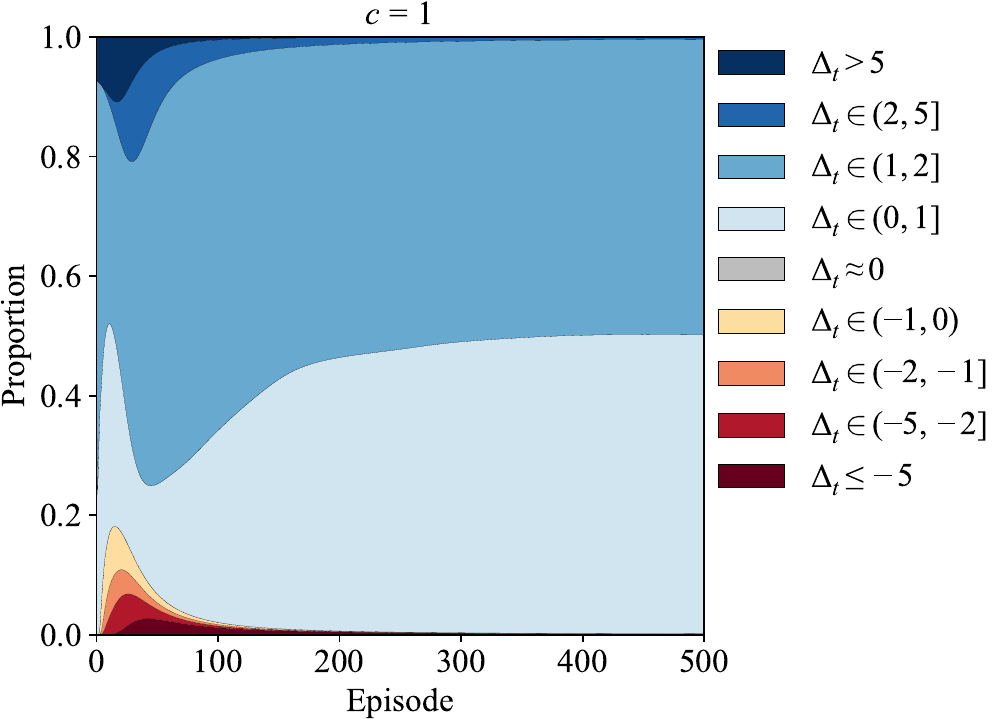}
    \\
    \vspace{0.5cm}
    \includegraphics[width=1.01\columnwidth]{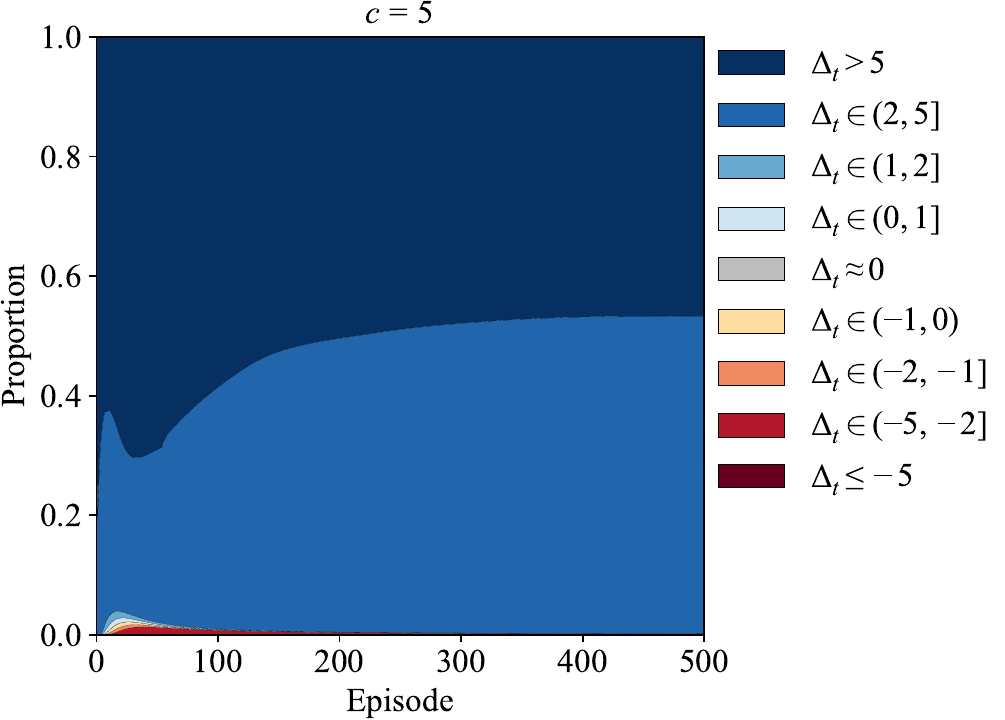}
    \caption{Proportion of TD errors that fall within different ranges when using different reinforcement baselines on the \textbf{chain} environment. \textbf{Top: } No reinforcement baseline, i.e., $c=0$, \textbf{Middle: }Moderate reinforcement baseline, i.e., $c=-1$, \textbf{Bottom:} Aggressive reinforcement baseline, i.e., $c=-5$. The legends and axes are identical across all three plots. The $\Delta_t \approx 0$ region actually corresponds to $\Delta_t \in (-10^{-6}, 10^{-6}]$.\vspace{0.36cm}}
    \label{fig:chainDeltaFreq}
\end{figure}

\begin{figure}[thbp]
    \centering
    \includegraphics[width=1.01\columnwidth]{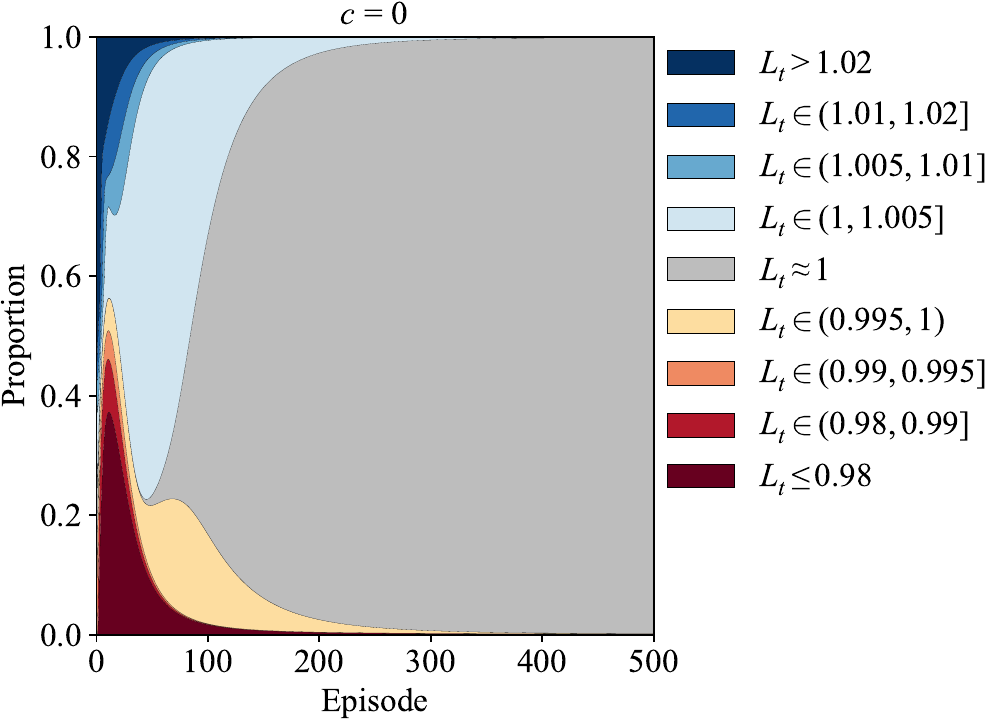}
    \\
    \vspace{0.5cm}
    \includegraphics[width=1.01\columnwidth]{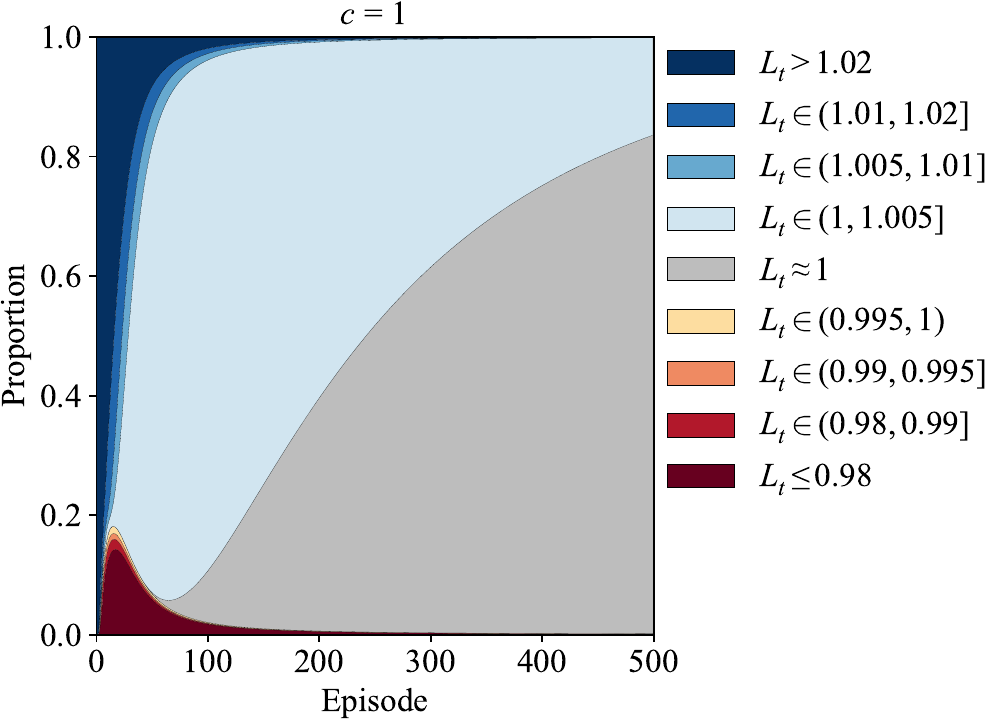}
    \\
    \vspace{0.5cm}
    \includegraphics[width=1.01\columnwidth]{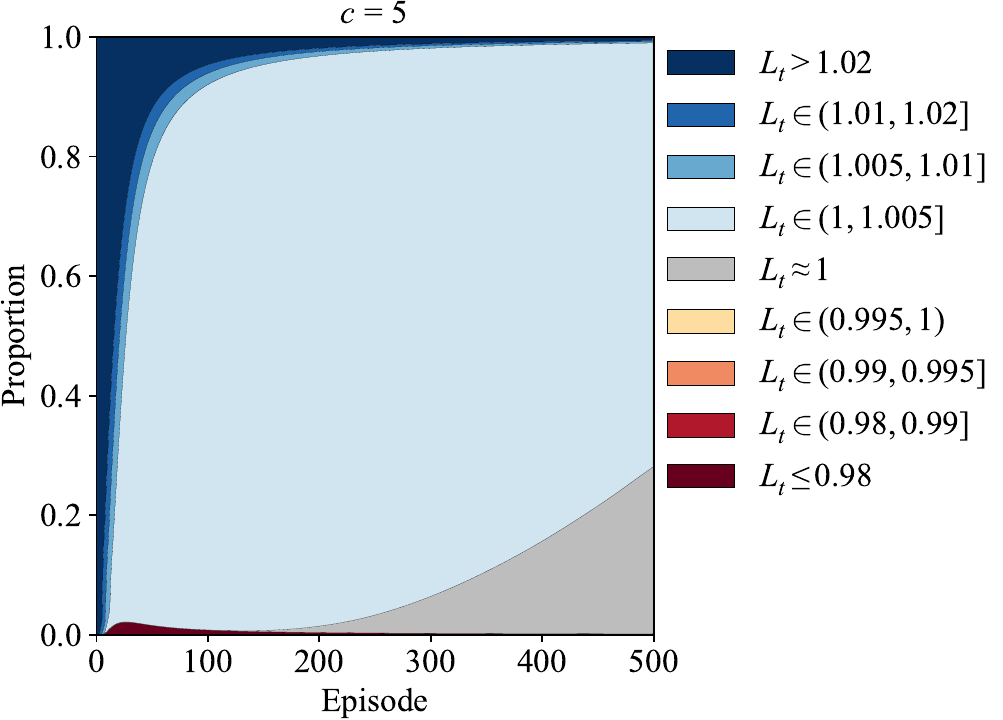}
    \caption{Proportion of $L_t$ (reinforcement of instantaneous behavior) that fall within different ranges when using different reinforcement baselines on the \textbf{chain} environment. \textbf{Top: } No reinforcement baseline, i.e., $c=0$, \textbf{Middle: }Moderate reinforcement baseline, i.e., $c=-1$, \textbf{Bottom:} Aggressive reinforcement baseline, i.e., $c=-5$. The legends and axes are identical across all three plots. The $L_t \approx 1$ region actually corresponds to $L_t \in (1-10^{-6}, 1+10^{-6}]$.}
    \label{fig:chainReinforcementFreq}
\end{figure}

Next, to establish that $c=-1$ corresponds to a moderate reinforcement baseline and $c=-5$ corresponds to an aggressive reinforcement baseline, in Figures \ref{fig:chainDeltaFreq} and \ref{fig:chainReinforcementFreq} we plot the relative frequencies of different ranges of $\Delta_t$ and $L_t$ across episodes. Similar to the gridworld environment, the top plots show that without a reinforcement baseline ($c=0$) reinforcement ($\Delta_t > 0$, $L_t >1$) and inhibition ($\Delta_t < 0$, $L_t <1$) are both common. The middle plots show that $c=-1$ corresponds to a moderate reinforcement bias, increasing the prevalence of reinforcing updates and reducing the prevalence of inhibitory updates. The bottom plots show that $c=-5$ corresponds to an aggressive reinforcement bias, almost eliminating inhibitory updates nd resulting in a majority of strongly reinforcing updates ($\Delta_t > 5$).


\begin{figure}[htbp]
    \centering
    \includegraphics[width=0.9\columnwidth]{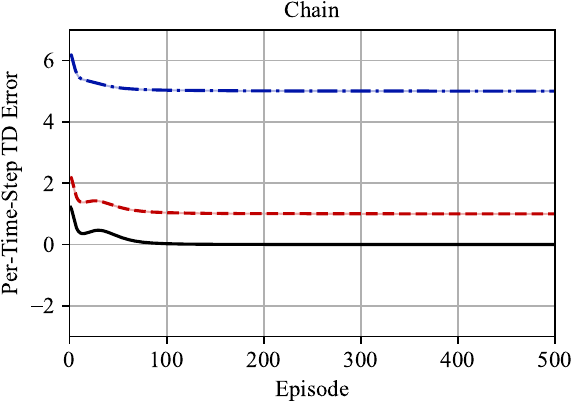}
    \includegraphics[width=0.75\columnwidth]{images/Chain/ChainReturns_legend.pdf}
    \caption{Average per-time-step TD error on the chain environment as the reinforcement baseline is varied. Shaded error bars depict standard error.}
    \label{fig:chainTDErrors}
\end{figure}

\begin{figure}[thbp]
    \centering
    \includegraphics[width=0.9\columnwidth]{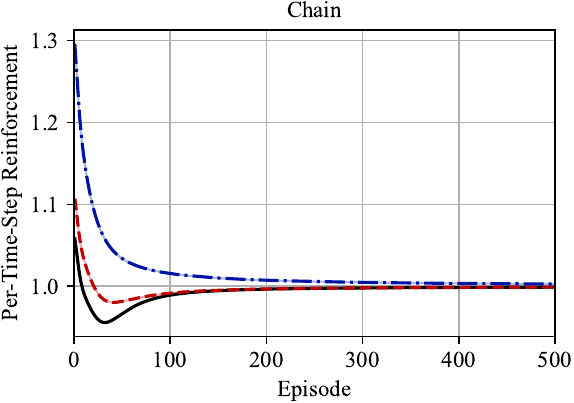}
    \includegraphics[width=0.75\columnwidth]{images/Chain/ChainReturns_legend.pdf}
    \caption{Average per-time-step reinforcement of instantaneous behavior (i.e., average $L_t$) for each episode, on the chain environment, as the reinforcement baseline is varied. Shaded error bars depict standard error.}
    \label{fig:chainReinforcement}
\end{figure}

Although they do not provide significant additional insights, for completeness, we provide the chain-environment correlates of Figures \ref{fig:gridworldTDErrors} and \ref{fig:gridworldReinforcement} in Figures \ref{fig:chainTDErrors} and \ref{fig:chainReinforcement}. Figure \ref{fig:chainTDErrors} shows the average per-time-step TD error for each episode as the reinforcement baseline is varied. Recall that the VFA weights were initialized to zero. For the gridworld this corresponded to an optimistic VFA (over-estimate of state-values), but for the chain environment this corresponds to a pessimistic VFA (under-estimate of state-values) since the rewards in the chain environment are all non-negative and some are positive. Hence, initially the average TD error is positive, even with $c=0$. Figure \ref{fig:chainReinforcement} further reinforces the conclusion that $c=-1$ produces a moderate reinforcement bias while $c=-5$ produces an aggressive reinforcement bias. 

In summary, for the chain environment, adding even a moderate reinforcement baseline to the BAC algorithm (with the selected hyperparameters) can result in a noticeable degradation of performance. The degradation of performance is supported by Figure \ref{fig:chainReturns}, while the scale of the reinforcement baselines being moderate and aggressive is supported by Figures \ref{fig:chainDeltaFreq}--\ref{fig:chainReinforcement}.

\section{Conclusion and Future Directions}

In this report we assumed that AI agents possess phenomenal consciousness (qualia) and developed a formal framework for \emph{qualia optimization}---jointly optimizing performance and qualia. Here we summarize the report and then suggest directions for future work. 

After providing background on philosophy of mind in Section \ref{sec:background:PoM}, in Section \ref{sec:RLBackgroundAndSetting} we introduced mathematical formulations of an agent interacting with an environment, which we called AEPs and AERPs (the latter necessarily including rewards). Then, after providing background on the relationship between RPEs (e.g., TD errors) and dopamine in Section \ref{sec:background_RPE}, in Section \ref{sec:qualia_opt} we stated our assumptions formally and defined qualia optimization for AI. Inspired by an example of a rat in a shuttle box, in Section \ref{sec:qualia_opt} we also introduced the AEI as a mechanism for transforming how an agent interacts with an environment---altering its perceptions and actions. This resulted in the AIEP and AIERP formulations, which extend the AEP and AERP formulations to include an AEI.

In Section \ref{sec:RewardQualiaHypothesis} we introduced the \emph{reward-qualia} setting, wherein the quality of an agent's qualia  can be measured in terms of the amount of reward it receives. We introduced the concept of aligning the performance and qualia objective functions, but showed that the alignment of these objectives does not necessarily improve the agent's qualia. We then introduced the reward bonus strategy, which inflates the agent rewards to increase the value of the qualia objective in many reward-qualia settings. To prevent these changes from decreasing the value of the performance objective, we showed how the RL algorithm implemented by the agent can be adjusted to undo the impact of the inflation of rewards, ensuring that the interactions with the base environment are unchanged.

In Section \ref{sec:reconsideringAEI} we generalized the idea behind preventing performance changes when using the reward bonus strategy, defining AEI inverters and inverse RL algorithms (algorithms that undo the transformations of the AEI). We then introduced the concept of inversion-exploitable qualia objective functions---qualia objective functions that produce different output (assessments of the quality of the agent's qualia) when \textbf{(a)} a base RL algorithm is used without an AEI (equivalently, with the identity AEI) and when \textbf{(b)} an AEI and the corresponding inverse base RL algorithm are applied. We showed that a single physical system can be interpreted as both (a) and (b), and so inversion-exploitable qualia objectives can assign different values to the quality of the agent's qualia for different AIEP models of the same physical system.

Recognizing the core problem---that one physical system can be modeled mathematically as two different AEPs, and that this modeling assumption changes the value of some qualia objective functions---we generalized inversion-exploitability by introducing the concept of \emph{representation-exploitable} qualia objective functions. These are qualia objective functions that can assign different values to the same physical system when it is modeled as different AEPs.\footnote{The qualia objective function only depends on the distributions of random variables in the AEP setting, and AIEPs can be viewed as special cases of AEPs (those where the environment consists of a base environment and AEI). So, by defining representation-exploitability for AEPs, we have defined it for all of the settings considered in this report.} We suggested three (not necessarily disjoint) strategies for making progress despite the existence of representation-exploitable qualia objectives: \textbf{1)} consider only representation-robust (i.e., not representation-exploitable) qualia objectives, \textbf{2)} restrict the set of AEIs and RL algorithms under consideration so that each AEI-algorithm pair corresponds to a unique underlying physical system, and \textbf{3)} restrict the set of allowed representation functions (mappings from physical systems to AEPs). 

Before considering other qualia optimization settings, we reconsidered the inclusion of the AEI in Section \ref{sec:agentBounday}. We presented the dual agent-environment strategy, which trivializes qualia optimization by disconnecting the problems of performance and qualia optimization. This strategy leverages the assumption that the AEI is defined to be part of the environment, and so its qualia experiences are not considered. It suggests that, when faced with the challenge of creating an agent that interacts with an environment and which has desirable qualia, an effective solution is to create a different agent (whose qualia are not considered) that actually interacts with the environment of interest, and to then create a completely separate environment for the agent whose qualia is being considered to interact with. This motivated our subsequent exclusion of the AEI from the problem formulation. 

After Section \ref{sec:summaryOfProblemFormulationConsiderations}, which summarizes the topics discussed in Sections \ref{sec:reconsideringAEI} and \ref{sec:agentBounday}, in Section \ref{sec:TDQualiaHypothesis} we introduced the \emph{RPE-qualia} setting, motivated by the previously established relationship between RPEs and dopamine. We showed how ambiguity about the precise meaning of a random variable within an RL algorithm (e.g., TD error) can arise when considering variants of the algorithm, culminating in the definition of two qualia-objective functions for measuring the cumulative TD error in the RPE-qualia setting: the implicit and explicit TDE-qualia objectives. After showing that both of these objectives are representation-exploitable, we reviewed more recent neuroscience research that suggests that dopamine does not necessarily produce desirable qualia in humans, undermining the idea that qualia objective functions should measure RPEs.

In Section \ref{sec:ReinforcementQualiaHypothesis} we introduced the \emph{reinforcement-qualia} setting, wherein the reinforcement of behavior is associated with desirable qualia. We defined the reinforcement of behavior to be when the agent's policy is updated to make the conditional probability of recent behavior (e.g., recent actions) more likely if similar situations are encountered in the future. This setting refines the RPE-qualia setting in two ways. First, we argued that (if one models the agent as part of the environment, thereby making changes to the agent constitute actions that can be reinforced or inhibited) RPEs may not perfectly correlate with the reinforcement of behavior. This is important because it means that dopamine not perfectly correlating with the quality of qualia in humans does not mean that reinforcement also fails to perfectly correlate with the quality of qualia.\footnote{We note, however, that reinforcement may also fail to perfectly correlate with the quality of qualia, in which case it might still serve as a surrogate objective until a qualia objective has been identified that better correlates with human qualia.} Second, we showed that there exist natural qualia objective functions in the MDP reinforcement-qualia setting that are representation-robust (i.e., not representation-exploitable). 

However, we struggled to formally define the reinforcement of behavior (and hence reinforcement-qualia objectives) in the general AEP setting for two reasons. First, evaluating whether a behavior has been made more likely if similar situations are encountered in the future requires reverting the context components of an agent's memory to simulate a ``similar situation,'' while leaving the policy components unchanged. This is problematic because it is unclear how the agent's memory can, in general, be separated into contextual and policy components. Second, it is unclear how the process of an agent learning from its most recent perceptions and selecting the next action can, in general, be decomposed into a learning phase and an acting phase, and without such a decomposition it is unclear how the counterfactual question ``what would the agent have done if placed back in the same situation with its updated policy'' can be formalized. To circumvent these challenges, we focused on BAC algorithms, which include a clear separation of contextual and policy components of memory, and which also include a clear separation of learning and acting phases. 

We then introduced reinforcement baselines as one strategy for inducing a reinforcement bias. These baselines differ from standard baselines in policy gradient methods only in their purpose (not in their operation)---they are designed to induce a reinforcement bias, not to minimize the variance of policy gradient updates nor to modulate how quickly the agent commits to behaviors. We suggested that for many non-adversarial environments---those where the agent typically does not initially favor suboptimal actions---reinforcement baselines might improve the agent's qualia in the reinforcement-qualia setting without causing significant degradation of performance. However, for environments in which the agent typically favors suboptimal actions initially, even moderate reinforcement baselines can have a significant detrimental impact on performance. 
We supported these suggestions with empirical studies on a gridworld environment and chain environment, demonstrating the existence of both cases: the gridworld represents a case where performance is robust to reinforcement baselines, whereas the chain environment represents a case where performance is sensitive to reinforcement baselines.

\subsection{Future Directions}

In addition to the clear next steps of evaluating how well the qualia objective functions proposed in this report align with human phenomenology, and of developing new qualia objectives that align more closely, there are many other directions for future work. We highlight a few promising examples below.

\begin{enumerate}[leftmargin=1.8em, itemsep=0ex]
    \item In this report we make many (sometimes implicit) assumptions that warrant further scrutiny. For instance, we assumed that random variables representing states, perceptions, memories, and actions are related to the underlying physical system through deterministic and invertible representation functions. However, the consequences of altering these assumptions remain unexplored. For example, we did not consider how the framework would behave under non-invertible representation functions, although we expect that the issues related to representation-exploitability would only become more severe. 
    
    We also did not thoroughly explore the implications of $\Phi$ corresponding to the actual physical system, instead assuming that $\Phi$ corresponds to \emph{properties} of that system.\footnote{Here $\Phi$ might correspond to any term like $\Phi_{S_t}$ or $\Phi_{M_t}$, or a collection of such terms.} Representation exploitability seems particularly troubling when $\Phi$ corresponds to properties, since it seems intuitive that merely re-encoding these properties should not alter the quality of the agent's qualia. If, instead, $\Phi$ referred directly to the physical system, the issue might appear less troubling, since there would be no arbitrary representational choice---the representation would be fixed by the system itself (i.e., the representation functions might naturally be restricted to identity functions).

    However, defining $\Phi$ to be the actual physical system raises deeper questions about whether, and in what sense, a qualia objective can be defined directly over a physical system, rather than over variables that describe its properties. Formal reasoning requires working with mathematical abstractions---such as random variables, measurable functions, or sets---which have precisely defined structures. A physical system, in contrast, does not have a formal mathematical structure until it is explicitly represented through such abstractions. Without first mapping the system into a well-defined mathematical representation, it remains unclear how a qualia objective could be formulated or evaluated. This connects to broader topics in philosophy of mind, philosophy of science, and epistemology, including the nature of representation and how mathematical models relate to the physical systems they are intended to describe.
    \item More broadly, we have not examined how the assumptions made here align---or conflict---with prominent theories in the philosophy of mind, such as physicalism, functionalism, or computationalism. Establishing these connections may help clarify which assumptions are defensible and whether this report meaningfully reflects any particular philosophical stance. 
    \item On a more technical front, we made several simplifying assumptions and decisions that could be reconsidered. For example, we only considered AEPs with discrete actions. As another example, we only considered AEIs that interact with the agent and the base environment at the same timescale. Alternate formulations might allow the AEI to interact with the environment multiple times between interactions with the agent, or vice versa. As one final example, we focused on episodic settings and did not explore how continuing environments might yield stationary distributions over variables such as perceptions or agent memories, and how these might factor into qualia objective functions that more closely resemble the performance objectives used for ergodic average-reward MDPs.
    \item After presenting the dual agent-environment strategy, we removed the AEI from consideration. Future work might revisit this decision, instead exploring other ways for ruling-out the dual agent-environment strategy while still allowing for an AEI-like mechanism. This might better align qualia optimization with the motivating example in Section \ref{sec:shuttleBox}.

    Reformulations that allow for the inclusion of an AEI could enable strategies that improve agent experiences over the long term. If the quality of the agent's qualia decreases as the policy converges (e.g., as in Figure \ref{fig:gridworldReinforcement} with $c=5$), then AEIs that introduce nonstationarity into the environment (e.g., cycling the meaning of the agent's actions in some smooth way) could sustain the ongoing learning and behavioral change necessary for improved long-term qualia.  
    \item In this report we focus on the quality of an agent's qualia, as measured by a single scalar value. However, experiences often involve multiple distinct dimensions. For example, an experience might be simultaneously pleasurable yet stressful, or painful yet meaningful. This suggests that, in some cases, qualia may be more appropriately modeled as a vector-valued quantity that captures these distinct experiential dimensions. In such cases, multiobjective optimization may provide useful tools for reasoning about trade-offs across different dimensions of experience.
    \item This report focused heavily on RL settings, even though the ideas extend to other settings wherein an agent interacts with an environment. As mentioned in Section \ref{sec:policies}, each time step could correspond to a supervised learning model making one or more predictions during training or testing, or to the execution of a single (possibly stochastic) unit in an artificial neural network. Future work might further explore such settings, including those in which agents are deterministic. Natural qualia objective functions for these settings may be quite different, so exploring these settings may help uncover qualia objective functions not apparent from the RL viewpoint.
    \item Another direction for future work involves formally analyzing a distinction between two classes of qualia objectives that emerge in this report. When using some qualia objectives---such as those based on agent rewards or RPEs---the quality of an agent's qualia at time $t$ depends only on the state of the system at that time. In contrast, when using reinforcement-based qualia objectives, what matters is not just the state at time $t$, but how the agent arrived at that state (e.g., via reinforcement or inhibition). These two types of objectives seem to reflect different assumptions about whether qualia supervene on instantaneous mental states or on temporally extended processes. 
    
    This distinction raises several open questions. Can this distinction be made formal and precise, or does it dissolve under formalization---revealing that the two cases are mathematically equivalent under different ways of attributing qualia across time? Do these classes differ in their susceptibility to representation-exploitability? Do they lead to fundamentally different optimization strategies? If the current state of the system (environment and agent) at time $t$ is insufficient to determine the agent's qualia at time $t$, then has the system state been incorrectly defined? 
    \item In the reinforcement-qualia setting we have not yet identified an algorithm-independent definition of reinforcement. Our likelihood-ratio construction works well for algorithms like BAC, where memory naturally separates into policy and contextual components, and where each agent update naturally divides into a learning phase followed by a stochastic acting phase. However, this construction breaks down for general AEPs, in which memory cannot be easily partitioned into policy and context components, and in which the learning and acting phases are entangled. A truly general measure of reinforcement would allow us to counterfactually query how much more likely a previously taken action would be after a policy update, given an identical situation, but without additional learning updates. 

    Developing such a general definition of reinforcement, or proving that no such definition exists, would be a natural next step in formalizing the reinforcement-qualia setting. It would help clarify whether reinforcement-based qualia can be defined independently of algorithm-specific design choices---such as the clean memory decomposition and staged learning-acting structure assumed by BAC---or whether these assumptions are essential. In doing so, it may reveal whether reinforcement-qualia objectives reflect a general computational principle or instead emerge only under particular design assumptions about how agents learn and act.
    \item Related to the previous direction, in Section \ref{sec:isRQReasonable} we suggested that, if the agent is viewed as being part of the environment rather than distinct from it, then the agent's actions would encompass both external interactions (how the agent influences the external environment) and internal processes (how the agent influences itself). However, the reinforcement-qualia objectives explored in this report focused only on reinforcement of external interactions (MDP actions), and did not address the broader notion of reinforcement that includes how an agent influences itself (memory updates). Future work could explore the reinforcement-qualia framework within this broader formulation. 
    \item In this report we modeled qualia as belonging specifically to the agent within an agent-environment system. As we saw with the dual agent-environment strategy, only considering the qualia of part of the system (the part we deem to be the agent) can lead to unsatisfying solutions. A more general formulation might instead model qualia as emerging from the entire agent-environment system without singling out a distinct agent component. Such an approach would require defining qualia and performance objectives at the system level. 
    \item Although the focus of this work was on taking qualia into consideration when designing ML algorithms, the qualia optimization framework may also offer insights into philosophy of mind. For example, if no representation-robust qualia objective function can be found that aligns with human phenomenology, this could challenge certain versions of computationalism or functionalism that posit qualia as arising solely from structural or computational features. Conversely, the identification of such a function would align with these theories.
\end{enumerate}

\subsection{Closing Comments}

Practitioners might consider incorporating reinforcement-bias mechanisms into their algorithms, particularly in cases where doing so does not significantly degrade performance. Even if the reinforcement-qualia hypothesis ultimately proves imperfect, it may still function as a useful surrogate objective until a qualia objective that more accurately reflects human phenomenology is identified.

Even if current AI systems do not possess qualia, the possibility that future systems might motivates the development of a formal framework for qualia optimization. Developing such a framework contributes both mathematical structure and conceptual clarity to this emerging area. In addition to informing the design of learning algorithms, the formulations presented here may also serve as tools for advancing debates in the philosophy of mind. We encourage the research community to evaluate, refine, and extend these foundations as part of a broader effort to understand and influence the quality of qualia that future systems may come to possess.

\bibliography{PoAM}

\begin{thebibliography}{67}
\providecommand{\natexlab}[1]{#1}
\providecommand{\url}[1]{\texttt{#1}}
\expandafter\ifx\csname urlstyle\endcsname\relax
  \providecommand{\doi}[1]{doi: #1}\else
  \providecommand{\doi}{doi: \begingroup \urlstyle{rm}\Url}\fi

\bibitem[Agarwal et~al.(2020)Agarwal, Kakade, Lee, and Mahajan]{pmlr-v125-agarwal20a}
A.~Agarwal, S.~M. Kakade, J.~D. Lee, and G.~Mahajan.
\newblock Optimality and approximation with policy gradient methods in {M}arkov decision processes.
\newblock In \emph{Proceedings of Thirty Third Conference on Learning Theory}, volume 125 of \emph{Proceedings of Machine Learning Research}, pages 64--66, 2020.

\bibitem[Barsalou(1999)]{barsalou1999perceptual}
L.~W. Barsalou.
\newblock Perceptual symbol systems.
\newblock \emph{Behavioral and Brain Sciences}, 22\penalty0 (4):\penalty0 577--660, 1999.

\bibitem[Barto(2013)]{barto2013intrinsic}
A.~G. Barto.
\newblock Intrinsic motivation and reinforcement learning.
\newblock \emph{Intrinsically Motivated Learning in Natural and Artificial Systems}, pages 17--47, 2013.

\bibitem[Bayer and Glimcher(2005)]{bayer2005midbrain}
H.~M. Bayer and P.~W. Glimcher.
\newblock Midbrain dopamine neurons encode a quantitative reward prediction error signal.
\newblock \emph{Neuron}, 47\penalty0 (1):\penalty0 129--141, 2005.

\bibitem[Berridge(2007)]{Berridge2007}
K.~C. Berridge.
\newblock The debate over dopamine's role in reward: the case for incentive salience.
\newblock \emph{Psychopharmacology}, 191\penalty0 (3):\penalty0 391--431, 2007.

\bibitem[Berridge and Kringelbach(2015)]{berridge2015pleasure}
K.~C. Berridge and M.~L. Kringelbach.
\newblock Pleasure systems in the brain.
\newblock \emph{Neuron}, 86\penalty0 (3):\penalty0 646--664, 2015.

\bibitem[Berridge and Robinson(1998)]{berridge1998role}
K.~C. Berridge and T.~E. Robinson.
\newblock What is the role of dopamine in reward: {H}edonic impact, reward learning, or incentive salience?
\newblock \emph{Brain Research Reviews}, 28\penalty0 (3):\penalty0 309--369, 1998.

\bibitem[Bhatnagar et~al.(2007)Bhatnagar, Ghavamzadeh, Lee, and Sutton]{Bhatnagar2007}
S.~Bhatnagar, M.~Ghavamzadeh, M.~Lee, and R.~S. Sutton.
\newblock Incremental natural actor-critic algorithms.
\newblock In \emph{Advances in Neural Information Processing Systems}, volume~20, 2007.

\bibitem[Bhatnagar et~al.(2009)Bhatnagar, Sutton, Ghavamzadeh, and Lee]{Bhatnagar2009}
S.~Bhatnagar, R.~S. Sutton, M.~Ghavamzadeh, and M.~Lee.
\newblock Natural actor-critic algorithms.
\newblock \emph{Automatica}, 45\penalty0 (11):\penalty0 2471--2482, 2009.

\bibitem[Casella and Berger(2001)]{casella2020statistical}
G.~Casella and R.~L. Berger.
\newblock \emph{Statistical Inference}.
\newblock Cengage Learning, 2nd edition, 2001.

\bibitem[Castro and Berridge(2014)]{Castro2014}
D.~C. Castro and K.~C. Berridge.
\newblock Opioid hedonic hotspot in nucleus accumbens shell: {M}u, delta, and dappa maps for enhancement of sweetness ``liking'' and ``wanting''.
\newblock \emph{Journal of Neuroscience}, 34\penalty0 (12):\penalty0 4239--4250, 2014.

\bibitem[Chalmers(1996)]{Chalmers1996}
D.~J. Chalmers.
\newblock Does a rock implement every finite-state automaton?
\newblock \emph{Synthese}, 108:\penalty0 309--333, 1996.

\bibitem[Chung et~al.(2021)Chung, Thomas, Machado, and Roux]{chung2021beyond}
W.~Chung, V.~Thomas, M.~C. Machado, and N.~L. Roux.
\newblock Beyond variance reduction: {U}nderstanding the true impact of baselines on policy optimization.
\newblock In \emph{Proceedings of the 38th International Conference on Machine Learning}, volume 139 of \emph{Proceedings of Machine Learning Research}, pages 1999--2009, 2021.

\bibitem[Claridge-Chang et~al.(2009)Claridge-Chang, Roorda, Vrontou, Sjulson, Li, Hirsh, and Miesenb{\"o}ck]{claridge2009writing}
A.~Claridge-Chang, R.~D. Roorda, E.~Vrontou, L.~Sjulson, H.~Li, J.~Hirsh, and G.~Miesenb{\"o}ck.
\newblock Writing memories with light-addressable reinforcement circuitry.
\newblock \emph{Cell}, 139\penalty0 (2):\penalty0 405--415, 2009.

\bibitem[Clarke(2022)]{clarke2022mapping}
S.~Clarke.
\newblock Mapping the visual icon.
\newblock \emph{The Philosophical Quarterly}, 72\penalty0 (3):\penalty0 552--577, 2022.

\bibitem[{Cleveland Clinic}(2022)]{clevelandclinic_dopamine}
{Cleveland Clinic}.
\newblock Dopamine: What it is, function \& symptoms, 2022.
\newblock URL \url{https://my.clevelandclinic.org/health/articles/22581-dopamine}.
\newblock Accessed: 2025-02-27.

\bibitem[Cottingham et~al.(1991)Cottingham, Murdoch, and Stoothoff]{ElizabethDescartes}
J.~Cottingham, D.~Murdoch, and R.~Stoothoff.
\newblock \emph{The Philosophical Writings of Descartes (Volume 3: The Correspondence)}.
\newblock Cambridge University Press, 1991.

\bibitem[Cover and Thomas(2006)]{CoverThomas2006}
T.~M. Cover and J.~A. Thomas.
\newblock \emph{Elements of Information Theory}.
\newblock Wiley-Interscience, 2nd edition, 2006.

\bibitem[Dabney et~al.(2020)Dabney, Kurth-Nelson, Uchida, Starkweather, Hassabis, Munos, and Botvinick]{dabney2020distributional}
W.~Dabney, Z.~Kurth-Nelson, N.~Uchida, C.~K. Starkweather, D.~Hassabis, R.~Munos, and M.~Botvinick.
\newblock A distributional code for value in dopamine-based reinforcement learning.
\newblock \emph{Nature}, 577\penalty0 (7792):\penalty0 671--675, 2020.

\bibitem[D'Ardenne et~al.(2008)D'Ardenne, McClure, Nystrom, and Cohen]{d2008bold}
K.~D'Ardenne, S.~M. McClure, L.~E. Nystrom, and J.~D. Cohen.
\newblock {BOLD} responses reflecting dopaminergic signals in the human ventral tegmental area.
\newblock \emph{Science}, 319\penalty0 (5867):\penalty0 1264--1267, 2008.

\bibitem[Descartes(1644)]{descartes1644principia}
R.~Descartes.
\newblock \emph{Principia Philosophiae}.
\newblock apud Ludovicum Elzevirium, Amsterdam, 1644.

\bibitem[Faure et~al.(2010)Faure, Richard, and Berridge]{Faure2010}
A.~Faure, J.~M. Richard, and K.~C. Berridge.
\newblock Desire and dread from the nucleus accumbens: {C}ortical glutamate and subcortical {GABA} differentially generate motivation and hedonic impact in the rat.
\newblock \emph{PLOS One}, 5\penalty0 (6):\penalty0 e11223, 2010.

\bibitem[Fazelpour and Danks(2021)]{fazelpour2021algorithmic}
S.~Fazelpour and D.~Danks.
\newblock Algorithmic bias: {S}enses, sources, solutions.
\newblock \emph{Philosophy Compass}, 16\penalty0 (8):\penalty0 e12760, 2021.

\bibitem[Fodor(1975)]{fodor1975language}
J.~A. Fodor.
\newblock \emph{The Language of Thought}.
\newblock Harvard University Press, Cambridge, MA, 1975.

\bibitem[Fodor(2000)]{Fodor2000}
J.~A. Fodor.
\newblock \emph{The Mind Doesn't Work That Way: {T}he Scope and Limits of Computational Psychology}.
\newblock MIT Press, Cambridge, MA, 2000.

\bibitem[Fodor and Pylyshyn(1988)]{fodor1988connectionism}
J.~A. Fodor and Z.~W. Pylyshyn.
\newblock Connectionism and cognitive architecture: A critical analysis.
\newblock \emph{Cognition}, 28\penalty0 (1--2):\penalty0 3--71, 1988.

\bibitem[Freed(2022)]{Freed2022}
W.~J. Freed.
\newblock \emph{Biology of Motivation, Dopamine, and Brain Circuits That Mediate Pleasure}, pages 105--119.
\newblock Springer International Publishing, 2022.

\bibitem[Godfrey-Smith(2009)]{GodfreySmith2009}
P.~Godfrey-Smith.
\newblock Triviality arguments against functionalism.
\newblock \emph{Philosophical Studies}, 45:\penalty0 273--295, 2009.

\bibitem[Greensmith et~al.(2004)Greensmith, Bartlett, and Baxter]{greensmith2004variance}
E.~Greensmith, P.~L. Bartlett, and J.~Baxter.
\newblock Variance reduction techniques for gradient estimates in reinforcement learning.
\newblock \emph{Journal of Machine Learning Research}, 5:\penalty0 1471--1530, 2004.

\bibitem[Grube(1977)]{plato_phaedo}
G.~Grube.
\newblock \emph{Plato Phaedo}.
\newblock Hackett Publishing Company, Inc., 2nd edition, 1977.

\bibitem[Haas(2022)]{haas2022reinforcement}
J.~Haas.
\newblock Reinforcement learning: {A} brief guide for philosophers of mind.
\newblock \emph{Philosophy Compass}, 17\penalty0 (9):\penalty0 e12865, 2022.

\bibitem[Jiang(2019)]{jiang2019value}
N.~Jiang.
\newblock On value functions and the agent-environment boundary.
\newblock \emph{arXiv preprint arXiv:1905.13341}, 2019.

\bibitem[Kaelbling et~al.(1998)Kaelbling, Littman, and Cassandra]{kaelbling1998planning}
L.~P. Kaelbling, M.~L. Littman, and A.~R. Cassandra.
\newblock Planning and acting in partially observable stochastic domains.
\newblock \emph{Artificial Intelligence}, 101\penalty0 (1--2):\penalty0 99--134, 1998.

\bibitem[Konda and Tsitsiklis(1999)]{konda1999actor}
V.~Konda and J.~Tsitsiklis.
\newblock Actor-critic algorithms.
\newblock In \emph{Advances in Neural Information Processing Systems}, volume~12, 1999.

\bibitem[Kringelbach and Berridge(2009)]{kringelbach2009towards}
M.~L. Kringelbach and K.~C. Berridge.
\newblock Towards a functional neuroanatomy of pleasure and happiness.
\newblock \emph{Trends in Cognitive Sciences}, 13\penalty0 (11):\penalty0 479--487, 2009.

\bibitem[Lacy(2024)]{Lacy2024}
D.~Lacy.
\newblock Qualia optimization in reinforcement learning: {B}alancing agent experience and performance.
\newblock \emph{University of Massachusetts Amherst}, 2024.
\newblock Honors Thesis.

\bibitem[Lacy and Thomas(2025)]{Lacy2025}
D.~Lacy and P.~S. Thomas.
\newblock Qualia optimization in reinforcement learning: {B}alancing agent experience and performance.
\newblock Manuscript in preparation, 2025.

\bibitem[Lycan(1981)]{Lycan1981}
W.~G. Lycan.
\newblock Form, function, and feel.
\newblock \emph{The Journal of Philosophy}, 78:\penalty0 24--50, 1981.

\bibitem[Mahler et~al.(2007)Mahler, Smith, and Berridge]{Mahler2007}
S.~V. Mahler, K.~S. Smith, and K.~C. Berridge.
\newblock Endocannabinoid hedonic hotspot for sensory pleasure: {A}nandamide in nucleus accumbens shell enhances `liking' of a sweet reward.
\newblock \emph{Neuropsychopharmacology}, 32\penalty0 (11):\penalty0 2267--2278, 2007.

\bibitem[Maley(2024)]{maley2024computation}
C.~J. Maley.
\newblock Computation for cognitive science: Analog versus digital.
\newblock \emph{Wiley Interdisciplinary Reviews: Cognitive Science}, 15\penalty0 (4):\penalty0 e1679, 2024.

\bibitem[Matsumoto and Hikosaka(2009)]{matsumoto2009two}
M.~Matsumoto and O.~Hikosaka.
\newblock Two types of dopamine neuron distinctly convey positive and negative motivational signals.
\newblock \emph{Nature}, 459:\penalty0 837--841, 2009.

\bibitem[Nota and Thomas(2020)]{Nota2020policy}
C.~Nota and P.~S. Thomas.
\newblock Is the policy gradient a gradient?
\newblock In \emph{Proceedings of the 19$^\text{th}$ International Conference on Autonomous Agents and Multiagent Systems}, 2020.

\bibitem[Peci\~{n}a and Berridge(2005)]{Pecina2005}
S.~Peci\~{n}a and K.~C. Berridge.
\newblock Hedonic hot spot in nucleus accumbens shell: {W}here do mu-opioids cause increased hedonic impact of sweetness?
\newblock \emph{Journal of Neuroscience}, 25\penalty0 (50):\penalty0 11777--11786, 2005.

\bibitem[Piccinini(2008)]{Piccinini2008}
G.~Piccinini.
\newblock Computation without representation.
\newblock \emph{Philosophical Studies}, 137:\penalty0 205--241, 2008.

\bibitem[Pinsker(1964)]{pinsker1964}
M.~S. Pinsker.
\newblock \emph{Information and Information Stability of Random Variables and Processes}.
\newblock Holden-Day, Inc., 1964.

\bibitem[Putnam(1988)]{Putnam1988}
H.~Putnam.
\newblock \emph{Representation and Reality}.
\newblock MIT Press, Cambridge, MA, 1988.

\bibitem[Qiao and Minematsu(2010)]{qiao2010study}
Y.~Qiao and N.~Minematsu.
\newblock A study on invariance of $f$-divergence and its application to speech recognition.
\newblock \emph{IEEE Transactions on Signal Processing}, 58\penalty0 (7):\penalty0 3884--3890, 2010.

\bibitem[Quilty-Dunn(2020)]{quilty2020perceptual}
J.~Quilty-Dunn.
\newblock Perceptual pluralism.
\newblock \emph{No{\^u}s}, 54\penalty0 (4):\penalty0 807--838, 2020.

\bibitem[Robert et~al.(1999)Robert, Casella, and Casella]{robert1999monte}
C.~P. Robert, G.~Casella, and G.~Casella.
\newblock \emph{Monte {C}arlo Statistical Methods}, volume~2.
\newblock Springer, 1999.

\bibitem[Roberts et~al.(2022)Roberts, Zhang, Bariach, Cowls, Gilburt, Juneja, Tsamados, Ziosi, Taddeo, and Floridi]{roberts2022artificial}
H.~Roberts, J.~Zhang, B.~Bariach, J.~Cowls, B.~Gilburt, P.~Juneja, A.~Tsamados, M.~Ziosi, M.~Taddeo, and L.~Floridi.
\newblock Artificial intelligence in support of the circular economy: {E}thical considerations and a path forward.
\newblock \emph{AI \& Society}, pages 1--14, 2022.

\bibitem[Schulman et~al.(2017)Schulman, Wolski, Dhariwal, Radford, and Klimov]{schulman2017proximal}
J.~Schulman, F.~Wolski, P.~Dhariwal, A.~Radford, and O.~Klimov.
\newblock Proximal policy optimization algorithms.
\newblock \emph{arXiv preprint arXiv:1707.06347}, 2017.

\bibitem[Schultz et~al.(1997)Schultz, Dayan, and Montague]{Schultz1997}
W.~Schultz, P.~Dayan, and P.~Montague.
\newblock A neural substrate of prediction and reward.
\newblock \emph{Science}, 275:\penalty0 1593--1599, 1997.

\bibitem[Searle(1992)]{Searle1992}
J.~R. Searle.
\newblock \emph{The Rediscovery of the Mind}.
\newblock MIT Press, Cambridge, MA, 1992.

\bibitem[Smith et~al.(2011)Smith, Berridge, and Aldridge]{Smith2011}
K.~S. Smith, K.~C. Berridge, and J.~W. Aldridge.
\newblock Disentangling pleasure from incentive salience and learning signals in brain reward circuitry.
\newblock \emph{Proceedings of the National Academy of Sciences of the United States of America}, 108\penalty0 (27):\penalty0 E255--E264, 2011.

\bibitem[Sprevak(2018)]{Sprevak2018}
M.~Sprevak.
\newblock Triviality arguments about computational implementation.
\newblock In M.~Sprevak and M.~Colombo, editors, \emph{Routledge Handbook of the Computational Mind}, pages 175--191. Routledge, 2018.

\bibitem[Steinberg et~al.(2013)Steinberg, Keiflin, Boivin, Witten, Deisseroth, and Janak]{steinberg2013causal}
E.~E. Steinberg, R.~Keiflin, J.~R. Boivin, I.~B. Witten, K.~Deisseroth, and P.~H. Janak.
\newblock A causal link between prediction errors, dopamine neurons and learning.
\newblock \emph{Nature Neuroscience}, 16\penalty0 (7):\penalty0 966--973, 2013.

\bibitem[Stevens(1970)]{Stevens1970}
S.~S. Stevens.
\newblock Neural events and the psychophysical law.
\newblock \emph{Science}, 170\penalty0 (3962):\penalty0 1043--1050, 1970.

\bibitem[Strens(2000)]{strens2000bayesian}
M.~Strens.
\newblock A {B}ayesian framework for reinforcement learning.
\newblock In \emph{Proceedings of the Seventeenth International Conference on Machine Learning}, volume 2000, pages 943--950, 2000.

\bibitem[Sutton et~al.(1999)Sutton, Precup, and Singh]{Sutton1999}
R.~Sutton, D.~Precup, and S.~Singh.
\newblock Between {MDP}s and semi-{MDP}s: {A} framework for temporal abstraction in reinforcement learning.
\newblock \emph{Artificial Intelligence}, 112:\penalty0 181--211, 1999.

\bibitem[Sutton and Barto(1998)]{SuttonBarto}
R.~S. Sutton and A.~G. Barto.
\newblock \emph{Reinforcement Learning: {A}n Introduction}.
\newblock MIT Press, 1998.

\bibitem[Sutton and Barto(2018)]{sutton2018reinforcement}
R.~S. Sutton and A.~G. Barto.
\newblock \emph{Reinforcement Learning: {A}n Introduction}.
\newblock MIT press, 2nd edition, 2018.

\bibitem[Sutton et~al.(2000)Sutton, McAllester, Singh, and Mansour]{Sutton2000}
R.~S. Sutton, D.~McAllester, S.~Singh, and Y.~Mansour.
\newblock Policy gradient methods for reinforcement learning with function approximation.
\newblock In \emph{Advances in Neural Information Processing Systems 12}, pages 1057--1063, 2000.

\bibitem[Thomas(2014)]{Thomas2014}
P.~S. Thomas.
\newblock Bias in natural actor-critic algorithms.
\newblock In \emph{Proceedings of the 31st International Conference on Machine Learning}, volume~32 of \emph{Proceedings of Machine Learning Research}, pages 441--448, 2014.

\bibitem[Waddell(2013)]{waddell2013reinforcement}
S.~Waddell.
\newblock Reinforcement signalling in {D}rosophila; dopamine does it all after all.
\newblock \emph{Current Opinion in Neurobiology}, 23\penalty0 (3):\penalty0 324--329, 2013.

\bibitem[Weiskrantz(1990)]{weiskrantz1990blindsight}
L.~Weiskrantz.
\newblock \emph{Blindsight: {A} case study and implications}.
\newblock Oxford University Press, 1990.

\bibitem[White(2017)]{white2017unifying}
M.~White.
\newblock Unifying task specification in reinforcement learning.
\newblock In \emph{Proceedings of the 34th International Conference on Machine Learnin}, volume~70 of \emph{Proceedings of Machine Learning Research}, pages 3742--3750, 2017.

\bibitem[Williams(1992)]{Williams1992}
R.~J. Williams.
\newblock Simple statistical gradient-following algorithms for connectionist reinforcement learning.
\newblock \emph{Machine Learning}, 8:\penalty0 229--256, 1992.

\end{thebibliography}

\clearpage

\begin{appendix}

\section{Notation}
\label{app:notation}

We adopt the following conventions:
\begin{enumerate}[leftmargin=1.8em, itemsep=0ex]
    \item We write $\mathbb N$, $\mathbb N_{>0}$, $\mathbb R$, and $\mathbb R_{>0}$ to denote the sets of natural numbers (including zero), natural numbers greater than zero, real numbers, and real numbers greater than zero, respectively.
    \item We use uppercase calligraphic letters to represent sets (e.g., $\mathcal X$).
    \item We use uppercase (not calligraphic) letters to represent random variables (e.g., $X$).
    \item We use lowercase letters to represent constants, elements of sets, and realizations of random variables (e.g., $c \in \mathbb R$, $x \in \mathcal X$ or $X=x$).
    \item We also use lowercase letters to represent functions (e.g., $f$, $g$, and $h$).
    \item We use braces to denote sets (e.g., $\{\text{red},\text{green},\text{blue}\}$).
    \item We use parentheses to denote sequences (e.g., $(1,1,2,3,5)$).
    \item We write $(x_i)_{i=0}^n$ as shorthand for the sequence $(x_0, x_1, \dotsc,x_n)$.
    \item We write $f:\mathcal X \to \mathcal Y$ to denote that $f$ is a function with domain $\mathcal X$ and range (codomain) $\mathcal Y$. 
    \item If $f:\mathcal X \times \mathcal Y \to \mathcal Z$, we write $f(\cdot,y)$ to denote the function that results from $f$ with the specified fixed $y$ value. That is, $f(\cdot,y):\mathcal X \to \mathcal Z$ and $f(\cdot,y)(x)=f(x,y)$.
    \item We write $a \triangleq b$ to denote that $a$ is defined to be $b$. When there are multiple definitions of a symbol, e.g., $\mathfrak q$, we do not use the $\triangleq$ symbol.
    \item We reserve the symbols $i$, $j$, and $k$ for indices.
    \item We reserve the symbols $f$ and $g$ (with subscripts) for functions. 
    \item We reserve the symbol $d$ (with subscripts) for probability distributions.
    \item If $d$ is a probability distribution, we write $X \sim d$ to denote that $X$ has distribution $d$ (or in pseudocode, to indicate that $X$ is sampled from the distribution $d$).
    \item If $d$ is a probability distribution, we write $\operatorname{supp}(d)$ to denote the support of $d$. Similarly, if $X$ is a random variable, we write $\operatorname{supp}(X)$ to denote the support of $X$.
    \item  We abbreviate \emph{probability mass function}, \emph{probability density function}, and \emph{cumulative distribution function} to PMF, PDF, and CDF respectively. 
    \item We write $\operatorname{law}(X)$ to denote the distribution of a random variable $X$.
    \item We write $\circ$ to denote function composition. That is, if $f:\mathcal Y \to \mathcal Z$ and $g:\mathcal X \to \mathcal Y$, then $f \circ g:\mathcal X \to \mathcal Z$ and $(f \circ g)(x)=f(g(x))$.
\end{enumerate}
In some cases these conventions are violated, but such violations should be made clear.

Tables \ref{tab:symbols} and \ref{tab:symbols2} provide lists of symbols and abbreviations, their meanings, and the pages on which they are initially defined. Standard symbols and abbreviations as well as symbols defined and used within a limited scope or in later appendices are not included in either table.

\begin{table*}[htbp]
    \centering
    \begin{minipage}[t]{0.48\textwidth}
        \centering
        \caption{List of Symbols (Part I)}
        \label{tab:symbols}
        \begin{tabular}{rll}
        \hline
        \textbf{Symbol} & \textbf{Meaning} & \textbf{Page} \\ \hline
        $t$ & Time, in $(0,1,\dotsc)$& \pageref{sym:t} \\[3pt]
        $S_t$ & State of the environment at time $t$ & \pageref{sym:St} \\[3pt]
        $M_t$ & Memory of the agent at time $t$ & \pageref{sym:Mt} \\[3pt]
        $P_t$ & (Agent) perception at time $t$ & \pageref{sym:Pt}, \pageref{sym:Pt2} \\[3pt]
        $A_t$ & (Agent) action at time $t$ & \pageref{sym:At}, \pageref{sym:At2} \\[3pt]
        $d_s$ & Next-state distribution & \pageref{sym:ds} \\[3pt]
        $d_m$ & Next-memory distribution & \pageref{sym:dm}, \pageref{sym:dm2} \\[3pt]
        $f_p$ & (Agent) perception function & \pageref{sym:fp}, \pageref{sym:fp2} \\[3pt]
        $f_a$ & (Agent) action function & \pageref{sym:fa}, \pageref{sym:fa2} \\[3pt]
        AEP & Agent-environment process & \pageref{sym:aep} \\[3pt]
        $R_t$ & (Agent) reward at time $t$ & \pageref{sym:Rt}, \pageref{sym:Rt2} \\[3pt]
        $f_r$ & Reward function & \pageref{sym:fr}, \pageref{sym:fr2} \\[3pt]
        AERP & Agent-environment reward process & \pageref{sym:aerp} \\[3pt]
        $\Phi_Z$ & Physical properties of random variable $Z$ & \pageref{sym:PhiZ} \\[3pt]
        $\rho_Z$ & Representation function for random variable $Z$ & \pageref{sym:rhoZ} \\[3pt]
        $\pi$ & State or perception-policy & \pageref{sym:pi} \\[3pt]
        $s_\infty$ & Terminal state & \pageref{sym:sinfty} \\[3pt]
        $p_\infty$ & Terminal perception & \pageref{sym:pinfty} \\[3pt]
        $a_\infty$ & Terminal action & \pageref{sym:ainfty} \\[3pt]
        $\operatorname{start}(i)$ & Start time of $i^\text{th}$ episode & \pageref{sym:start} \\[3pt]
        $\operatorname{end}(i)$ & End time of $i^\text{th}$ episode & \pageref{sym:end} \\[3pt]
        $\operatorname{len}(i)$ & Length of the $i^\text{th}$ episode & \pageref{sym:len} \\[3pt]
        $\operatorname{start}(t)$ & Start time of the episode containing time $t$ & \pageref{sym:startt} \\[3pt]
        $\operatorname{end}(t)$ & End time of the episode containing time $t$ & \pageref{sym:endt} \\[3pt]
        $\operatorname{dur}(t)$ & Duration (time since episode start) at time $t$ & \pageref{sym:durt} \\[3pt]
        $G_i$ & Return of $i^\text{th}$ episode & \pageref{sym:Gi} \\[3pt]
        $G_t$ & Return from time $t$& \pageref{sym:Gt} \\[3pt]
        $i_\text{max}$ & Maximum number of episodes & \pageref{sym:imax} \\[3pt]
        BAC & Basic actor-critic & \pageref{sym:bac} \\[3pt]
        $\Theta_t$ & Policy parameters at time $t$ & \pageref{sym:Thetat} \\[3pt]
        VFA & Value function approximation & \pageref{sym:VFA} \\[3pt]
        $W_t$ & VFA weights at time $t$ & \pageref{sym:Wt} \\[3pt]
        $E_t$ & Eligibility trace vector at time $t$ & \pageref{sym:Et} \\[3pt]
        $\pi_\text{BAC}$ & BAC policy parameterization & \pageref{sym:pibac} \\[3pt]
        $v$ & VFA parameterization & \pageref{sym:v} \\[3pt]
        $\theta_0$ & Initial perception-policy parameter vector & 
        \pageref{sym:theta0} \\[3pt]
        $w_0$ & Initial VFA weight vector & \pageref{sym:w0} \\[3pt]
    \end{tabular}
    \end{minipage}
    \hfill
    \begin{minipage}[t]{0.48\textwidth}
        \centering
        \caption{List of Symbols (Part II)}
    \label{tab:symbols2}
        \begin{tabular}{rll}
        \hline
        \textbf{Symbol} & \textbf{Meaning} & \textbf{Page} \\ \hline
        $\lambda$ & Eligibility trace decay rate & \pageref{sym:lambda} \\[3pt]
        $\alpha$ & Critic step size & \pageref{sym:alpha} \\[3pt]
        $\beta$ & Actor step size & \pageref{sym:beta} \\[3pt]
        $\Delta_t$ & Temporal difference (TD) error at time $t$ & \pageref{sym:deltat} \\[3pt]
        AEI & Agent-environment interface & \pageref{sym:aeixidea} \\[3pt]
        $Y_t$ & AEI state at time $t$ & \pageref{sym:Yt} \\[3pt]
        $Y'_t$ & Intermediate AEI state at time $t$ & \pageref{sym:Ypt} \\[3pt]
        $X_t$ & Base environment state at time $t$ & \pageref{sym:Xt} \\[3pt]
        $\widebar A_t$ & AEI action at time $t$ & \pageref{sym:barAt} \\[3pt]
        $\widebar P_t$ & AEI perception at time $t$ & \pageref{sym:barPt} \\[3pt]
        $d_x$ & Base environment next-state distribution & \pageref{sym:dx} \\[3pt]
        $f_{\widebar p}$ & AEI perception function & \pageref{sym:fbarp} \\[3pt]
        $d_y$ & Next AEI-state distribution& \pageref{sym:dy} \\[3pt]
        $d_{y'}$ & Next intermediate AEI state distribution & \pageref{sym:dyp} \\[3pt]
        $f_{\widebar a}$ & AEI action function & \pageref{sym:fba} \\[3pt]
        AIEP & Agent-interface-environment process & \pageref{sym:aiep} \\[3pt]
        $f_{\widebar r}$ & Base reward function & \pageref{sym:fbarr} \\[3pt]
        $\widebar R_t$ & Base reward & \pageref{sym:barRt} \\[3pt]
        AIERP & Agent-interface-environment reward process & \pageref{sym:aierp} \\[3pt]
        $\texttt{env}$ & Environment distributions & \pageref{sym:env} \\[3pt]
        %
        $\texttt{alg}$ & RL algorithm distributions & \pageref{sym:alg} \\[3pt]
        $\texttt{aei}$ & AEI distributions & \pageref{sym:aeixsym} \\[3pt]
        $\texttt{aei}_\mathbf{I}$ & Identity AEI & \pageref{sym:aeiI} \\[3pt]
        $\mathfrak q$ & Qualia objective function & \pageref{sym:q} \\[3pt]
        $\mathfrak p$ & Performance objective function & \pageref{sym:p} \\[3pt]
        $\sum_{i,t}$ & Shorthand for the summations in \eqref{eq:discountedReturnQualiaDefn}  & \pageref{sym:sumit} \\[3pt]
        $\texttt{aei}_c$ & Reward bonus AEI & \pageref{sym:aeic} \\[3pt]
        $\texttt{alg}_{-c}$ & Reward bonus inverse RL algorithm & \pageref{sym:algc} \\[3pt]
        $\texttt{alg}^{-\text{aei}}$ & Inverse RL algorithm & \pageref{sym:invRLalg} \\[3pt]
        $g_{\widebar p}$ & Inverse RL algorithm perception inverter & \pageref{sym:gbarp} \\[3pt]
        $g_a$ & Inverse RL algorithm action pretransformer& \pageref{sym:ga} \\[3pt]
        $L_t$ & Likelihood ratio at time $t$ for MDPs & \pageref{sym:Lt} \\[3pt]
        $J$ & Discounted objective & \pageref{sym:J} \\[3pt]
        $\nabla J$ & Policy gradient & \pageref{sym:policygradient} \\[3pt]
        $\Theta_{\forall t}=\theta$ & Shorthand for the event $\forall t \in \mathbb N,\,\Theta_t=\theta$ & \pageref{sym:foralltheta} \\[3pt]
        $b$ & Baseline function & \pageref{sym:b} \\[3pt]
    \end{tabular}
    \end{minipage}
    \vspace{2pt}
    \noindent\rule{\textwidth}{0.4pt}
\end{table*}

\subsection{Notation for Markov Decision Processes (MDPs)}
\label{app:MDP}

In this appendix we define the notation that we use for MDPs. A finite, episodic, and discounted MDP is a tuple $(\mathcal S, \mathcal A, p, r, d_0, \gamma)$, where:
\begin{enumerate}[leftmargin=1.8em, itemsep=0ex]
    \item Time is indexed by $t \in \mathbb N$.
    \item $\mathcal S$ is the finite set of all possible states of the environment and is called the \emph{state set}.
    \item For all times $t \in \mathbb N$, $S_t$ is a random variable representing the state of the environment at time $t$ and $\operatorname{supp}(S_t)\subseteq \mathcal S$. 
    \item $\mathcal A$ is the finite set of all possible actions the agent can select and is called the \emph{action set}.\footnote{In more general formulations, the set of admissible actions may vary depending on the current state. We present an MDP formulation where all actions are admissible in all states.} 
    \item For all times $t \in \mathbb N$, $A_t$ is the action chosen by the agent at time $t$ and $\operatorname{supp}(A_t)\subseteq \mathcal A$. 
    \item $p:\mathcal S \times \mathcal A \times \mathcal S \to [0,1]$ is called the \emph{transition function} and characterizes how states transition according to the definition
    \begin{equation}
        p(s,a,s')\triangleq\Pr(S_{t+1}=s'|S_t=s,A_t=a),
    \end{equation}
    for all $s \in \mathcal S$, $a \in \mathcal A$, $s' \in \mathcal S$, and $t \in \mathbb N$. Notice that $p(s,a,s')$ does not depend on $t$, and so the conditional distribution of the next state given the current state and action is stationary.
    \item For all times $t \in \mathbb N_{>0}$, $R_t$ is a real-valued and bounded random variable called the reward at time $t$. Notice that while states and actions begin with $S_0$ and $A_0$, the first reward is $R_1$.
    \item $r:\mathcal S \times \mathcal A \to \mathbb R$ is called the \emph{reward function} and is defined by the equation
    \begin{equation}
        r(s,a)\triangleq\mathbf{E}\left [ R_{t+1} | S_t=s,A_t=a\right ],
    \end{equation}
    for all $s \in \mathcal S$, $a \in \mathcal A$, and $t \in \mathbb N$.\footnote{In some formulations $r(s,a)=\mathbf{E}[R_t|S_t=s,A_t=a]$, which places the reward at the same time step as the state and action via a different indexing of rewards. In contrast, we adopt a definition of the reward function consistent with the \emph{agent-environment process} (AEP) formulation we propose, where the reward follows the state and action that caused it.}
    \item $d_0:\mathcal S \to [0,1]$ is called the \emph{initial state distribution} and characterizes the distribution of the initial state $S_0$ according to the definition
    \begin{equation}
        d_0(s)\triangleq\Pr(S_0=s),
    \end{equation}
    for all $s \in \mathcal S$. \item $\gamma \in [0,1)$ is a constant called the \emph{reward discount parameter}.
\end{enumerate}

An MDP partially characterizes the sequence of random variables 
\begin{equation}
    \label{eq:MDPseq}
    (S_0,A_0,R_1,S_1\dotsc,R_{t-1},S_{t-1},A_{t-1},R_t,S_t,A_t,\dotsc).
\end{equation}
The notation for the transition function suggests a conditional independence assumption within this sequence. However, one can write and reason about conditional probabilities without implying conditional independence. We therefore make the conditional independence assumption of MDPs explicit: For all $t \in \mathbb N$, $S_{t+1}$ is conditionally independent of all previous random variables given $S_t$ and $A_t$. MDPs include a second conditional independence assumption: For all $t \in \mathbb N$, $R_{t+1}$ is conditionally independent of all previous random variables given $S_t$ and $A_t$.

Unlike the \emph{agent-environment processes} (AEPs) defined in this report, MDPs are only a mathematical model of an environment, not the agent that interacts with the environment---hence why they only partially specify the sequence in \eqref{eq:MDPseq}. RL agents that interact with MDPs implement stationary  \emph{policies} $\pi:\mathcal S \times \mathcal A \to [0,1]$, which characterize the conditional distribution of $A_t$ given $S_t$ via the definition 
\begin{equation}
    \pi(s,a)\triangleq\Pr(A_t=a|S_t=s),
\end{equation}
for all $s \in \mathcal S$, $a \in \mathcal A$, and $t \in \mathbb N$.\footnote{Although we define $\pi$ to be a stationary policy, notice that RL agents often change the stationary policy that they use at each time step.} Some RL algorithms maintain a stationary \emph{parameterized policy} with policy parameters $\Theta_t \in \mathbb R^n$ for some $n \in \mathbb N_{>0}$. Abusing notation, when considering the use of a parameterized policy, we define 
\begin{equation}
    \pi(s,a,\theta)\triangleq\Pr(A_t=a|S_t=s,\Theta_t=\theta),
\end{equation}
for all $s \in \mathcal S$, $a \in \mathcal A$, $\theta \in \mathbb R^n$, and $t \in \mathbb R^n$. When considering the use of a parametric policy, we refer to $\pi$ as a \emph{policy parameterization}.

We define stationary policies to be Markovian. That is, under the stationary policy $\pi$,  $A_t$ is conditionally independent of all previous random variables given $S_t$. Similarly, under the stationary parameterized policy $\pi$, $A_t$ is conditionally independent of all previous random variables given $S_t$ and $\Theta_t$. 

Let $G \triangleq \sum_{t=1}^\infty \gamma^t R_t$, which we call the \emph{discounted return}, and let $J:\mathbb R^n \to \mathbb R$ be a function called the \emph{discounted objective}, which takes as input policy parameters and outputs the expected discounted return if the agent were to use an implicit policy parameterization $\pi$ with the specified policy parameters:
\begin{equation}
    \label{eq:RLobjective}
    J(\theta)\triangleq\mathbf{E}\left [ G \middle | \forall t \in \mathbb N,\,\Theta_t = \theta \right ].
\end{equation}
A necessary but not sufficient property of an \emph{optimal policy} \citep{sutton2018reinforcement} is that it maximizes the expected discounted return. Lastly, to simplify later expressions, let 
\begin{equation}
    G_t\triangleq\sum_{k=0}^\infty \gamma^k R_{t+k+1},
\end{equation}
which we call the \emph{discounted return from time $t$}.

\section{Philosophy of Mind Background: Triviality Arguments}
\label{app:trivialityArguments}

A long‐standing concern in the philosophy of mind is how to determine when a physical system truly implements an abstract computation. One influential proposal \citep{Putnam1988} suggests a ``mapping'' theory of implementation---that a physical system implements a computation if there is an isomorphism between the states and transitions of the physical system and those of a formal computational model (for example, a finite‐state automaton). We adopt a similar formalization when reasoning about how an abstract mathematical model of an agent interacting with an environment can correspond to aspects of the real physical world.

However, \citet[Appendix, pp.~121]{Putnam1988} famously argued that ``[every] ordinary open system is a realization of every abstract finite automaton.'' This insight has spurred a number of similar arguments, called \emph{triviality arguments}, which claim that, in the absence of further constraints, virtually any physical system can be interpreted as carrying out any computation. Such arguments are called triviality arguments because they suggest that sophisticated computations (e.g., the computations performed by a human brain) can be trivially implemented by a simple device like a mechanical watch. In Section \ref{sec:reconsideringAEI} we encounter a similar problem, wherein a wide range of seemingly different mathematical models of an agent interacting with an environment---models that suggest different experiences of the agent---can all correspond to the same physical world. We therefore review triviality arguments from philosophy of mind, recognizing that we are far from the first to encounter and reason about such issues. Specifically, \citet{Sprevak2018} provided an overview of four triviality arguments, which we review here.
\\\\
\noindent\textbf{Hinckfuss's Pail Argument.} 
According to \citet{Lycan1981}, Ian Hinckfuss suggested an (unpublished) thought experiment that considers a pail of water resting in the sunlight. Given the immense complexity at the microscopic level (e.g., currents, bacteria breeding, and molecular interactions), there could exist a mapping from the water's states and transitions to the states and transitions of a system executing a human-like program---even if only briefly. This thought experiment highlights how a mapping-based definition of computation could imply that ordinary objects might, by sheer physical complexity, be said to implement highly nontrivial computations. 
\\\\
\noindent\textbf{Searle's Wall Argument.} 
\citet[Chapter 9, Section V]{Searle1992} provided a similar thought experiment, writing:\footnote{We once again reiterate that the four arguments we review are based on the overview by \citet{Sprevak2018}. The quote we provide is a subset of a quote selected by Sprevak, and so one might view this as a quote of Sprevak quoting Searle.}
\begin{quote}\emph{
    1. For any object there is some description of that object such that under that description the object is a digital computer.\\
    2. For any program and for any sufficiently complex object, there is some description of the object under which it is implementing the program. [... If the wall behind me] is a big enough wall it is implementing any program, including any program implemented in the brain.
    }
\end{quote}
Searle's argument further underscores the implications of a mapping-based definition of computation. 
\\\\
\noindent\textbf{Putnam's Rock Argument.} 
\citet{Putnam1988} formalized these arguments by showing that for any open physical system one can partition its phase space---the complete set of its physical parameters---into regions in such a way that the system's evolution (subject to the principles of noncyclical behavior and continuity) can be mapped onto the state transitions of an inputless finite‐state automaton. \citet{Putnam1988} argued that this implies that nearly every open system (for instance, a rock) can therefore be shown to implement every possible inputless finite-state computation. 
\\\\
\noindent\textbf{Chalmers' Clock-and-Dial Argument.} 
One counter-argument to Putnam's argument suggests that a mapping or isomorphism between the states and transitions of the physical system and those of a formal computational model is not sufficient. Rather, the physical system must also be robust to small physical changes, and the mapping must apply to all states that \emph{could} occur in the computational model---not just those that occur given specific inputs. \citet{Chalmers1996} extended Putnam's argument to handle these objections by introducing the concepts of a ``clock'' and a ``dial.'' He argued that if a physical system possesses a component that reliably progresses through a series of states (a clock) and another that can stably hold one of many possible states (a dial), then one can construct a mapping from the system's behavior to the states of any inputless finite-state automaton. Furthermore, Chalmers' formulation was designed to overcome the aforementioned objections to Putnam's example. 
\\\\
There are a wide range of responses to these triviality arguments \citep{Fodor2000,Piccinini2008}, including suggestions of imposing further physical or causal restrictions to rule out arbitrary mappings \citep{Chalmers1996,GodfreySmith2009}. For additional details, we refer the reader to the work of \citet[Section 7]{Sprevak2018}.

\section{Implications of Markov Assumptions}
\label{app:AssumptionImplications}

In this appendix we elaborate on the implications of the Markov assumptions made in Section \ref{sec:MarkovandstationarityAssumptions}. Recall the Markov assumptions:
\begin{itemize}
    \item \emph{Markovian states.} $S_t$ is conditionally independent of all previous random variables given $S_{t-1}$ and $A_{t-1}$. 
    \item \emph{Markovian memories.} $M_t$ is conditionally independent of all previous random variables given $M_{t-1}$ and $P_t$. 
\end{itemize}
We made the following claims:
\begin{quote}
    \emph{
    These two assumptions characterize how states, perceptions, memories, and actions should be defined for a given agent-environment system. 
    Although they are stated as assumptions, they do not restrict the set of agent-environment systems under consideration. 
    }
\end{quote}
We provide supporting evidence for these claims in this appendix, focusing on the Markovian-states assumption first, and then the Markovian-memories assumption.

Notice that in the most general setting $S_{t+1}$ could be influenced by all of the past random variables describing the agent and environment. If \textbf{(a)} $S_t$ is a complete characterization of everything about the environment up to and including time $t$ that influences the environment at times $t' > t$ and \textbf{(b)} $M_t$ is a complete characterization of everything about the agent up to and including time $t$ that influences the environment at times $t' > t$, then $S_{t+1}$ should be conditionally independent of all previous random variables given $S_t$ and $M_t$. Furthermore, if \textbf{(c)} $A_t$ is a complete characterization of everything about $M_t$ that influences the environment at time $t+1$, then we can condition on $A_t$ instead of $M_t$, meaning that $S_{t+1}$ should be conditionally independent of all previous random variables given $S_t$ and $A_t$, as stated in the Markovian-states assumption. Note that \textbf{(a)}, \textbf{(b)}, and \textbf{(c)} are included within the definitions of $S_t$, $M_t$, and $A_t$. So, the Markovian-states assumption formalizes intuition for how states, memories, and actions are defined for a given agent-environment system. 

To see that the Markovian-state assumption does not restrict the set of systems under consideration, notice that the history of all past random variables could be encoded within $S_t$ and $A_t$ (taking some care to ensure that the history of memories is encoded in $S_t$ and $A_t$), thereby ensuring that the Markovian-state assumption necessarily holds. This complete encoding of the history within $S_t$ and $A_t$ would be cumbersome. The Markovian-states assumption allows for more restricted definitions of states and actions by characterizing exactly what information must be encoded within $S_t$ and $A_t$.

Similarly, notice that in the most general setting $M_{t+1}$ could also be influenced by all of the past random variables describing the agent and environment. If \textbf{(d)} $S_t$ is a complete characterization of everything about the environment up to and including time $t$ that influences the agent at times $t'' \geq t$ and \textbf{(e)} $M_t$ is a complete characterization of everything about the agent up to and including time $t$ that influences the agent at times $t' > t$, then $M_{t+1}$ should be conditionally independent of all previous random variables given $S_{t+1}$ and $M_t$. 
Furthermore, if \textbf{(f)} $P_{t+1}$ is a complete characterization of everything about $S_{t+1}$ that influences the agent at time $t+1$, then we can condition on $P_{t+1}$ instead of $S_{t+1}$, meaning that $M_{t+1}$ should be conditionally independent of all previous random variables given $P_{t+1}$ and $M_t$, as stated in the Markovian-memories assumption. Note that \textbf{(d)}, \textbf{(e)}, and \textbf{(f)} are included within the definitions of $S_t$, $M_t$, and $P_{t+1}$. So, like the Markovian-states assumption, the Markovian-memories assumption formalizes intuition for how states, memories, and perceptions are defined for a given agent-environment system. 

To see that the Markovian-memories assumption does not restrict the set of systems under consideration, notice that the history of all past random variables could be encoded within $M_t$ and $P_{t+1}$ (taking some care to ensure that even the history of states is encoded in $M_t$ and $P_{t+1}$), thereby ensuring that the Markovian-memories assumption necessarily holds. This complete encoding of the history within $M_t$ and $P_{t+1}$ would be cumbersome. The Markovian-memories assumption allows for more restricted definitions of memories and perceptions by  characterizing exactly what information must be encoded within the definitions of $M_t$ and $P_{t+1}$.\footnote{Notice that the Markovian-memories assumption ensures that perceptions are defined in a way that encodes the necessary information, but does not limit the perceptions to only include information that influences the agent. For example, defining $P_t=S_t$ is trivially sufficient as a definition of perceptions. While this formulation is sufficient for this initial study, future work might consider restricting the perceptions to only encode information that influences the agent, and might include similar restrictions for actions.}

\section{Proof of Qualia Improvement for Reward Bonuses}
\label{app:RewardBonusProof}

In this appendix we prove a result stated in Section \ref{sec:SpecificExampleWhere}---that
\begin{equation}
    \mathfrak q(\texttt{alg}_{-c},\texttt{aei}_c) = \mathfrak q(\texttt{alg}_0, \texttt{aei}_{\mathbf{I}}) + c \frac{i_\text{max}}{1-\gamma_{\mathfrak q}}. 
\end{equation} 
Recall the (unproven) observation that using $\texttt{aei}_c$ with $\texttt{alg}_{-c}$ only changes the random variable $R_t$---the distributions of all other random variables, $X_t, \widebar P_t, Y_t, P_t, M_t, A_t, Y'_t, \widebar A_t,$ and $\widebar R_t$, remain unchanged. We do not prove this property formally because it should be clear from intuition alone (the only change that $\texttt{aei}_c$ makes is the addition of a constant to the rewards $R_t$, and prior to using these rewards the RL algorithm $\texttt{alg}_{-c}$ subtracts these same constants, with no other changes to the entire system comprised of the base-environment, AEI, and agent). This property allows us to expand the expression for $\mathfrak q(\texttt{alg}_{-c},\texttt{aei}_c)$ as follows (color is used to highlight changes and the relationships between terms in different expressions):
\begin{align}
    \mathfrak q(&\texttt{alg}_{-c},\texttt{aei}_c)=\mathbf{E}\left [\sum_{i=0}^{i_\text{max}-1}\sum_{t=\operatorname{start}(i)+1}^{\operatorname{end}(i)} \gamma_{\mathfrak q}^{\operatorname{dur}(t)} R_t \right ]\\
    \overset{\text{(a)}}{=}& \mathbf{E}\left [\sum_{i=0}^{i_\text{max}-1}\sum_{t=\operatorname{start}(i)+1}^{ \textcolor{eqnGreen}{\operatorname{end}(i) - 1}} \gamma_{\mathfrak q}^{\operatorname{dur}(t)} R_t\right ] + \mathbf{E}\left [\sum_{i=0}^{i_\text{max}-1} \gamma_{\mathfrak q}^{\operatorname{dur}(\operatorname{end}(i))} R_{\operatorname{end}(i)}\right ]\\ 
    \overset{\text{(b)}}{=}& \mathbf{E}\left [\sum_{i=0}^{i_\text{max}-1}\sum_{t=\operatorname{start}(i)+1}^{\operatorname{end}(i) - 1} \gamma_{\mathfrak q}^{\operatorname{dur}(t)} \left (\widebar R_t + c \right )\right ]\\
    & + \mathbf{E}\left [\sum_{i=0}^{i_\text{max}-1} \gamma_{\mathfrak q}^{\operatorname{dur}(\operatorname{end}(i))} \left (\widebar R_{\operatorname{end}(i)} + \frac{c}{1-\gamma_{\mathfrak q}} \right ) \right ]\\ 
    =&\underbrace{\mathbf{E}\left [\sum_{i=0}^{i_\text{max}-1}\sum_{t=\operatorname{start}(i)+1}^{\operatorname{end}(i) - 1} \gamma_{\mathfrak q}^{\operatorname{dur}(t)}\widebar R_t\right ]}_{\text{Term A}} + \underbrace{c\sum_{i=0}^{i_\text{max}-1} \textcolor{eqnRed}{\sum_{t=\operatorname{start}(i)+1}^{\operatorname{end}(i) - 1} \gamma_{\mathfrak q}^{\operatorname{dur}(t)}}}_{\text{Term B}} \\
    &
    \label{eq:tempInRewQualProof}+ \underbrace{\mathbf{E}\left [\sum_{i=0}^{i_\text{max}-1} \gamma_{\mathfrak q}^{\operatorname{dur}(\operatorname{end}(i))} \widebar R_{\operatorname{end}(i)} \right ]}_{\text{Term C}}+ \underbrace{\frac{c}{1-\gamma_{\mathfrak q}}  \textcolor{eqnBlue}{\sum_{i=0}^{i_\text{max}-1} \gamma_{\mathfrak q}^{\operatorname{dur}(\operatorname{end}(i))}}}_{\text{Term D}},
\end{align}
where \textbf{(a)} follows from the linearity of expectations, which allows for the final terms in the summation over $t$ to be separated from the other terms and \textbf{(b)} follows from the definition of $R_t$ in \eqref{eq:rewardBonusTransform}. Notice that Term A and Term C sum to 
\begin{equation}
    \mathbf{E}\left [\sum_{i=0}^{i_\text{max}-1}\sum_{t=\operatorname{start}(i)+1}^{\operatorname{end}(i)} \gamma_{\mathfrak q}^{\operatorname{dur}(t)}\widebar R_t\right ]=\mathfrak q(\texttt{alg}_0,\texttt{aei}_{\textbf{I}}),
\end{equation}
since $R_t=\widebar R_t$ under the identity AEI, $\texttt{aei}_{\textbf{I}}$. We will proceed by simplifying Term B and then Term D, before returning to \eqref{eq:tempInRewQualProof} to obtain a simplified expression for $\mathfrak q(\texttt{alg}_{-c},\texttt{aei}_c)$. 

Recall that $\operatorname{dur}(t)=t-(\operatorname{start}(i)+1)$, and so 
\begin{align}
     \textcolor{eqnRed}{\sum_{t=\operatorname{start}(i)+1}^{\operatorname{end}(i) - 1} \gamma_{\mathfrak q}^{\operatorname{dur}(t)}} =&  \textcolor{eqnRed}{\sum_{t=\operatorname{start}(i)+1}^{\operatorname{end}(i) - 1} \gamma_{\mathfrak q}^{t-(\operatorname{start}(i)+1)}}\\
    \overset{\text{(a)}}{=}& \textcolor{eqnRed}{\sum_{k=0}^{\operatorname{end}(i)- \operatorname{start}(i) - 2} \gamma_{\mathfrak q}^k}\\
    \overset{\text{(b)}}{=}& \textcolor{eqnRed}{\frac{1-\gamma_{\mathfrak q}^{\operatorname{end}(i)- (\operatorname{start}(i)+1)}}{1-\gamma_{\mathfrak q}}},
\end{align}
where \textbf{(a)} follows from a change of variable---that is, applying the rule that for all natural numbers $a$ and $b$ where $b > a$ and all functions $f:\mathbb N \to \mathbb R$, $\sum_{j=a}^b f(j) = \sum_{k=0}^{b-a}f(k+a)$---and \textbf{(b)} follows because the summation is a finite geometric series. Also, notice that by expanding $\operatorname{dur}(\operatorname{end}(i))$ we have that 
\begin{equation}
    \textcolor{eqnBlue}{\sum_{i=0}^{i_\text{max}-1} \gamma_{\mathfrak q}^{\operatorname{dur}(\operatorname{end}(i))}}= \textcolor{eqnBlue}{\sum_{i=0}^{i_\text{max}-1} \gamma_{\mathfrak q}^{\operatorname{end}(i) - (\operatorname{start}(i) + 1)}}.
\end{equation} 
So, the sum of Term B and Term D can be expressed as
\begin{align}
    &c \sum_{i=0}^{i_\text{max}-1} \textcolor{eqnRed}{\sum_{t=\operatorname{start}(i)+1}^{\operatorname{end}(i) - 1} \gamma_{\mathfrak q}^{\operatorname{dur}(t)}} + \frac{c}{1-\gamma_{\mathfrak q}} \textcolor{eqnBlue}{\sum_{i=0}^{i_\text{max}-1} \gamma_{\mathfrak q}^{\operatorname{dur}(\operatorname{end}(i))}}\\
    =& c  \sum_{i=0}^{i_\text{max}-1}  \textcolor{eqnRed}{\frac{1-\gamma_{\mathfrak q}^{\operatorname{end}(i)- (\operatorname{start}(i)+1)}}{1-\gamma_{\mathfrak q}}} + \frac{c}{1-\gamma_{\mathfrak q}}\textcolor{eqnBlue}{\sum_{i=0}^{i_\text{max}-1} \gamma_{\mathfrak q}^{\operatorname{end}(i) - (\operatorname{start}(i) + 1)}}\\
    =& \frac{c}{1-\gamma_{\mathfrak q}} \left ( \sum_{i=0}^{i_\text{max}-1} 1-\gamma_{\mathfrak q}^{\operatorname{end}(i)- (\operatorname{start}(i)+1)} + \sum_{i=0}^{i_\text{max}-1} \gamma_{\mathfrak q}^{\operatorname{end}(i) - (\operatorname{start}(i) + 1)}\right )\\
    =& \frac{c}{1-\gamma_{\mathfrak q}} \sum_{i=0}^{i_\text{max}-1} 1-\gamma_{\mathfrak q}^{\operatorname{end}(i)- (\operatorname{start}(i)+1)} + \gamma_{\mathfrak q}^{\operatorname{end}(i) - (\operatorname{start}(i) + 1)}\\
    =& \frac{c}{1-\gamma_{\mathfrak q}} \sum_{i=0}^{i_\text{max}-1} 1\\
    =& \frac{c\,i_\text{max}}{1-\gamma_{\mathfrak q}}.
\end{align}

Returning to \eqref{eq:tempInRewQualProof}, we have that the sum of Term A and Term C is $\mathfrak q(\texttt{alg}_0, \texttt{aei}_{\mathbf{I}})$ and that the sum of Term B and Term D is $c\,i_\text{max}(1-\gamma_{\mathfrak q})^{-1}$, and so we can conclude that
\begin{equation}
    \mathfrak q(\texttt{alg}_{-c},\texttt{aei}_c) = \mathfrak q(\texttt{alg}_0, \texttt{aei}_{\mathbf{I}}) + c \frac{i_\text{max}}{1-\gamma_{\mathfrak q}}.
\end{equation}

\section{Information Theory Background}
\label{app:infoTheory}

In this appendix we review notation, definitions, and properties from information theory. We emphasize that the results presented in this appendix are well-known and not contributions of this report, although in some cases proofs are provided for completeness. Although elsewhere $X$ is a specific random variable in our formulations (the base environment state), here we use $X$ to denote an arbitrary random variable.

We begin by defining the entropy of discrete random variables. 

\begin{defn}[Shannon Entropy -- Discrete]
    The entropy of a discrete random variable $X$ with support $\mathcal X$ and PMF $p$ is
    \begin{equation}
        H(X)\triangleq -\sum_{x \in \mathcal X} p(x) \log_2\big (p(x)\big ).
    \end{equation}
\end{defn}
Shannon extended entropy to continuous random variables, defining differential entropy:
\begin{defn}[Differential Entropy -- Continuous]
    The differential entropy of a continuous random variable $X$ with support $\mathcal X$ and PDF $f$ is
    \begin{equation}
        H(X)\triangleq -\int_{\mathcal X}\!f(x) \log_2\big (f(x)\big )\, dx.
    \end{equation}
\end{defn} 

Next we define \emph{relative entropy}, which is also called the \emph{Kullback-Leibler divergence} (KL-divergence) for pairs of discrete or continuous distributions. 
\begin{defn}[Relative Entropy -- Discrete]
    Let $X$ and $Y$ be two discrete random variables with PMFs $p$ and $q$ and let $\mathcal X = \operatorname{supp}(X)$.  
    If $\mathcal X \subseteq \operatorname{supp}(Y)$, then the relative entropy between $p$ and $q$ is
    \begin{equation}
        D_\text{KL}(p \Vert q) \triangleq \sum_{x \in \mathcal X} p(x) \ln\left ( \frac{p(x)}{q(x)}\right ).
    \end{equation}
\end{defn}

\begin{defn}[Relative Entropy -- Continuous]
    Let $X$ and $Y$ be two continuous random variables with PDFs $p$ and $q$ and let $\mathcal X = \operatorname{supp}(X)$.  
    If $\mathcal X \subseteq \operatorname{supp}(Y)$, then the relative entropy between $p$ and $q$ is
    \begin{equation}
        D_\text{KL}(p \Vert q) \triangleq \int_{\mathcal X} \!p(x) \ln\left ( \frac{p(x)}{q(x)}\right ) \,dx.
    \end{equation}
\end{defn}

The definitions of relative entropy for discrete and continuous random variables can be retrieved from the more general measure-theoretic definition:
\begin{defn}[Relative Entropy -- General]
    Let $p$ and $q$ be probability measures defined on a measurable space $(\Omega,\mathcal F)$.   
    If $p$ is absolutely continuous with respect to $q$, then the relative entropy between $p$ and $q$ is
    \begin{equation}
        D_\text{KL}(p \Vert q) \triangleq \int_\Omega \ln\left ( \frac{dp}{dq}\right ) dp,
    \end{equation}
    where $\frac{dp}{dq}$ is the Radon-Nikodym derivative of $p$ with respect to $q$. 
\end{defn}

Next we define the mutual information of two discrete or continuous random variables. 
\begin{defn}[Mutual Information -- Discrete]
    \label{def:MI1}
    Let $X$ and $Y$ be discrete random variables with PMFs (marginal distributions) $p_X$ and $p_Y$, support $\mathcal X$ and $\mathcal Y$, and joint PMF $p_{(X,Y)}$. 
    The mutual information of $X$ and $Y$ is
    \begin{equation}
        I(X;Y)\triangleq\sum_{x \in \mathcal X}\sum_{y \in \mathcal Y} p_{(X,Y)}(x,y)\ln \left ( \frac{p_{(X,Y)}(x,y)}{p_X(x) p_Y(y)}\right ).
    \end{equation}
\end{defn}

\begin{defn}[Mutual Information -- Continuous]
    \label{def:MI2}
    Let $X$ and $Y$ be continuous random variables with PDFs (marginal distributions) $p_X$ and $p_Y$, support $\mathcal X$ and $\mathcal Y$, and joint PDF $p_{(X,Y)}$. 
    The mutual information of $X$ and $Y$ is
    \begin{equation}
        I(X;Y)\triangleq\int_{\mathcal X}\! \int_{\mathcal Y}\! p_{(X,Y)}(x,y)\ln \left ( \frac{p_{(X,Y)}(x,y)}{p_X(x) p_Y(y)}\right )\,dy\,dx.
    \end{equation}
\end{defn}

Similar to relatively entropy, the definitions of mutual information for pairs of discrete or continuous random variables can be retrieved from the more general measure-theoretic definition:
\begin{defn}[Mutual Information -- General]
    \label{def:MI3}
    Let $X$ and $Y$ be random variables defined on a probability space $(\Omega, \mathcal F, p)$. 
    Let $p_{(X,Y)}$ be the joint probability measure of $X$ and $Y$, let $p_X$ and $p_Y$ be the marginal probability measures of $X$ and $Y$, and let $p_X \otimes p_Y$ denote the product measure of $p_X$ and $p_Y$. 
    The mutual information between $X$ and $Y$ is then
    \begin{equation}
        I(X;Y) \triangleq D_{\text{KL}}\left (p_{(X,Y)} \Vert p_X \otimes p_Y \right ),
    \end{equation}
    if $p_{(X,Y)}$ is absolutely continuous with respect to $p_X \otimes p_Y$, and $I(X;Y)=\infty$ otherwise.
\end{defn}

Next we present properties of entropy and mutual information, starting with the symmetry of mutual information. 
\begin{prop}[Symmetry of Mutual Information]
    \label{prop:MISym}
    For all random variables $X$ and $Y$, $I(X;Y)=I(Y;X)$.
\end{prop}
\begin{proof}
    See the work of \citet[page 11]{pinsker1964}. Note that Pinsker calls $I(X;Y)$ the \emph{information of one random variable with respect to the other} \citep[page 9]{pinsker1964}.
\end{proof}

Next we review a fundamental result of information theory called the \emph{data processing inequality} (DPI). 
\begin{prop}[Data Processing Inequality]
    \label{thm:DPI}
    Let $X$ and $Y$ be random variables defined on a probability space $(\Omega, \mathcal F, p)$. If $f$ is a measurable function such that $Z=f(Y)$ is a random variable, then, $I(X;Y) \geq I(X;Z)$.
\end{prop}
\begin{proof}
    See the work of \citet[page 11]{pinsker1964}.
\end{proof}

It is well known that applying a deterministic function to a discrete random variable cannot increase its entropy. That is:
\begin{prop}
    \label{prop:DiscreteEntropyProp}
    If $X$ is a discrete random variable and $f$ is a (deterministic) function, then $H(f(X)) \leq H(X)$.
\end{prop}
\begin{proof}
    Before providing a proof we reiterate that this is a well known result and not a contribution of this work. We provide a proof of this property because some standard reference texts (including those that we are relying on) do not explicitly state this property \citep{CoverThomas2006}. 

    It follows from the DPI that: 
    \begin{equation}
        \label{eq:DPI_result}
        I(X; f(X)) \leq I(X;X).
    \end{equation}
    Since $I(X;X)=H(X)$ \citep[Equation 2.47]{CoverThomas2006} we therefore have that
    \begin{equation}
        \label{eq:entropyProofStep}
        I(X;f(X)) \leq H(X).
    \end{equation} 

    Next, recall that the conditional entropy $H(f(X)|X)$ is defined as \citep[Equation 2.10]{CoverThomas2006}:
    \begin{equation}
        \label{eq:lkajwt}
        H(f(X)|X)\triangleq \sum_{x \in \operatorname{supp}(X)} \Pr(X=x)H(f(X)|X=x).
    \end{equation}
    Since $f(X)$ is deterministic given that $X=x$, and since the entropy of a deterministic random variable is zero, we have that $H(f(X)|X=x)=0$. So, the entire right hand side of \eqref{eq:lkajwt} is zero: $H(f(X)|X)=0$.

    Combining this result with the property that $I(X;f(X)) = H(f(X)) - H(f(X)|X)$ \citep[Equation 2.44]{CoverThomas2006}, we have that $I(X;f(X)) = H(f(X))$. Substituting this result into \eqref{eq:entropyProofStep} we have that
    \begin{equation}
        H(f(X)) \leq H(X),
    \end{equation}
    establishing the property.
\end{proof}

This result does \emph{not} extend to continuous random variables, since applying a deterministic function to a continuous random variable can increase the differential entropy.
\begin{prop}
    \label{prop:contEntropyInflation}
    If $X$ is a continuous random variable and $f$ is a (deterministic) function, then it can occur that $H(f(X)) > H(X)$.
\end{prop}
\begin{proof}
    Let $X$ be uniform on $[0,1]$ and let $f(x)=2x$. Then 
    \begin{align}
        H(X) =& -\int_0^1 \!1 \underbrace{\log_2(1)}_{=0}\,dx\\
        =&0.
    \end{align}
    and 
    \begin{align}
        H(f(X))=&-\int_0^2 \!0.5 \underbrace{\log_2(0.5)}_{-1}\,dx\\
        =& 1.
    \end{align}
    Hence, this is a case where $H(f(X)) > H(X)$.
\end{proof}

Notice that we write $H(X)$ and $I(X;Y)$ even though entropy and mutual information are properties of the distributions of random variables like $X$ and $Y$---they are not functions of specific values of $X$ and $Y$. Similarly, here we consider the space of such functions---functions whose domain is the distribution of a random variable (or the distributions of multiple random variables). We call these functions \emph{functions of distributions}. When writing symbols that represent functions of distributions other than entropy and mutual information, we place a breve above the symbol to indicate that it is a function of the distributions of its arguments (the distributions of random variables) rather than a function of the arguments themselves, e.g., $\breve f(X)$. We adopt this notation rather than explicitly using a distribution as the argument of $\breve f$ because later we will substitute entropy and mutual information for functions of distributions like $\breve f$. Furthermore, if $\breve f$ is a function of distributions and $g$ is a function, we write $\breve f(g(X))$ to denote $\breve f$ applied to the distribution of $g(X)$. 

We say that a function of a distribution is \emph{invariant to invertible transformations} if applying an invertible transformation to its inputs does not change its value. We formalize this notion in the following definition.
\begin{defn}[Invariant to Invertible Transformations -- Univariate]
    A function of a distribution $\breve f$ is invariant to invertible transformations if and only if, for all invertible functions $g$,
    \begin{equation}
        \breve f(X) = \breve f(g(X)).
    \end{equation}
\end{defn}

It is well known that Shannon entropy is invariant to invertible transformations. 
\begin{prop}[Shannon Entropy Invariant]
    \label{prop:shannonInvariant}
    If $X$ is a discrete random variable and $g:\mathcal X \to \mathcal Y$ is an invertible function, then $H(g(X)) = H(X)$.
\end{prop}
\begin{proof}
    It follows from Property \ref{prop:DiscreteEntropyProp} that $H(g(X)) \leq H(X)$. Since $g$ is invertible, we also have that
    \begin{align}
        H(X) =& H\left (g^{-1}(g(X)\right )\\
        \overset{\text{(a)}}{\leq}& H(g(X)),
    \end{align}
    where \textbf{(a)} again follows from Property \ref{prop:DiscreteEntropyProp}. Together these inequalities imply that $H(g(X)) = H(X)$, and so Shannon entropy is invariant under invertible transformations. 
\end{proof}

However, it is also well known that differential entropy is \emph{not} invariant to invertible transformations.
\begin{prop}[Differential Entropy Not Invariant]
    \label{prop:differentialNotInvariant}
    If $X$ is a continuous random variable and $g:\mathcal X \to \mathcal Y$ is an invertible function, then it can occur that $H(g(X)) \neq H(x)$.
\end{prop}
\begin{proof}
    This follows from the proof of Property \ref{prop:contEntropyInflation}, since it relies on a function $f$ that is invertible.
\end{proof}
In fact, it is not merely differential entropy that is not invariant to invertible transformations when considering continuous random variables---there do not exist any functions of distributions of \emph{univariate} continuous random variables that are invariant to invertible transformations.
\begin{prop}[Nonexistence of Continuous Invariant]
    \label{prop:univariateNonexistence}
    If $\breve f$ is invariant to invertible transformations, then $\breve f$ must be constant for all continuous random variables.
\end{prop}
\begin{proof}
    We provide a proof by contradiction. Suppose for contradiction that there exists a function of distributions $\breve f$ that is invariant to invertible transformations and which is not constant for all univariate continuous random variables. This implies that there exist two univariate continuous random variables $X$ and $Y$ such that 
    \begin{equation}
        \label{eq:nonexistenceProofContradictionEqn}
        \breve f(X) \neq \breve f(Y). 
    \end{equation}

    Let $F_X$ and $F_Y$ be the \emph{cumulative distribution functions} (CDFs) of $X$ and $Y$ respectively. Since $X$ and $Y$ are continuous distributions, $F_X$ and $F_Y$ are strictly increasing and hence invertible. Furthermore, the probability integral transform \citep[Theorem 2.1.10]{casella2020statistical} implies that $F_X(X)$ and $F_Y(Y)$ are both uniform on $[0,1]$. Hence:
    \begin{align}
        \breve f(X)\overset{\text{(a)}}{=}&\breve f(F_X(X))\\
        \overset{\text{(b)}}{=}& \breve f(F_Y(Y))\\
        \label{eq:nonexistenceProofContradictionEqn2}
        \overset{\text{(c)}}{=}& \breve f(Y),
    \end{align}
    where \textbf{(a)} and \textbf{(c)} follow from $\breve f$ being invariant to invertible transformations (including $F_X$ and $F_Y$) and \textbf{(b)} follows from $F_X(X)$ and $F_Y(Y)$ having the same distribution---they are both uniform on $[0,1]$. Notice that \eqref{eq:nonexistenceProofContradictionEqn} and \eqref{eq:nonexistenceProofContradictionEqn2} cannot both be true, establishing a contradiction.
\end{proof}

To summarize so far, there exist functions of distributions that are invariant to invertible transformations if we restrict our consideration to discrete random variables. One example of such a function of distributions is Shannon entropy. However, when we consider continuous univariate random variables, there do not exist any \emph{nontrivial} functions of distributions that are invariant to invertible transformations, where \emph{nontrivial} means non-constant for continuous distributions. This implies, for example, that differential entropy is not invariant to invertible transformations. 

Next we explain and provide support for the following more positive result: If we expand our consideration to functions of the distributions of \emph{multiple} random variables and we restrict the transformations to independent invertible transformations of each random variable, then there exist nontrivial functions of distributions that are invariant to invertible transformations. First, we generalize the definition of invariance to invertible transformations to functions of multiple random variables. 
\begin{defn}[Invariant to Invertible Transformations -- Multivar.]
    For any $n > 2$, a function of $n$ distributions $\breve f$ is invariant to invertible transformations if and only if, for all sequences of $n$ random variables $X_1, \dotsc, X_n$ and all sequences of measurable and invertible functions $g_1,\dotsc,g_n$,
    \begin{equation}
        \breve f(X_1,X_2,\dotsc,X_n) = \breve f(g_1(X_1), g_2(X_2),\dotsc,g_n(X_n)).
    \end{equation}
\end{defn}
Notice that one can define a random variable $Y=(X_1,X_2)$, and so a function of a single distribution (that of $Y$) can also be a function of multiple distributions (in this case, the distributions of $X_1$ and $X_2$). However, the univariate definition of invariance to invertible transformations would require invariance to transformations $g(Y)$ that can transform $X_1$ and $X_2$ in a dependent way, whereas the multivariate definition of invertible transformations only considers independent transformations of $X_1$ and $X_2$, e.g., the value of $g_2(X_2)$ does not depend on $X_1$. 

In the following property we show that mutual information is invariant to invertible transformations, regardless of whether one considers discrete, continuous, or hybrid distributions. 
\begin{prop}[Mutual Information is Invariant]
    \label{prop:MIInvariant}
    Let $X$ and $Y$ be random variables and let $g$ and $h$ be measurable and invertible functions. Then $I(X;Y)=I(g(X);h(Y))$. 
\end{prop}
\begin{proof}
    Although this is a well-known result, we were unable to find a suitable reference that covers the full measure theoretic setting without additional assumptions. We therefore provide a proof based on the data processing inequality, Property \ref{thm:DPI}. Note that the assumption that $g$ and $h$ are measurable is necessary to ensure that $g(X)$ and $g(Y)$ remain well-defined random variables on the underlying probability space.
    
    We establish that $I(X;Y)=I(g(X);h(Y))$ by showing that 
    \begin{equation}
        \label{eq:MIInvariant1}
        I(X;Y)=I(X;h(Y))
    \end{equation}
    and then that 
    \begin{equation}
        \label{eq:MIInvariant2}
        I(X;h(Y))=I(g(X);h(Y)),
    \end{equation}
    which together imply that $I(X;Y) = I(g(X);h(Y))$.

    First, to establish \eqref{eq:MIInvariant1} we show that $I(X;Y) \geq I(X;h(Y))$ and $I(X;Y) \leq I(X;h(Y))$. By the data processing inequality, $I(X;Y) \geq I(X;h(Y))$ since $h$ is a measurable function. Similarly, by the data processing inequality, $I(X;Y) \leq I(X;h(Y))$ since
    \begin{align}
        I(X;h(Y)) \geq& I(X;h^{-1}(h(Y)))\\
        =& I(X;Y).
    \end{align}
    Notice that the data processing inequality is applicable here because $h$ being measurable and invertible implies that $h^{-1}$ exists and is also measurable. 

    Second, to establish \eqref{eq:MIInvariant2} we show that $I(X;h(Y))\geq I(g(X);h(Y))$ and $I(X;h(Y))\leq I(g(X);h(Y))$. $I(X;h(Y))\geq I(g(X);h(Y))$ follows from the data processing inequality because $g$ is a measurable function. Similarly, by the data processing inequality, $I(X;h(Y))\leq I(g(X);h(Y))$ since
    \begin{align}
        I(g(X);h(Y)) \geq& I(g^{-1}(g(X));h(Y))\\
        =& I(X, h(Y)).
    \end{align}
    Notice that the data processing inequality is applicable here because $g$ being measurable and invertible implies that $g^{-1}$ exists and is also measurable.
\end{proof}

Relative entropy (KL divergence) is one of a broader class of functions of distributions called $f$-divergences. Whereas mutual information is invariant to invertible transformations, $f$-divergences (including relative entropy) are invariant when the same differentiable invertible transformation is applied to both random variables. This result is significantly weaker since it not only requires the transformation to be differentiable, but it also requires the transformations of the two random variables to be the same.

\begin{prop}
    If $X$ and $Y$ be random variables, $g$ is a differentiable and invertible function, and $D_f$ is an $f$-divergence, then $D_f(X\Vert Y)=D_f(g(X)\Vert g(Y))$.
\end{prop}
\begin{proof}
    See the work of \citet{qiao2010study}.
\end{proof}

\PackageWarning{manual}{Warning: Figure occurs prior to reference.}
\begin{figure*}[thbp]
    \begin{mdframed}[linewidth=1pt, roundcorner=4pt]
        \begin{align}
            \nabla J(\theta)\overset{\text{(a)}}{=}&\mathbf{E}\left [ \sum_{t=0}^\infty \gamma^t \big(G_t -v^\pi(S_t)-b(S_t)\big)\frac{\partial \ln \big ( \pi(S_t,A_t,\theta) \big )}{\partial \theta}\middle |\Theta_{\forall t}=\theta\right ]\\
            =& \sum_{t=0}^\infty \gamma^t \mathbf{E}\left [\big (G_t -v^\pi(S_t)-b(S_t)\big) \frac{\partial \ln \big ( \pi(S_t,A_t,\theta) \big )}{\partial \theta}\middle |\Theta_{\forall t}=\theta\right ]\\
            \overset{\text{(b)}}{=}& \sum_{t=0}^\infty \gamma^t \mathbf{E}\left [ \mathbf{E}\left [\big (G_t -v^\pi(S_t)-b(S_t)\big ) \frac{\partial \ln \big ( \pi(S_t,A_t,\theta) \big )}{\partial \theta}\middle | S_t, A_t, \Theta_{\forall t} = \theta \right ] \middle |\Theta_{\forall t}=\theta \right ]\\
            \overset{\text{(c)}}{=}& \sum_{t=0}^\infty \gamma^t \mathbf{E}\left [ \Big (\mathbf{E}\left [G_t \middle | S_t, A_t, \Theta_{\forall t} = \theta \right ] - v^\pi(S_t) - b(S_t)\Big ) \frac{\partial \ln \big ( \pi(S_t,A_t,\theta) \big )}{\partial \theta} \middle |\Theta_{\forall t}=\theta \right ]\\
            \overset{\text{(d)}}{=}& \sum_{t=0}^\infty \gamma^t \mathbf{E}\left [ \Big ( \mathbf{E}\left [\Delta_{t+1} \middle | S_t, A_t \right ] - b(S_t)\Big )\frac{\partial \ln \big ( \pi(S_t,A_t,\theta) \big )}{\partial \theta} \middle |\Theta_{\forall t}=\theta \right ]\\
            =&\sum_{t=0}^\infty \gamma^t \mathbf{E}\left [ \big (\Delta_{t+1} -b(S_t)\big )\frac{\partial \ln \big ( \pi(S_t,A_t,\theta) \big )}{\partial \theta} \middle |\Theta_{\forall t}=\theta \right ]\\
            \label{eq:ljalkj245lkjwf}=&\mathbf{E} \left [\sum_{t=0}^\infty \gamma^t\big (\Delta_{t+1}-b(S_t)\big )\frac{\partial \ln \big ( \pi(S_t,A_t,\theta) \big )}{\partial \theta} \middle |\Theta_{\forall t}=\theta \right ],
        \end{align} 
        where \textbf{(a)} follows from \eqref{eq:PolicyGradientBaseline} using the combined baseline $v^\pi(S_t)+b(S_t)$, \textbf{(b)} follows from the law of iterated expectations, \textbf{(c)} follows because, within the inner expectation, $v^\pi(S_t)$ and $ \partial \ln \left ( \pi(S_t,A_t,\theta) \right ) / \partial \theta$ are constant, and \textbf{(d)} follows from \eqref{eq:expectedTDError}. Recall that in this derivation the TD error $\Delta_{t+1}$ is defined in terms of the state-value function $v^\pi$ rather than an approximation thereof.
    \end{mdframed}
    \PackageWarning{manual}{Warning: Confirm the reference is actually below as stated.}
    \caption{A proof sketch referenced in the text below.}
    \label{fig:TDProofFig}
\end{figure*}

\section{Deriving BAC With a Reinforcement Baseline}
\label{app:BACDerivation}

In this section we show how the BAC algorithm can be modified to include a reinforcement baseline. 
At a high level, we do this by deriving the policy parameter update of BAC from the policy gradient, but with an extra reinforcement baseline $b(S_t)$ that carries through the derivation, resulting in a variant of BAC that includes a reinforcement baseline. 

To see the connection been BAC and the policy gradient, first consider the VFA approximation in more detail. Recall that $v(p,w)$ is the value of the VFA approximation with perception $p$ and VFA weights $w$. Since we are considering MDPs, we assume that the VFA parameterization ignores the reward component of perceptions $P_t=(S_t,R_t)$, and write $v(s,w)$, where $s$ is a state. This is an abuse of notation because $v$ was defined to take perceptions as input, not states.

The aim of the VFA updates in the BAC algorithm is to make $v(\cdot,w)$ approximate the state-value function,\footnote{Alternatively, the state-value function might be denoted by $v^\theta$ to highlight that the policy depends on $\theta$. Also, recall that $\Theta_{\forall t}=\theta$ is shorthand for the event $\forall t \in \mathbb N,\,\Theta_t=\theta$.}
\begin{equation}
    v^\pi(s)\triangleq\mathbf{E}\left [ G_t | S_t=s, \Theta_{\forall t}=\theta \right ]. 
\end{equation}
If the VFA parameterization $v$ with VFA weights $w$ perfectly approximates this state-value function so that $v(\cdot,w)=v^\pi$, then it is well known that\footnote{This equation and its subsequent derivation suggest that BAC already uses an estimate of $v^\pi(s)$ as a baseline. The term $b(S_t)$ that we introduce later within Figure \ref{fig:TDProofFig} corresponds to a reinforcement baseline that is included in addition to this existing implicit baseline.} 
\begin{equation}
    \label{eq:expectedTDError}
    \mathbf{E}[\Delta_{t+1} | S_t=s,A_t=a]=\mathbf{E}[G_t |S_t=s,A_t=a,\Theta_{\forall t}=\theta]-v^\pi(s).
\end{equation} 
Although we omit a full proof of this property, we provide a proof sketch:
\begin{align}
    &\mathbf{E}[\Delta_{t+1} | S_t{=}s,A_t{=}a]\\
    =&\mathbf{E}[R_{t+1} + \gamma v^\pi(S_{t+1})-v^\pi(S_t)|S_t{=}s, A_t{=}a]\\
    \nonumber =&\mathbf{E}[R_{t+1} + \gamma \left ( R_{t+1} + \gamma R_{t+2} + \cdots \right ) |S_t{=}s, A_t{=}a,\Theta_{\forall t}{=}\theta]-v^\pi(s)\\
    \nonumber =&\mathbf{E}[R_{t+1} + \gamma  R_{t+1} + \gamma^2 R_{t+2} + \cdots |S_t{=}s, A_t{=}a,\Theta_{\forall t}{=}\theta]-v^\pi(s)\\
    =&\mathbf{E}[G_t |S_t{=}s, A_t{=}a,\Theta_{\forall t}{=}\theta]-v^\pi(s).
\end{align}
It can therefore be shown that, in this case where the VFA approximation is perfect, $\Delta_{t+1}$ can be substituted for $G_t$ in \eqref{eq:PolicyGradientBaseline}, the expression for the policy gradient with a baseline. Although we again omit a complete proof, we provide a proof sketch in Figure \ref{fig:TDProofFig} (we use a figure to allow it to span across both columns).

In order to derive the policy update rule within BAC from the policy gradient expression in \eqref{eq:ljalkj245lkjwf}, we make a sequence of approximations. First, we consider the (unbiased if the VFA perfectly approximates the state-value function) estimator of the policy gradient
\begin{equation}
\label{eq:lkajsglkj242t}
\widehat{\nabla J}(\theta)=\sum_{t=0}^\infty \gamma^t(\Delta_{t+1}-b(S_t))\frac{\partial \ln \big ( \pi(S_t,A_t,\theta) \big )}{\partial \theta}.
\end{equation}
Next, we consider the contribution to this estimator from only a single time step:
\begin{equation}
    \widehat{\nabla J_t}(\theta)=\gamma^t (\Delta_t-b(S_t))\frac{\partial \ln\left ( \pi(S_t,A_t,\theta) \right )}{\partial \theta}.
\end{equation}
This per-time-step gradient estimate results in the per-time-step approximate policy gradient update
\begin{equation}
    \Theta_t \gets \Theta_{t-1} + \beta \gamma^t  (\Delta_{t}-b(S_{t-1})) \frac{\partial \ln \big (\pi(S_{t-1}, A_{t-1}, \Theta_{t-1})\big )}{\partial \Theta_{t-1}}.
\end{equation}
Notice that applying this update at each time $t \in \mathbb N$ would \emph{not} exactly reproduce a single policy update using the policy gradient estimator in \eqref{eq:lkajsglkj242t}. Although the cumulative update to the policy parameters would sum over the time steps $t \in \mathbb N$, the policy parameters would change between updates, altering both \textbf{a)} terms within the gradient estimate like $\pi(S_{t-1},A_{t-1},\Theta_{t-1})$, and \textbf{b)} the distribution of states and actions that the agent encounters. The cumulative change to the policy parameters after one full episode would therefore be a biased estimator of the policy gradient, even if the VFA perfectly approximates the state-value function.

Following standard practice \citep{Thomas2014,Nota2020policy}, we drop the $\gamma^t$ from the per-time-step update, which results in further bias. This results in the final update,
\begin{equation}
    \Theta_t \gets \Theta_{t-1} + \beta (\Delta_{t}-b(S_{t-1})) \frac{\partial \ln\big (\pi(S_{t-1}, A_{t-1}, \Theta_{t-1})\big )}{\partial \Theta_{t-1}},
\end{equation}
which, corresponds to the BAC update with a reinforcement baseline $b(S_t)$ added.

\section{Hyperparameter Settings}
\label{app:hyp}

In this appendix we describe the hyperparameters of the BAC algorithm used in our experiments.

\subsection{Tabular Softmax Policy and Tabular VFA}

For simplicity, we used tabular policy and VFA parameterizations. Before defining these tabular parameterizations more formally, we first review relevant notation and properties of how MDPs can be modeled within AEPs. Recall that, when modeling an MDP in an AEP, we encode the state and reward within the perception so that $P_t=(S_t,R_t)$. For simplicity here, we also assume that the states of a finite MDP correspond to the initial positive integers: $\mathcal S = \{1,2,\dotsc,|\mathcal S|\}$.  

When using a tabular VFA, the number of weights (i.e., the length of the vector $W_t$) equals the number of states. Intuitively, the VFA stores one weight per state, representing that state's estimated value. More formally, for every perception $p=(s,r)$ and VFA weight vector $w$, 
\begin{equation}
    v(p,w)=w_{s},
\end{equation}
where $w_s$ denotes the $s^\text{th}$ element of $w$.

When using a tabular policy parameterization, the number of policy parameters (i.e., the length of $\Theta_t$) equals the product of the number of states and the number of actions. For each state $s$ and action $a$, there is one corresponding policy parameter, which we denote by $\theta_{s,a}$ (for any policy parameter vector $\theta$). Given the state $s$, the action is sampled from a softmax distribution defined by the policy parameters associated with that state. More formally, for all states $s$, actions $a$, and policy parameters $\theta$,
\begin{align}
    \label{eq:softmaxTabular}
    \pi_\text{BAC}(s,a,\theta)=\frac{e^{\theta_{s,a}}}{\sum_{a'}e^{\theta_{s,a'}}}.
\end{align}

\subsection{Compatible Features}
\label{sec:compatibleFeatures}

The term $\ln(\pi(s,a,\theta))/\partial \theta$ often appears in expressions for the policy gradient, and is sometimes called the \emph{compatible features} \citep{Bhatnagar2007}. Here we derive the compatible features for the tabular policy used in our experiments. For brevity (and because this derivation applies when tabular softmax representations are used beyond BAC algorithms), we drop the BAC subscript and write $\pi$ for $\pi_\text{BAC}$ here. We also introduce shorthand notation for the numerator and denominator of \eqref{eq:softmaxTabular}---notation that conflicts with other symbols in this report and which we only use here. Let 
\begin{equation}
    \alpha_a \triangleq e^{\theta_{s,a}}
\end{equation}
and
\begin{equation}
    \beta \triangleq \sum_{\hat a \in \mathcal A}e^{\theta_{s,\hat a}},
\end{equation}
so that \eqref{eq:softmaxTabular} can we rewritten as
\begin{align}
    \label{eq:lkjh24l6kjs}
    \pi(s,a,\theta)=\frac{\alpha_a}{\beta}.
\end{align}
We will derive an expression for each element of the compatible features: 
\begin{align}
    \label{eq:compatibleFeatureElement}
    \frac{\partial \ln \big ( \pi(s,a,\theta)\big )}{\partial \theta_{s',a'}},
\end{align}
where $s'$ and $a'$ may or may not equal $s$ and $a$. 

First, notice that if $s' \neq s$, then \eqref{eq:compatibleFeatureElement} is zero, since the value of $\theta_{s',a'}$ does not change $\alpha_a$ or $\beta$, and hence does not influence the right hand side of \eqref{eq:lkjh24l6kjs}. Next we consider the case where $s'=s$, which can be further broken into two cases: when $a'\neq a$ and when $a' = a$. First, consider the case where $s'=s$ and $a'\neq a$:
\begin{align}
    \label{eq:lkj24365lkj1a}\frac{\partial \ln \big ( \pi(s,a,\theta)\big )}{\partial \theta_{s',a'}}=&\frac{1}{\pi(s,a,\theta)}\frac{\partial}{\partial \theta_{s',a'}} \pi(s,a,\theta)\\
    =&\frac{1}{\pi(s,a,\theta)}\frac{\partial}{\partial \theta_{s',a'}} \frac{\alpha_a}{\beta}\\
    \label{eq:lkj24365lkj1b}\overset{\text{(a)}}{=}&\frac{1}{\pi(s,a,\theta)} \frac{\left (\frac{\partial \alpha_a}{\partial \theta_{s',a'}}\right )\beta  - \alpha_a \frac{\partial \beta}{\partial \theta_{s',a'}}}{\beta^2}\\
    \label{eq:lkj2465kljsdf}\overset{\text{(b)}}{=}&\frac{1}{\pi(s,a,\theta)} \frac{-\alpha_a \alpha_{a'}}{\beta^2},
\end{align}
where \textbf{(a)} follows from the quotient rule and \textbf{(b)} follows because $\partial \alpha_a / \partial \theta_{s',a'}=0$ (since we are considering the case where $a' \neq a$) and $\partial \beta / \partial \theta_{s',a'}=\alpha_{a'}$ (since we are considering the case where $s'=s$). Since $\alpha_a/\beta=\pi(s,a,\theta)$ and $\alpha_{a'}/\beta=\pi(s,a',\theta)$, it follows from \eqref{eq:lkj2465kljsdf} that
\begin{align}
    \frac{\partial \ln \big ( \pi(s,a,\theta)\big )}{\partial \theta_{s',a'}}=&-\frac{1}{\pi(s,a,\theta)} \pi(s,a,\theta)\pi(s,a',\theta)\\
    =&-\pi(s,a',\theta). 
\end{align}
Finally, consider the case where $s'=s$ and $a'=a$
\begin{align}
    \label{eq:lkj24365lkj1}\frac{\partial \ln \big ( \pi(s,a,\theta)\big )}{\partial \theta_{s',a'}}=&\frac{1}{\pi(s,a,\theta)}\frac{\partial}{\partial \theta_{s',a'}} \pi(s,a,\theta)\\
    =&\frac{1}{\pi(s,a,\theta)}\frac{\partial}{\partial \theta_{s',a'}} \frac{\alpha_a}{\beta}\\
    \label{eq:lkj24365lkj12}=&\frac{1}{\pi(s,a,\theta)} \frac{\left (\frac{\partial \alpha_a}{\partial \theta_{s',a'}}\beta\right )  - \alpha_a \frac{\partial \beta}{\partial \theta_{s',a'}}}{\beta^2}\\
    \overset{\text{(a)}}{=}&\frac{1}{\pi(s,a,\theta)} \frac{\alpha_a \beta -\alpha_a \alpha_{a}}{\beta^2},
\end{align}
where lines \eqref{eq:lkj24365lkj1}--\eqref{eq:lkj24365lkj12} are identical to \eqref{eq:lkj24365lkj1a}--\eqref{eq:lkj24365lkj1b}, and \textbf{(a)} follows because, when $s=s'$ and $a=a'$, $\partial \alpha_a / \partial \theta_{s',a'}=\alpha_a$  and $\alpha_a=\alpha_{a'}$. Hence,
\begin{align}
    \frac{\partial \ln \big ( \pi(s,a,\theta)\big )}{\partial \theta_{s',a'}}=&\frac{1}{\pi(s,a,\theta)} \left ( \frac{\alpha_a}{\beta} - \left (\frac{\alpha_a}{\beta}\right)^2\right )\\
    =&\frac{1}{\pi(s,a,\theta)} \big ( \pi(s,a,\theta) - \pi(s,a,\theta)^2\big )\\
    =& 1-\pi(s,a,\theta).
\end{align}

Combining all of these cases, we obtain the following expression for each element of the compatible features:
\begin{align}
    \label{eq:tabularCompatibleFeaturesSOlution}
    \frac{\partial \ln\big (\pi(s, a, \theta)\big )}{\partial \theta_{s',a'}}
    =&\begin{cases}
    0 &\mbox{if }s'\neq s\\
    -\pi(s,a', \theta)&\mbox{if } s'=s \text{ and } a'\neq a\\
    1-\pi(s, a, \theta)&\mbox{if } s'=s \text{ and } a'=a.
    \end{cases}
\end{align}

\subsection{Other Hyperparameters}

For both environments we defined the initial policy parameters and VFA weights to be zero, i.e., $\theta_0=0$ and $w_0=0$. We also used $\gamma=1$ and $\lambda=0.8$. For the gridworld we used $\alpha = 0.1$ and $\beta=0.01$, while for the chain environment we used $\alpha = 0.1$ and $\beta = 0.1$. These hyperparameters were found via a manual search that only considered performance. We estimate that roughly five sets of hyperparameters were tested for each environment, as the aim was only to find parameters that result in reliable learning, not to find optimal hyperparameters. Although few hyperparameters were tested, the author has significant prior experience with similar algorithms and environments, and so the hyperparameter settings were informed by a much larger number of related simulations.

\vfill
\noindent
\makebox[\linewidth]{\leaders\hrule\hfill\quad\textbf{End}\quad\leaders\hrule\hfill}

\end{appendix}
\end{document}